\documentclass[10pt,twocolumn,letterpaper]{article}

\usepackage{cvpr}              
\definecolor{cvprblue}{rgb}{0.21,0.49,0.74}
\usepackage[pagebackref,breaklinks,colorlinks,allcolors=cvprblue]{hyperref}
\usepackage{diagbox}

\def\paperID{33797} 
\def\confName{CVPR}
\def\confYear{2026}

\title{FMMC: Harnessing the Power of Foundation Models for Accurate Material Classification}

\author{
Qingran Lin$^{1}$\textsuperscript{*} \quad
Fengwei Yang$^{2}$\textsuperscript{*} \quad
Chaolun Zhu$^{3}$ \\
\\
$^{1}$ Georgia Institute of Technology \quad
$^{2}$ Duke University \quad
$^{3}$ Waseda University
}

\begin{document}
\twocolumn[{%
	\renewcommand\twocolumn[1][]{#1}%
    
	\maketitle

            \resizebox{\textwidth}{!}{%
            \definecolor{red}{RGB}{255, 0, 0}
            \definecolor{green}{RGB}{0,255,0}
            \definecolor{blue}{RGB}{0,0,255}
            \definecolor{yellow}{RGB}{255,255,0}
            \definecolor{magenta}{RGB}{255,0,255}
            \definecolor{cyan}{RGB}{0,255,255}
            \definecolor{gray}{RGB}{128,128,128}
            \definecolor{orange}{RGB}{255,165,0}
                \begin{tabular}
                {cccccccc}
                    \includegraphics[width=7cm]{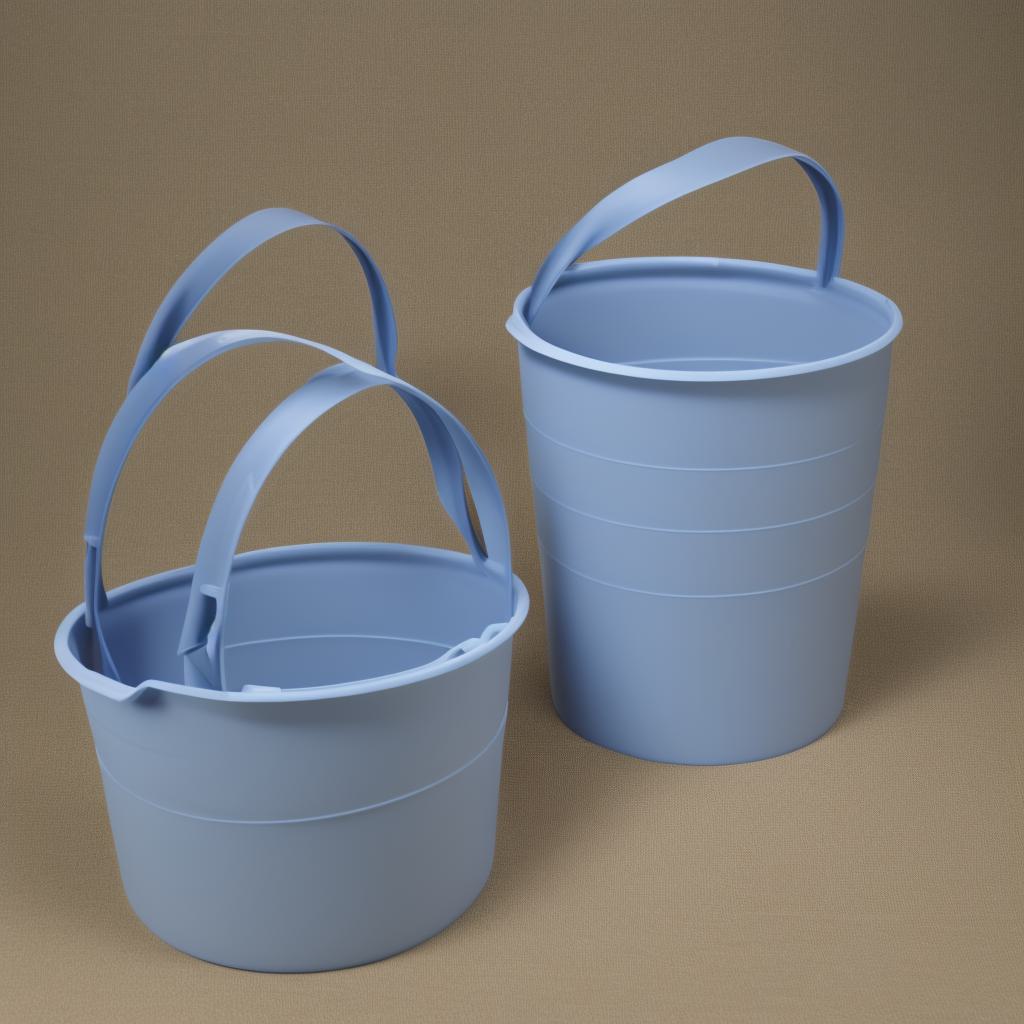}
                    & \includegraphics[width=7cm]{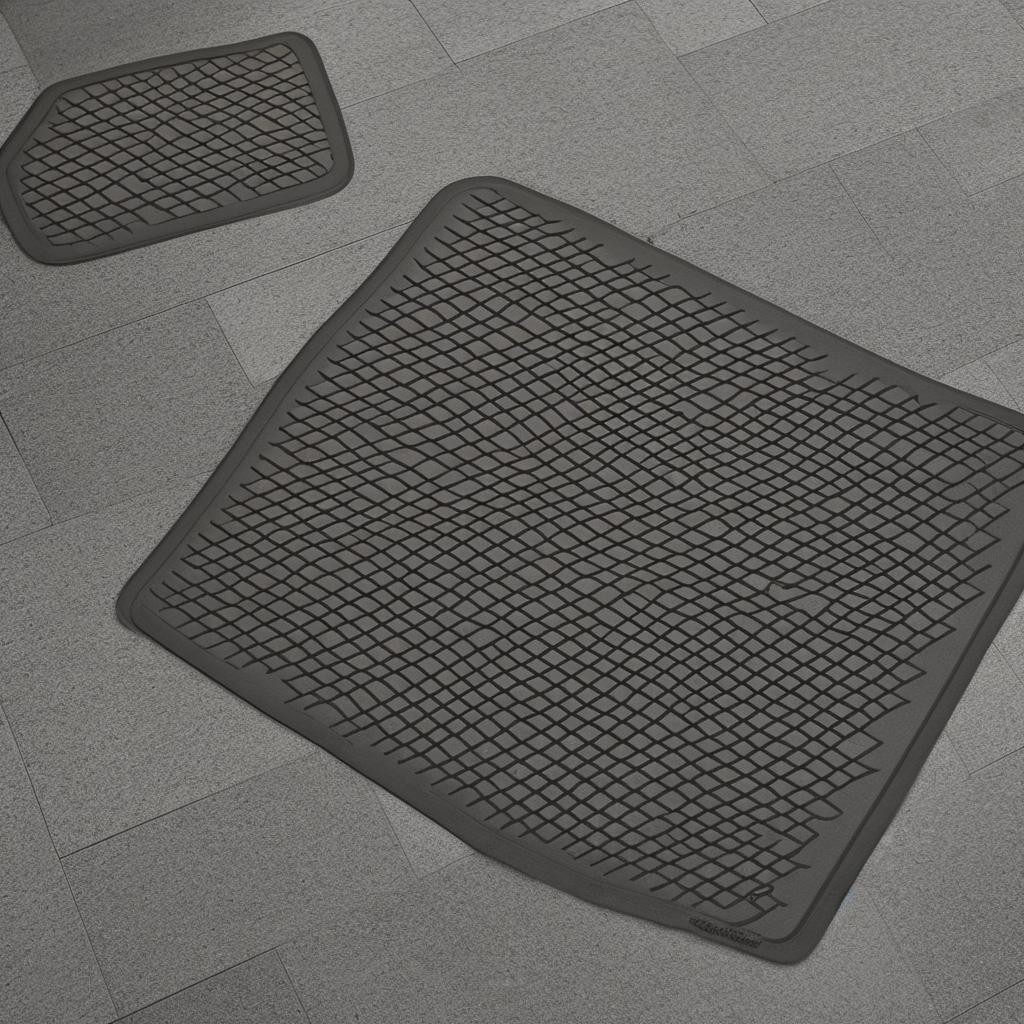} 
                    & \includegraphics[width=7cm]{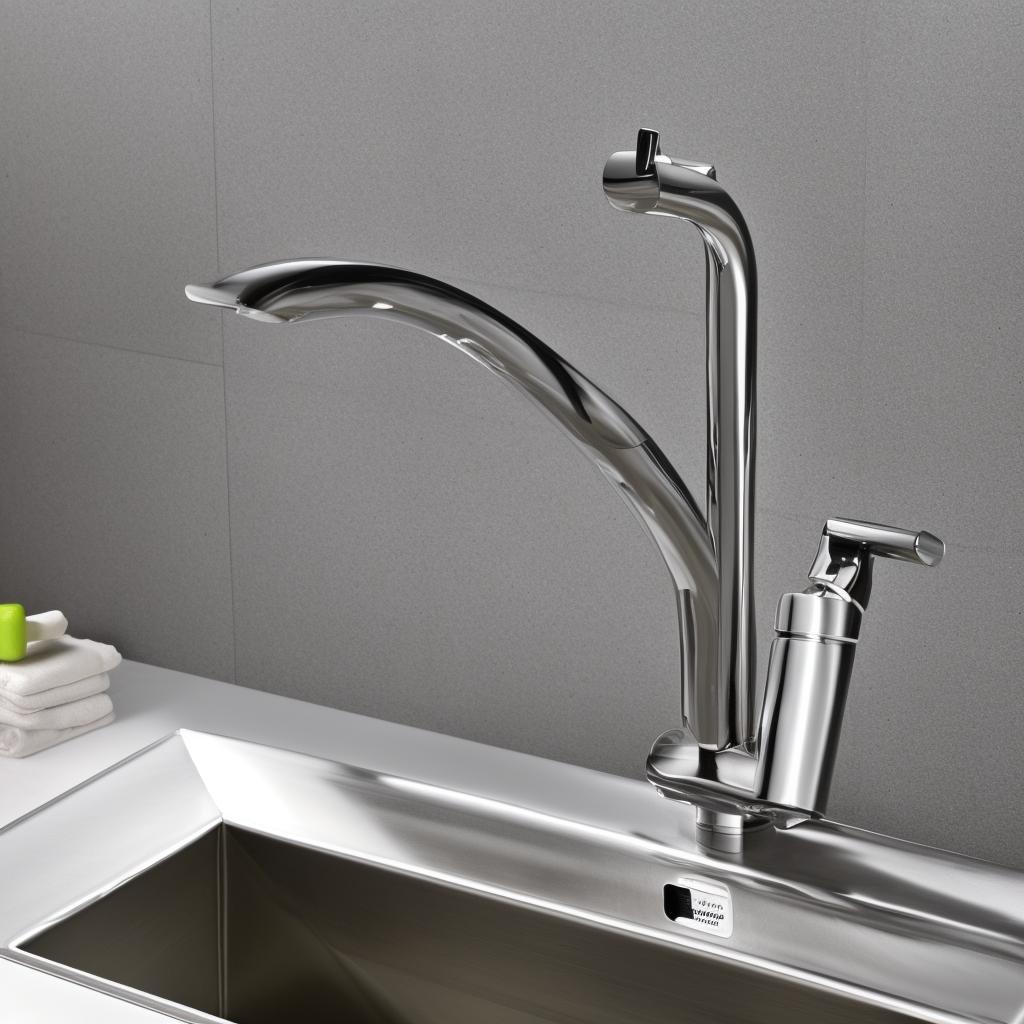} 
                    & \includegraphics[width=7cm]{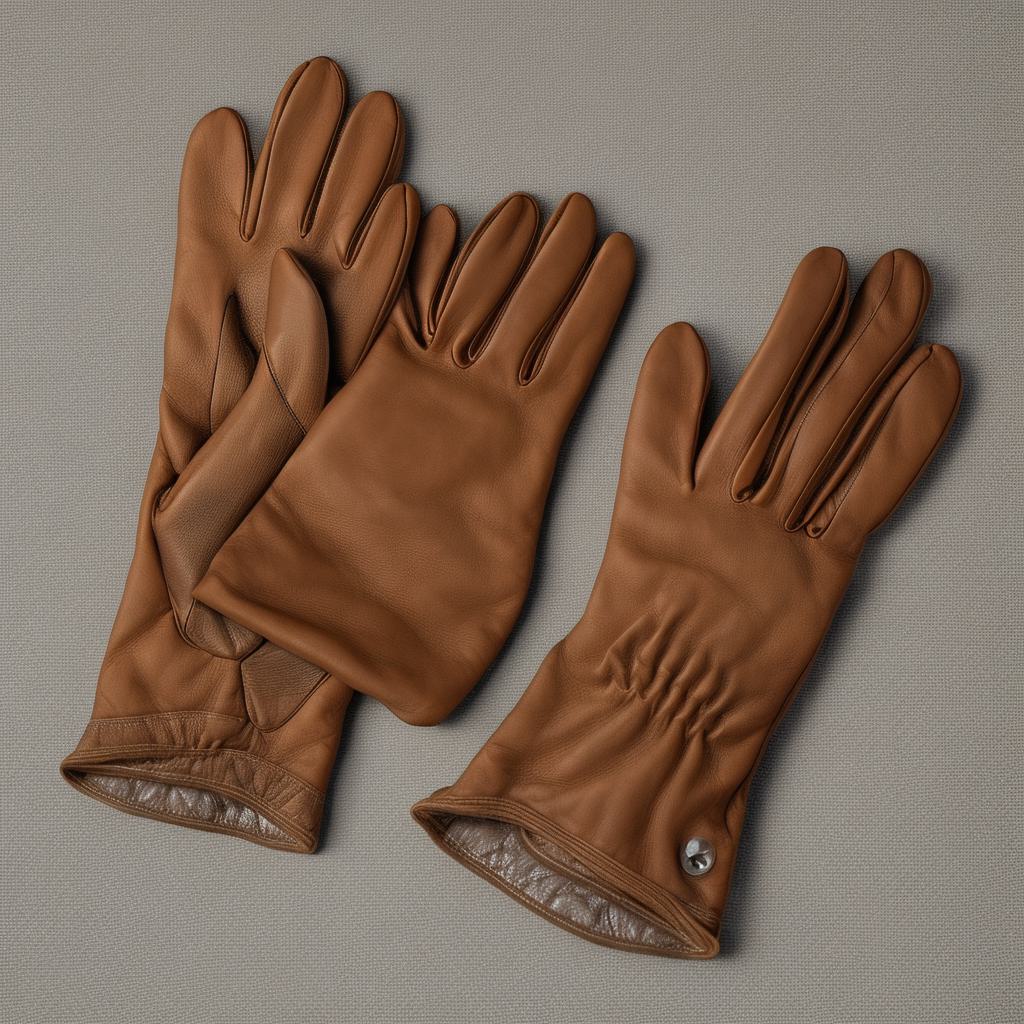} 
                    & \includegraphics[width=7cm]{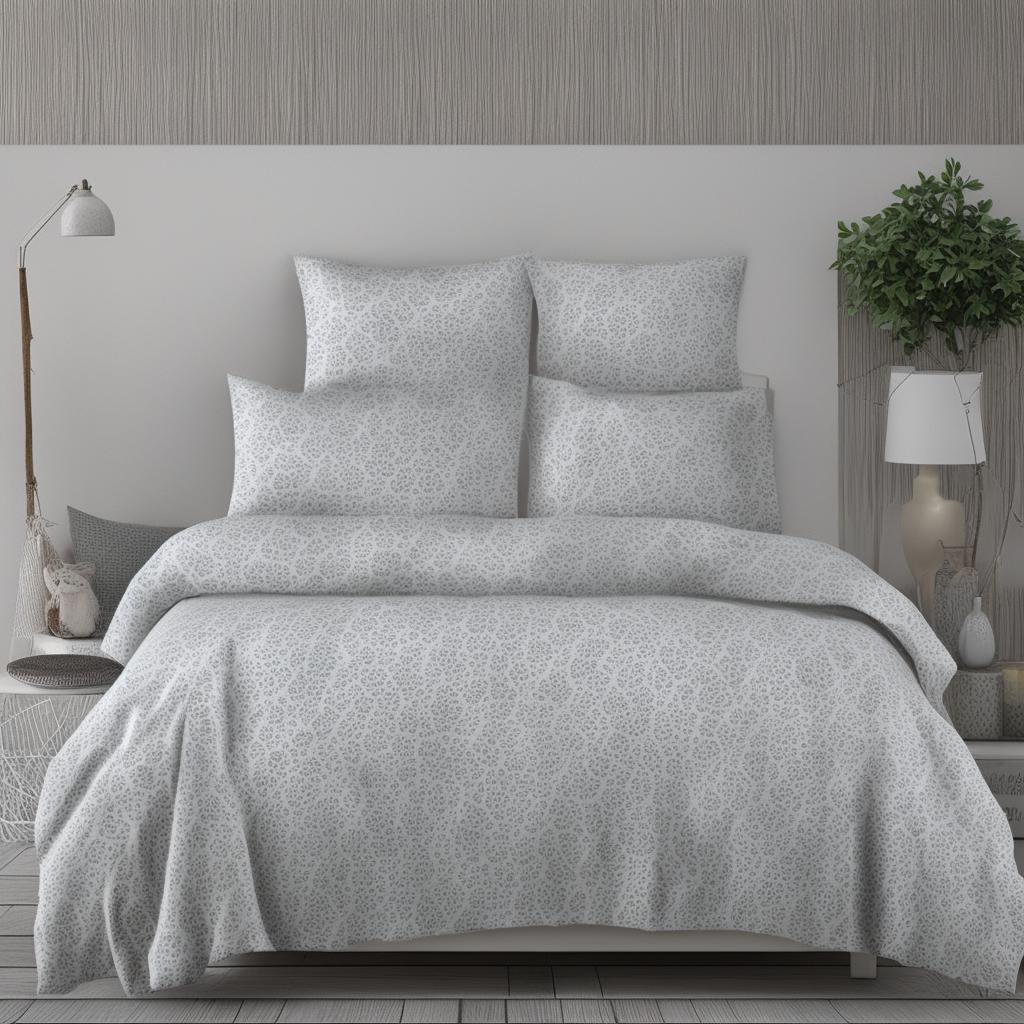} 
                    & \includegraphics[width=7cm]{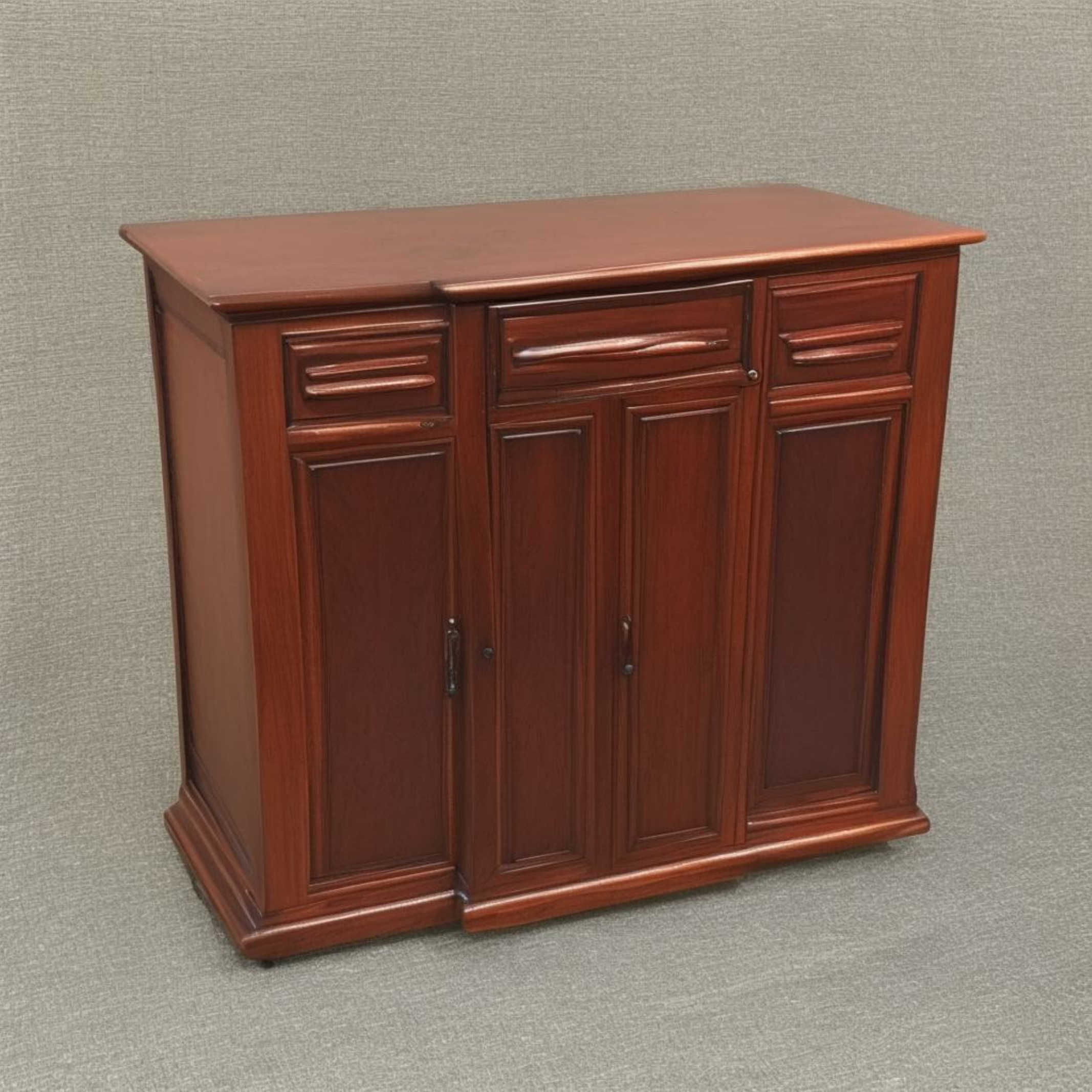}
                    & \includegraphics[width=7cm]{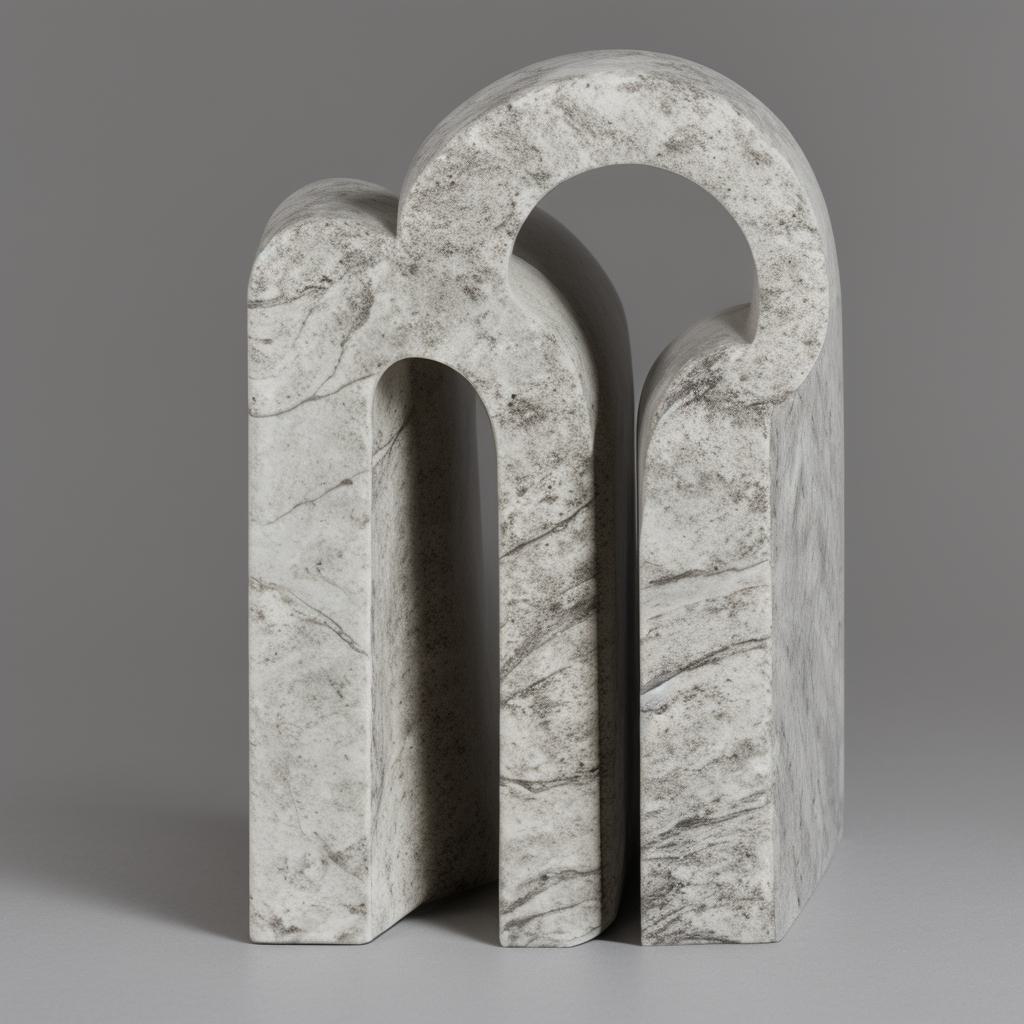} 
                    & \includegraphics[width=7cm]{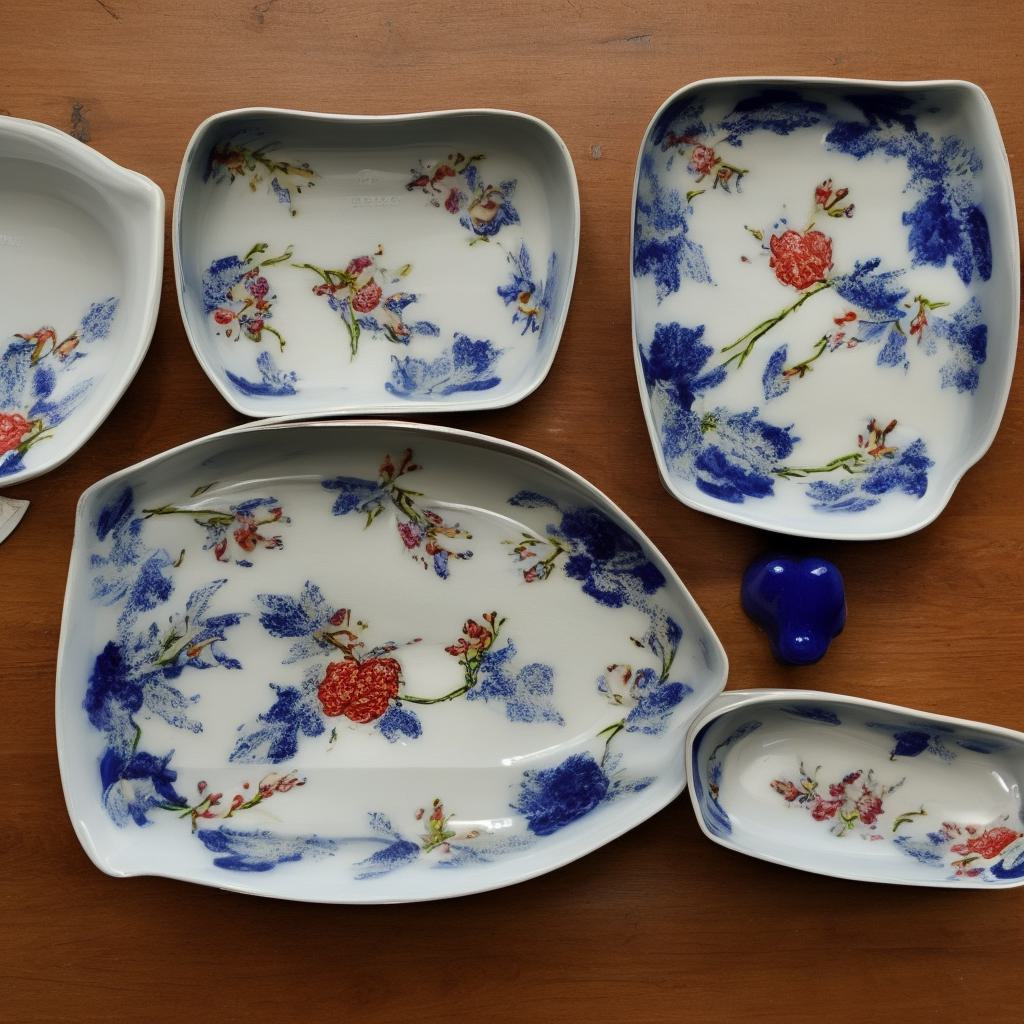} \\

                  \fontsize{28}{28}\selectfont{\emph{Bucket,plastic,}} & \fontsize{28}{28}\selectfont{\emph{Floormat,rubber,}} & \fontsize{28}{28}\selectfont{\emph{Faucet,metal,}}& \fontsize{28}{28}\selectfont{\emph{Gloves,leather,}} & \fontsize{28}{28}\selectfont{\emph{Duvetcover,fabric,}} & \fontsize{28}{28}\selectfont{\emph{Cabinet,wood,}} & 
                \fontsize{28}{28}\selectfont{\emph{Bookend,stone,}} & \fontsize{28}{28}\selectfont{\emph{Dish,ceramic,}}\\
              \fontsize{28}{28}\selectfont{\emph{polypropylene}}&\fontsize{28}{28}\selectfont{\emph{natural}}&\fontsize{28}{28}\selectfont{\emph{stainless-steel}}&\fontsize{28}{28}\selectfont{\emph{deerskin}}&\fontsize{28}{28}\selectfont{\emph{woven}}&\fontsize{28}{28}\selectfont{\emph{mahogany}}&\fontsize{28}{28}\selectfont{\emph{marble}}&\fontsize{28}{28}\selectfont{\emph{porcelain}}\\
               &&&&&&&\\
                 \multicolumn{8}{c}{\fontsize{36}{36}\selectfont{(a) Generated images with text  prompts}}\\
                     &&&&&&&\\

                    \includegraphics[width=7cm]{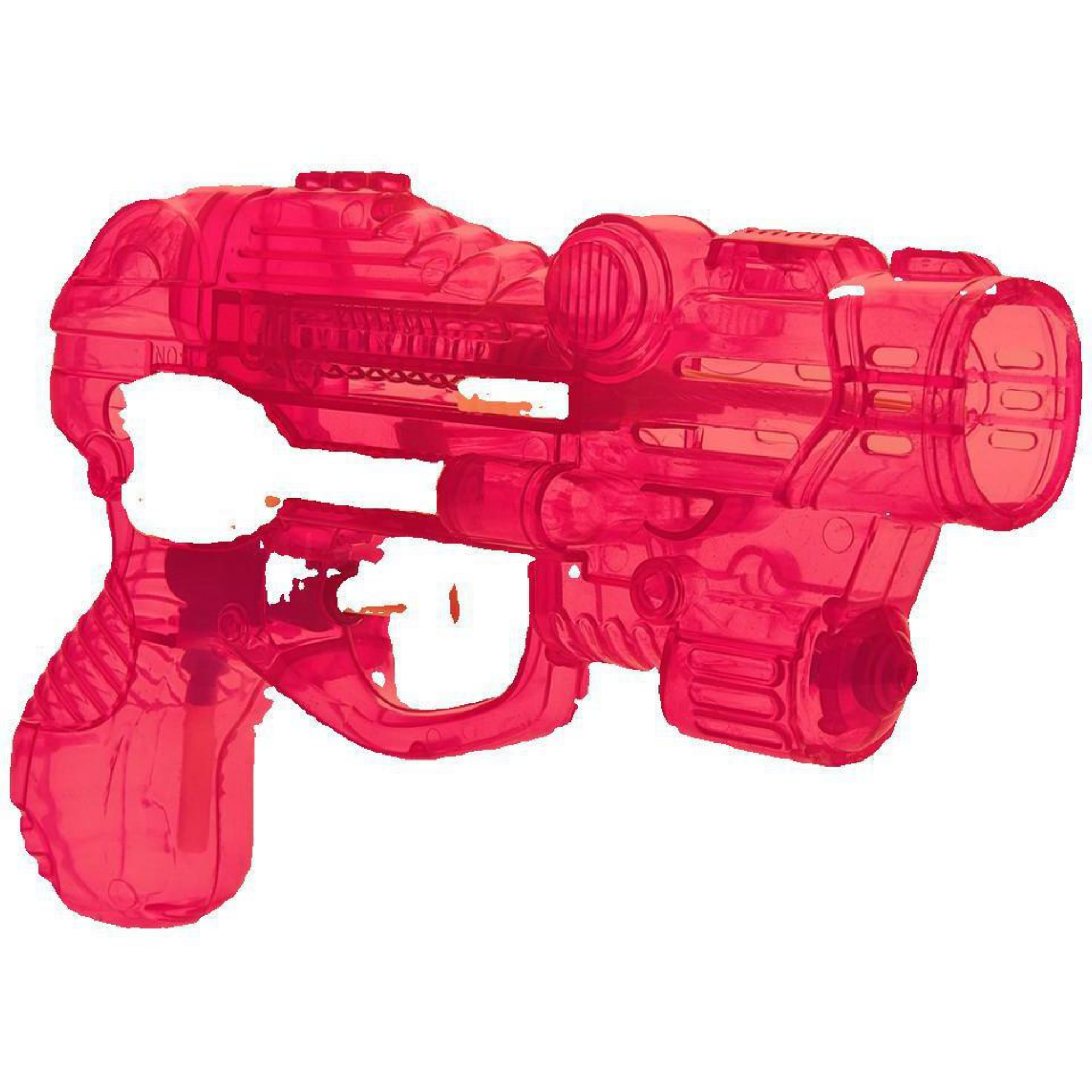} 
                    & \includegraphics[width=7cm]{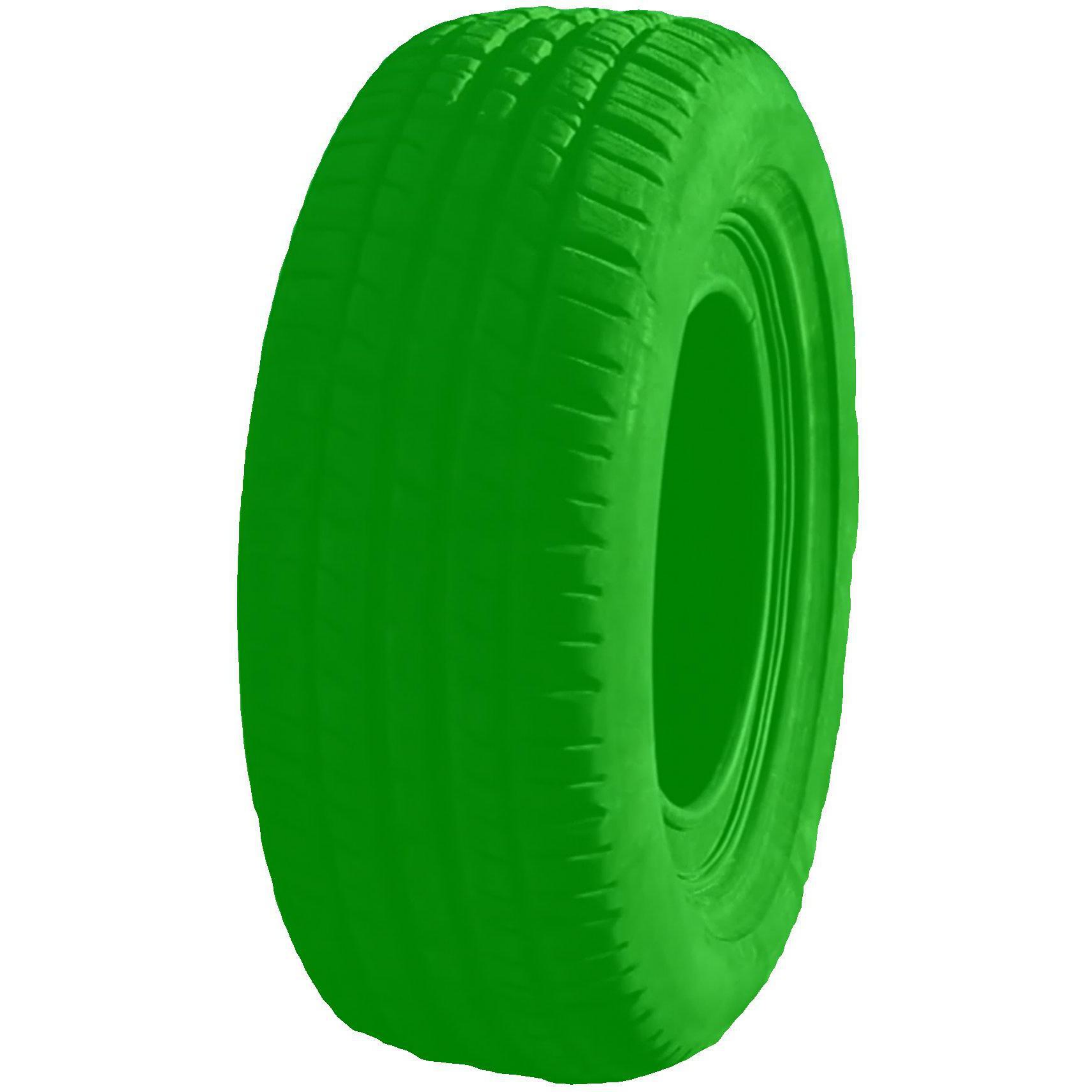} 
                    & \includegraphics[width=7cm]{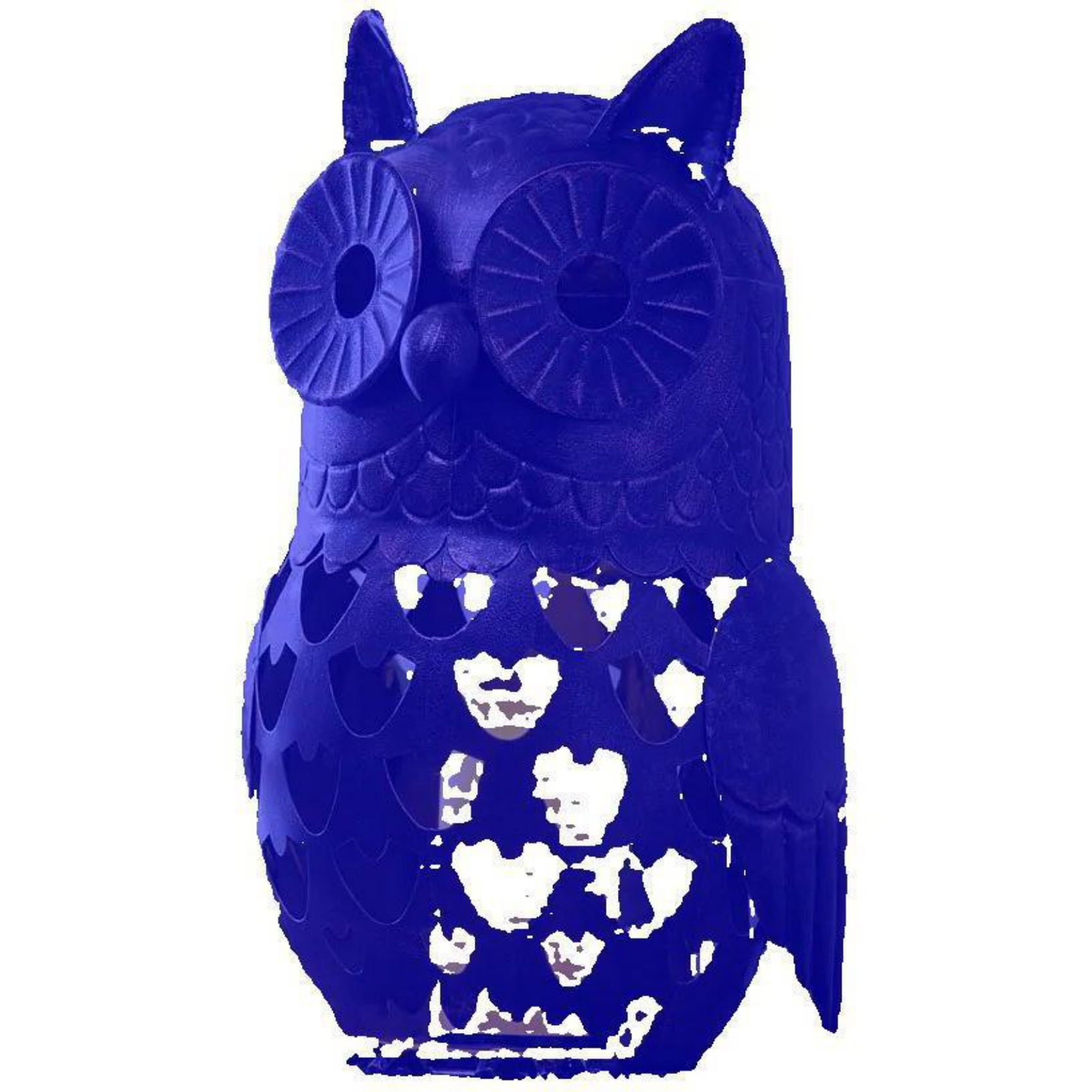} 
                    & \includegraphics[width=7cm]{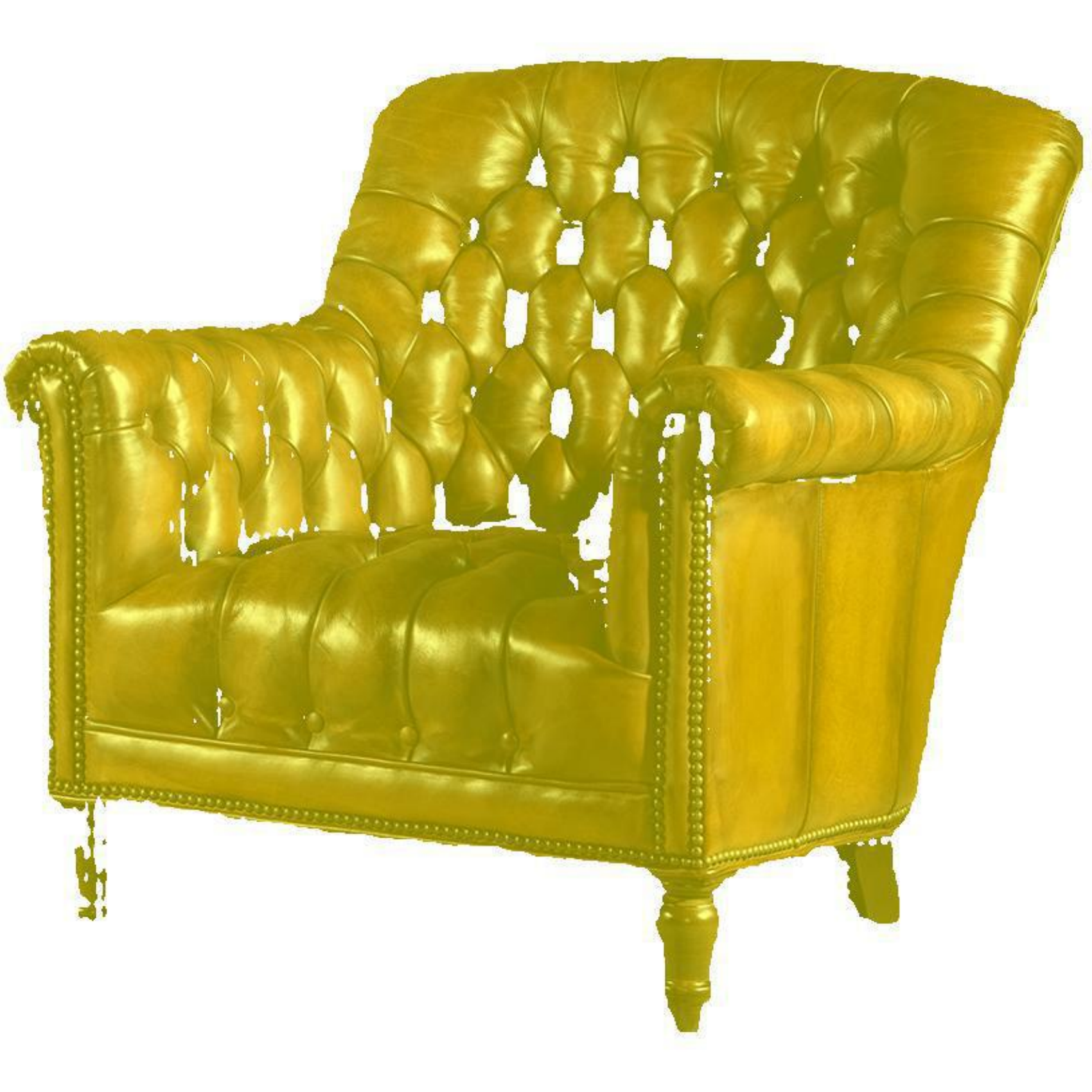}
                    & \includegraphics[width=7cm]{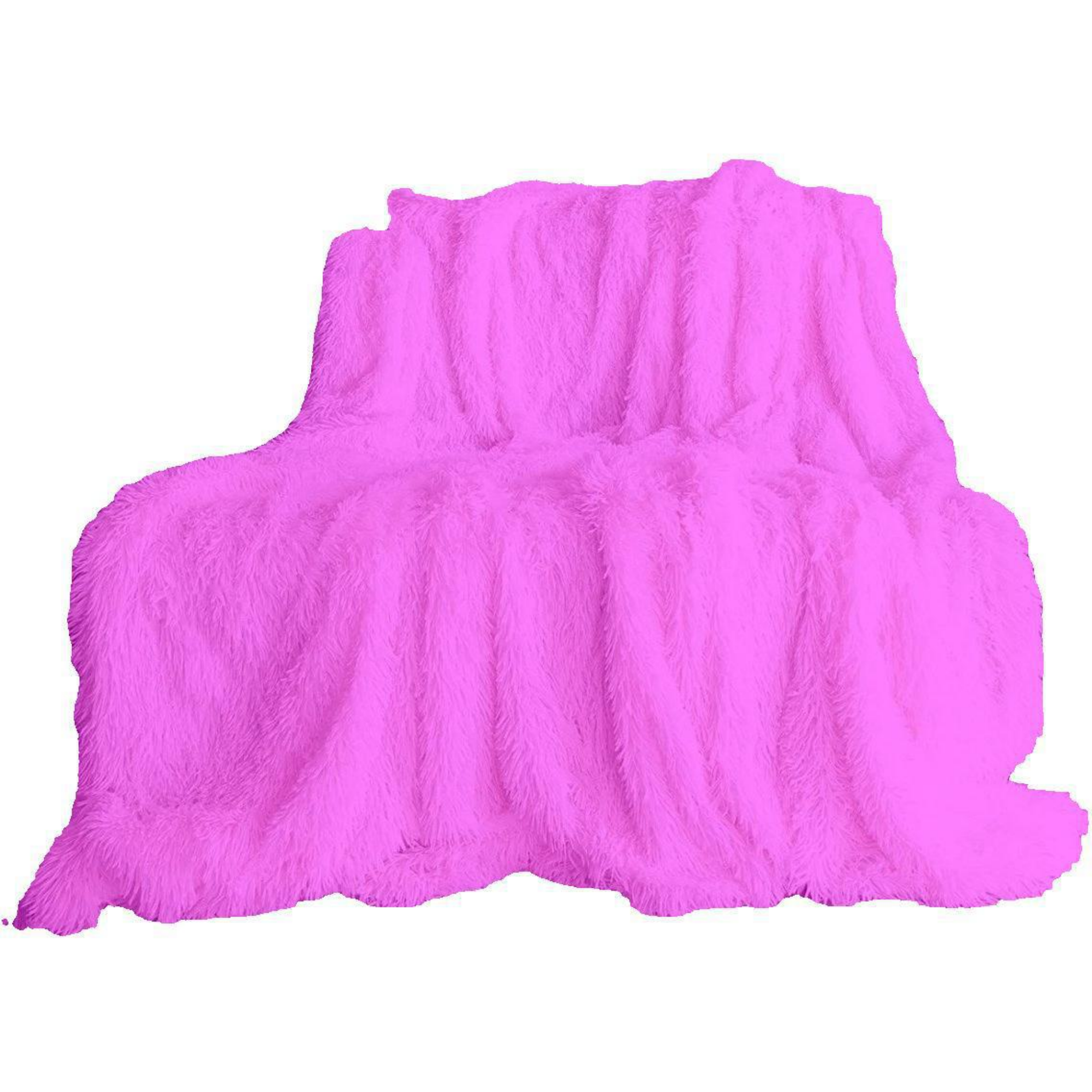} 
                    & \includegraphics[width=7cm]{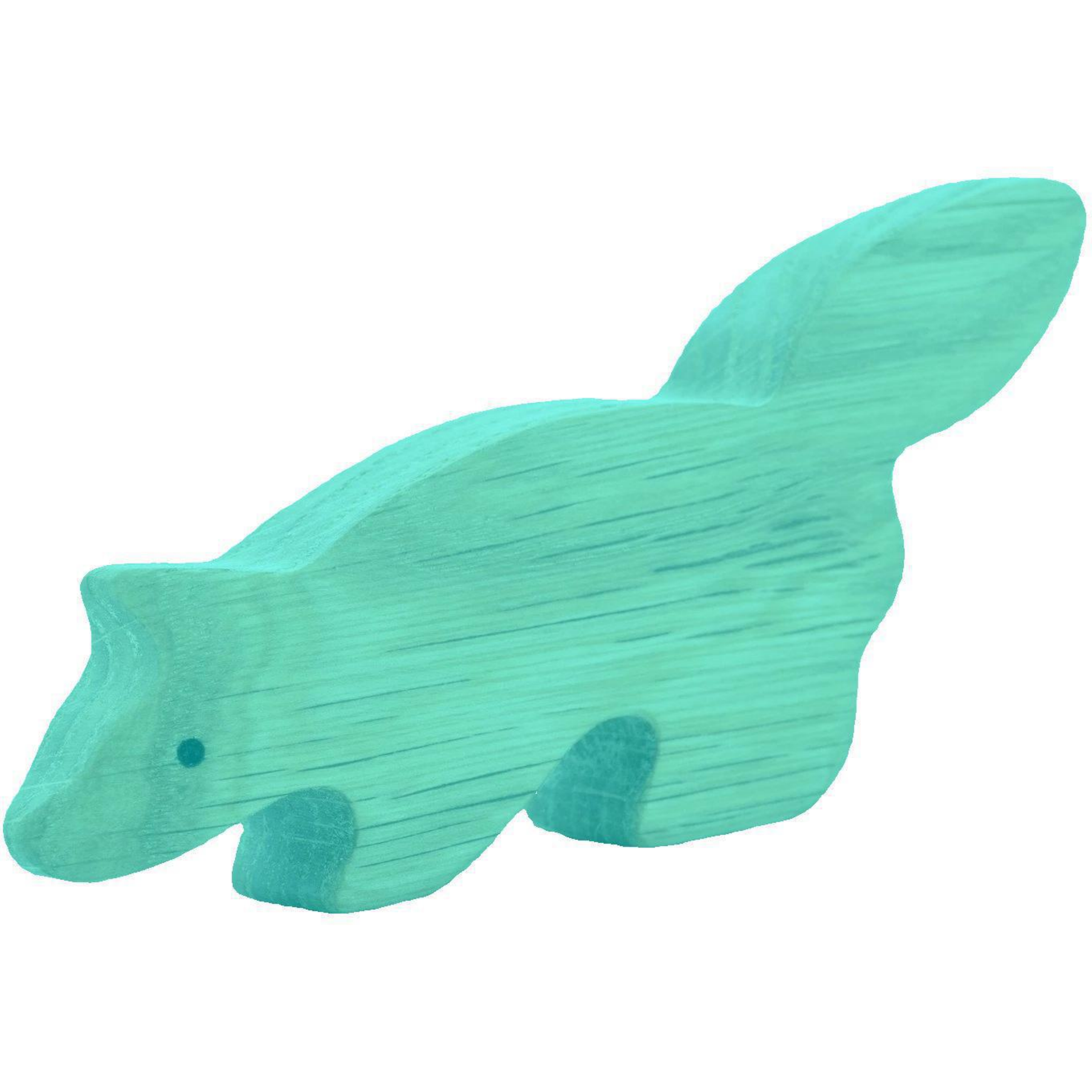} 
                    & \includegraphics[width=7cm]{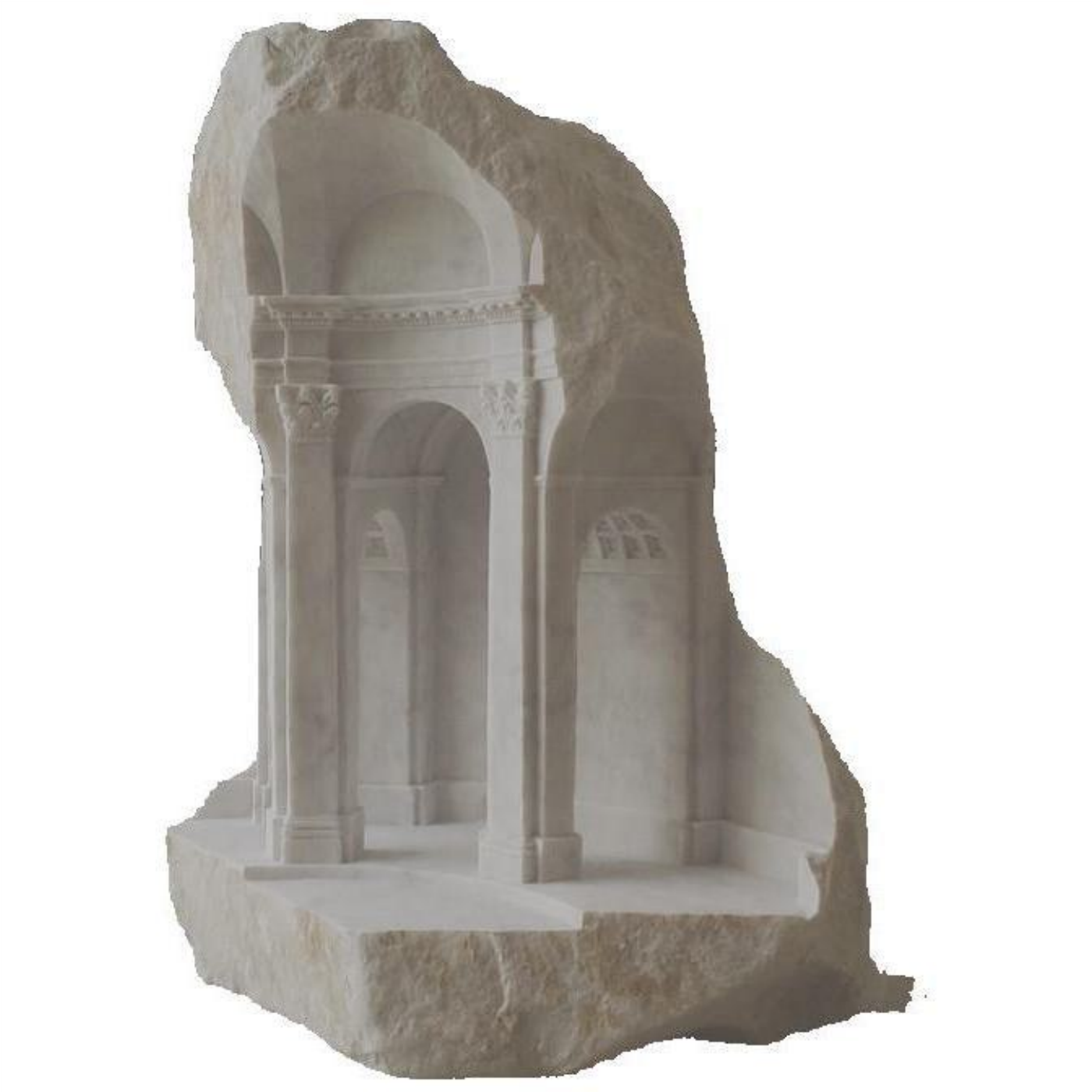} 
                    & \includegraphics[width=7cm]{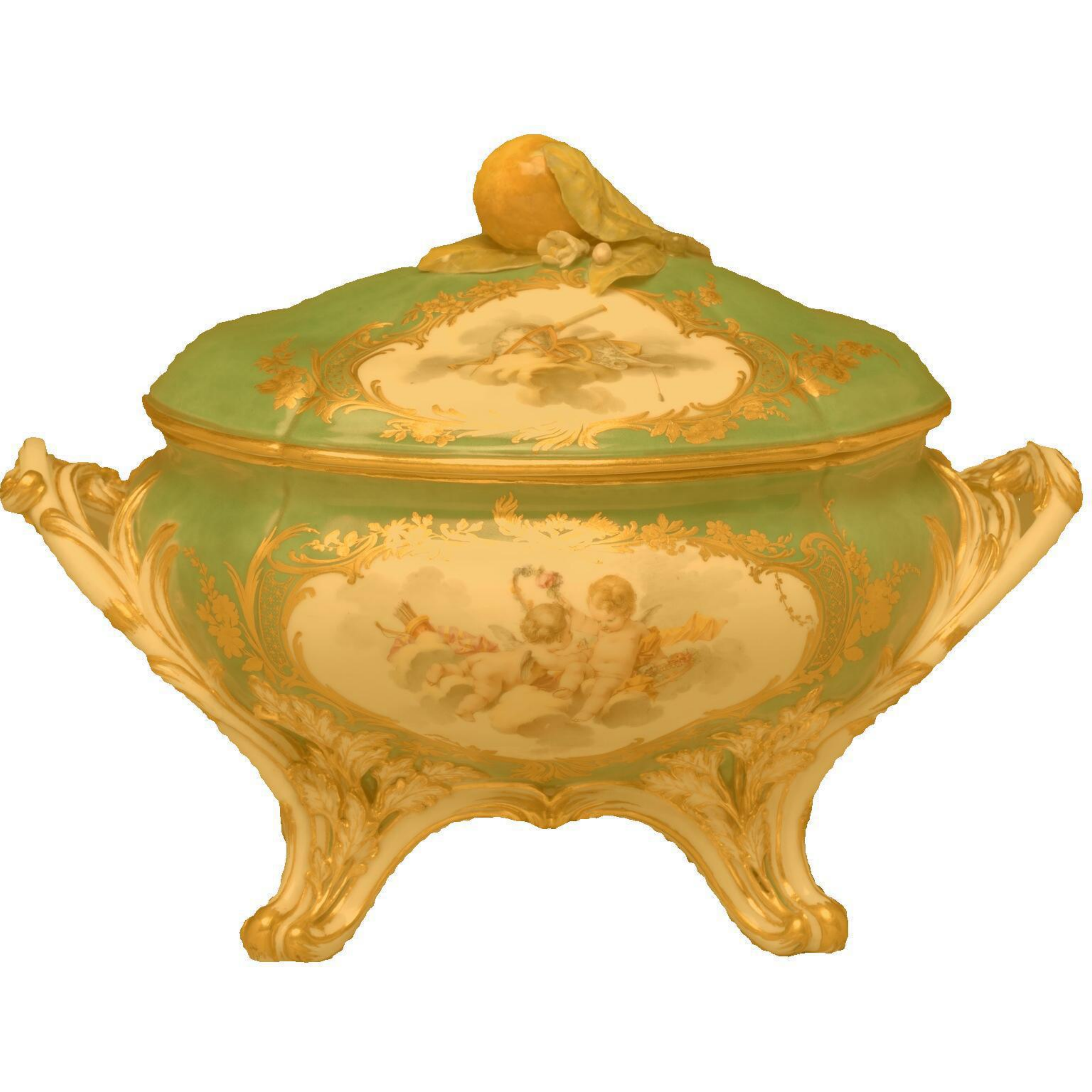} \\

                       \fontsize{28}{28}\selectfont{\textbf{plastic} \colorbox{red}{\phantom{X}}} &  \fontsize{28}{28}\selectfont{\textbf{rubber} \colorbox{green} 
                    {\phantom{X}}} & \fontsize{28}{28}\selectfont{\textbf{metal}  \colorbox{blue}{\phantom{X}}} &  \fontsize{28}{28}\selectfont{\textbf{leather} \colorbox{yellow}{\phantom{X}}}
                     &  \fontsize{28}{28}\selectfont{\textbf{fabric} \colorbox{magenta}{\phantom{X}}} &  \fontsize{28}{28}\selectfont{\textbf{wood} \colorbox{cyan}{\phantom{X}}} &  \fontsize{28}{28}\selectfont{\textbf{stone} \colorbox{gray}{\phantom{X}}} &  \fontsize{28}{28}\selectfont{\textbf{ceramic} \colorbox{orange}{\phantom{X}}} \\ 
                       &&&&&&&\\
                       \multicolumn{8}{c}{\fontsize{36}{36}\selectfont{(b) Material prediction on real images}}\\

                \end{tabular}
            }

		\captionof{figure}{With careful knowledge distillation from image generation network, we built generated image material dataset and propose an method for accurate material classification}
		\label{fig:teaser}
        \vspace{0.1in}
       }]
\let\thefootnote\relax\footnotetext{* Equal contribution.}

\begin{abstract}
Material classification has emerged as a critical task in computer vision and graphics, supporting the assignment of accurate material properties to a wide range of digital and real-world applications. While traditionally framed as an image classification task, this domain faces significant challenges due to the scarcity of annotated data, limiting the accuracy and generalizability of trained models. Recent advances in vision-language foundation models (VLMs) offer promising avenues to address these issues, yet existing solutions leveraging these models still exhibit unsatisfying results in material recognition tasks. In this work, we propose a novel framework that effectively harnesses foundation models to overcome data limitations and enhance classification accuracy. Our method integrates two key innovations:(a) a robust image generation and auto-labeling pipeline that creates a diverse and high-quality training dataset with material-centric images, and automatically assigns labels by fusing object semantics and material attributes in text prompts;(b) a prior incorporation strategy to distill information from VLMs, combined with a joint fine-tuning method that optimizes a pre-trained vision foundation model alongside VLM-derived priors, preserving broad generalizability while adapting to material-specific features. Extensive experiments demonstrate significant improvements on multiple datasets. We show that our synthetic dataset effectively captures the characteristics of real world materials, and the integration of priors from vision-language models significantly enhances the final performance.
\end{abstract}    
\section{Introduction}
\label{sec:intro}

Accurate material classification is a fundamental problem in visual understanding, bridging raw observations to actionable material properties that are critical for rendering, simulation, and content generation. Recent advances in material modeling span three key directions including texture-space synthesis~\cite{guo2020materialgan,guerrero2022matformer,zhou2022tilegen,hu2023generating,zhou2023photomat}, image-space editing~\cite{hu2022controlling,sharma2024alchemist,lopes2024material,cheng2025zest}, and material-aware 3D generation~\cite{li2024materialseg3d,fang2024make,zhang2024mapa}. However, these workflows critically depend on robust material classification to map observations to actionable material parameters. For example, \cite{zhang2024mapa} leverage image-based classification to retrieve optimal procedural materials from a database, underscoring its foundational role. Yet, despite material classification being framed as an image recognition task, existing solutions trained on limited annotated data struggle with reasonable distinctions (e.g., non-metallic vs. metallic surfaces). Even state-of-the-art vision-language models (VLMs) such as CLIP~\cite{radford2021learning} or GPT-4v~\cite{achiam2023gpt} pre-trained on web-scale data, exhibit a significant drop in material recognition accuracy compared to object classification, revealing a critical gap in domain-specific understanding. We attribute this to two factors: (1) the scarcity of high-quality, material-focused annotations for training, and (2) the underutilization of cross-modal priors in VLMs during training.

To address the scarcity of annotated material data, we propose synthesizing a labeled dataset using generative models. While modern diffusion-based methods~\cite{ramesh2021zero,rombach2022high} can produce realistic images at scale, naively generating images with material prompts introduces two critical issues:
\begin{itemize}
    \item \textbf{Material Ambiguity}: Generated images often depict \textit{incorrect material correlations}---e.g., a foreground object labeled ``ceramic'' may appear alongside conflicting materials in the background (wooden tables, fabric textures).
    
    \item \textbf{Diversity Limitations}: Prompt engineering alone fails to ensure sufficient visual variation (lighting, viewpoints, occlusions), leading to poor generalization on real-world images.
\end{itemize}
To resolve these challenges, we introduce semantic grounding to enforce precise material-to-region associations. By augmenting generation prompts with \textit{object semantics} and leveraging zero-shot segmentation models~\cite{liu2023grounding,ren2024grounded}, we isolate the target object (vase) and propagate its material label (ceramic) while filtering out irrelevant regions (marble table). This approach not only ensures auto-labeled accuracy but also enriches diversity by decoupling material from scene context.Through this design, we overcome the challenges of ambiguity and diversity in generative data, and produce a large-scale, auto-labeled dataset encompassing 21 material classes.

To harness cross-modal priors for improved generalization, we design a dual-stream framework that synergizes vision-based and language-based material representations. For visual grounding, we extract dense patch-level features using DINOv2~\cite{oquab2023dinov2}, a vision foundation model pretrained on unlabeled images, which captures fine-grained texture and reflectance patterns critical for material recognition. Simultaneously, we generate language descriptors of material appearance (e.g., ``a glossy, translucent plastic with faint subsurface scattering") using GPT-4v~\cite{achiam2023gpt}, then encode these textual descriptions into semantic embeddings via CLIP~\cite{radford2021learning}. Crucially, this language stream distills domain-agnostic knowledge about material properties that may not be explicitly discernible from visual features alone.
The two modalities are fused through a lightweight Multilayer Perceptron (MLP): DINOv2’s masked object features (aggregated via spatial averaging) are concatenated with CLIP’s language embeddings, then projected into a joint latent space for material prediction. This design ensures that visual cues are contextualized by semantic priors, enabling the model to disambiguate perceptually similar materials that can be identified from semantics. We finetune the head of DINOv2 together with the MLP during supervised training on our synthetic dataset to harness cross-modal priors while preserving the generalizability of both foundation models.

We validate our method on three datasets: the FMD ~\cite{Sharan-JoV-14} dataset, a classic 10-class benchmark for material classification; a 21-class DMS-test dataset constructed from the DMS~\cite{upchurch2022dense} dataset; and a 21-class self-collected Google-test dataset from Google Images capturing real-world material appearances. Our approach achieves 89\% accuracy on FMD ~\cite{Sharan-JoV-14} dataset and 92\% on the Google-test dataset, outperforming state-of-the-art material classifier~\cite{drehwald2023one} by 33\% and 29\%, respectively. Notably, off-the-shelf vision-language models (VLMs) like CLIP~\cite{radford2021learning} and GPT-4v~\cite{achiam2023gpt} achieve only 38\% and 43\% accuracy on DMS-test dataset, highlighting their inadequacy for domain-specific tasks. Ablation results highlight the importance of both modalities, with the absence of language priors (GPT-4v + CLIP) reducing accuracy by at most 4\%, and the removal of vision priors (DINOv2) causing a similar maximum drop.
Crucially, models trained on our synthetic dataset outperform the DMS-trained baselines by 9\% on the FMD~\cite{Sharan-JoV-14} dataset and by 18\% on the Google-test dataset.
These results underscore the effectiveness of our unified solution in bridging synthetic dataset for real-world material recognition.

Overall, our core research contributions are:
\begin{itemize}
\item A synthetic data generation and auto-labeling framework for material classification that creates diverse, high-fidelity images with reliable labels, addressing the scarcity of annotated material data.
\item A dual-stream framework for cross-modal prior fusion that combines vision-language and vision foundation models to improve generalizability and disambiguate perceptually similar materials.
\item A collaborative fine-tuning strategy that preserves foundational model priors while adapting to material-specific tasks, achieving state-of-the-art performance on multiple datasets.
\end{itemize}

\section{Related Work}

\paragraph{Material datasets} 
Material classification models are typically trained on datasets with annotated material images. Notable examples include FMD ~\cite{Sharan-JoV-14}, which features 10 material categories, and OpenSurface ~\cite{bell2013opensurfaces}, providing 19,000 indoor scene images covering 37 materials. MINC ~\cite{bell2015material} is larger, with 0.4 million images of 23 materials, while DMS \cite{upchurch2022dense} offers the most extensive collection, with 3.2 million segments in 44,560 images across 52 material classes. However, datasets such as OpenSurfaces \cite{bell2013opensurfaces} and DMS \cite{upchurch2022dense} exhibit a noticeable class imbalance, where common materials like wood and metal appear far more frequently than rare ones (e.g., wax, rubber). This unbalanced distribution hinders the model’s ability to generalize across diverse material types.  \cite{upchurch2022dense}.
To alleviate this issue, our generative dataset produces balanced samples across material categories, enabling more uniform and comprehensive supervision. Datasets like the Light-Field Material Dataset ~\cite{wang20164d} and Multi-Illumination dataset ~\cite{murmann2019dataset}, which account for varied lighting, are primarily designed for material parameter reconstruction rather than broad classification tasks due to their limited scale.

\paragraph{Generative dataset} As synthetic data generation methods continue to advance, there is a growing body of research leveraging these techniques to create datasets for learning purposes \cite{dunlap2023diversify,toker2024satsynth,nguyen2024dataset,peng2026lodlocv3generalizedaerial}. We follow this approach to generate our material image dataset, with a focus on overcoming specific challenges such as maintaining content diversity and ensuring automatic and accurate material labeling.

\paragraph{Material recognition}
Material recognition research has focused on classification, with key developments including the FV-CNN technique by Cimpoi et al. \cite{cimpoi2015deep}, which combines CNNs with IFV classifiers and has been tested on datasets like OpenSurfaces \cite{bell2013opensurfaces} and FMD \cite{Sharan-JoV-14}.
Recently, \cite{cai2022rgb} introduced the KITTI-Materials dataset and RMSNet, a hierarchical Transformer-based model for road scene material segmentation using RGB images. Mei et al. \cite{Mei_2022_CVPR} presented PGSNet, which uses RGB and polarization data to segment glass materials, outperforming previous models on the RGBP-Glass dataset.
Manuel et al. \cite{drehwald2023one} developed a contractive learning method for material classification trained on MatSim, a dataset which incorporates synthetic physical rendering images and natural images. Their approach rivals CLIP \cite{radford2021learning} in general material classification.

\paragraph{Harness Foundation Models for Vision Tasks} Recent works leverage foundation models for vision tasks through two paradigms: vision-only priors and vision-language alignment. For vision-centric learning, methods like DINOv2~\cite{oquab2023dinov2} exploit self-supervised pretraining to extract generalizable features for segmentation. \cite{yang2024depth} and \cite{huang2024material} further exploit such generalizable features for depth and PBR parameter estimation. More broadly, MossFuse~\cite{du2026unsupervised} highlights the
value of cross-modal interaction, while recent multi-modal reasoning and decision-making frameworks ~\cite{li2025spacedrive,yuan2025autodrive,yuan2025video,yuan2026if,hu2025mpcformerphysicsinformeddatadrivenapproach,lian2026finetuningenoughparallelframework}
underscore the importance of effectively leveraging multi-modal inputs. In our case, fine-grained material understanding requires both subtle visual appearance cues and semantic priors. Vision-language models (VLMs) like CLIP~\cite{radford2021learning} and GPT-4v~\cite{achiam2023gpt} enable zero-shot material classification in works such as MAPA~\cite{zhang2024mapa} and Make-It-Real~\cite{fang2024make}, which align material descriptors with images. While these methods bypass manual annotation, they exhibit limited accuracy on fine-grained materials due to (1) vague text prompts that fail to capture nuanced visual properties and (2) inadequate adaptation to small, domain-specific datasets. Instead, we finetune a dual-stream network architecture to better exploit both language and vision priors.
\label{sec:related}

\section{Method}
\label{sec:method}

Our framework addresses material classification through two synergistic components: (1) synthetic dataset generation with semantic-aware auto-labeling to overcome annotation scarcity, and (2) a dual-stream architecture that fuses vision-language priors for robust feature learning. For synthetic data generation (Section \ref{sec:dataset}),
we first mitigate data limitations by generating a diverse material-centric dataset using text-to-image diffusion models. The pipeline integrates object-aware prompt engineering with vision foundation models (DINOv2, Grounding DINO) for automatic region-material labeling. For learning material classification, we formulate the material classification as a cross-modal fusion task with both vision and language streams in Sections \ref{sec:formulation}. Finally, we present our dual-stream network architecture along with the training specifics in Section \ref{sec:approach-network}.

\subsection{Synthetic Dataset}
\label{sec:dataset}
\begin{figure*}
\centering
\includegraphics[width=\textwidth]{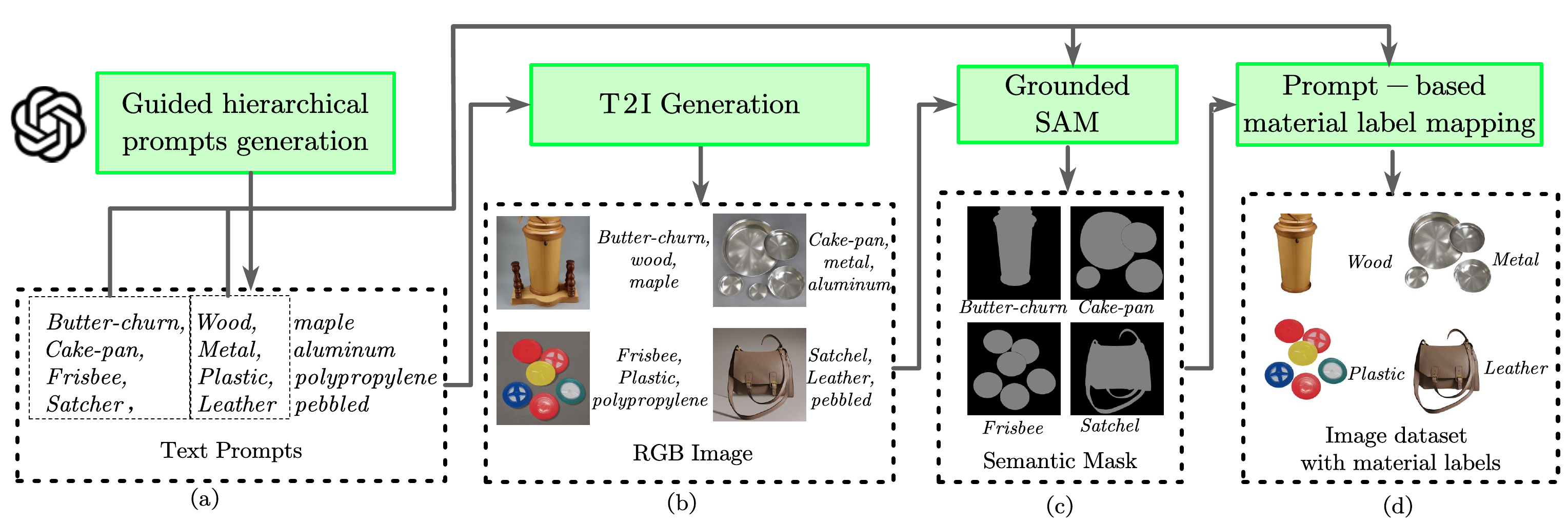}
\caption{Dataset generation workflow. Our pipeline synthesizes labeled material images through: (a) Hierarchical prompt engineering with LLM-guided plausibility filtering, (b) Diffusion-based image generation with model selection, (c) Semantic mask extraction, and (d) Region-aware material label assignment.}
\label{fig:dataset_generation}
\vspace{-3mm}
\end{figure*}

\begin{figure*}[ht]
    \resizebox{\textwidth}{!}{%
 \begin{tabular}
                {cccccccc}

                    \includegraphics[width=7cm]{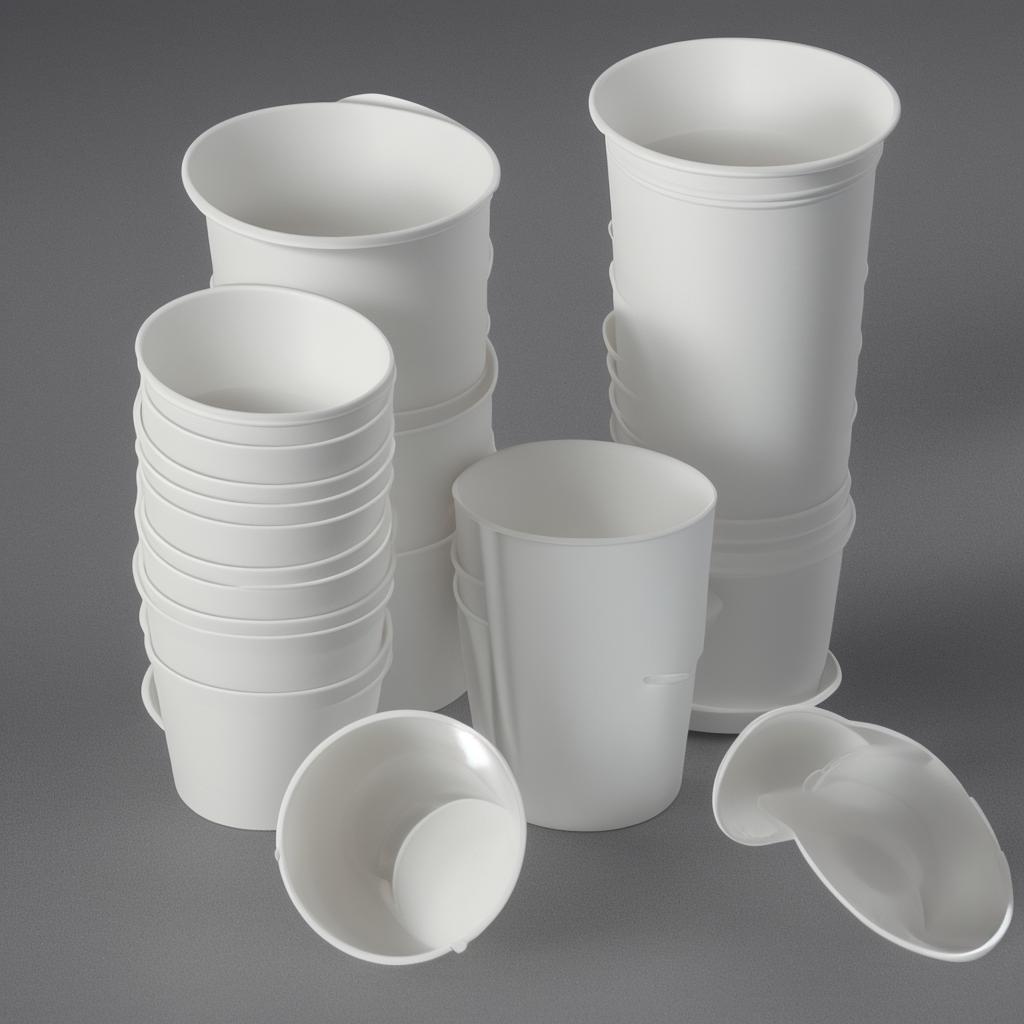} 
                    & \includegraphics[width=7cm]{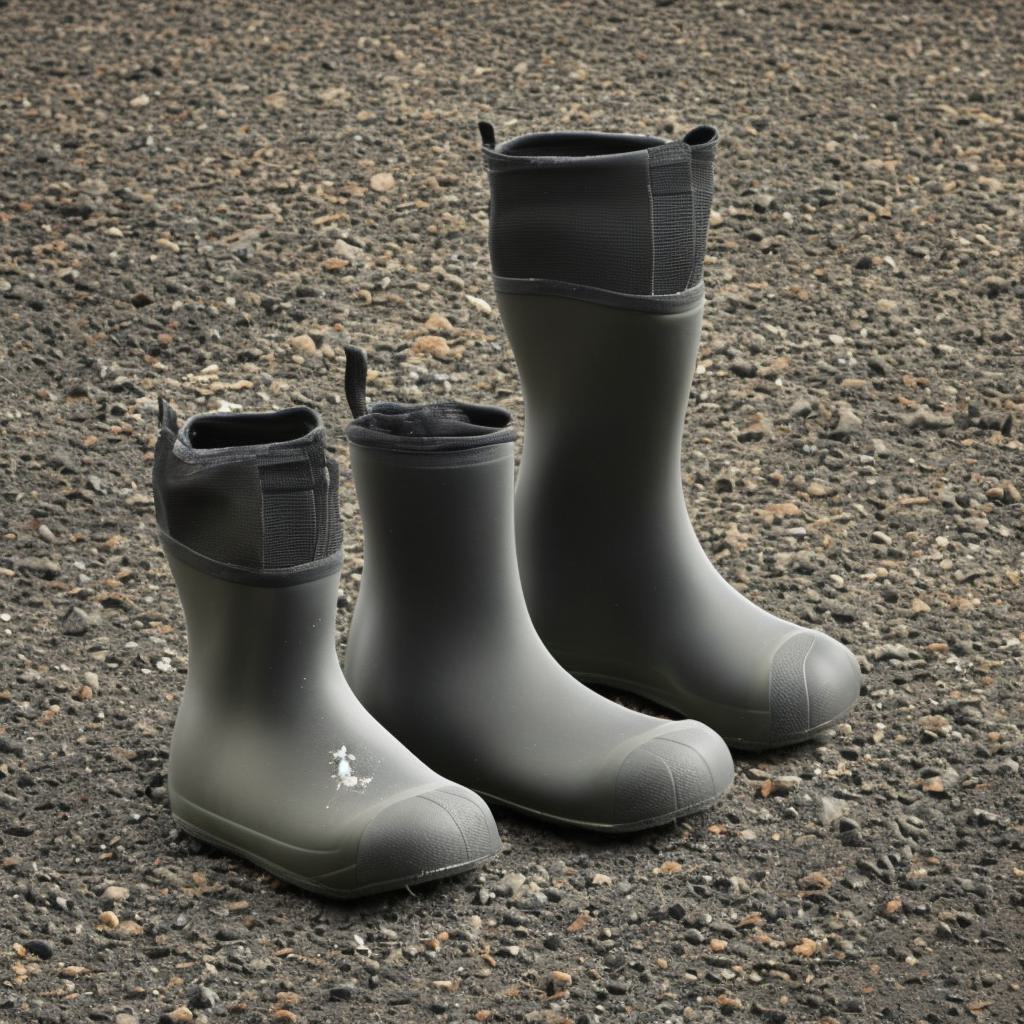} 
                    & \includegraphics[width=7cm]{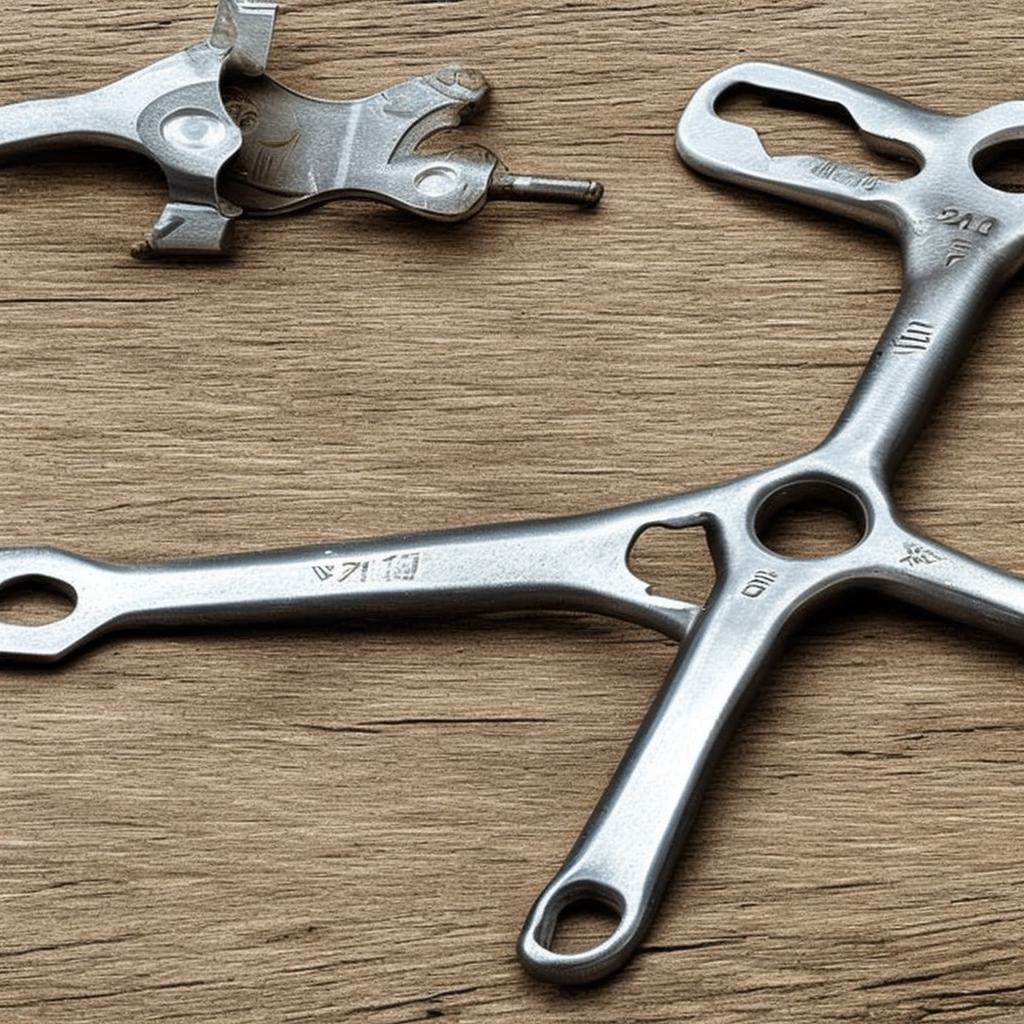} 
                    & \includegraphics[width=7cm]{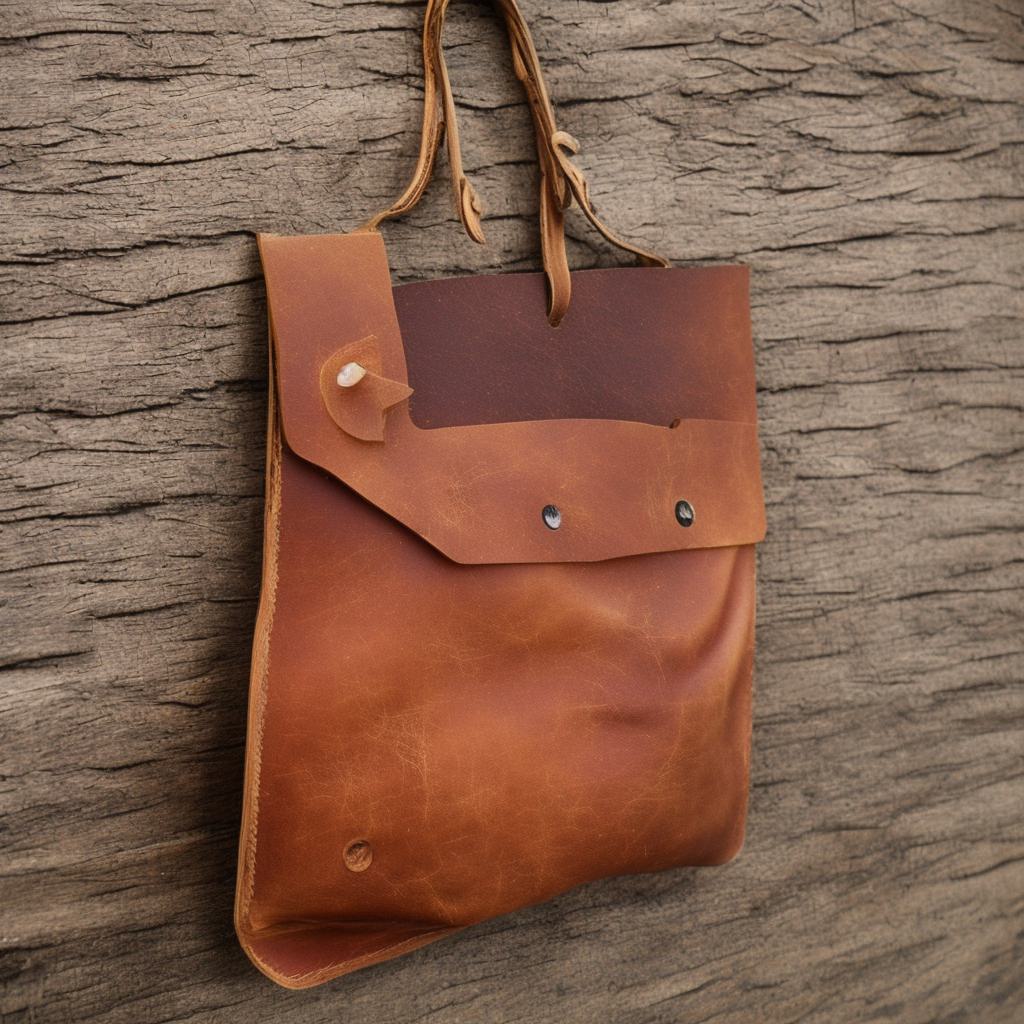} 
                    & \includegraphics[width=7cm]{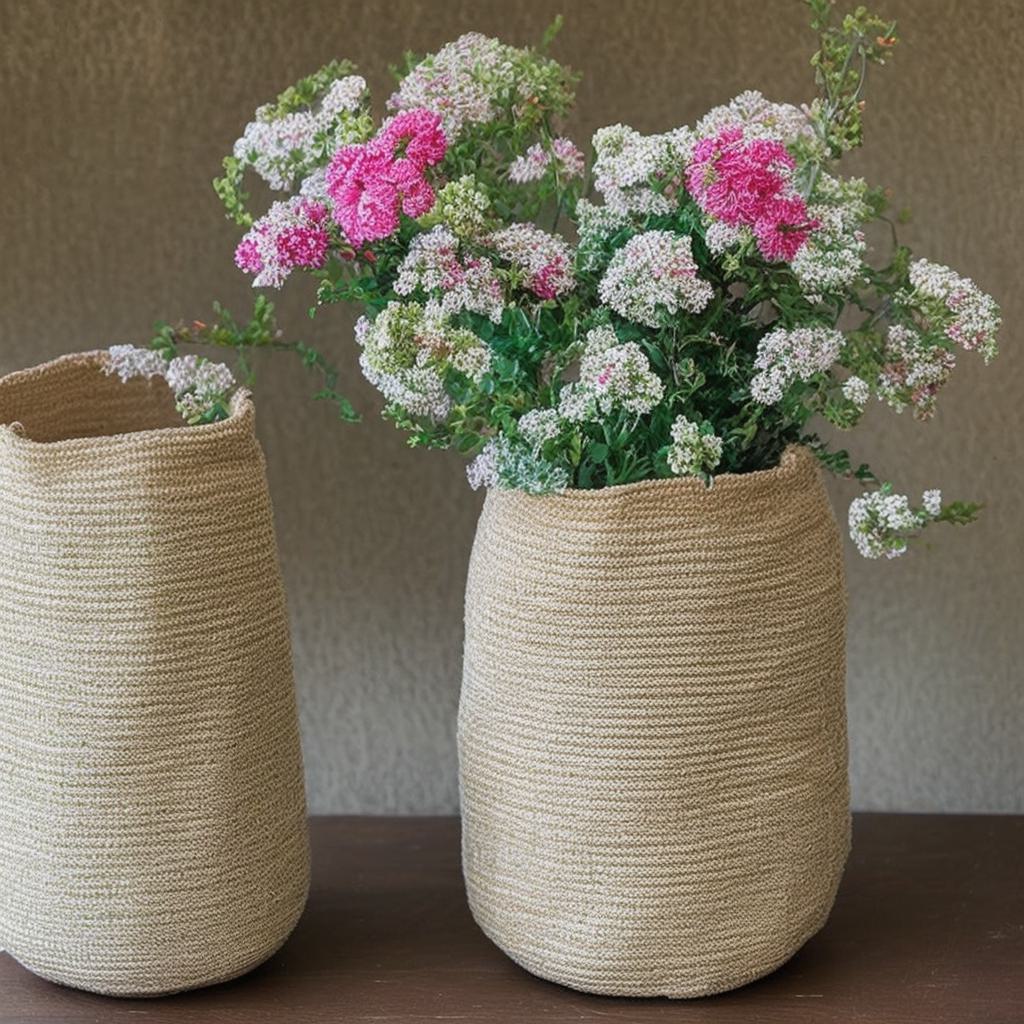} 
                    & \includegraphics[width=7cm]{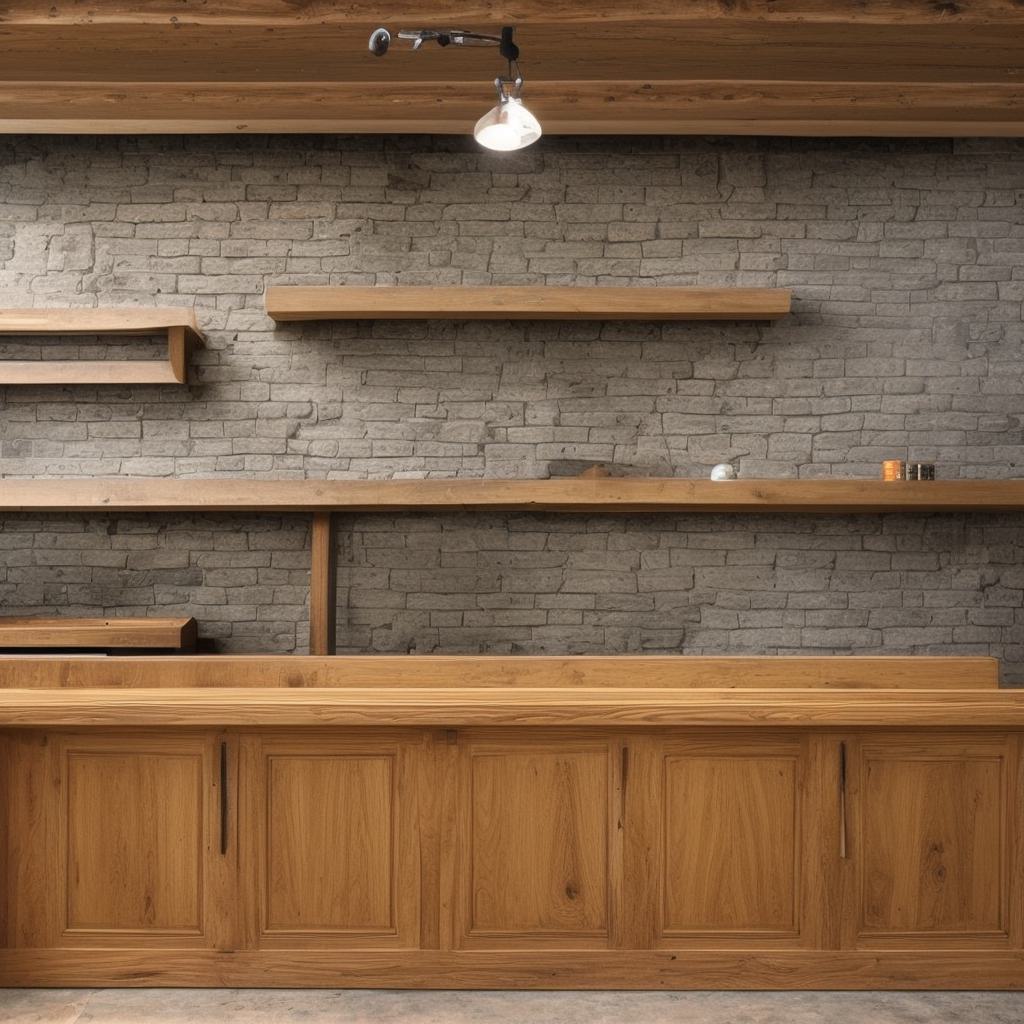} 
                    & \includegraphics[width=7cm]{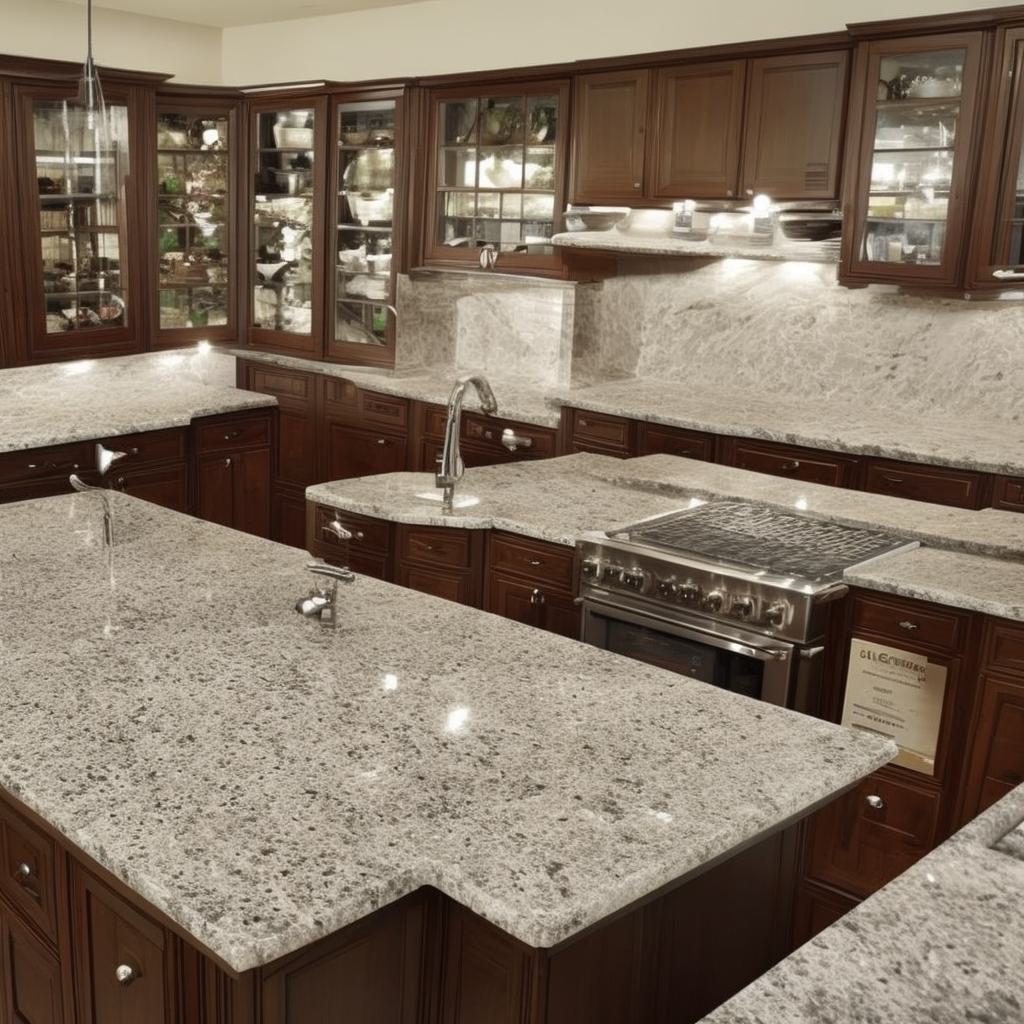} 
                    & \includegraphics[width=7cm]{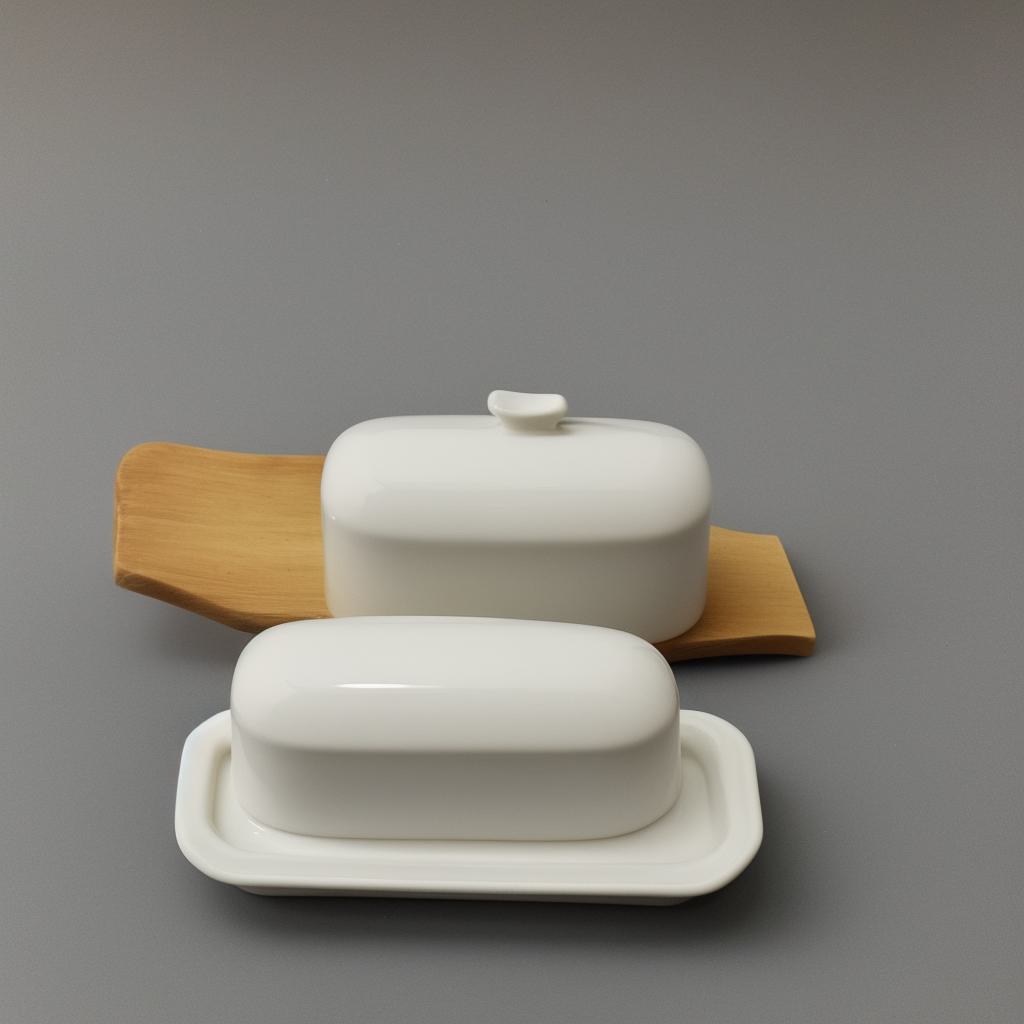} \\

                  \fontsize{26}{26}\selectfont{\emph{cup,Plastic,}} &  \fontsize{26}{26}\selectfont{\emph{boot,Rubber,}} &  \fontsize{26}{26}\selectfont{\emph{wrench,Metal,}}& \fontsize{26}{26}\selectfont{\emph{bag,Leather,}} & \fontsize{26}{26}\selectfont{\emph{vase,Fabric,}} & \fontsize{26}{26}\selectfont{\emph{counter,Wood,}} & 
                \fontsize{26}{26}\selectfont{\emph{countertop,Stone,}} & \fontsize{26}{26}\selectfont{\emph{butterdish,Ceramic,}}\\
                
                \fontsize{26}{26}\selectfont{\emph{polystyrene}}& \fontsize{26}{26}\selectfont{\emph{neoprene}}& \fontsize{26}{26}\selectfont{\emph{steel}}& \fontsize{26}{26}\selectfont{\emph{vegetable-tanned}}& \fontsize{26}{26}\selectfont{\emph{jute}}& \fontsize{26}{26}\selectfont{\emph{oak}} & \fontsize{26}{26}\selectfont{\emph{granite}}& \fontsize{26}{26}\selectfont{\emph{porcelain}}\\

                  \includegraphics[width=7cm]{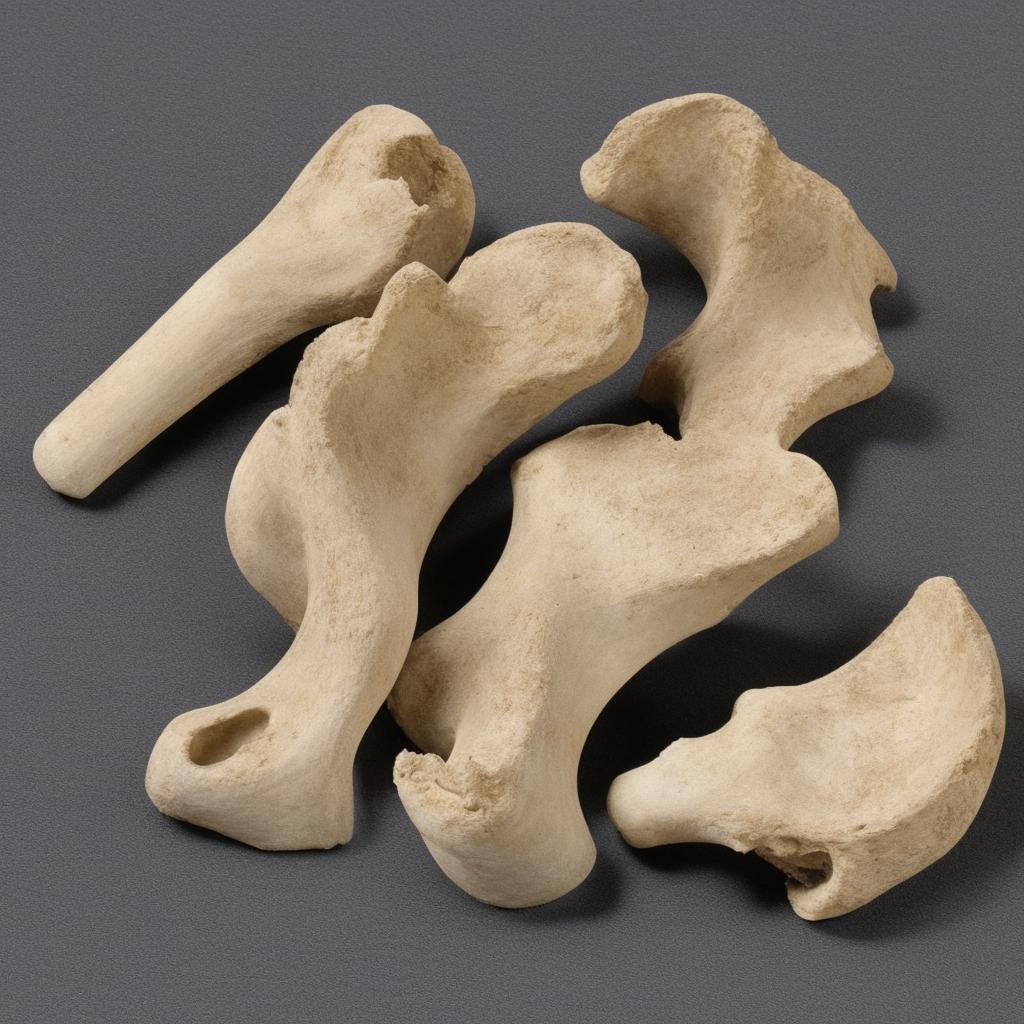} 
                 & \includegraphics[width=7cm]{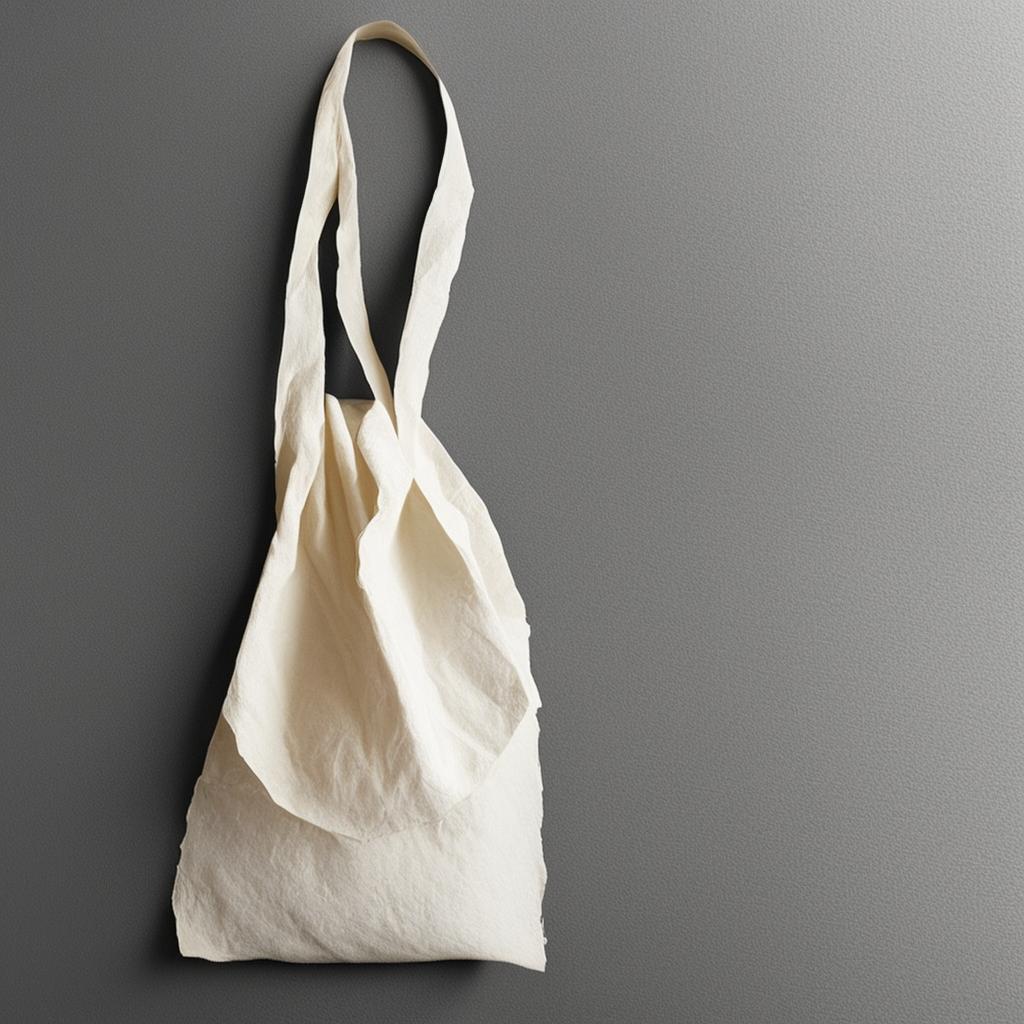} 
                    & \includegraphics[width=7cm]{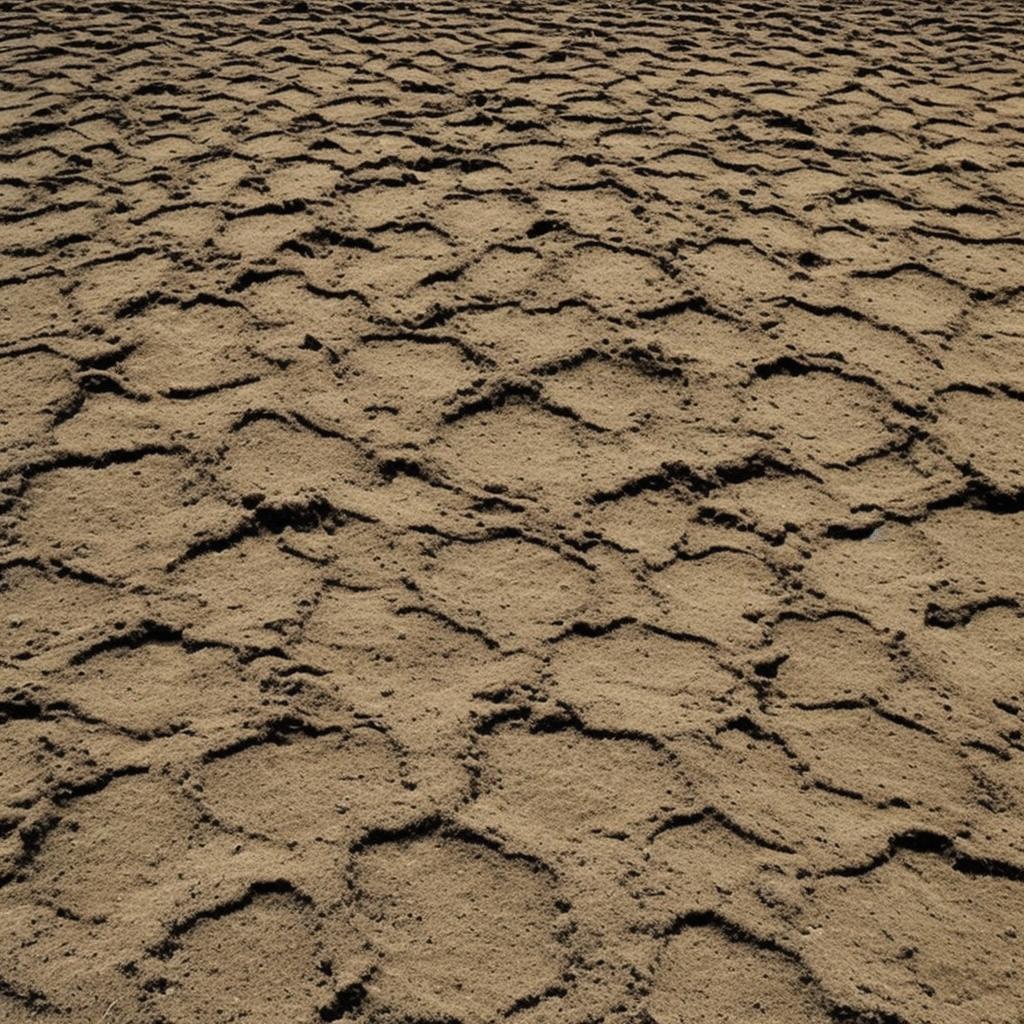} 
                    & \includegraphics[width=7cm]{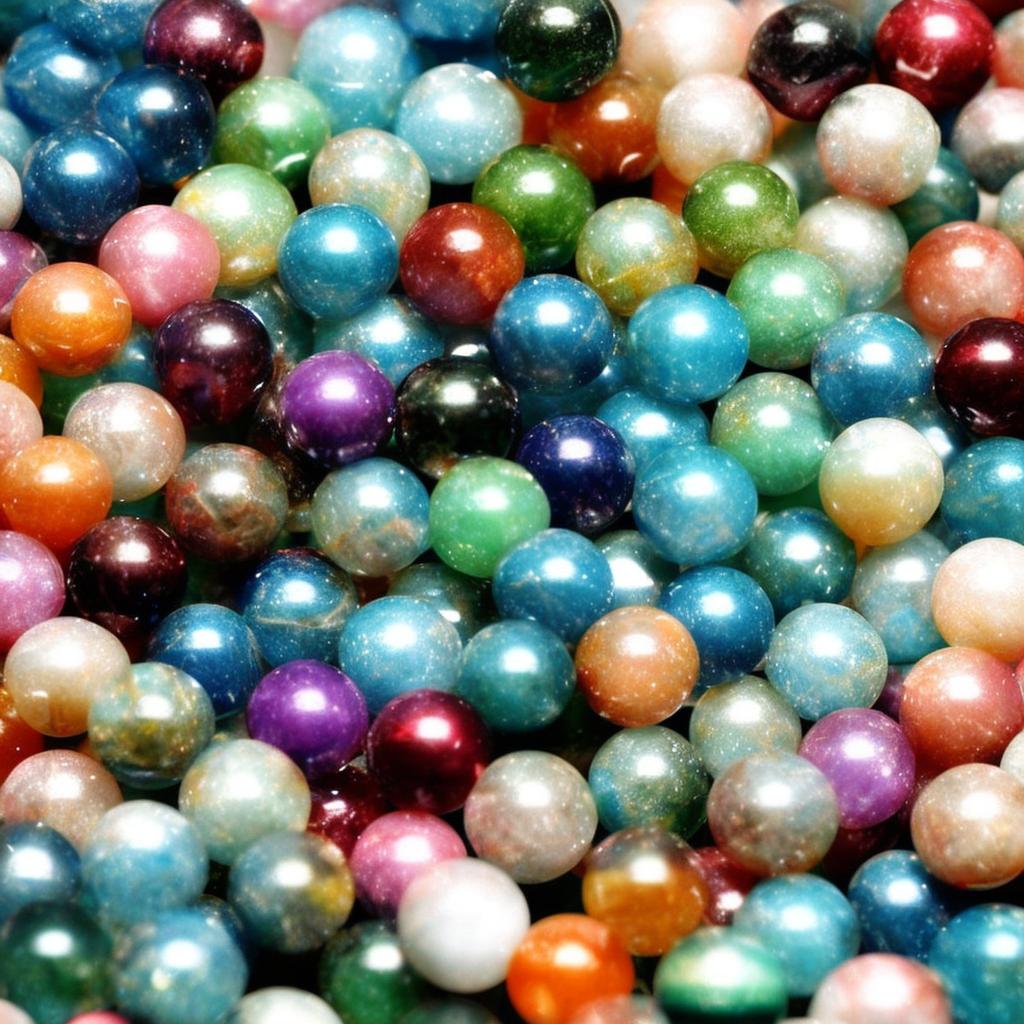} 
                    & \includegraphics[width=7cm]{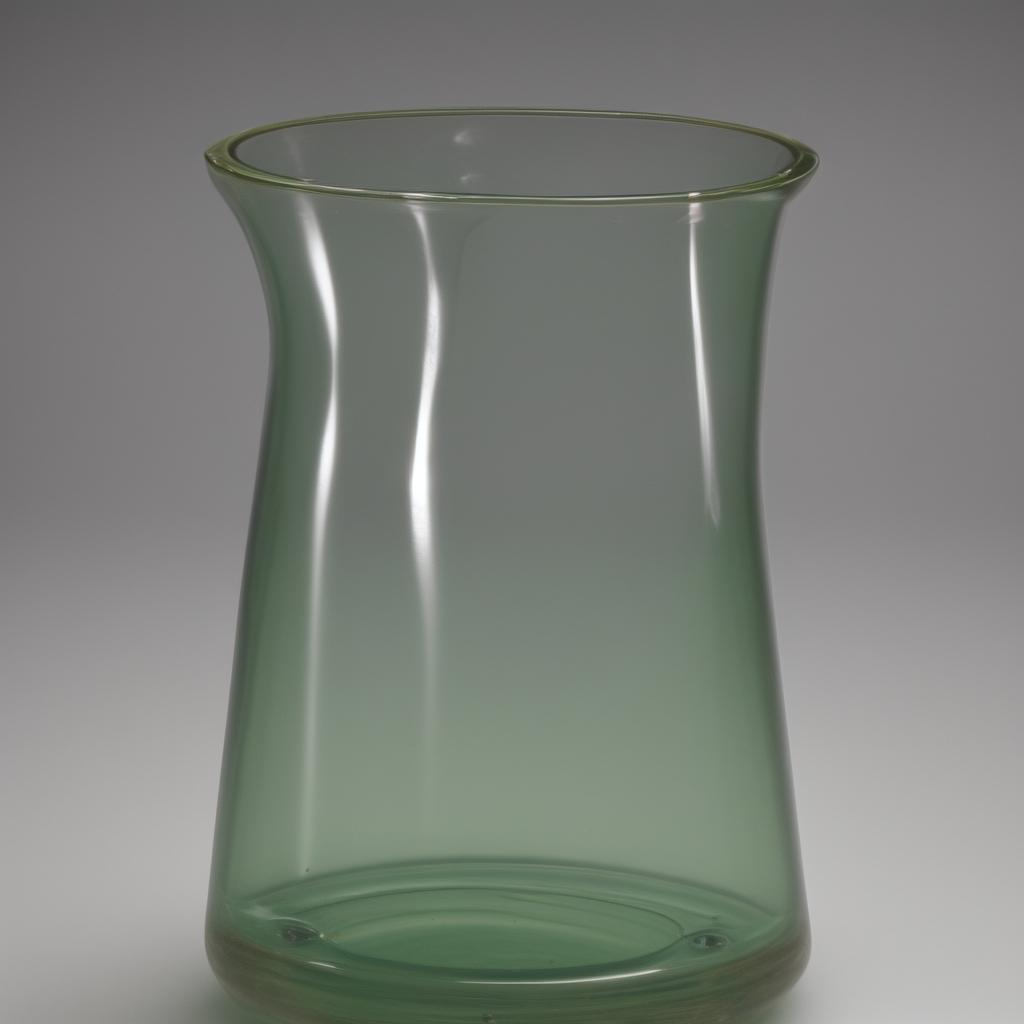} 
                    & \includegraphics[width=7cm]{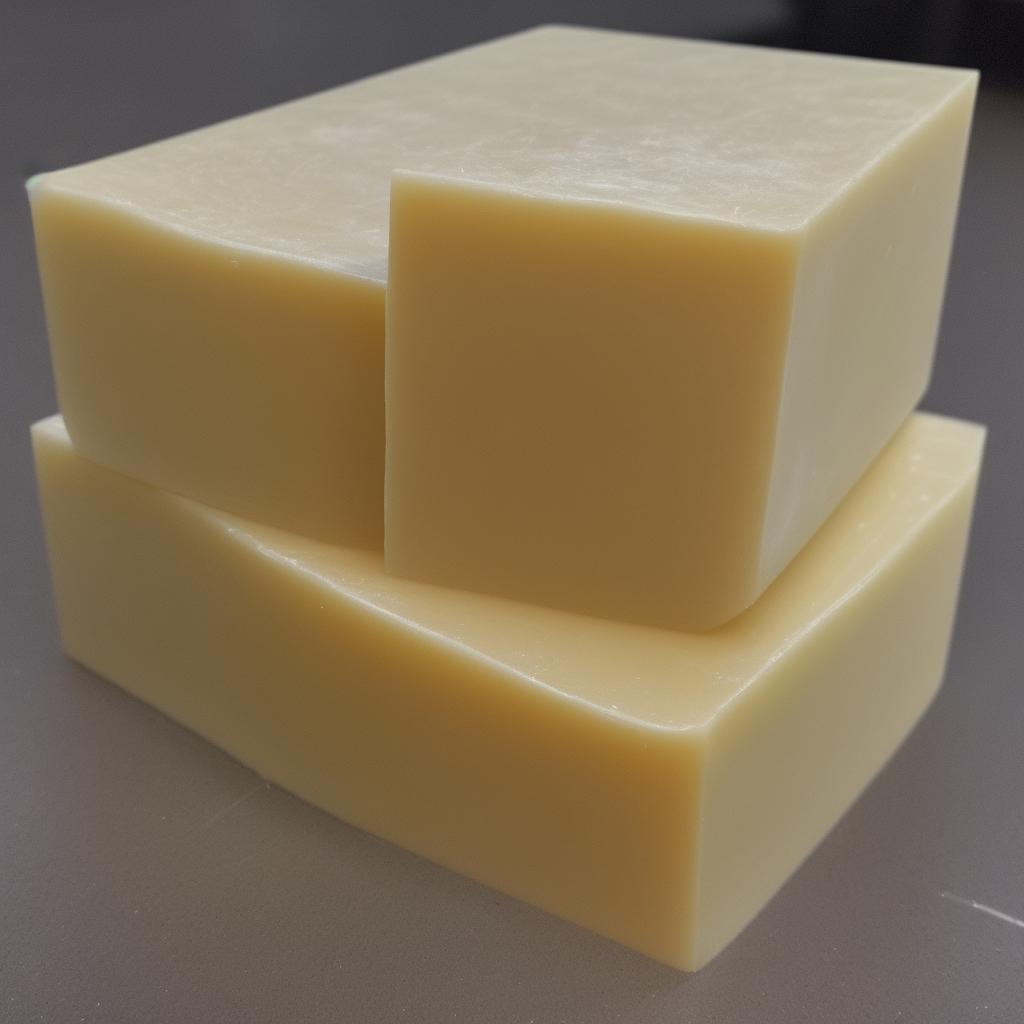} 
                    & \includegraphics[width=7cm]{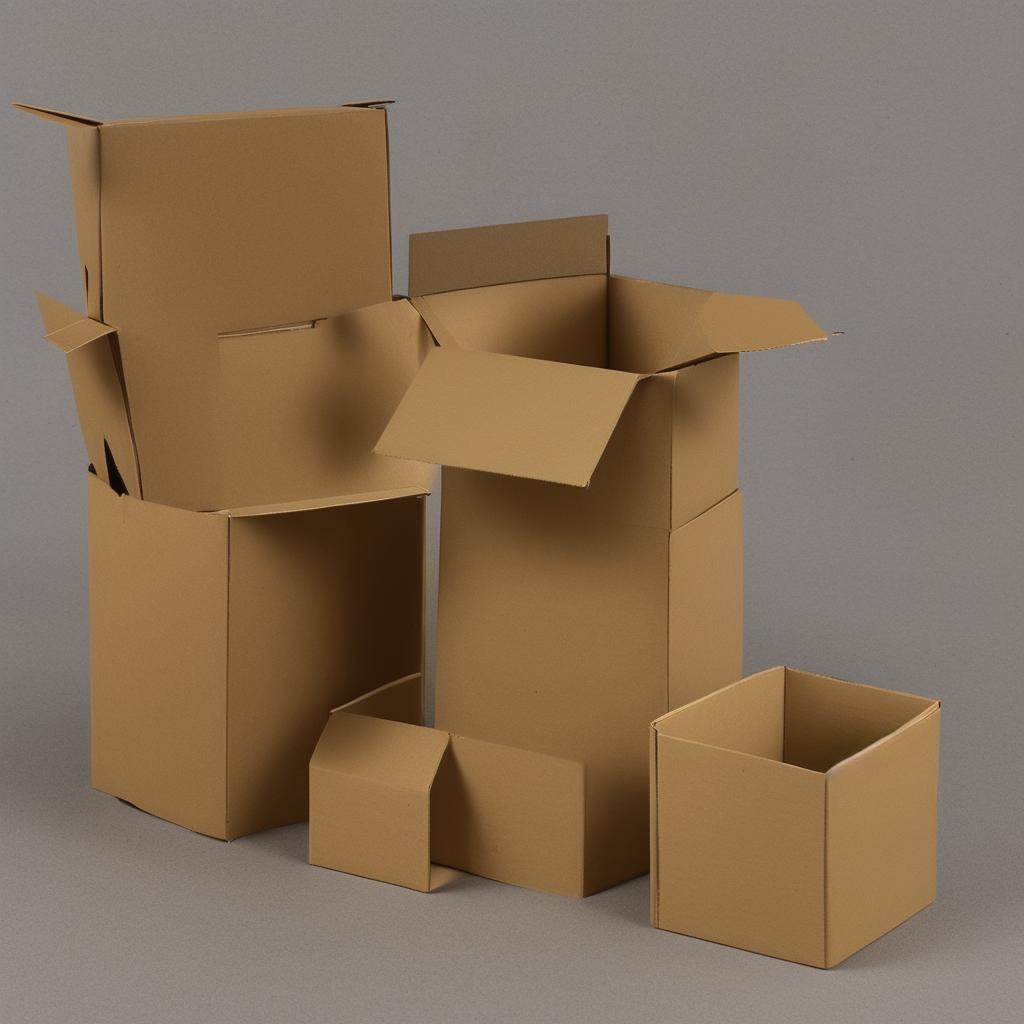} 
                    & \includegraphics[width=7cm]{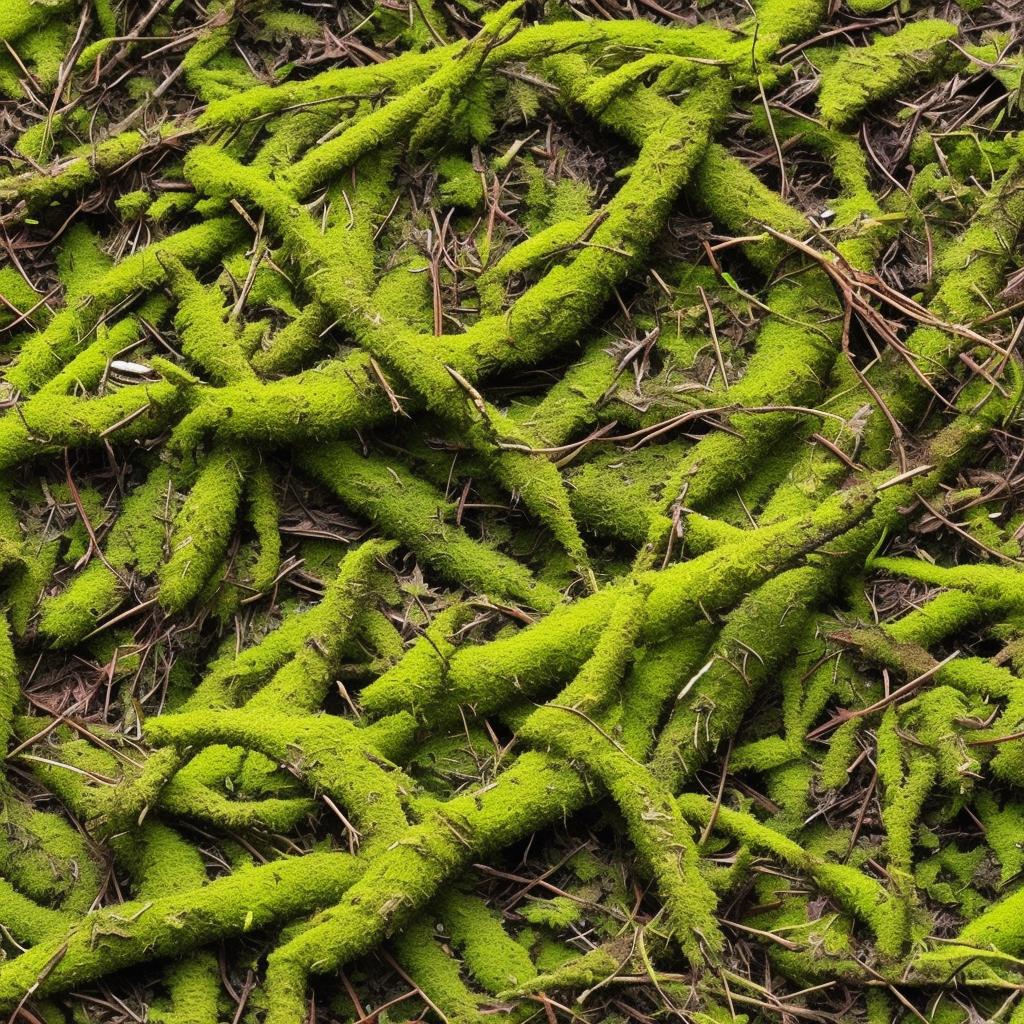} \\ 

                 \fontsize{26}{26}\selectfont{\emph{artifact,Bone,}} &  \fontsize{26}{26}\selectfont{\emph{bag,Paper,}} &  \fontsize{26}{26}\selectfont{\emph{bank,Soil,}}& \fontsize{26}{26}\selectfont{\emph{bead,Gemstone,}} & \fontsize{26}{26}\selectfont{\emph{beaker,Glass,}} & \fontsize{26}{26}\selectfont{\emph{block,Wax,}} & 
                \fontsize{26}{26}\selectfont{\emph{box,Cardboard,}} & \fontsize{26}{26}\selectfont{\emph{branch,Foliage,}}\\

                \fontsize{26}{26}\selectfont{\emph{natural}}& \fontsize{26}{26}\selectfont{\emph{towel}}& \fontsize{26}{26}\selectfont{\emph{eroded}}& \fontsize{26}{26}\selectfont{\emph{polished}}& \fontsize{26}{26}\selectfont{\emph{blown}}& \fontsize{26}{26}\selectfont{\emph{smooth}} & \fontsize{26}{26}\selectfont{\emph{fluted}}& \fontsize{26}{26}\selectfont{\emph{moss}}\\

                \end{tabular}

                }
               {\centering
                 \resizebox{0.625\textwidth}{!}{%
                \begin{tabular}
                {ccccc}
                \includegraphics[width=7cm]{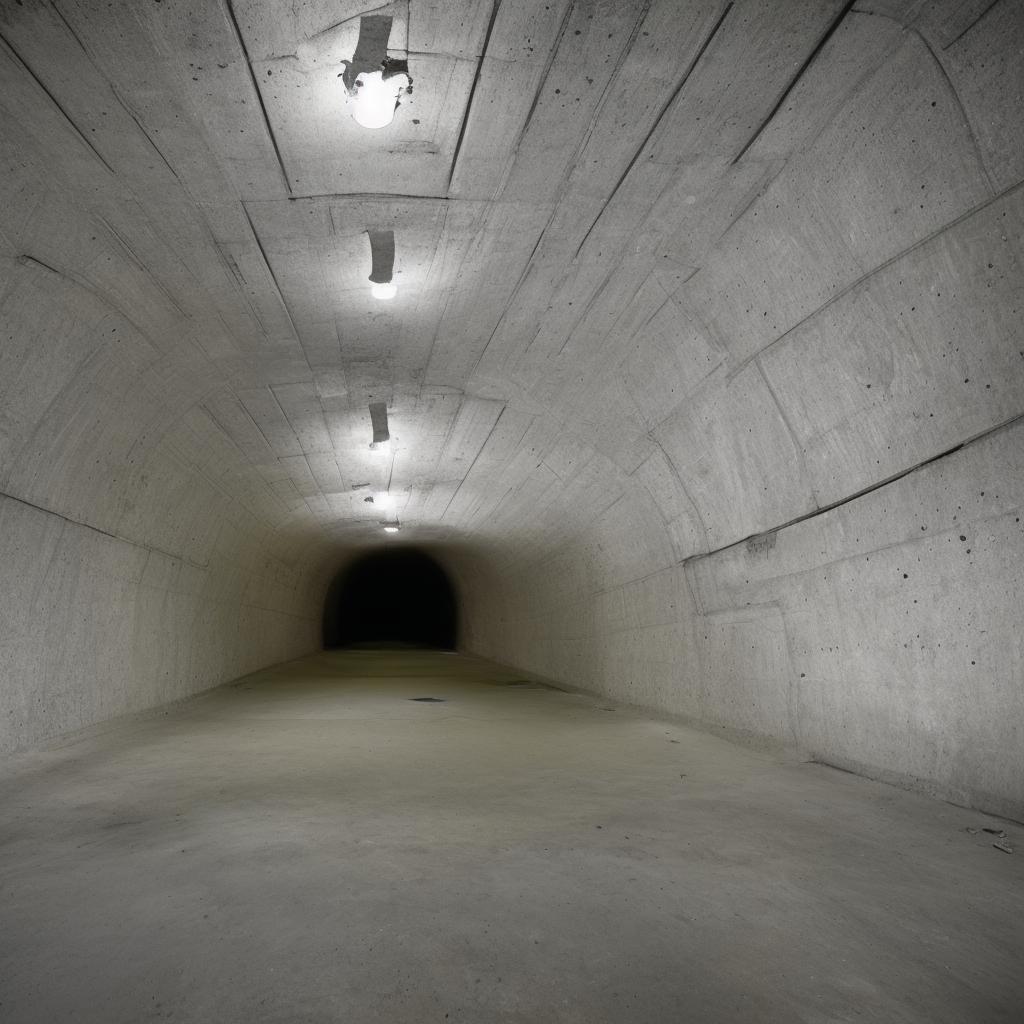} 
                    & \includegraphics[width=7cm]{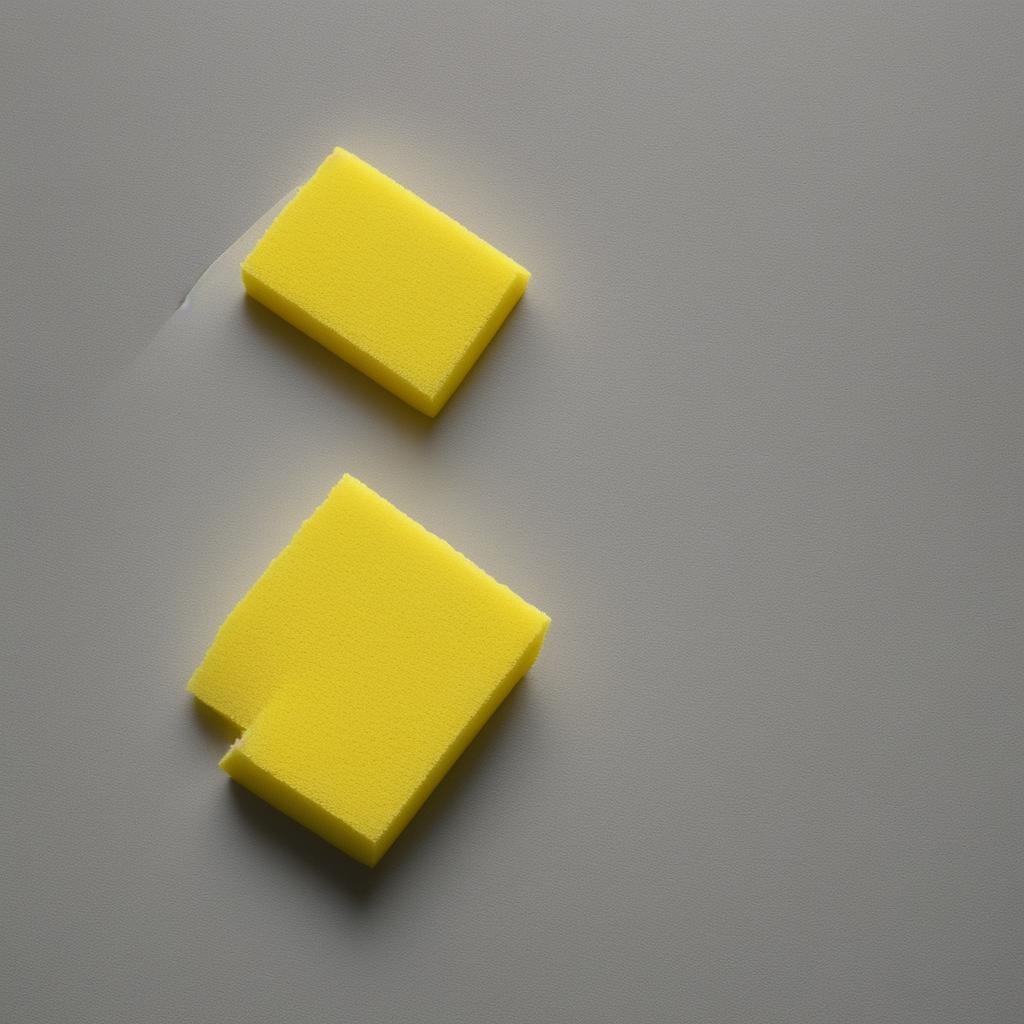} 
                    & \includegraphics[width=7cm]{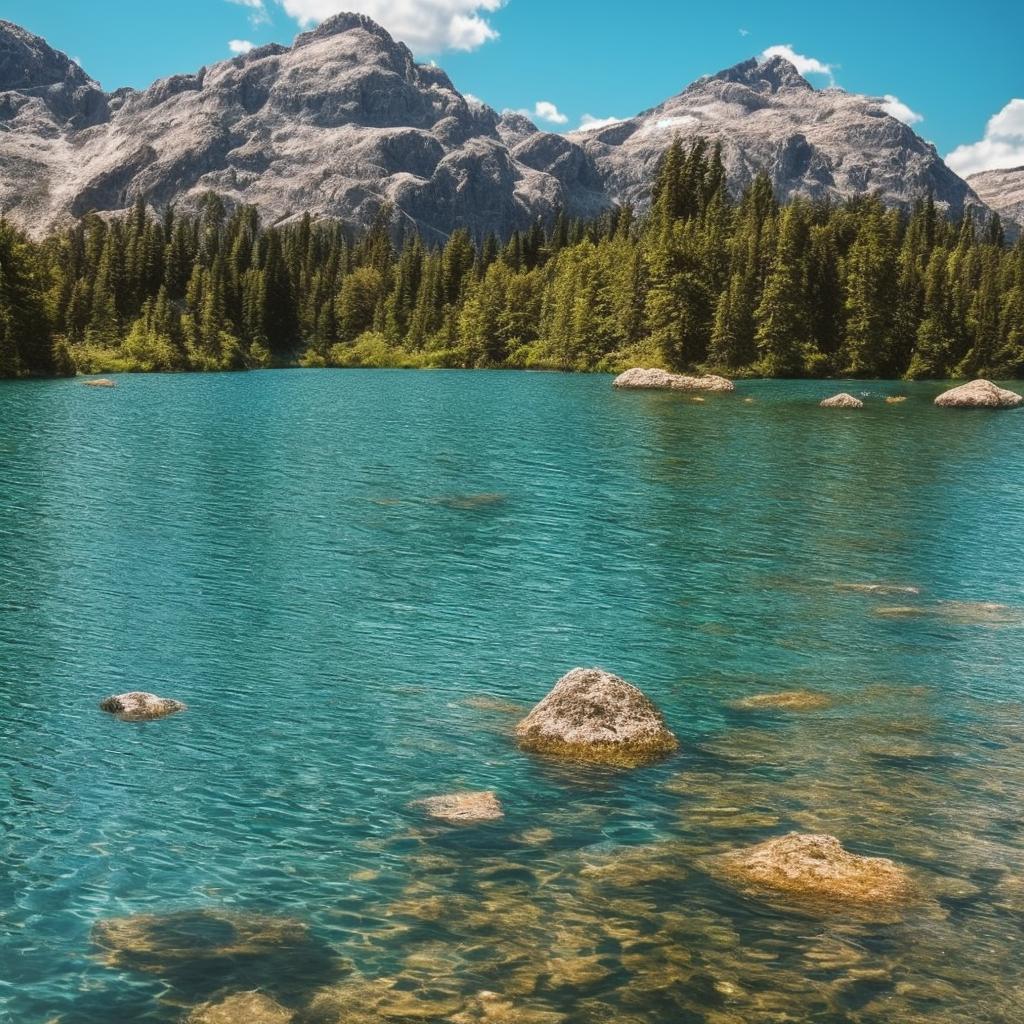} 
                    & \includegraphics[width=7cm]{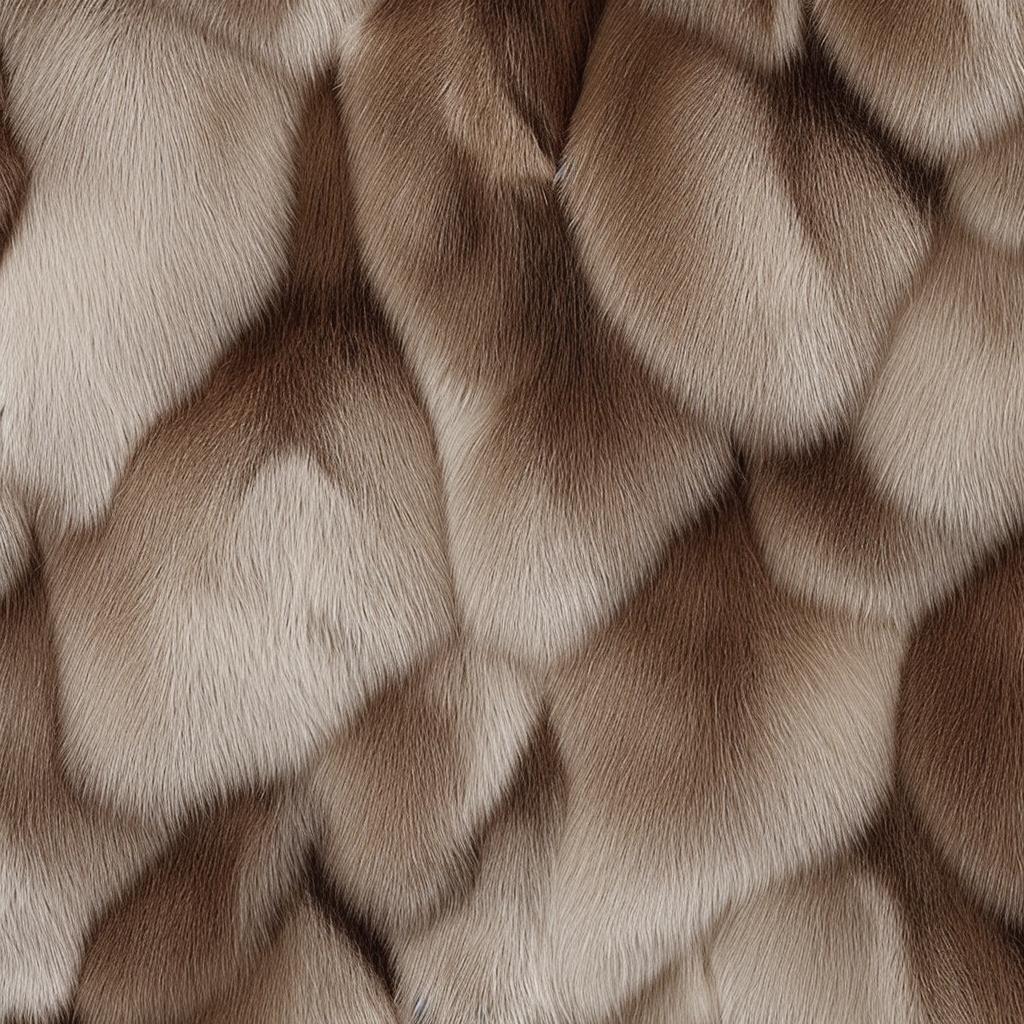} 
                    & \includegraphics[width=7cm]{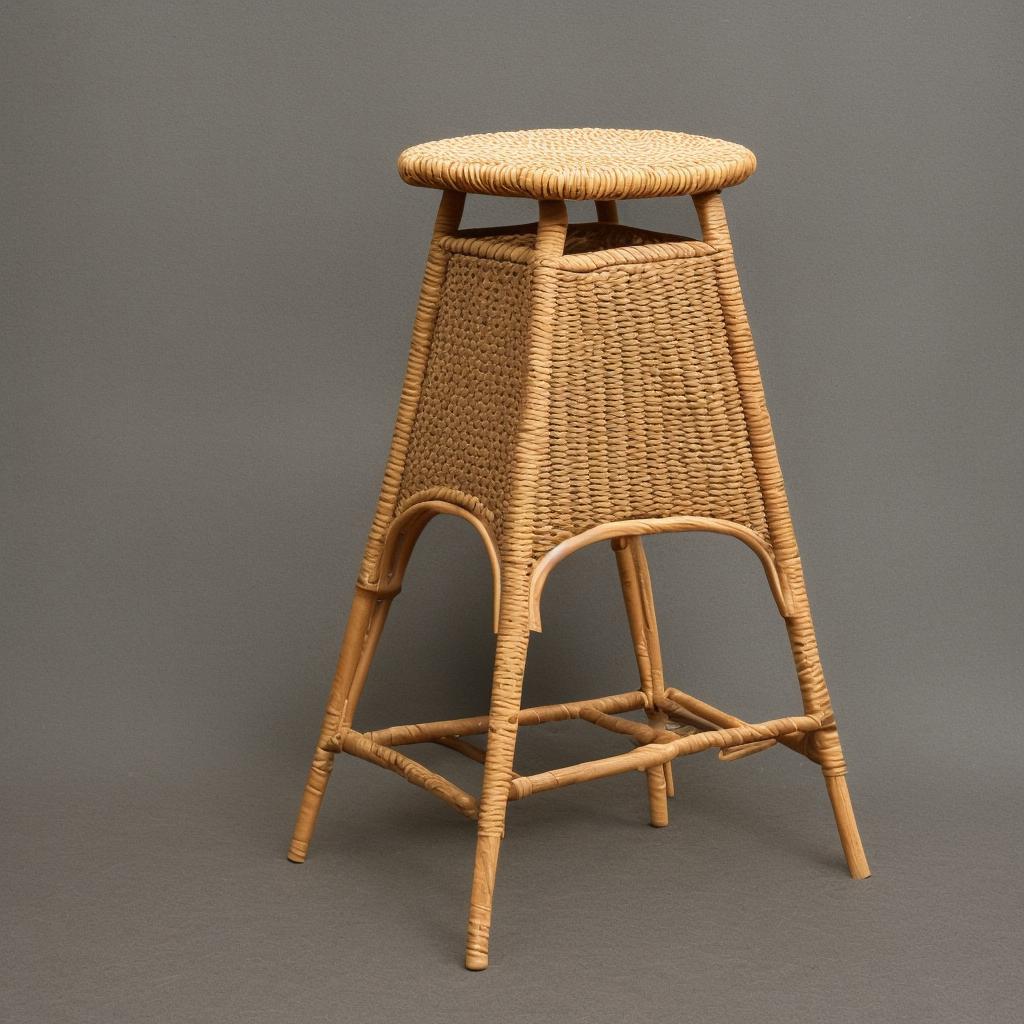} \\

                \fontsize{26}{26}\selectfont{\emph{bunker,Concrete,}} &  \fontsize{26}{26}\selectfont{\emph{foam,Sponge,}} &  \fontsize{26}{26}\selectfont{\emph{lake,Water,}}& \fontsize{26}{26}\selectfont{\emph{pelt,Fur,}} & \fontsize{26}{26}\selectfont{\emph{stool,Wicker,}}\\

                 \fontsize{26}{26}\selectfont{\emph{reinforced}}& \fontsize{26}{26}\selectfont{\emph{yellow}}& \fontsize{26}{26}\selectfont{\emph{clear}}& \fontsize{26}{26}\selectfont{\emph{natural}}& \fontsize{26}{26}\selectfont{\emph{open-weave}}

                 \end{tabular}
                 }
                \par}
                \captionof{figure}{Generated material images using diverse prompts across 21 material categories, with one image per category.}
    \label{fig:generated_imgs}
\vspace{-5mm}
\end{figure*}

Our synthetic dataset generation process addresses annotation scarcity by combining hierarchical prompt engineering, material image generation, and automated label validation. The workflow (Fig. \ref{fig:dataset_generation}) ensures high-quality material-image associations while maximizing diversity.

\paragraph{Prompt Design with Validation}
To generate plausible object-material pairs, we employ a three-tiered prompting strategy: \textit{object}, \textit{material category} (wood, metal, plastic, etc.), and \textit{sub-material} (e.g., ``stainless steel" under ``metal") or an \textit{adjective} describing the material (e.g., ``polished'' metal).. We first instruct ChatGPT to produce candidate triplets (e.g., ``vase, ceramic, porcelain"), filtering implausible combinations (e.g., ``sponge, metal") via human validation. By discarding implausible combinations, this step ensures both realism and diversity in the generated pairings. Compared to free-form prompting or attribute lists (e.g., ``a wooden object"), our approach improves text-to-image alignment since we can further perform semantic segmentation to align material and related image region.

\paragraph{Material Image Generation}
We qualitatively evaluate four diffusion models—SDXL~\cite{podell2023sdxl}, Stable Diffusion v2.1~\cite{rombach2022high}, PixArt-$\alpha$~\cite{chen2023pixart} and Playground v2.5~\cite{li2024playground} to assess the fidelity and diversity of generated material images. Through the inspection of generated images we find SDXL produces images  with marginally higher visual quality than SD v2.1 but introduces subtle artifacts in texture patterns (e.g. unrealistic wood grain). PixArt-$\alpha$ and Playground v2.5  exhibit superior artistic styling compared to SDXL, yet they share similar issues concerning texture realism. Given these findings, we select SDv2.1 for its balance of material accuracy and diversity, synthesizing 20k+ images across 21 categories: plastic, metal, leather, fabric, wood, stone, ceramic, water, sponge, bone, cardboard, concrete, foliage, fur, gemstone, glass, paper, soil, wax, wicker, rubber, thus offering broad coverage of real-world materials. As depicted in Figure \ref{fig:generated_imgs}, the generated images closely align with the prompts, and their quality is commendable.

\paragraph{Auto-Labeling via Semantic Grounding}
To resolve material ambiguity in generated scenes (e.g., a ``ceramic vase" on a ``wooden table"), we apply Grounding DINO~\cite{liu2023grounding} to segment the target object using the \textit{object} field from prompts. Material labels are assigned exclusively to the segmented region, filtering out conflicting materials in background areas. This approach achieves 98\% label accuracy on human-verified samples, which is hard to achieve based on pure material recognition, but reversely can be used to create a dataset that is helpful for improving material recognition.

\begin{figure*}
    \centering
    \includegraphics[width=\textwidth]{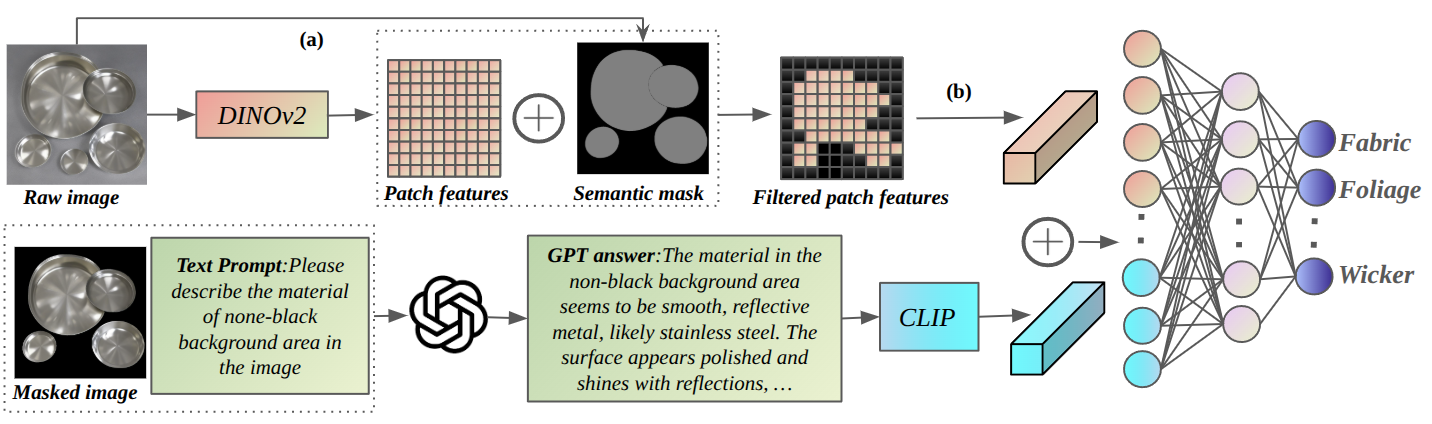}
    \caption{Dual-stream architecture. (1) \textbf{Vision Stream}: DINOv2 extracts patch features from the masked region, aggregated via max-pooling. (2) \textbf{Language Stream}: GPT-4v generates material descriptors encoded by CLIP. The fused features are classified via MLP. 
    \newline
      (a) Grounded SAM semantic mask extraction (b) Feature aggregation (max-pooling + flatten) }
    \label{fig:network_architecture}
    \vspace{-4mm}
\end{figure*}

\paragraph{Automatic material labeling.}
We've observed that diffusion models often produce images featuring multiple material categories, with the associated material typically appearing as the primary object or a significant part of the image. This can significantly impair learning performance if the entire image and its material label are used directly for training. A more substantial challenge is the absence of a method to directly identify the correct material regions, as this is the core problem we aim to solve. However, we've discovered an effective solution to this issue by employing semantic labels as a mediator.
As shown in Figure \ref{fig:dataset_generation} (c), we use Grounded SAM \cite{ren2024grounded} to incorporate object semantics from the text prompt as input to detect and segment the object areas within the image. For instance, if the text prompt is``Cake-pan", Grounded SAM can identify and segment the Cake-pan within the image. Since material and semantic features are usually accurately combined during the generation process, and semantic understanding solutions are well-established, we can label the material areas in the image with high precision through prompt-based material label mapping in Figure \ref{fig:dataset_generation} (d). This approach allows us to address the challenge of material region identification and enhance the accuracy of our material classification model.

\subsection{Formulation}
\label{sec:formulation}

Given an image \(\mathcal{I}\) and a binary mask \(\mathcal{M}\) denoting the region of interest, we predict the material label \(l\) by fusing \textit{vision} and \textit{language} foundation model priors. Let \(\mathcal{T}\) denote the image-specific material text description
generated by a vision-language model from the masked image $(\mathcal{I}\circ\mathcal{M})$
and the corresponding textual instruction. The prediction is formulated as:
\[
l = \mathcal{C}\big(\mathcal{E}_{\text{vis}}(\mathcal{I}, \mathcal{M}), \mathcal{E}_{\text{txt}}(\mathcal{T})\big),
\]
where \(\mathcal{E}_{\text{vis}}\) extracts visual features from the masked region and \(\mathcal{E}_{\text{txt}}\) encodes textual material descriptors.

\paragraph{Vision Prior Stream} extracts dense features using a frozen vision foundation model \(\phi_{\text{vis}}\):
\[
\mathbf{f}_{\text{vis}} = \mathcal{A}_{\text{vis}}\left(\phi_{\text{vis}}(\mathcal{I}) \circ \mathcal{M}\right),
\]
where \(\circ\) denotes element-wise masking and \(\mathcal{A}_{\text{vis}}: \mathbb{R}^{H' \times W' \times D} \to \mathbb{R}^{d_{\text{vis}}}\) aggregates features (e.g., masked average pooling).

\paragraph{Language Prior Stream} encodes material semantics using a frozen VLM \(\phi_{\text{txt}}\):
\[
\mathbf{f}_{\text{txt}} = \mathcal{A}_{\text{txt}}\left(\phi_{\text{txt}}(\mathcal{T})\right),
\]
where \(\mathcal{A}_{\text{txt}}: \mathbb{R}^{D_{\text{txt}}} \to \mathbb{R}^{d_{\text{txt}}}\) projects text embeddings to a latent space.

\paragraph{Cross-Modal Fusion} combines both streams via:
\[
l = \underset{k\in\{1,\dots,K\}}{\text{argmax}}\left(\text{MLP}\left(\mathbf{f}_{\text{vis}} \oplus \mathbf{f}_{\text{txt}}\right)\right),
\]
where $K$ is the number of material classes, and
$\text{MLP}(\mathbf{f}_{\text{vis}}\oplus \mathbf{f}_{\text{txt}})$ denotes
the predicted logit for class $k$.
\[
\mathcal{L} = -\sum_{k=1}^K \hat{l}_k \log p_k,
\]
where \(p_k\) is the predicted probability for class \(k\) and \(\hat{l}_k\) the ground truth label. Frozen \(\phi_{\text{vis}}\) and \(\phi_{\text{txt}}\) preserve pre-trained priors while \(\mathcal{A}_{\text{vis}}\), \(\mathcal{A}_{\text{txt}}\), and MLP are learnable.


\subsection{Network Architecture}
\label{sec:approach-network}

Our dual-stream architecture combines vision foundation model features with language-guided material semantics (Figure .~\ref{fig:network_architecture}).

\paragraph{Vision Stream} employs DINOv2~\cite{oquab2023dinov2} as the backbone encoder $\mathcal{E}_{\text{vis}}$. For an input image $\mathcal{I}$, it produces dense patch features:
\[
\{\mathbf{f}_{(i,j)}\} = \mathcal{E}_{\text{DINOv2}}(\mathcal{I}) \in \mathbb{R}^{32 \times 32 \times 768},
\]
where each $\mathbf{f}_{(i,j)}$ corresponds to a $32 \times 32$ image patch. We downsample the binary mask $\mathcal{M}$ to $32 \times 32$ resolution, retaining only features where $\mathcal{M}_{(i,j)} > 0$. These masked features are aggregated via max-pooling:
\[
\mathbf{f}_{\text{vis}} = \max_{i,j} \left(\{\mathbf{f}_{(i,j)}\} \circ \mathcal{M}\right) \in \mathbb{R}^{768}.
\]

\paragraph{Language Stream} encodes material semantics using CLIP~\cite{radford2021learning}. For each image to recognize, we ask GPT-4V~\cite{achiam2023gpt} to generate descriptive text $\mathcal{T}$ (e.g., ``appears polished and shines with reflections ..." shown in Figure \ref{fig:network_architecture}). CLIP’s text encoder $\phi_{\text{txt}}$ projects these into embeddings:
\[
\mathbf{f}_{\text{txt}} = \phi_{\text{txt}}(\mathcal{T}) \in \mathbb{R}^{512}.
\]
The resulting text embedding is fused with the visual feature for material classification.

\paragraph{Fusion \& Classification} concatenates both modalities:
\[
\mathbf{f}_{\text{fuse}} = \mathbf{f}_{\text{vis}} \oplus \mathbf{f}_{\text{txt}} \in \mathbb{R}^{768+512},
\]
where $\oplus$ denotes concatenation. A MLP maps $\mathbf{f}_{\text{fuse}}$ to class probabilities:
\[
p(l=k) = \text{MLP}(\mathbf{f}_{\text{fuse}})_k.
\]

\paragraph{Training Protocol} We freeze DINOv2 and CLIP to preserve pretrained priors and train only MLP parameters with AdamW (lr=5e-5). This ensures adaptation to material recognition without catastrophic forgetting of foundation model knowledge.
\section{Experiment}

\subsection{Overview}

We evaluate both the proposed material classification method and the quality of the datasets as our core contributions. In our experiments, we evaluate performance using mean Intersection over Union (mIoU) following the protocol in \cite{upchurch2022dense} and mean Accuracy (mAcc). For methods involving GPT-4v, however, we report accuracy only. In Section 4.2, we compare our method with state-of-the-art methods to demonstrate its effectiveness.
To ensure a fair comparison, we conduct experiments on three datasets: the FMD ~\cite{Sharan-JoV-14} dataset, a classic 10-class benchmark for material classification; the DMS-test dataset, a 21-class subset we constructed from the comprehensive DMS ~\cite{upchurch2022dense} dataset; and the Google-test dataset, a curated 21-class collection from Google Images that provides higher-quality samples and better reflects real-world material appearances as well as artist-created content. To evaluate the quality of our synthetic dataset relative to the public DMS ~\cite{upchurch2022dense} dataset, we train our vision module separately on two datasets and compare the material classification performance on DMS-test, Google-test and FMD~\cite{Sharan-JoV-14}.
In Section 4.3, we further conduct ablation studies to investigate the contributions of each component of our proposed method. This includes the dual-stream language and vision architecture, the selection of the vision backbone, and the impact of dataset scale and  with/without auto labeled semantics on our task.
 
\subsection{Comparison}
\label{sec:comparison}
\textbf{Compare with SOTA methods} Due to the scarcity of high-quality data in material classification tasks, recent state-of-the-art methods, such as Mapa \cite{zhang2024mapa} and Make-It-Real \cite{fang2024make}, have begun to incorporate priors from zero-shot foundation models, notably CLIP \cite{radford2021learning} and GPT-4v \cite{achiam2023gpt}, for material segmentation purposes. In our study, we compare our proposed approach against these models.

For CLIP, we adopt the ViT-H/14 model, which was trained on approximately two billion image–text pairs, to serve as a strong baseline for comparison. We first use CLIP ~\cite{radford2021learning} to encode the candidate material texts (plastic, foliage, glass etc.) and the input test image. We then select the material text whose encoded feature is the nearest neighbor to the feature encoded by the input image as the predicted material. For evaluation, we consider 10 classes that align with FMD~\cite{Sharan-JoV-14}, whereas for DMS-test and Google-test we use the full set of 21 classes. For GPT-4v~\cite{achiam2023gpt}, we evaluate its performance by prompting it with ``Please identify the material of the non-masked area.” A prediction is regarded as correct if the response contains the ground-truth class label of the input image.

In addition to comparing with zero-shot foundation models, we also evaluate our approach against baselines and the state-of-the-art method proposed in MatSim \cite{drehwald2023one} for general material classification. The MatSim method outputs a 512-dimensional descriptive vector for each region. To assess its performance, we follow this procedure: each image in the test dataset is matched against all other images based on the similarity of their feature vectors. If the best-matched image belongs to the same class, it is considered correctly classified.

The comparison results are shown in Table~\ref{tab:table1} (Material classification performance on the FMD~\cite{Sharan-JoV-14} dataset) and Table~\ref{tab:table2} (Material classification performance on the DMS-test dataset and Google test dataset). Our approach significantly outperforms both zero-shot CLIP \cite{radford2021learning} and zero-shot GPT-4v \cite{achiam2023gpt}. This result indicates that zero-shot foundation models struggle with material classification, likely due to limited exposure to labeled material data during training. Furthermore, our approach significantly outperforms MatSim \cite{drehwald2023one}, a state-of-the-art method specifically designed for material tasks. This advantage is likely due to our innovative dual-stream architecture and higher-quality training data. These results highlight the substantial superiority of our approach over existing methods.

\begin{table}[h!]
    \caption{Comparison with state-of-the-art methods on the FMD dataset (10 classes). All values denote classification accuracy (truncated to two decimal places), and the best result in each row is highlighted in \textbf{bold}.}
    \label{tab:table1}
    \resizebox{\columnwidth}{!}{%
    \begin{tabular}{l|c|c|c|c}
    \diagbox[width=6em]{Class}{Method}&CLIP\cite{radford2021learning} & GPT-4v \cite{achiam2023gpt}& MatSim\cite{drehwald2023one} & Ours\\
      \hline
      fabric&0.66&0.85&0.47 &\textbf{1.00}\\
      foliage& 0.93 & 0.19 &0.82 &\textbf{0.97}\\
      glass& 0.84 & 0.76 & 0.57 &\textbf{0.87}\\
      leather & 0.67 & \textbf{0.87} &0.57 &0.85\\
      metal & 0.62 & 0.61 & 0.47 &\textbf{0.94}\\
      paper & 0.83  & \textbf{0.88} &0.39  &0.84\\
      plastic & \textbf{0.90} & 0.87 &0.52 &\textbf{0.90}\\
      stone & 0.87 & 0.64 &0.62  &\textbf{0.91}\\
      water & 0.84 & 0.83 &0.57  &\textbf{0.92}\\
      wood & 0.91 &\textbf{0.95} &0.62  &0.75\\
      \hline
      {Average}& 0.80 & 0.74 &0.56& \textbf{0.89}\\
    \end{tabular}
    }
     \vspace{-3mm} 
\end{table}

\begin{table}[h!]
  \begin{center}
    \caption{Comparison of material classification on the DMS-test dataset (21 classes) and the Google-test dataset (21 classes). All values denote classification accuracy, and the best result in each row is highlighted in \textbf{bold}.}
    \label{tab:table2}
    \resizebox{\columnwidth}{!}{%
    \begin{tabular}{l|c|c|c|c}
    \diagbox{Class}{Method}&CLIP\cite{radford2021learning} & GPT-4v \cite{achiam2023gpt}& MatSim\cite{drehwald2023one} & Ours\\
    \hline
      DMS-test &0.38& 0.43 & 0.41 &\textbf{0.64}\\
      Google-test &0.81 & 0.74& 0.63 & \textbf{0.92}\\
    \end{tabular}
    }
    \vspace{-8mm} 
  \end{center}
\end{table}


\paragraph{Analysis of dataset quality} 
We evaluate the quality of our synthetic dataset by comparing it with the state-of-the-art (SOTA) dataset, DMS \cite{upchurch2022dense}, which contains 3.2 million dense material segments and is currently the largest material segmentation dataset available. We train our vision branch(fine-tune DINOv2) separately on DMS dataset and our synthetic dataset and compare the material classification performance on DMS-test, Google-test and FMD~\cite{Sharan-JoV-14}. 
As expected, in-domain evaluation, where the model is trained on the DMS dataset and tested on DMS-test, achieves relatively higher performance compared to models trained on our synthetic dataset. However, this advantage disappears in cross-dataset evaluation, and when tested on Google-test and FMD, the synthetic-trained model significantly outperforms the DMS-trained one as shown in Table~\ref{tab:table3}. These results suggest that while DMS provides moderate in-domain performance, it lacks sufficient generalization ability.
Figure \ref{fig:pca_grid_singlecol} provides further insights by visualizing PCA distributions of ten material classes,derived from max-pooled aggregated DINOv2 patch features. The FMD samples (blue) form compact clusters with small intra-class variation. In contrast, the DMS samples (orange) appear more dispersed and often shifted from FMD, revealing a clear domain gap between the two datasets. This gap is reflected not only in the mean shift but also in differences in variance and distributional shape, indicating that the two datasets capture distinct feature statistics. Our synthetic samples (green) not only overlap with DMS but also align more closely with FMD, thereby bridging the two distributions. This explains why synthetic training achieves superior cross-domain performance, as the synthesis pipeline combining validated prompt design, material image generation, semantic grounding, and automatic annotation introduces greater diversity that extends the representational coverage beyond existing benchmark.

\begin{table}[h!]
    \caption{Comparison on material classification with DMS dataset(mIoU\textbar mAcc). Best in \textbf{bold}.}
    \label{tab:table3}
    \resizebox{\columnwidth}{!}{%
    \begin{tabular}{l|c|c|c}
      \diagbox{Train data}{Test data} & DMS-test & Google-test & FMD~\cite{Sharan-JoV-14} \\
      \hline
      DMS\cite{upchurch2022dense}& \textbf{0.52\texttt{\textbar}0.67} & 0.57\texttt{\textbar}0.71 & 0.67\texttt{\textbar}0.79 \\
      Ours& 0.46\texttt{\textbar}0.60 & \textbf{0.81\texttt{\textbar}0.89}   & \textbf{0.79\texttt{\textbar}0.88} \\
    \end{tabular}
    }
    \vspace{-2mm} 
\end{table}


\definecolor{ds1blue}{HTML}{1f77b4}
\definecolor{ds2orange}{HTML}{ff7f0e}
\definecolor{ds3green}{HTML}{2ca02c}

\begin{figure}[t]
    \centering
    \setlength{\tabcolsep}{2pt} 
    \renewcommand{\arraystretch}{1.0}
    \begin{tabular}{cccc}
        \includegraphics[width=0.24\columnwidth]{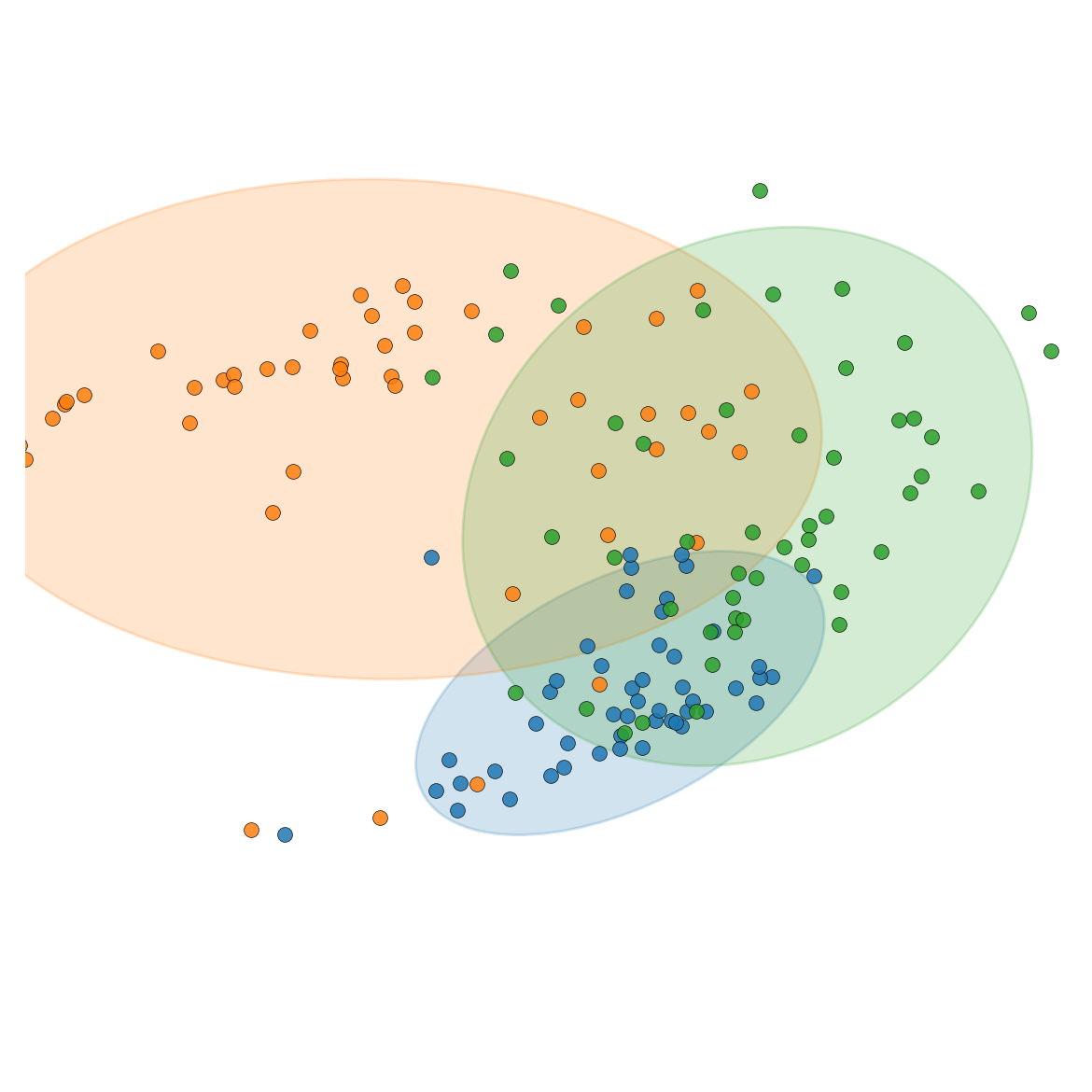} &
        \includegraphics[width=0.24\columnwidth]{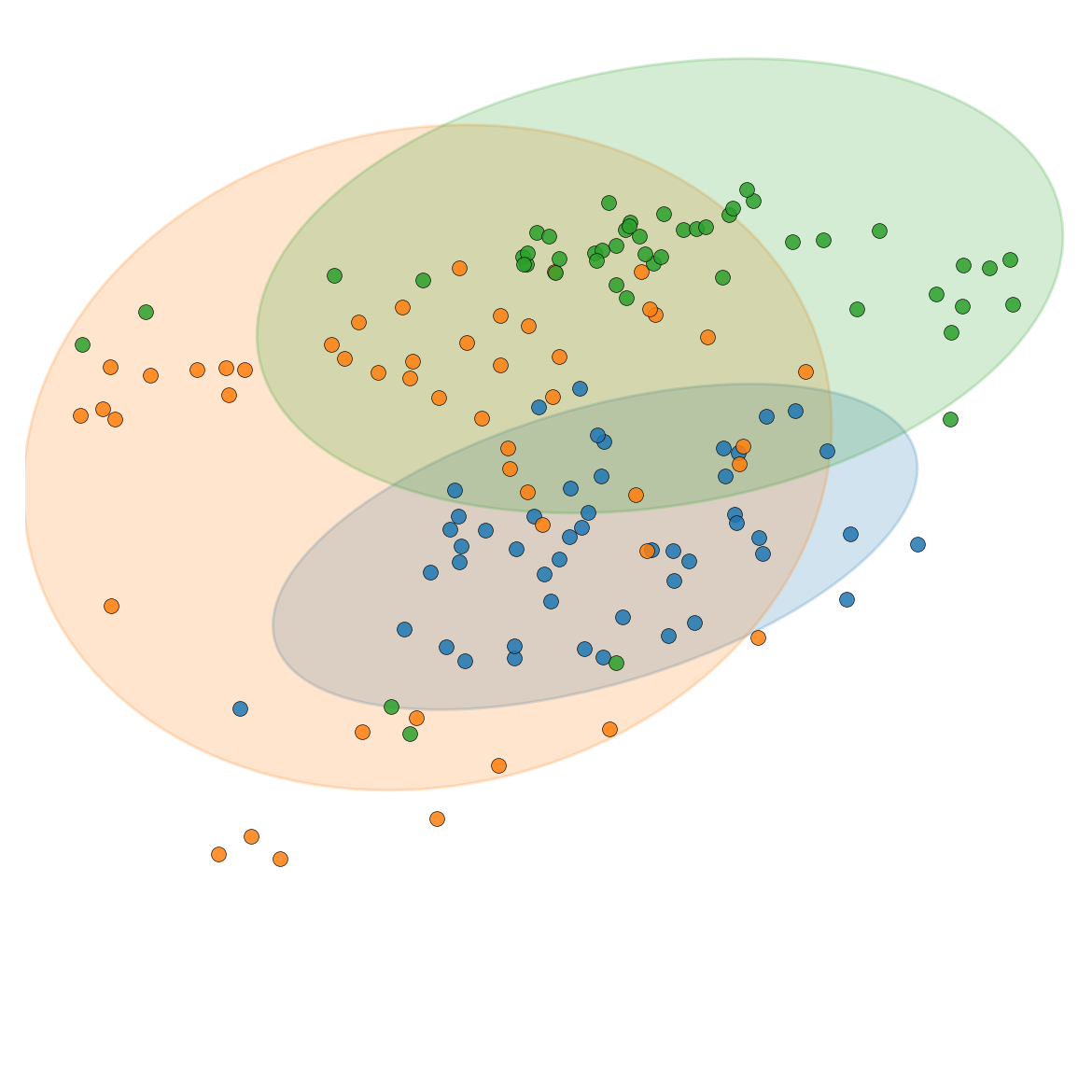} &
        \includegraphics[width=0.24\columnwidth]{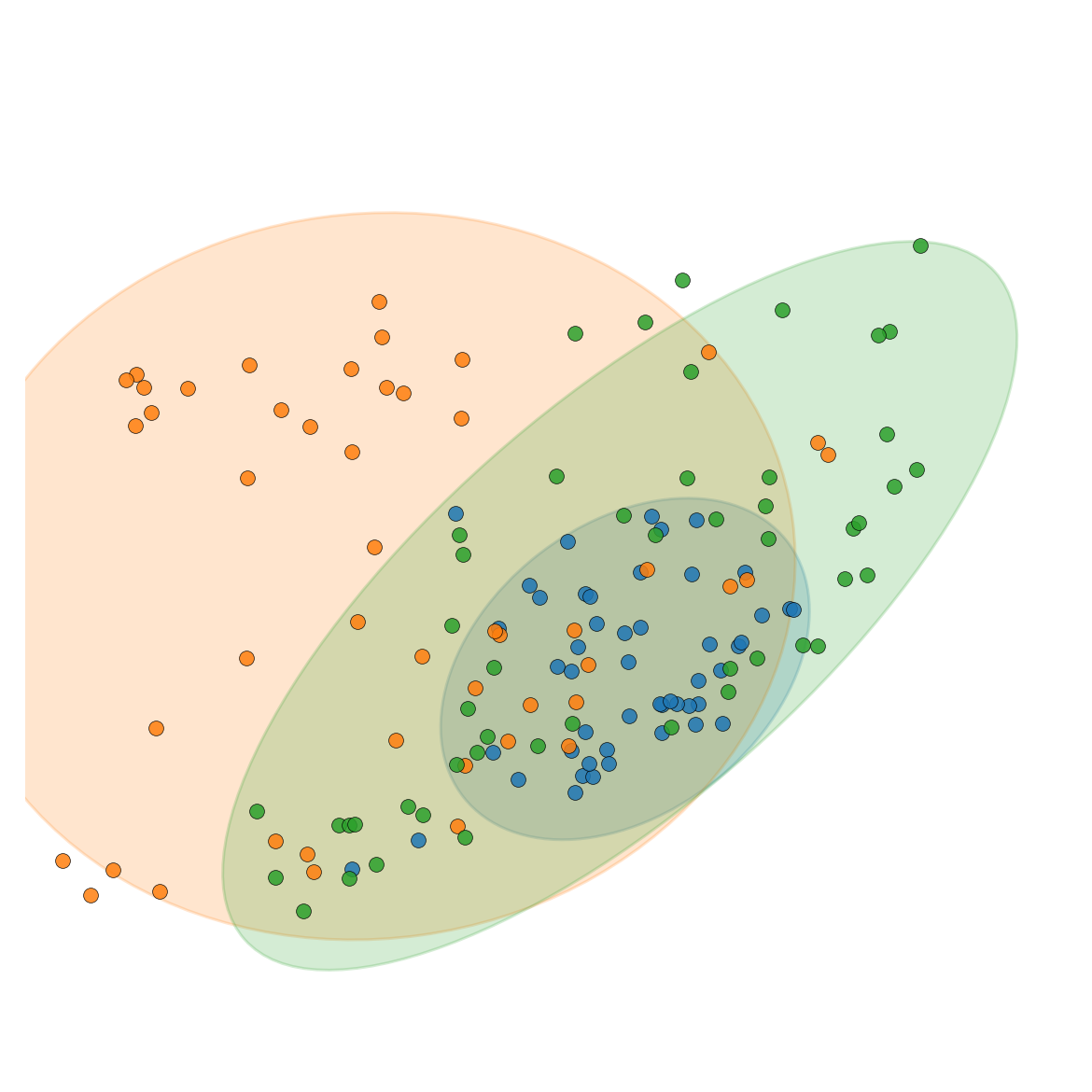} &
        \includegraphics[width=0.24\columnwidth]{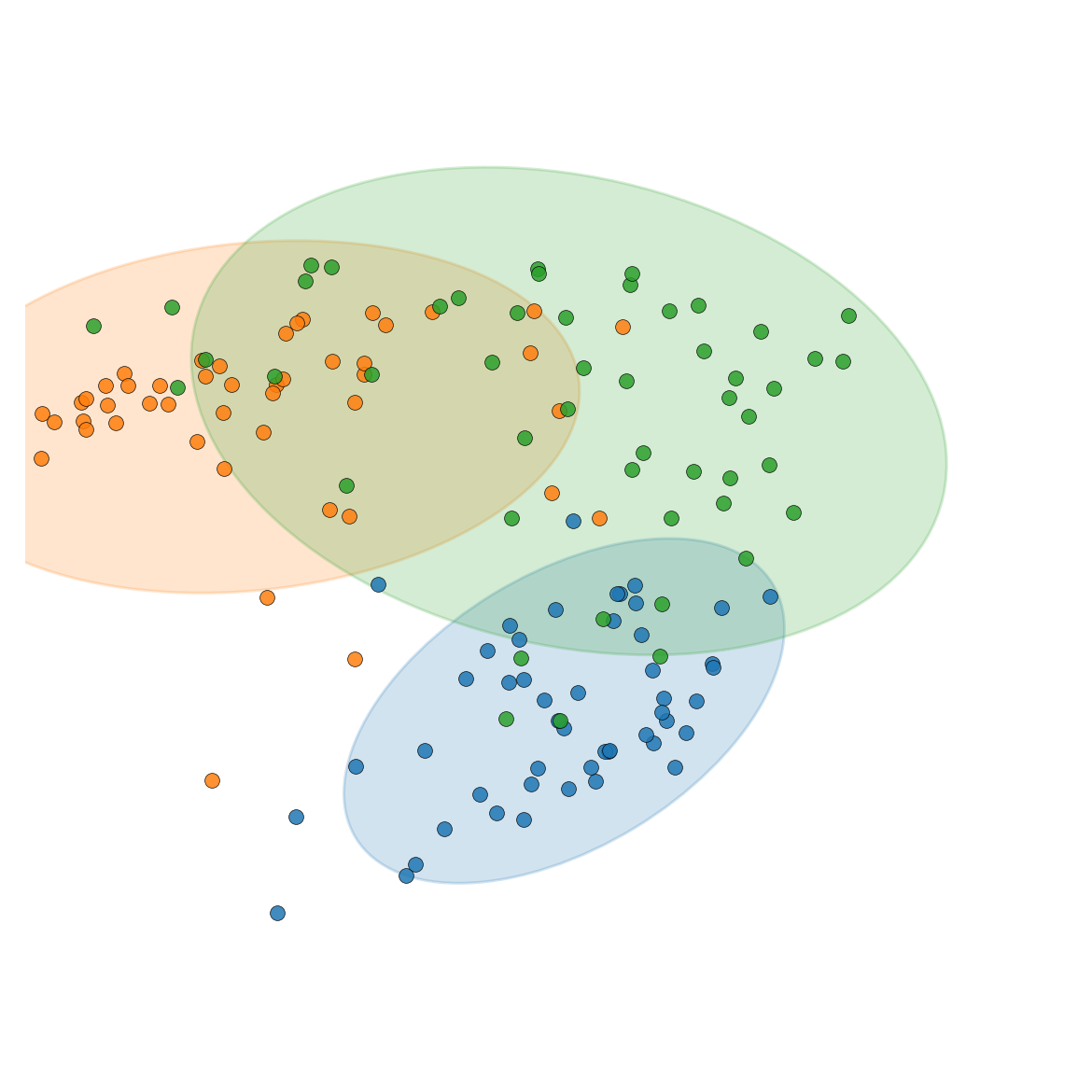} \\
        {\scriptsize \emph{Fabric}} &
        {\scriptsize \emph{Foliage}} &
        {\scriptsize \emph{Glass}} &
        {\scriptsize \emph{Leather}} \\
        \includegraphics[width=0.24\columnwidth]{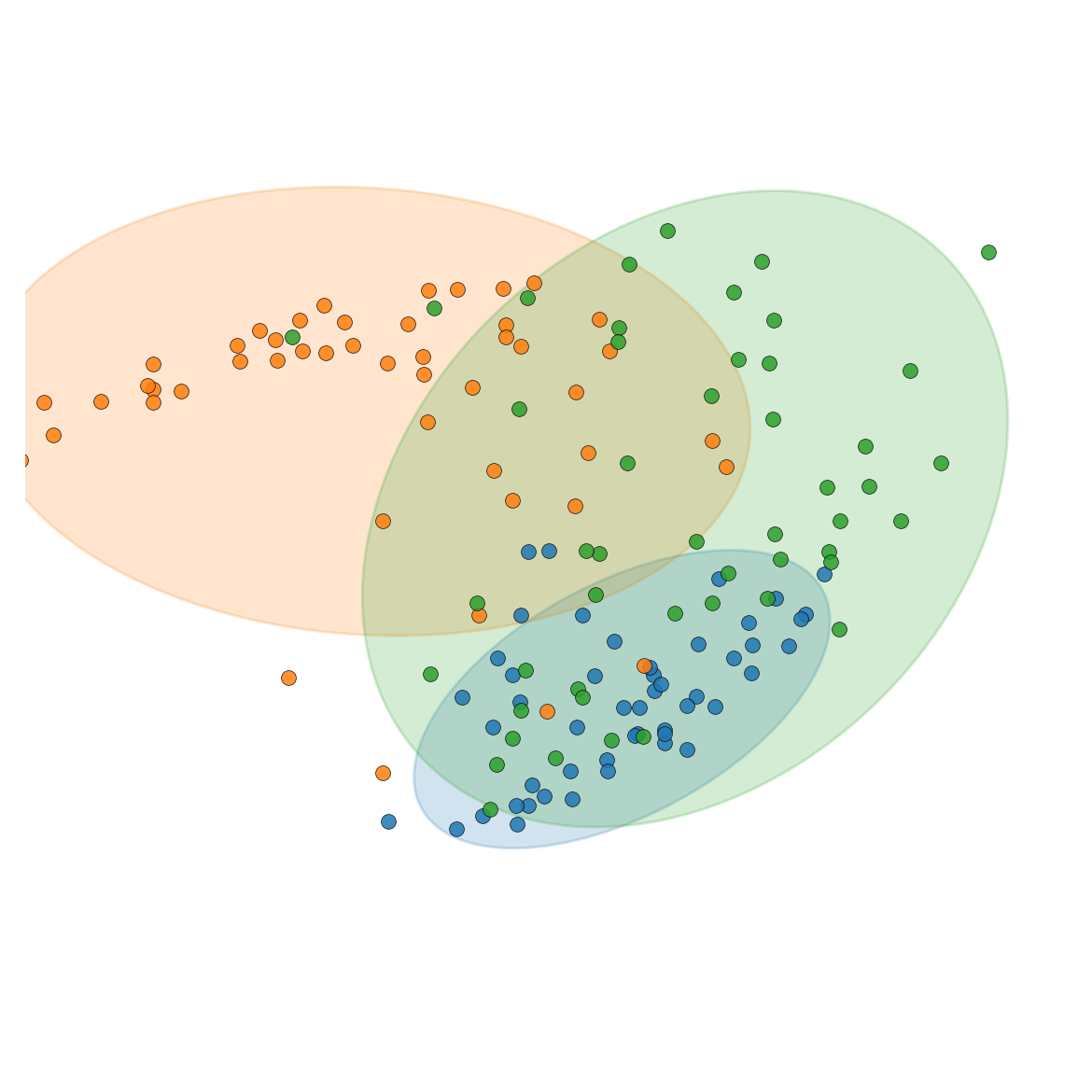} &
        \includegraphics[width=0.24\columnwidth]{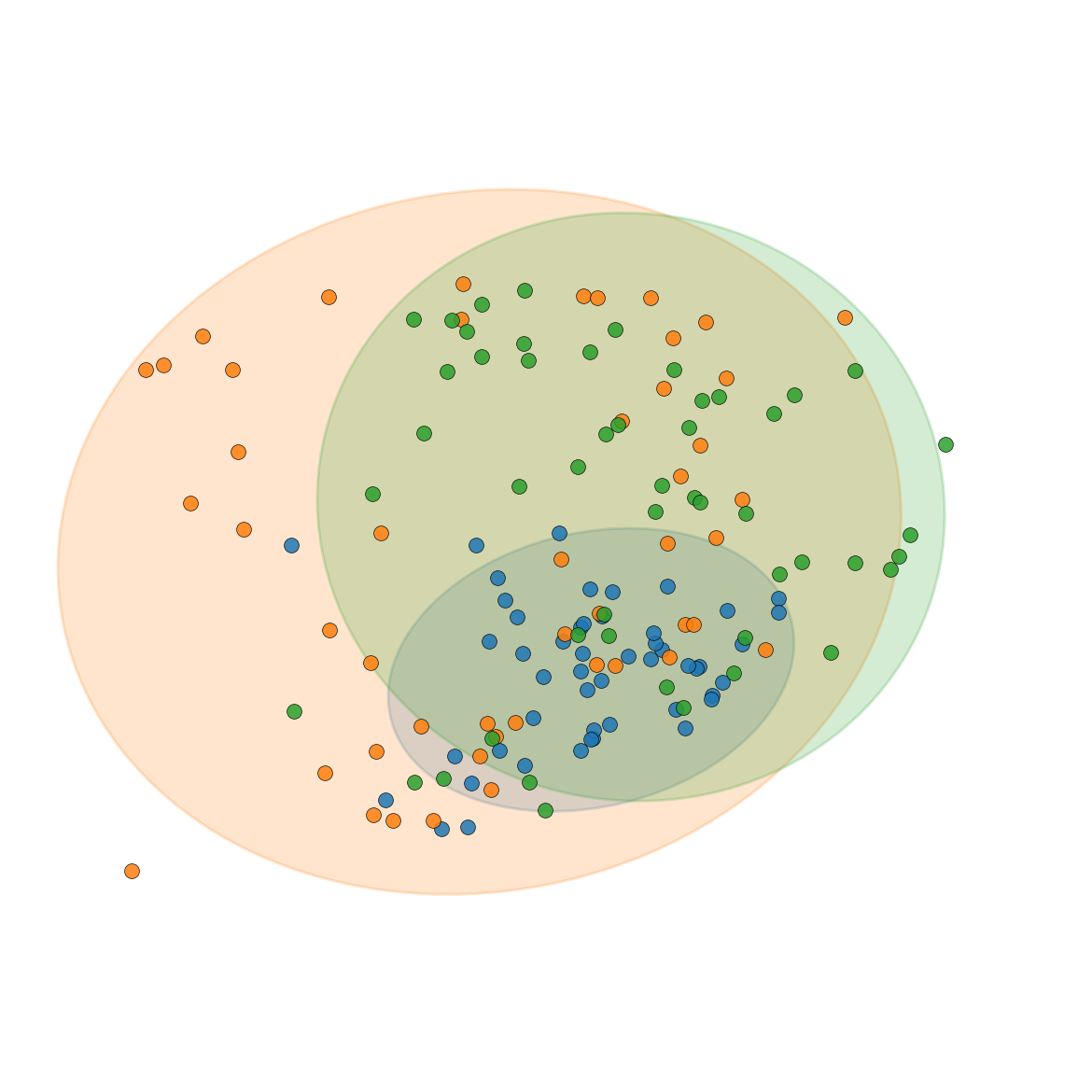} &
        \includegraphics[width=0.24\columnwidth]{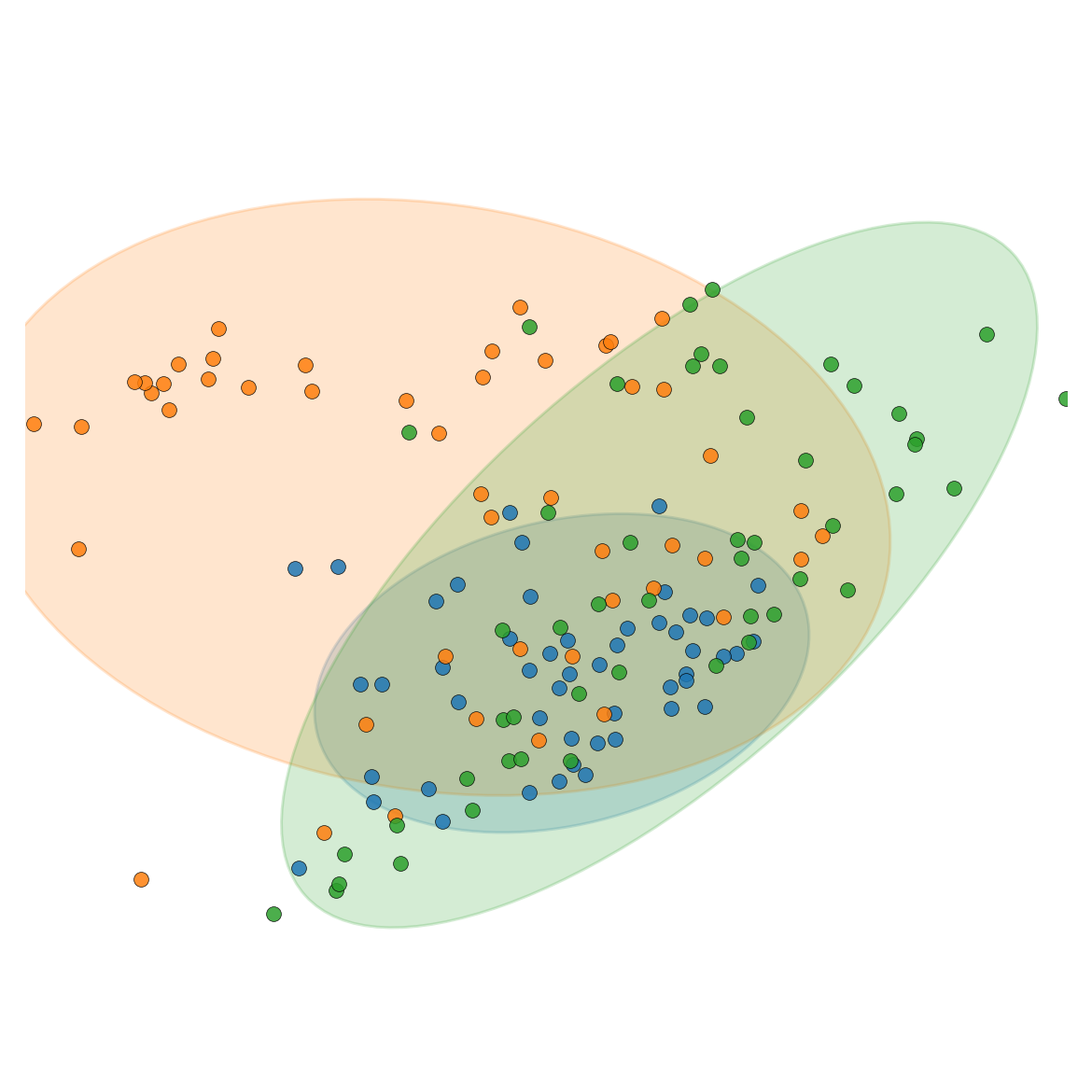} &
        \includegraphics[width=0.24\columnwidth]{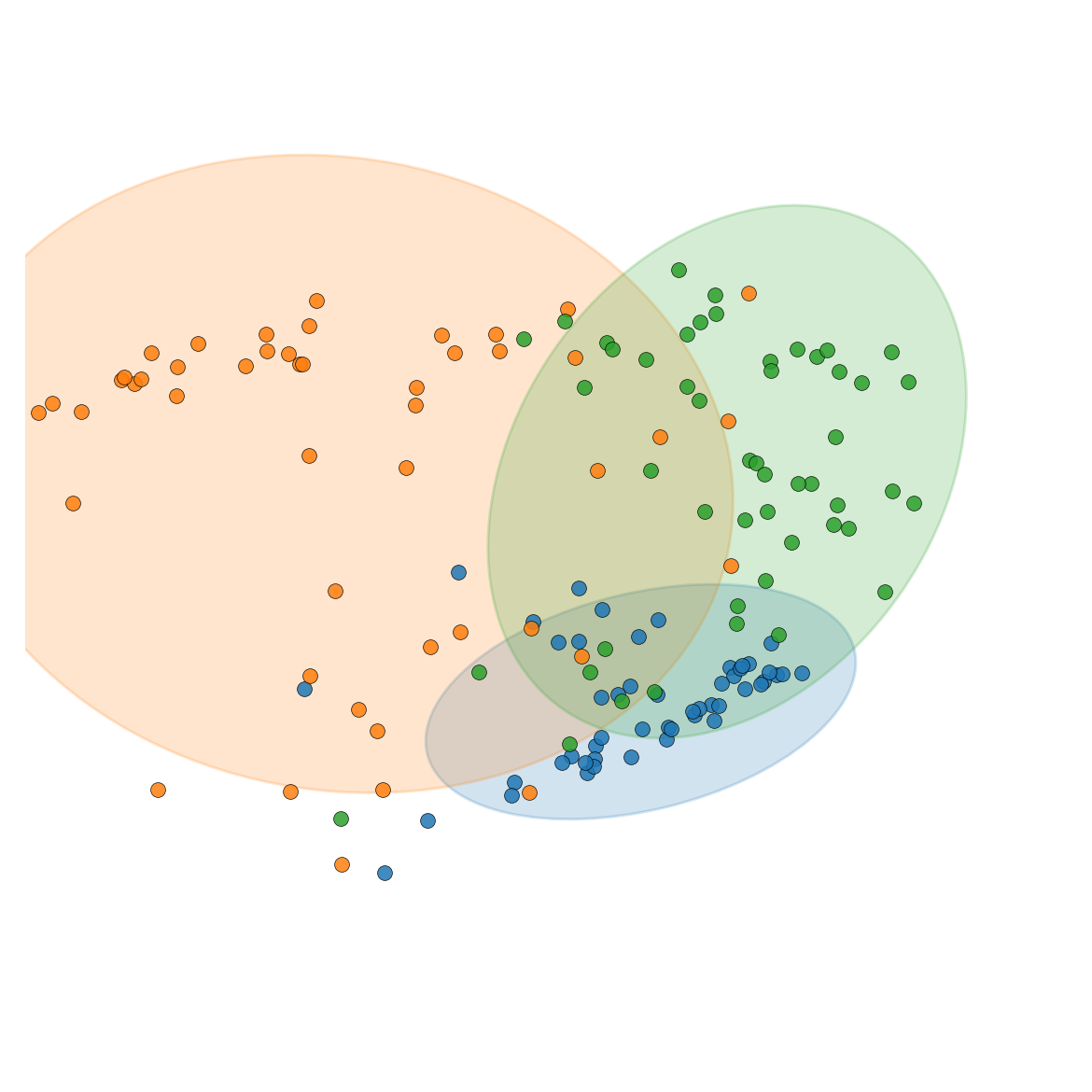} \\
        {\scriptsize \emph{Metal}} &
        {\scriptsize \emph{Paper}} &
        {\scriptsize \emph{Plastic}} &
        {\scriptsize \emph{Stone}} \\
    \end{tabular}
     {\centering
     \resizebox{0.25\textwidth}{!}{%
     \begin{tabular}{cc}
                \includegraphics[width=7cm]{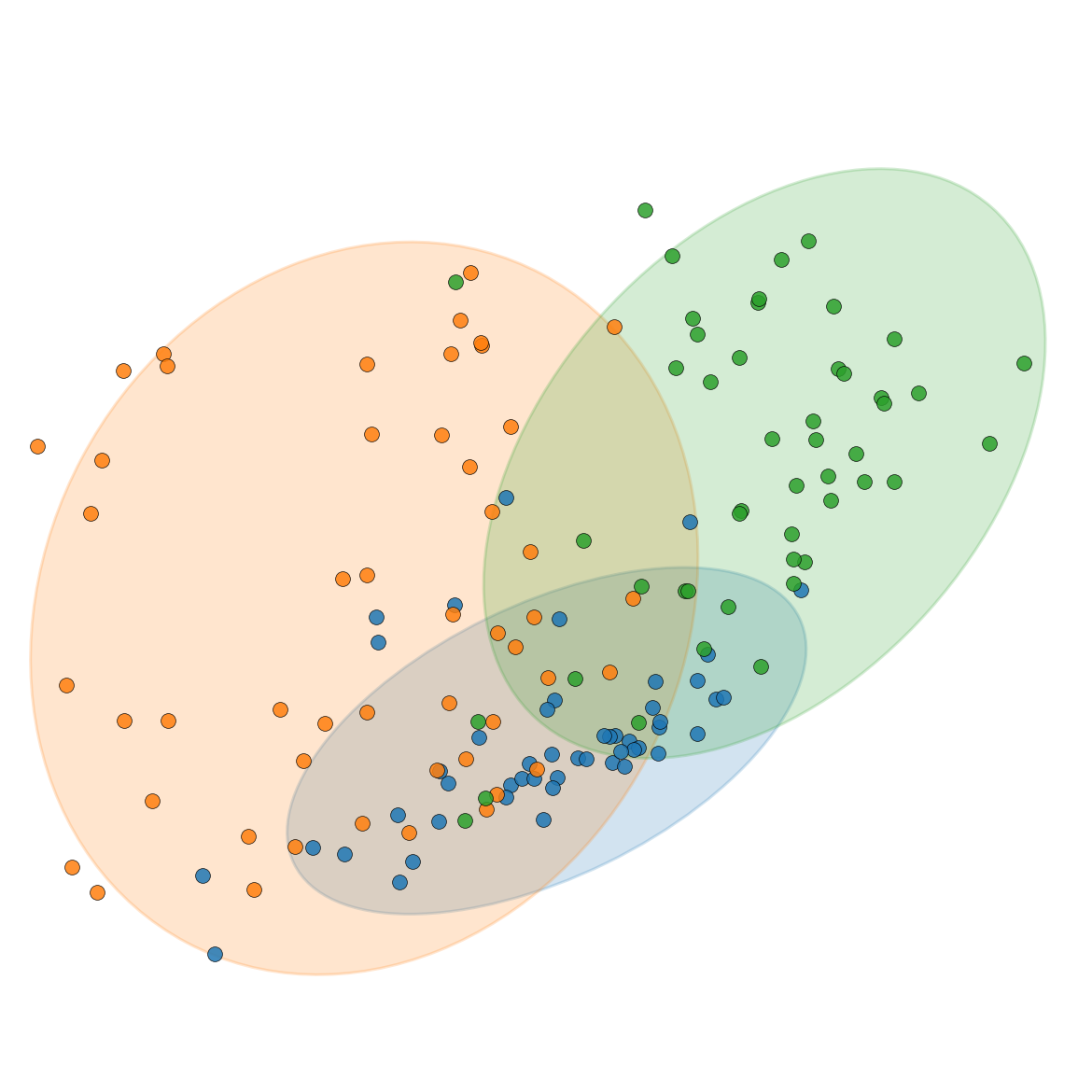} 
                    & \includegraphics[width=7cm]{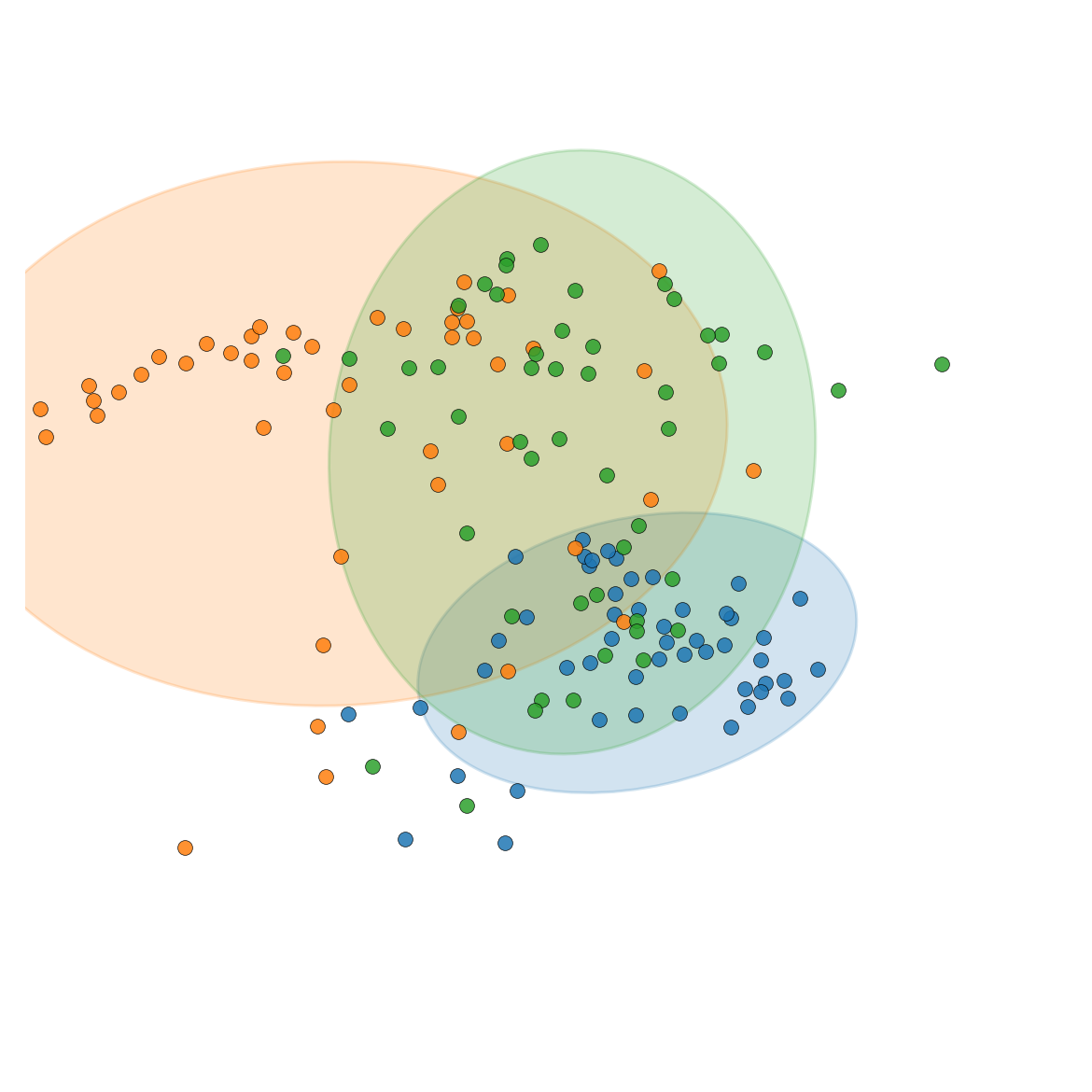} \\

                \fontsize{26}{26}\selectfont{\emph{Water}} &  \fontsize{26}{26}\selectfont{\emph{Wood}} \\

                 \end{tabular}
                 }
                \par}
    \caption{ PCA overlays of max-pooled DINOv2 patch features from 10 material classes across \textcolor{ds1blue}{FMD}, \textcolor{ds2orange}{DMS}, and \textcolor{ds3green}{Ours} datasets. Axes are fixed to $[-40,40]$ on both dimensions. Each dataset is shown with scatter points and its covariance ellipse (two standard deviations), highlighting intra-class variation.}
    \label{fig:pca_grid_singlecol}
\end{figure}

\subsection{Ablation test}
\label{sec:ablation}

\paragraph{Ablation study of dual modules} We evaluate the contributions of the language and vision streams in our network. To achieve this, we trained the vision stream (DINOv2) and the language stream (GPT-4v + CLIP) separately on our generated dataset and assessed their performance on three test datasets. As shown in Table~\ref{tab:table4}, our proposed dual-stream method achieved the best performance on all test datasets in both mIoU and mAcc. This result validates the effectiveness of fusing language and vision priors in our approach.
      
     

\begin{table}[h!]
    \begin{center}
    \caption{Influence of language (GPT-4v+CLIP) and vision (DINOv2) priors to the performance of material classification(mIoU\textbar mAcc). Best in \textbf{bold}.}
    \label{tab:table4}
    \resizebox{\columnwidth}{!}{%
    \begin{tabular}{cc|ccc}
      
       GPT-4v+CLIP& DINOv2&  Google-test& DMS-test &FMD~\cite{Sharan-JoV-14}\\
       \hline
       \checkmark& $\times$&0.83\texttt{\textbar}0.90 & 0.49\texttt{\textbar}0.64 &0.74\texttt{\textbar}0.85\\
       $\times$&\checkmark& 0.81\texttt{\textbar}0.89 & 0.46\texttt{\textbar}0.60 &0.79\texttt{\textbar}0.88\\
       \checkmark& \checkmark&\textbf{0.86\texttt{\textbar}0.92}& \textbf{0.50\texttt{\textbar}0.64}&\textbf{0.81\texttt{\textbar}0.89}\\
     
    \end{tabular}
    }
    \vspace{-8mm} 
     \end{center}
\end{table}

\paragraph{Compare across different vision backbones}
To analyze the impact of model capacity on the performance of the vision stream, we explored different vision backbone architectures by replacing our DINOv2 component with Vision Transformer (ViT) \cite{dosovitskiy2020image} and ResNet \cite{he2016deep}. All models are trained on our generated dataset and evaluated on FMD~\cite{Sharan-JoV-14}. We also investigated whether to release all parameters (full) or only the head parameters (head) for training. Table~\ref{tab:table5} shows that when training in ``full'' mode, the model with a ResNet backbone achieved the highest performance (0.44mIoU\textbar0.61mAcc) among the tested models, while the model based on a DINOv2 backbone obtained the lowest score (0.23mIoU\textbar0.38mAcc). This suggests that models with lighter parameter capacity perform better when trained in full mode. When training in ``head" mode (i.e. only fine-tuning the MLP layer), our vision stream approach (DINOv2-head) achieved the best performance (0.79mIoU\textbar0.88mAcc) among all tested models. This is likely due to the powerful pre-trained image features produced by DINOv2. In contrast, ResNet-head, which has the smallest backbone capacity, had the worst performance (0.48mIoU\textbar0.64mAcc) among the models trained in ``head" mode. In summary, we found that the DINOv2-head configuration performed the best and thus selected it as our vision backbone.
\begin{table}[h!]
    \caption{Comparison of material classification performance across different vision backbones(mIoU\textbar mAcc). Best in \textbf{bold}.}
    \label{tab:table5}
    \resizebox{\columnwidth}{!}{%
    \begin{tabular}{l|c|c|c}
      \diagbox{Train method}{Vision backbone}&  ResNet-101 &  ViT-L/16 & DINOv2\\
      \hline
       head & 0.48\texttt{\textbar}0.64& 0.50\texttt{\textbar}0.66& \textbf{0.79\texttt{\textbar}0.88}\\
       full &  \textbf{0.44\texttt{\textbar}0.61} & 0.34\texttt{\textbar}0.51& 0.23\texttt{\textbar}0.38\\
     
    \end{tabular}
     \vspace{-8mm} 
    }
\end{table}

\begin{figure}[ht]
     \centering
     \includegraphics[width=0.4\textwidth]{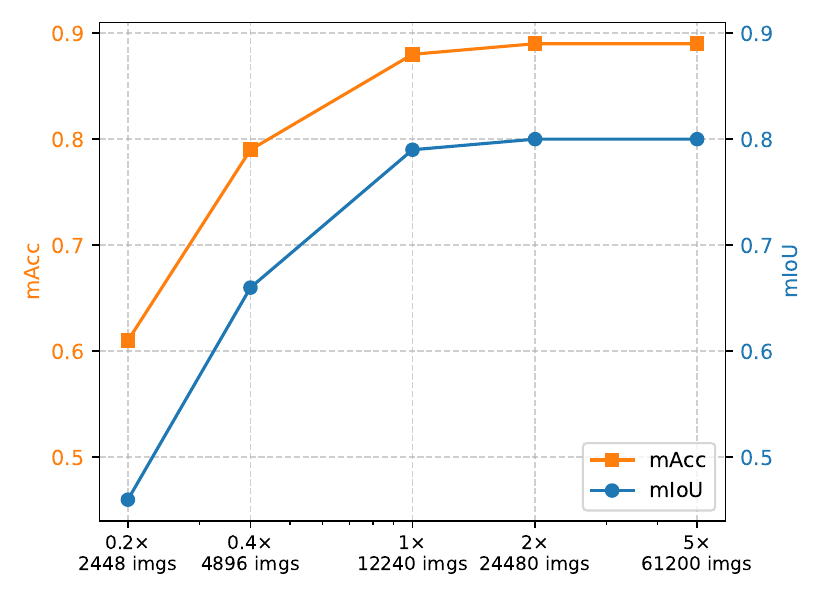}

    \captionof{figure}{Model performance in relation to the number of images.}
    \label{fig:num_miou_figure}
    \vspace{-4mm} 
\end{figure}

\paragraph{The influence of dataset scale} To explore how dataset scale affects model performance, we trained our network (using only the vision stream for simplicity) on datasets with varying numbers of images, and evaluated it on the FMD ~\cite{Sharan-JoV-14} dataset. The results are shown in Figure~\ref{fig:num_miou_figure}.
The baseline (1x) corresponds to a dataset generated from 2448 prompts distributed across the 10 classes, with each prompt producing 5 images, resulting in a total of 12,240 images. As shown in Figure~\ref{fig:num_miou_figure}, both mIoU and mAcc increase rapidly when scaling from 0.2x to 1x, confirming the benefit of enlarging the training set in the low-data regime. Although performance gains plateau beyond 2x (0.80mIoU\textbar0.89mAcc), this trend indicates that increasing the dataset scale consistently enhances performance, highlighting the promise of our generation pipeline for constructing larger datasets tailored to more complex material recognition tasks.

\paragraph{Comparison with/without auto labeled semantics}We investigated the effect of auto-labeled semantics by comparing our approach against a variant without semantic labels, in which all image patches were directly max-pooled during training. To this end, we constructed a dataset with empty masks, trained the vision branch on it, and evaluated performance on FMD~\cite{Sharan-JoV-14}. The auto-labeled model achieved (0.79mIoU\textbar0.88mAcc), compared to (0.75mIoU\textbar0.85mAcc) without semantic labels, corresponding to improvements of 4\% in mIoU and 3\% in mAcc.

        
        

\label{sec:experiment}
\section{Conclusion}
We introduce a novel framework for material classification that addresses limited annotated data by leveraging vision-language models and an auto-labeling pipeline. Our method enhances accuracy through joint fine-tuning and achieves superior performance across datasets. This work highlights the potential of synthetic data and VLM-guided priors for scalable material recognition.
{
    \small
    \bibliographystyle{ieeenat_fullname}
    \bibliography{main}
}

\clearpage
\setcounter{page}{1}
\maketitlesupplementary

\section{Class-wise Accuracy on DMS-test}
\label{sec:DMS}
\begin{figure*}[ht]
    \resizebox{\textwidth}{!}{%
 \begin{tabular}
                {cccccccc}

                    \includegraphics[width=7cm]{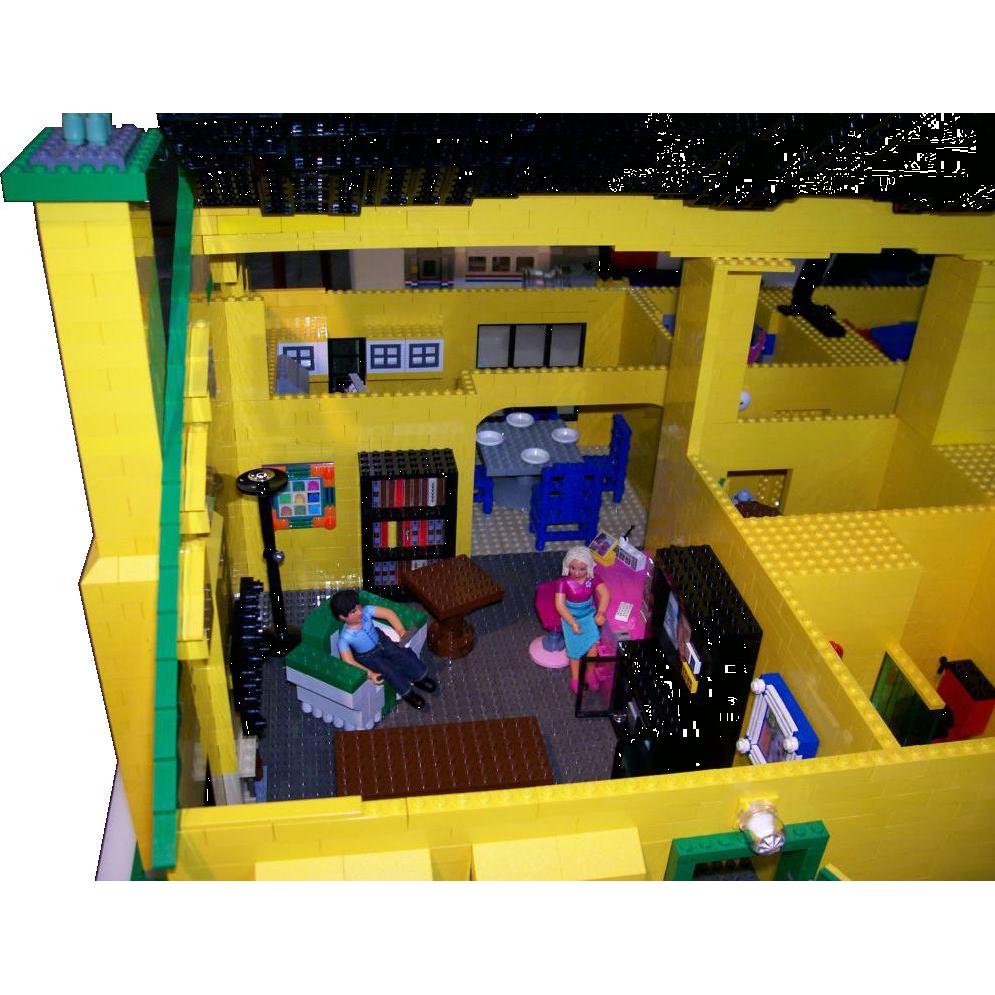} 
                    & \includegraphics[width=7cm]{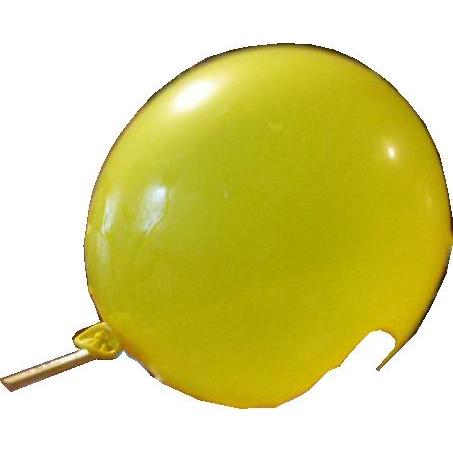} 
                    & \includegraphics[width=7cm]{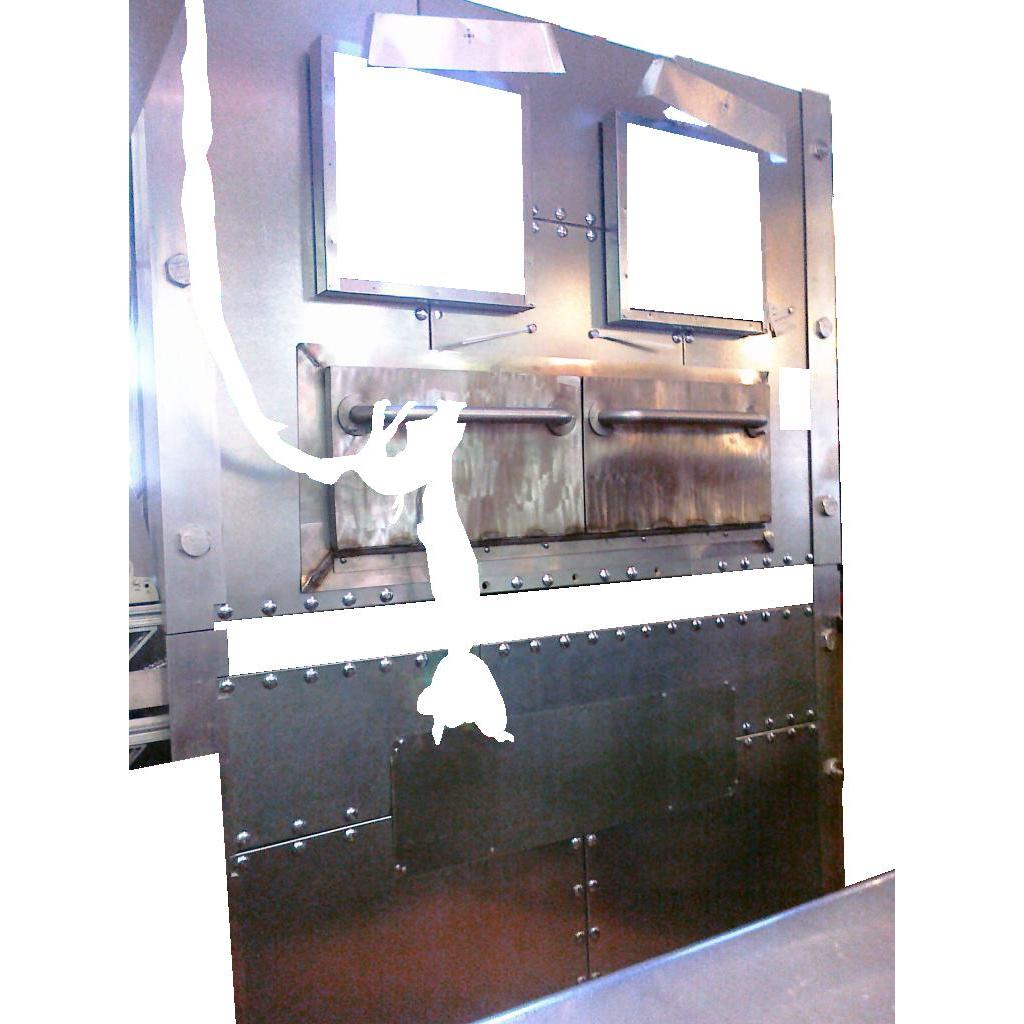} 
                    & \includegraphics[width=7cm]{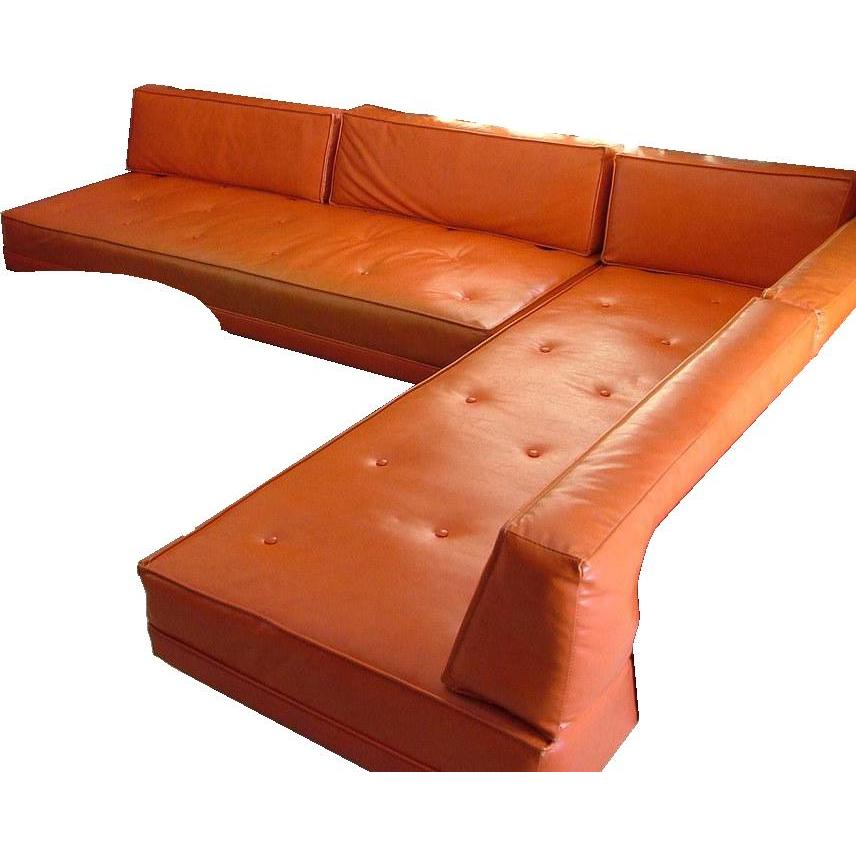} 
                    & \includegraphics[width=7cm]{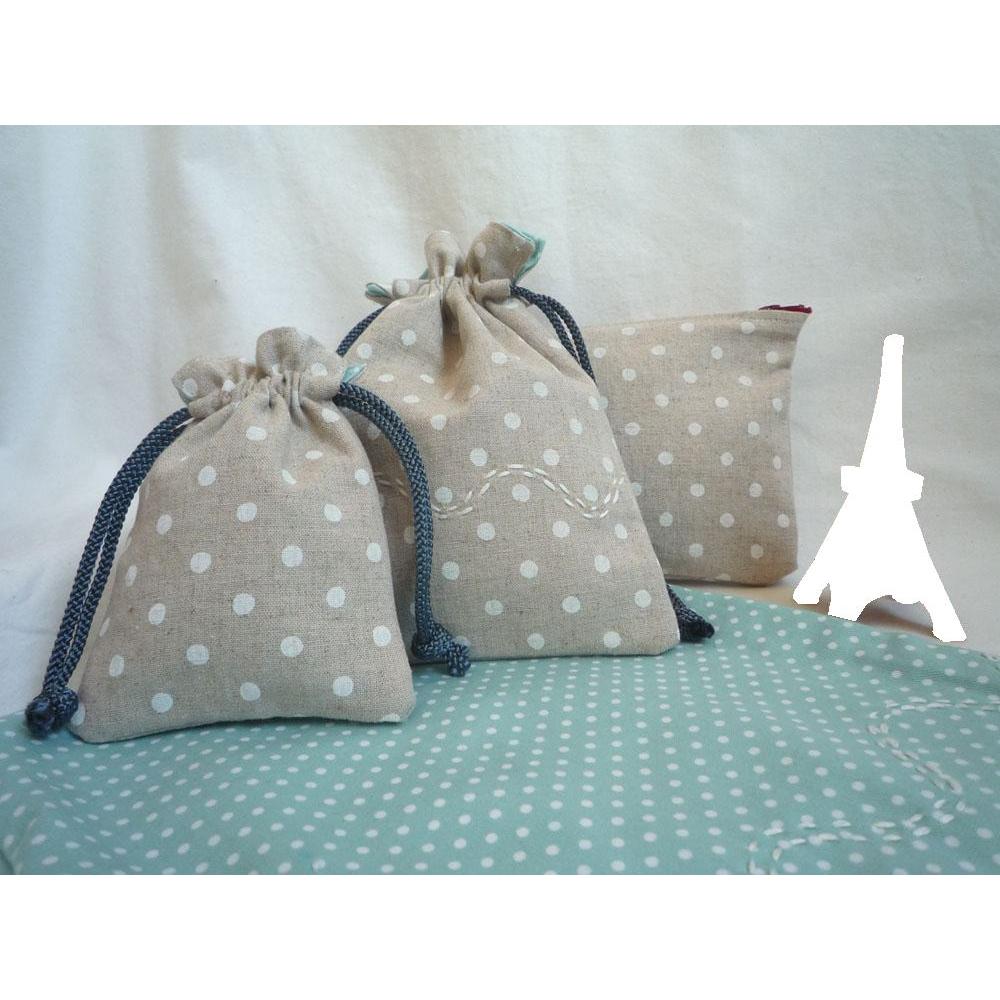} 
                    & \includegraphics[width=7cm]{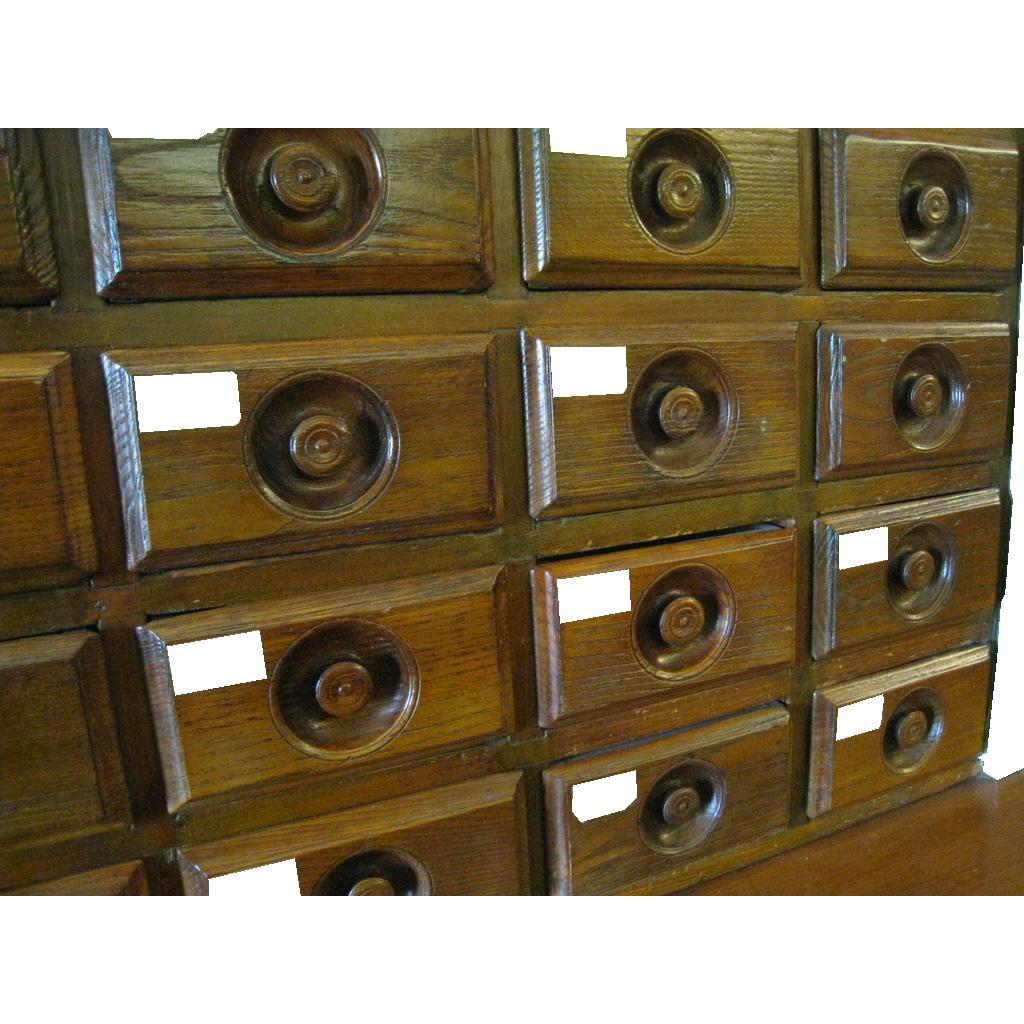} 
                    & \includegraphics[width=7cm]{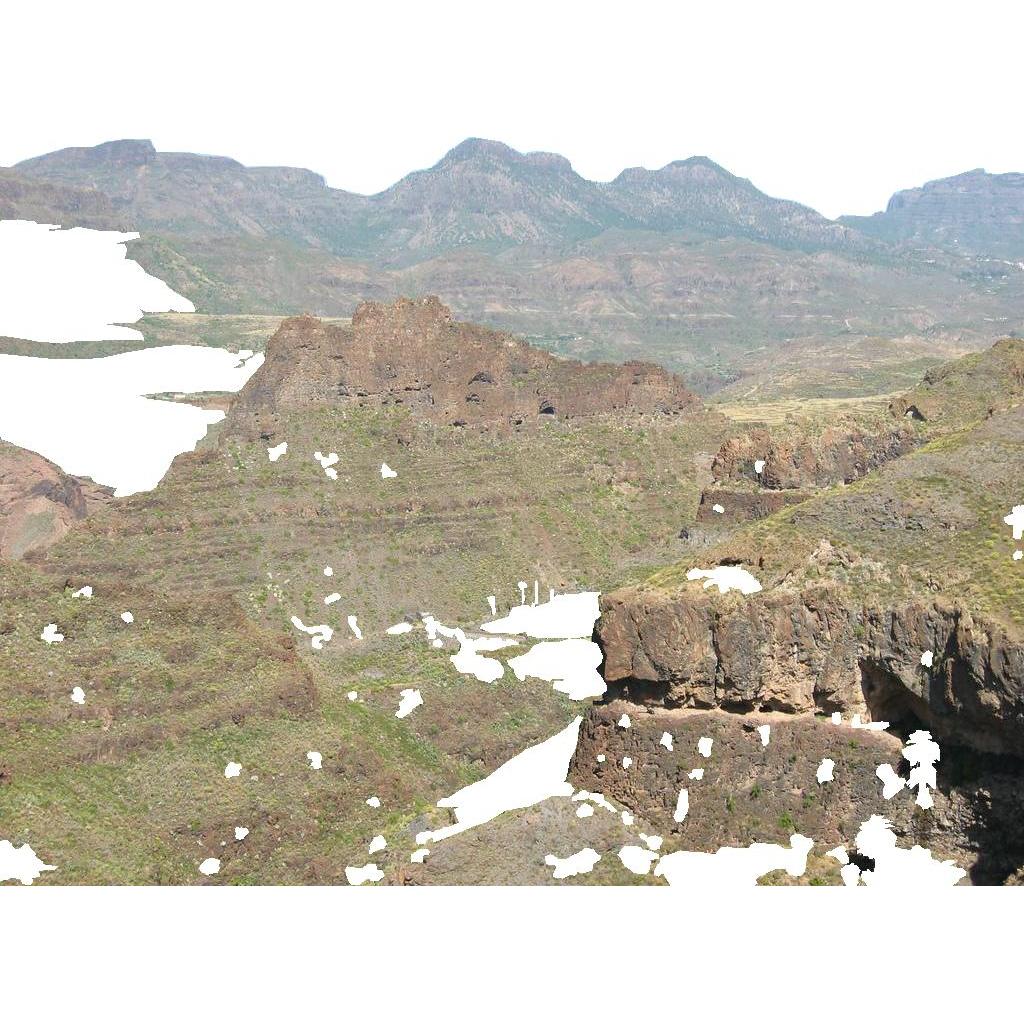} 
                    & \includegraphics[width=7cm]{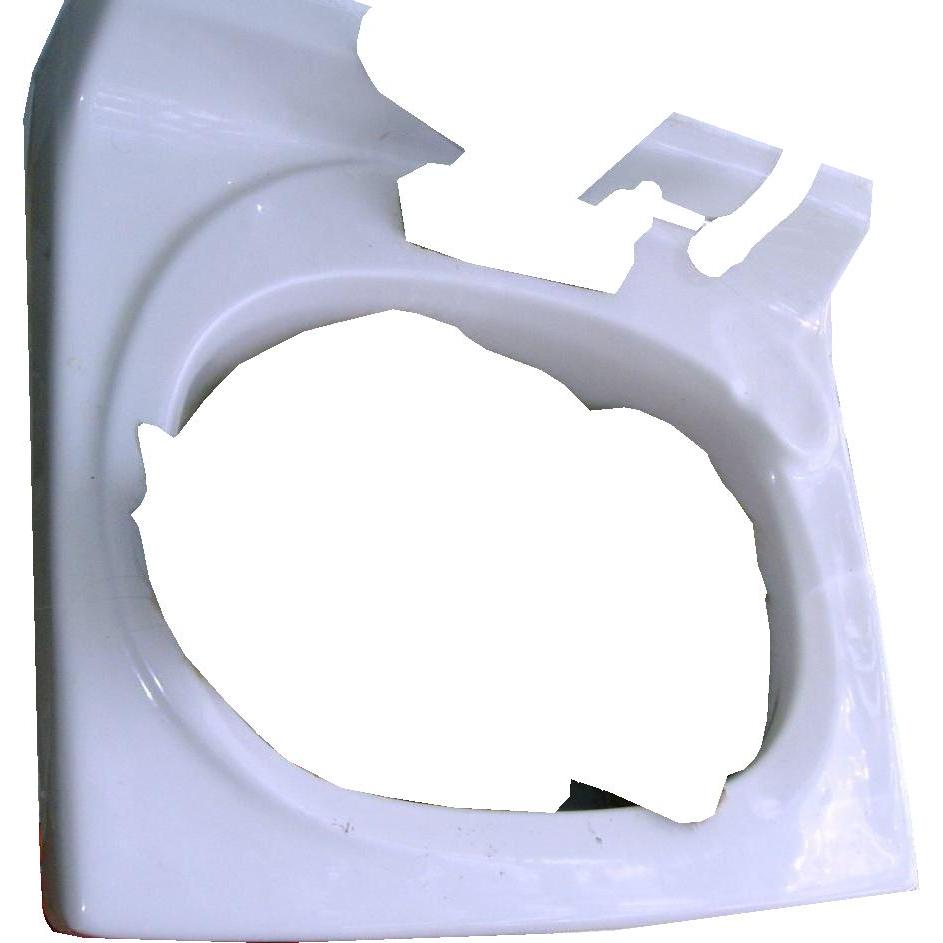} \\

                  \fontsize{26}{26}\selectfont{\emph{Plastic}} &  \fontsize{26}{26}\selectfont{\emph{Rubber}} &  \fontsize{26}{26}\selectfont{\emph{Metal}}& \fontsize{26}{26}\selectfont{\emph{Leather}} & \fontsize{26}{26}\selectfont{\emph{Fabric}} & \fontsize{26}{26}\selectfont{\emph{Wood}} & 
                \fontsize{26}{26}\selectfont{\emph{Stone}} & \fontsize{26}{26}\selectfont{\emph{Ceramic}}\\

                  \includegraphics[width=7cm]{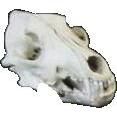} 
                 & \includegraphics[width=7cm]{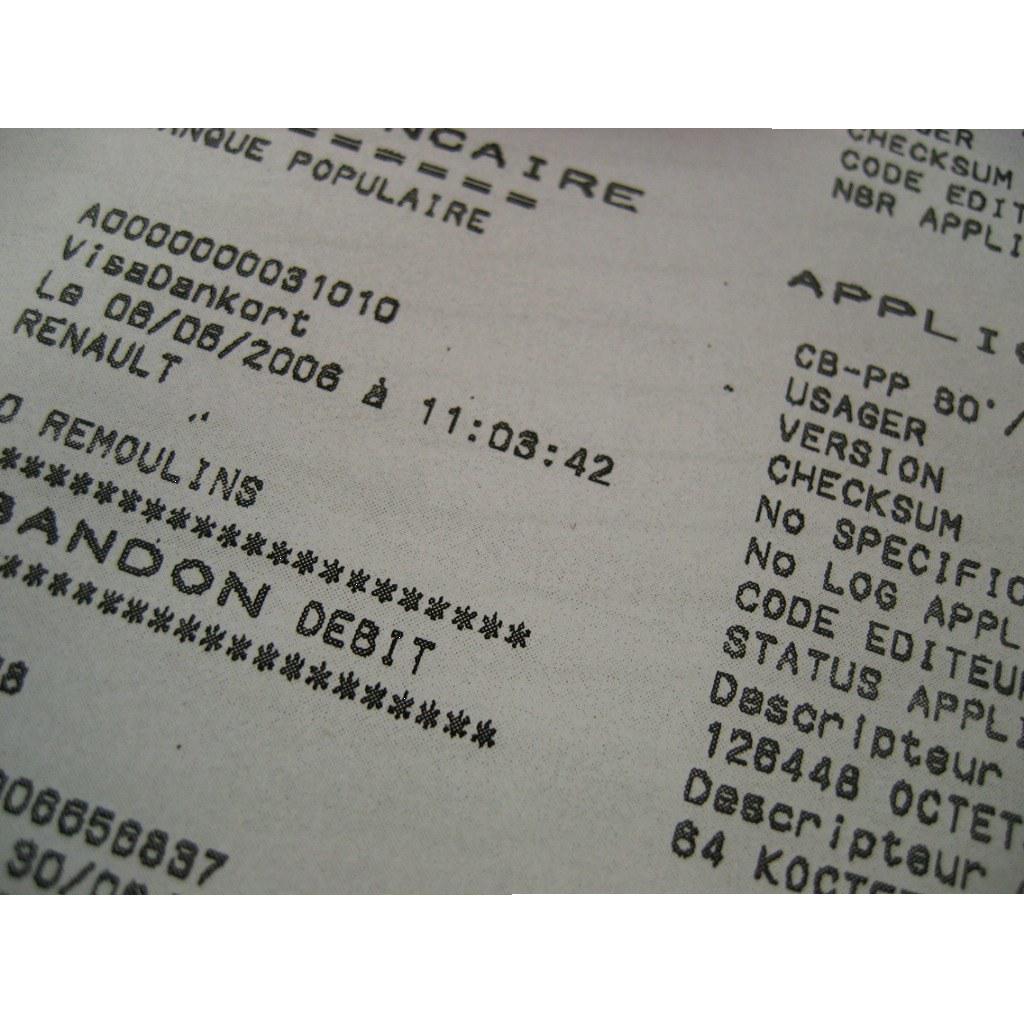} 
                    & \includegraphics[width=7cm]{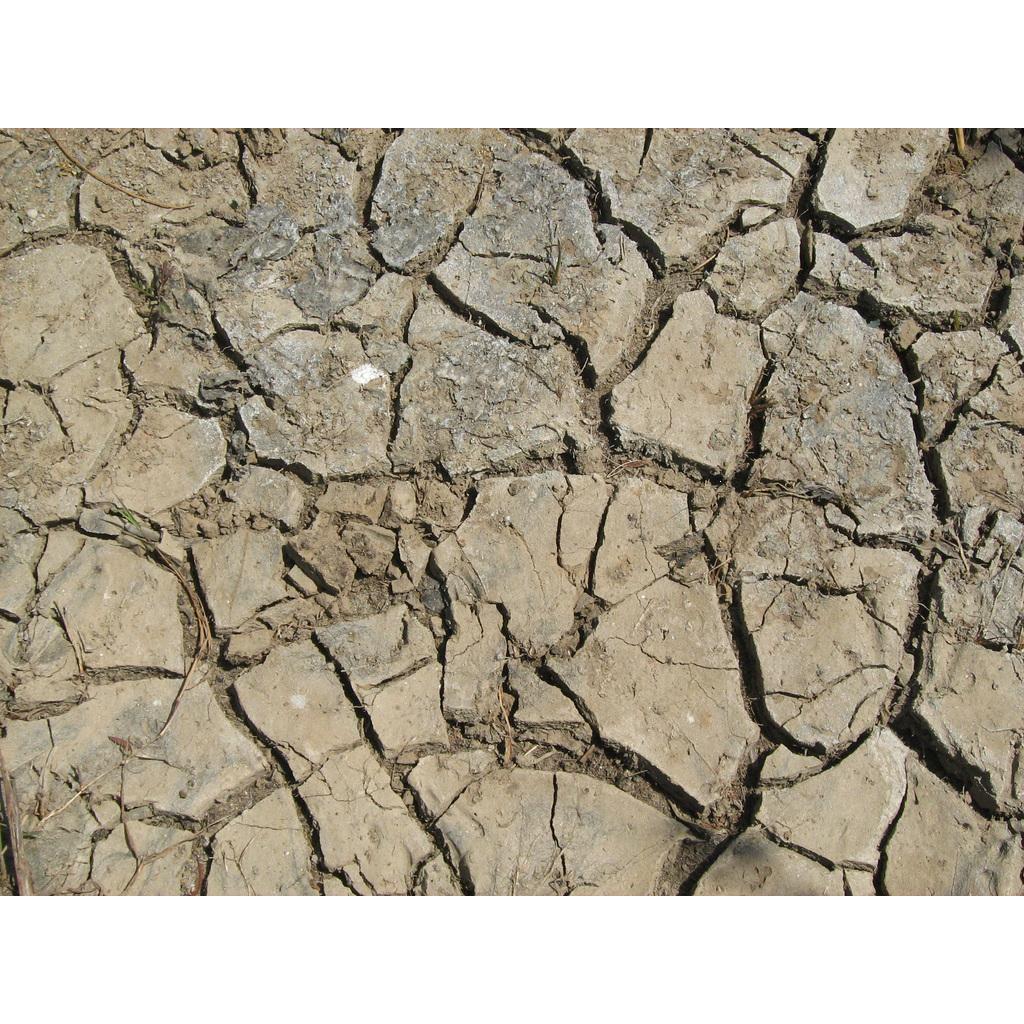} 
                    & \includegraphics[width=7cm]{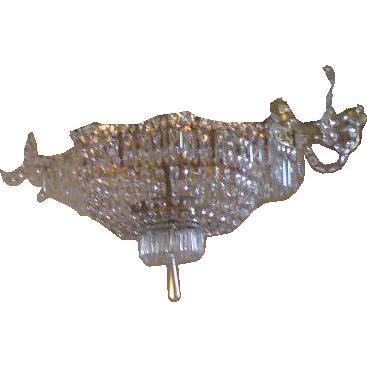} 
                    & \includegraphics[width=7cm]{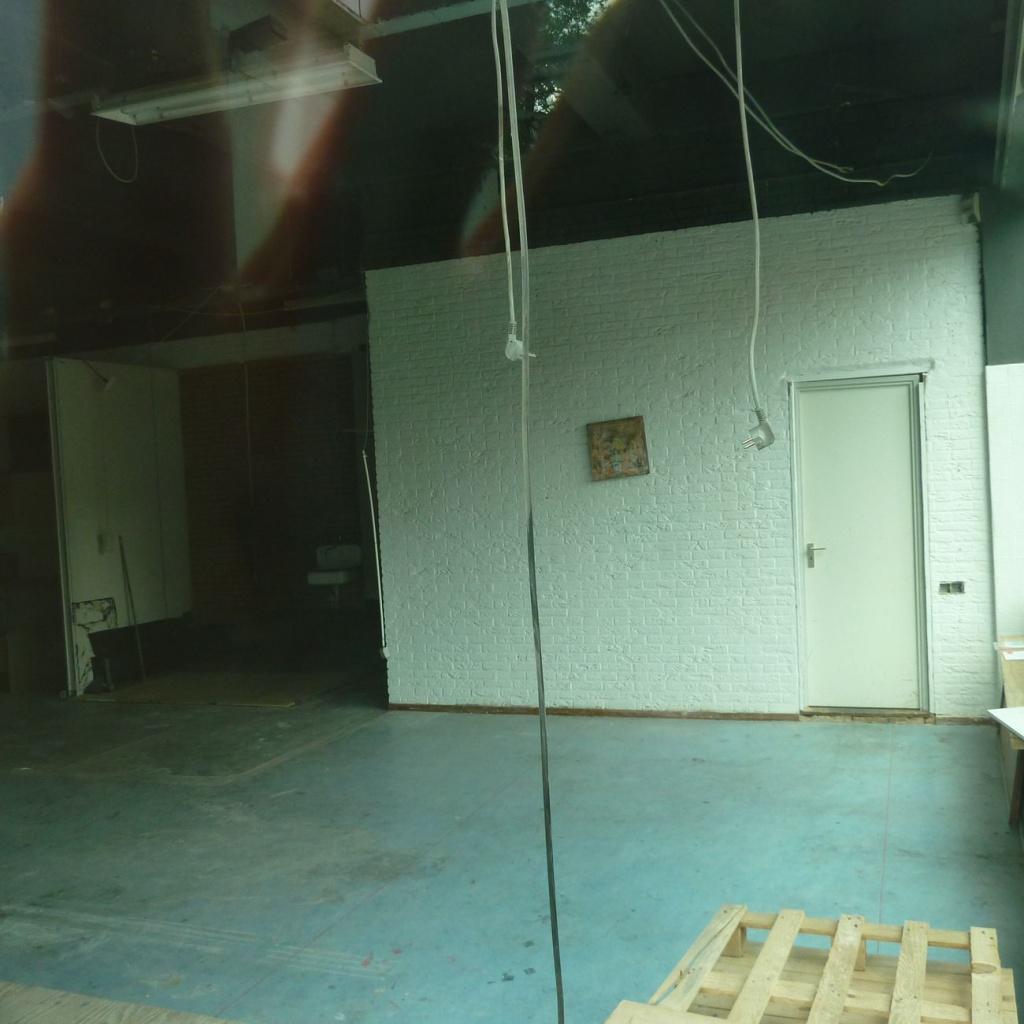} 
                    & \includegraphics[width=7cm]{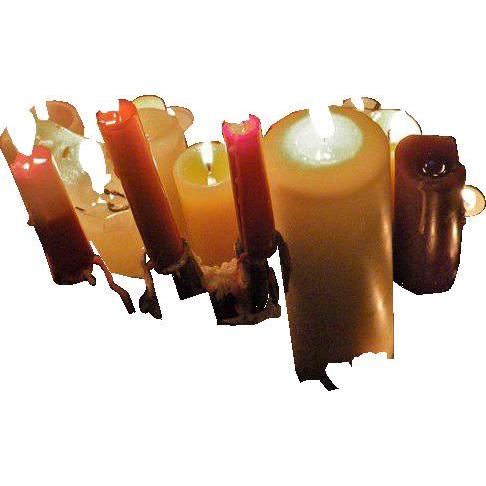} 
                    & \includegraphics[width=7cm]{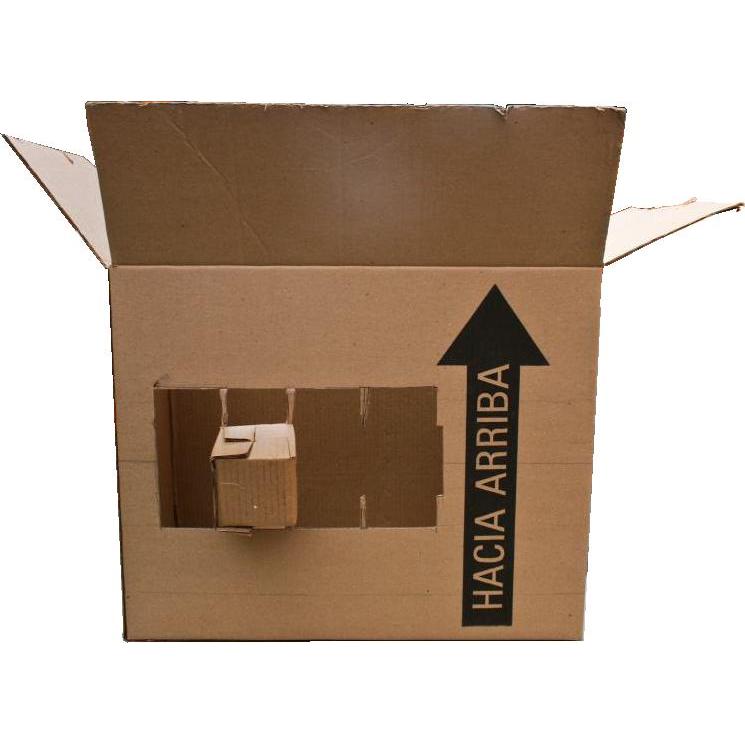} 
                    & \includegraphics[width=7cm]{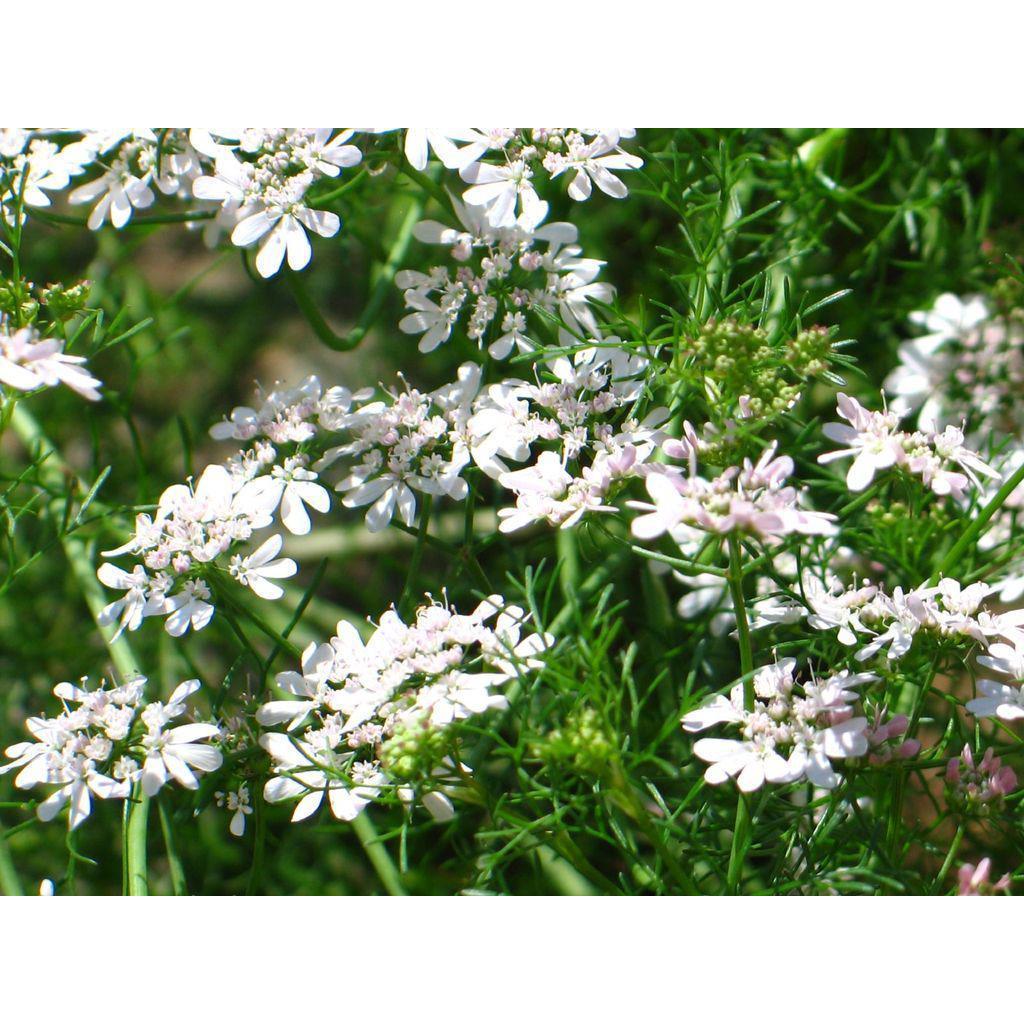} \\ 

                 \fontsize{26}{26}\selectfont{\emph{Bone}} &  \fontsize{26}{26}\selectfont{\emph{Paper}} &  \fontsize{26}{26}\selectfont{\emph{Soil}}& \fontsize{26}{26}\selectfont{\emph{Gemstone}} & \fontsize{26}{26}\selectfont{\emph{Glass}} & \fontsize{26}{26}\selectfont{\emph{Wax}} & 
                \fontsize{26}{26}\selectfont{\emph{Cardboard}} & \fontsize{26}{26}\selectfont{\emph{Foliage}}\\

                \end{tabular}

            }
               {\centering
                 \resizebox{0.625\textwidth}{!}{%
                \begin{tabular}
                {ccccc}
                \includegraphics[width=7cm]{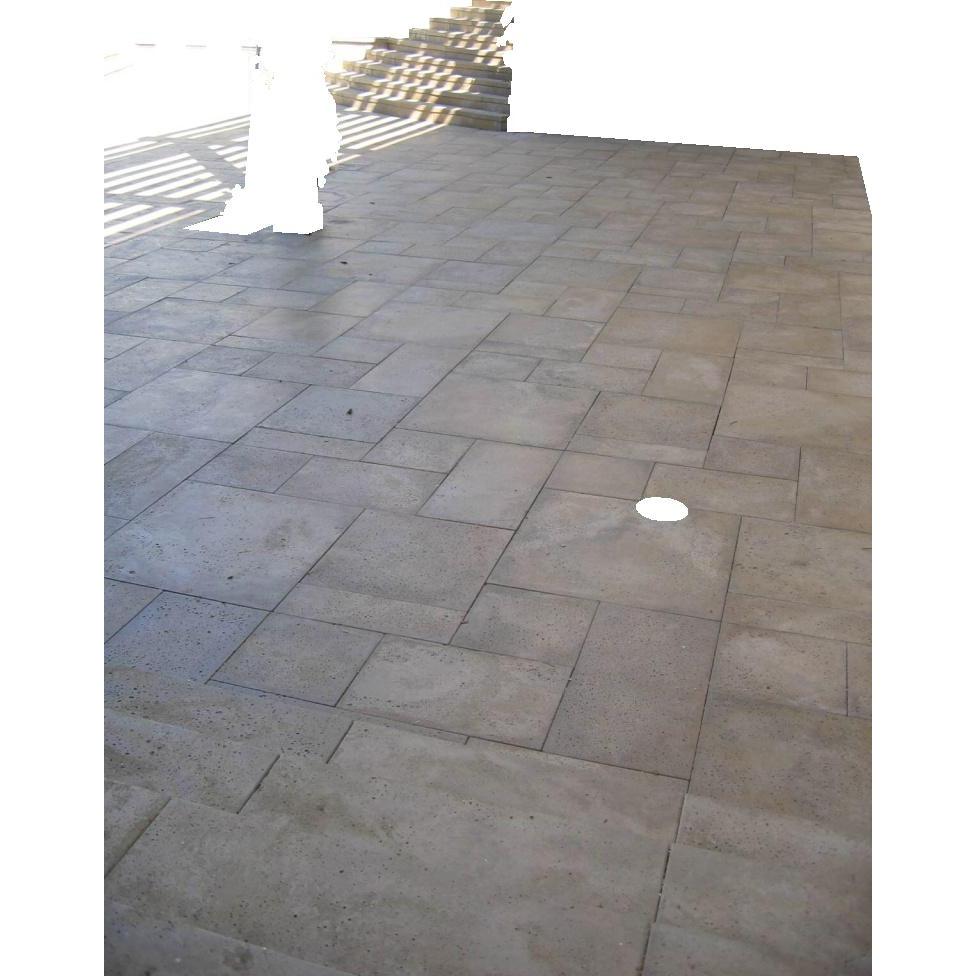} 
                    & \includegraphics[width=7cm]{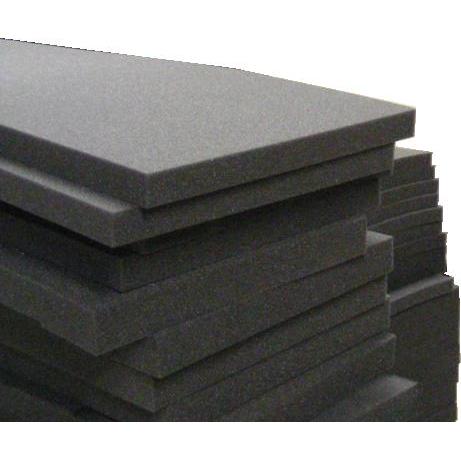} 
                    & \includegraphics[width=7cm]{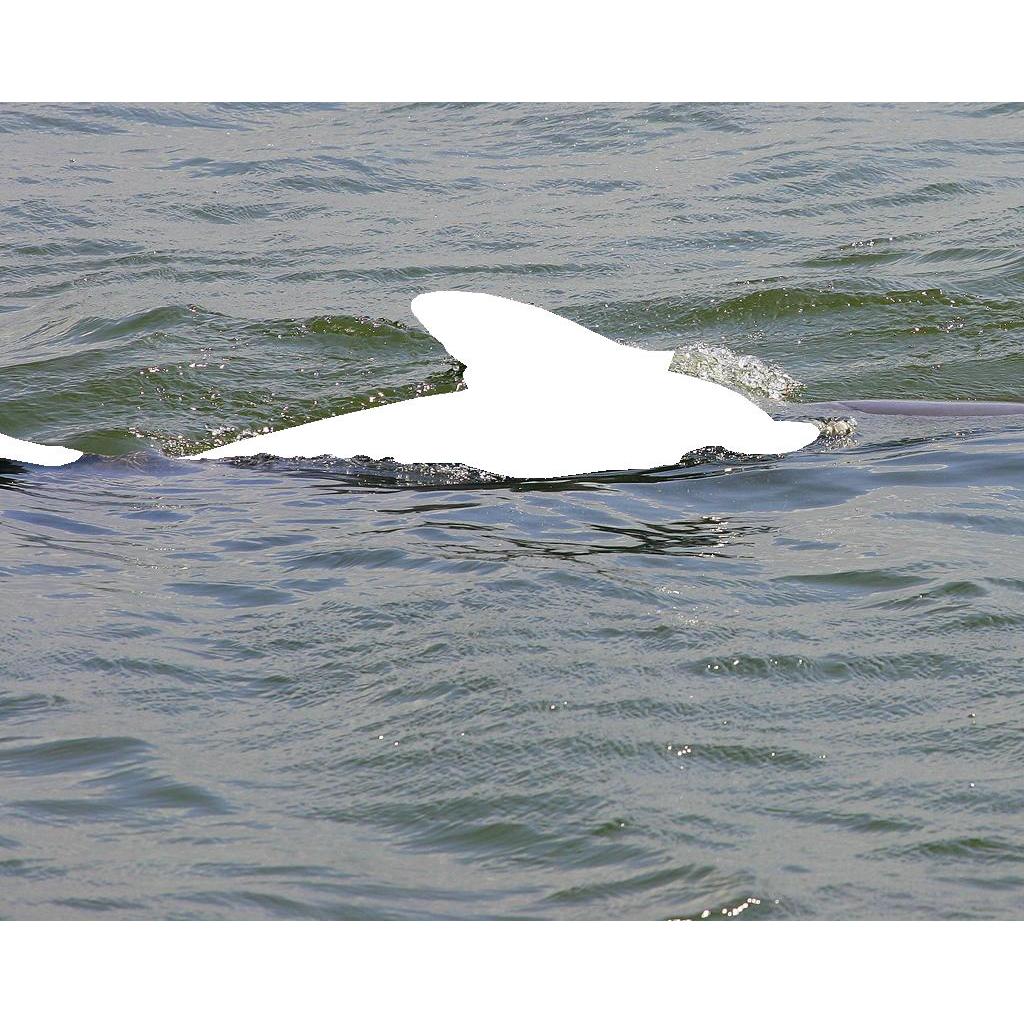} 
                    & \includegraphics[width=7cm]{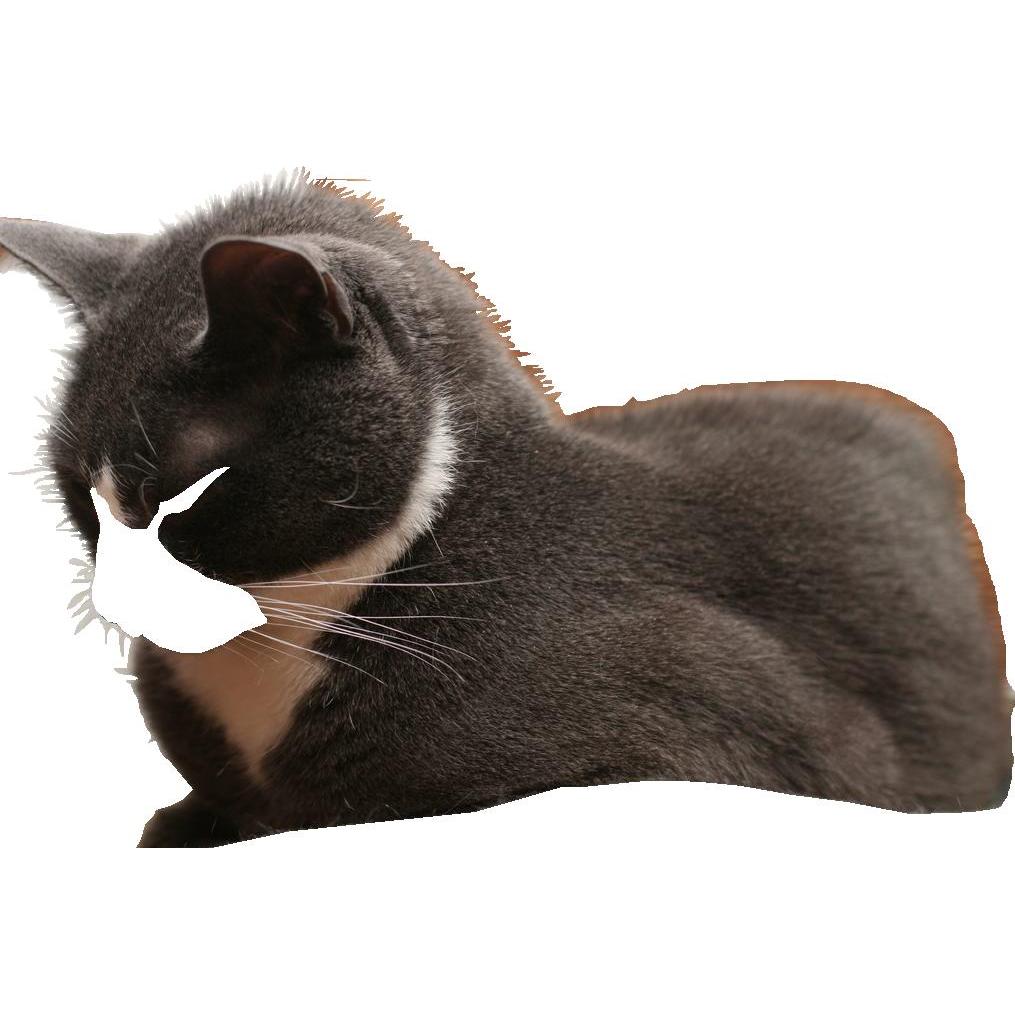} 
                    & \includegraphics[width=7cm]{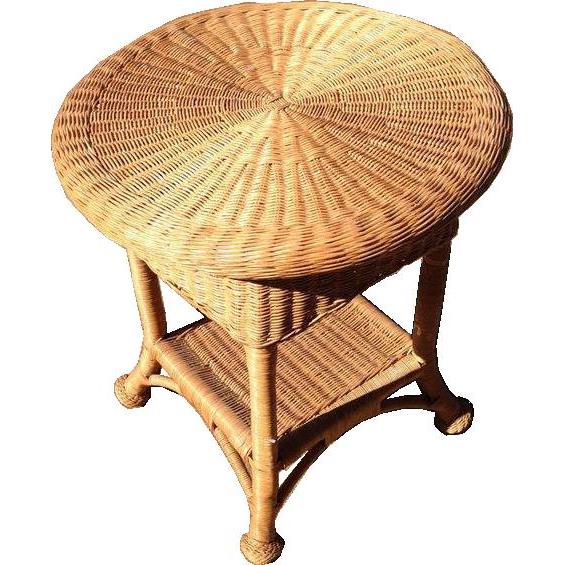} \\

                \fontsize{26}{26}\selectfont{\emph{Concrete}} &  \fontsize{26}{26}\selectfont{\emph{Sponge}} &  \fontsize{26}{26}\selectfont{\emph{Water}}& \fontsize{26}{26}\selectfont{\emph{Fur}} & \fontsize{26}{26}\selectfont{\emph{Wicker}}\\
     
                 \end{tabular}
                 }
                \par}
                \captionof{figure}{Illustration of the DMS-test dataset. One representative sample is shown for each of the 21 categories.}
    \label{fig:figure7}
\vspace{-5mm}
\end{figure*}

Table ~\ref{tab:table6}.  reports the per-class classification accuracy on the DMS-test dataset. While Table ~\ref{tab:table2} in the main text only provides the averaged results, this table further breaks down the performance into individual classes.Figure ~\ref{fig:figure7} illustrates one representative sample from each of the 21 categories in the DMS-test dataset. The full test set consists of 1,050 images, constructed by retaining connected regions larger than 10,000 pixels in each category and randomly sampling 50 images per category.
\begin{table}[h!]
    \caption{Comparison with state-of-the-art methods on the DMS-test dataset (21 classes). All values denote classification accuracy(truncated to two decimal places), and the best result in each row is highlighted in \textbf{bold}.}
    \label{tab:table6}
    \resizebox{\columnwidth}{!}{%
    \begin{tabular}{l|c|c|c|c}
    \diagbox[width=6em]{Class}{Method}&CLIP\cite{radford2021learning} & GPT-4v \cite{achiam2023gpt}& MatSim\cite{drehwald2023one} & Ours\\
      \hline
      fabric   & 0.34         & 0.54          & 0.28             & \textbf{0.64} \\
      foliage  & 0.60         & 0.30          & 0.60             & \textbf{0.74} \\
      glass    & \textbf{0.52}& 0.24          & 0.16             & 0.40          \\
      leather  & 0.32         & 0.42          & 0.52             & \textbf{0.54} \\
      metal    & 0.32         & 0.58          & 0.18             & \textbf{0.82} \\
      paper    & 0.44         & 0.38          & 0.36             & \textbf{0.56} \\
      plastic  & 0.24         & 0.36          & 0.06             & \textbf{0.74} \\
      stone    & 0.36         & 0.38          & 0.20             & \textbf{0.52} \\
      water    & 0.26         & 0.48          & 0.38             & \textbf{0.90} \\
      wood     & 0.36         & \textbf{0.80} & 0.32             & 0.76          \\
      rubber   & 0.18         & \textbf{0.42} & 0.30             & 0.34          \\
      ceramic  & 0.02         & 0.30          & \textbf{0.40}    & 0.20          \\
      sponge   & 0.32         & 0.34          & 0.58             & \textbf{0.66} \\
      bone     & \textbf{0.86}& 0.44          & 0.58             & 0.70          \\
      cardboard& 0.72         & 0.68          & 0.42             & \textbf{0.78} \\
      concrete & 0.22         & 0.34          & 0.18             & \textbf{0.56} \\
      fur      & 0.46         & \textbf{0.80} & 0.58             & \textbf{0.80} \\
      gemstone & 0.14         & 0.04          & \textbf{0.66}    & 0.24          \\
      soil     & 0.18         & 0.30          & 0.50             & \textbf{0.60} \\
      wax      & 0.50         & 0.74          & 0.72             & \textbf{1.00} \\
      wicker   & 0.72         & 0.36          & 0.66             & \textbf{0.94} \\
      \hline
      {Average}& 0.38         & 0.43          & 0.41            & \textbf{0.64}  \\
    \end{tabular}
    }
     \vspace{-3mm} 
\end{table}

\section{Class-wise Accuracy on Google-test} 
\begin{figure*}[ht]
    \resizebox{\textwidth}{!}{%
 \begin{tabular}
                {cccccccc}

                    \includegraphics[width=7cm]{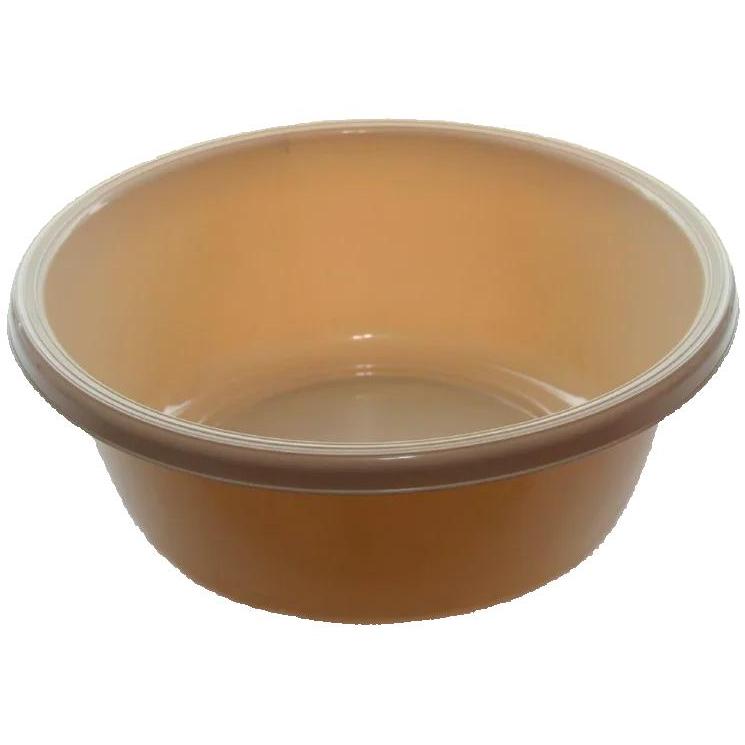} 
                    & \includegraphics[width=7cm]{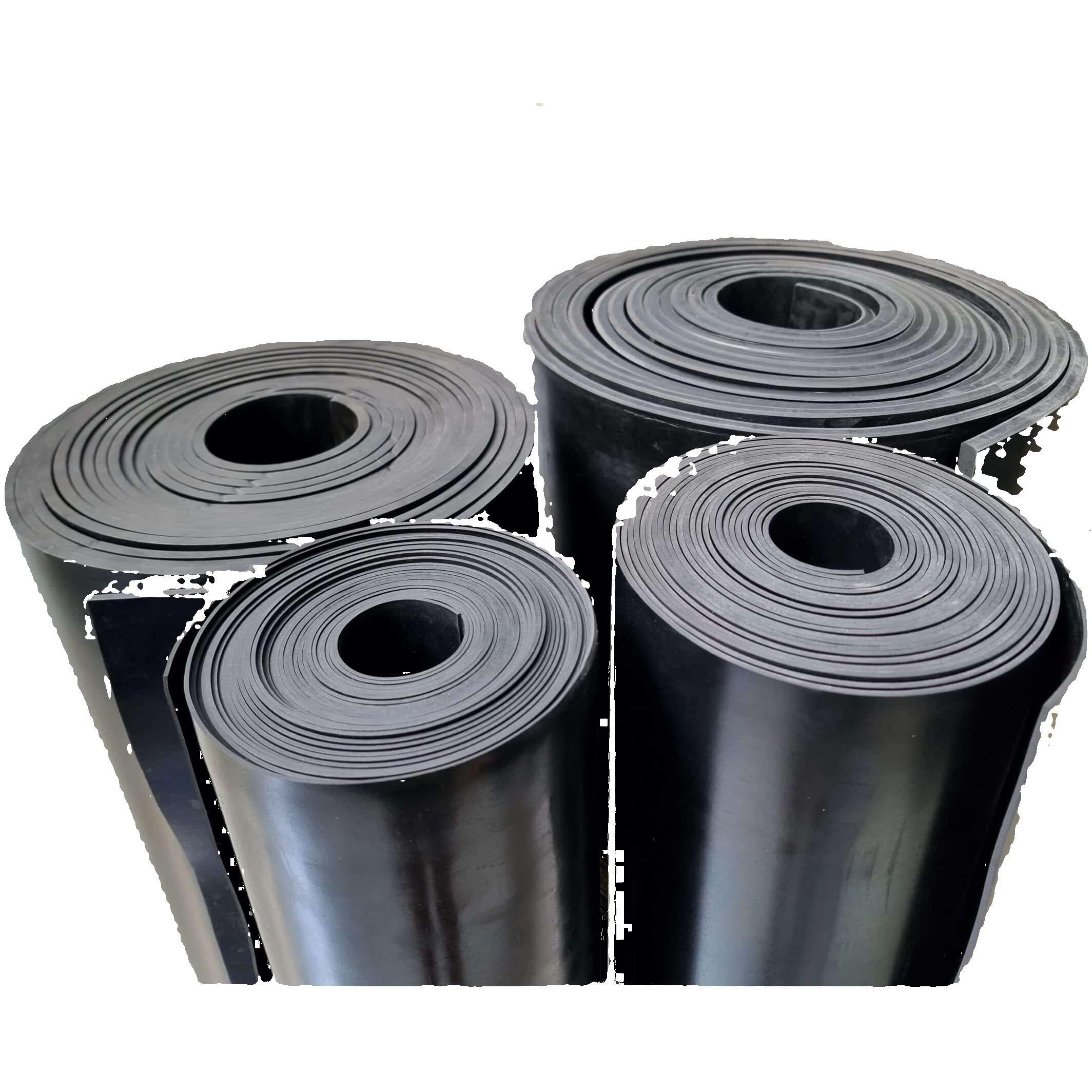} 
                    & \includegraphics[width=7cm]{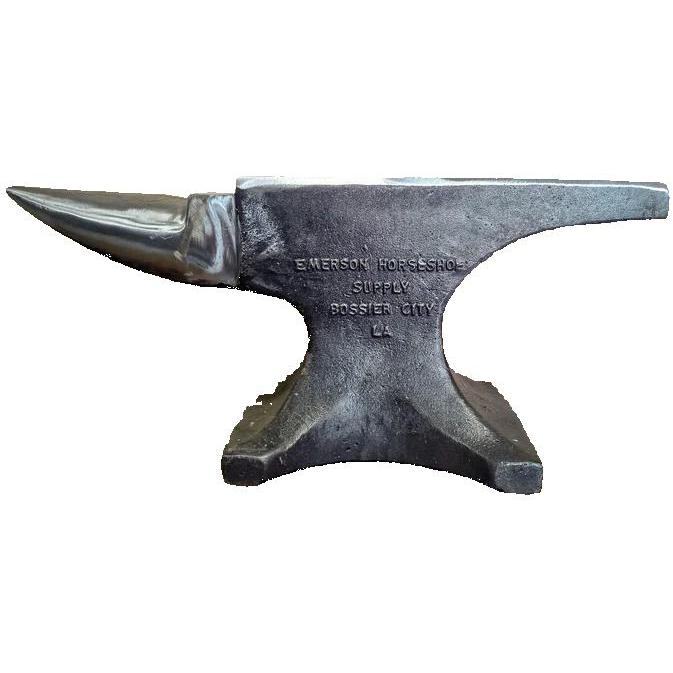} 
                    & \includegraphics[width=7cm]{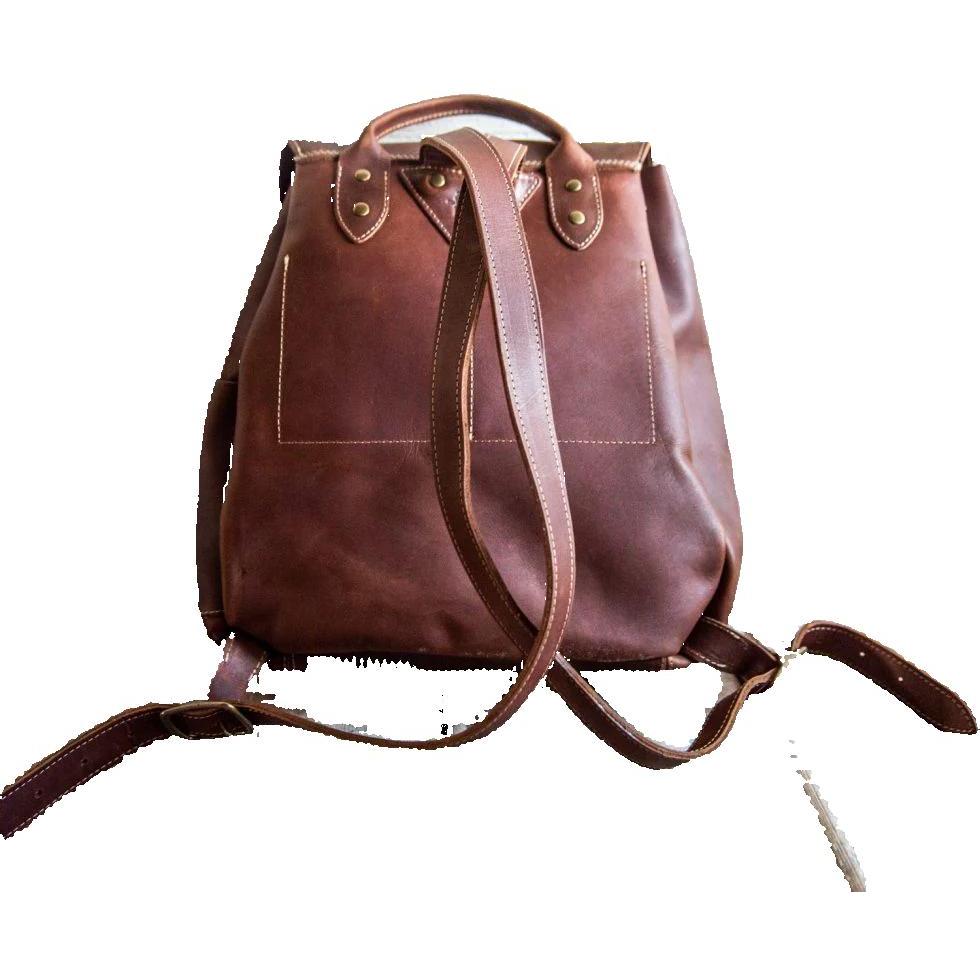} 
                    & \includegraphics[width=7cm]{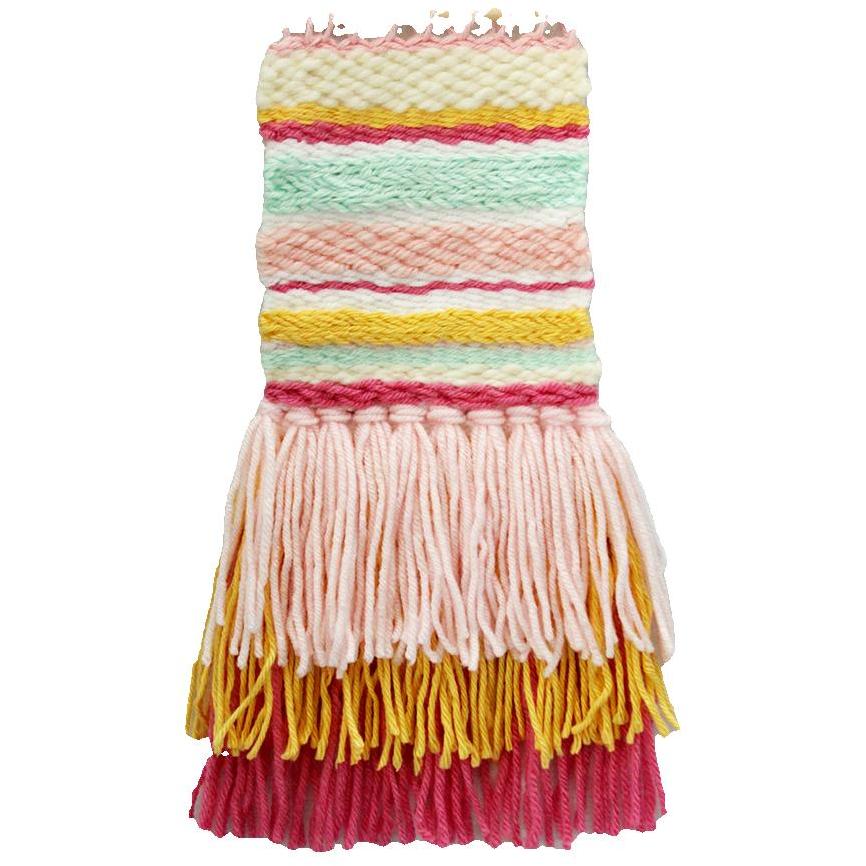} 
                    & \includegraphics[width=7cm]{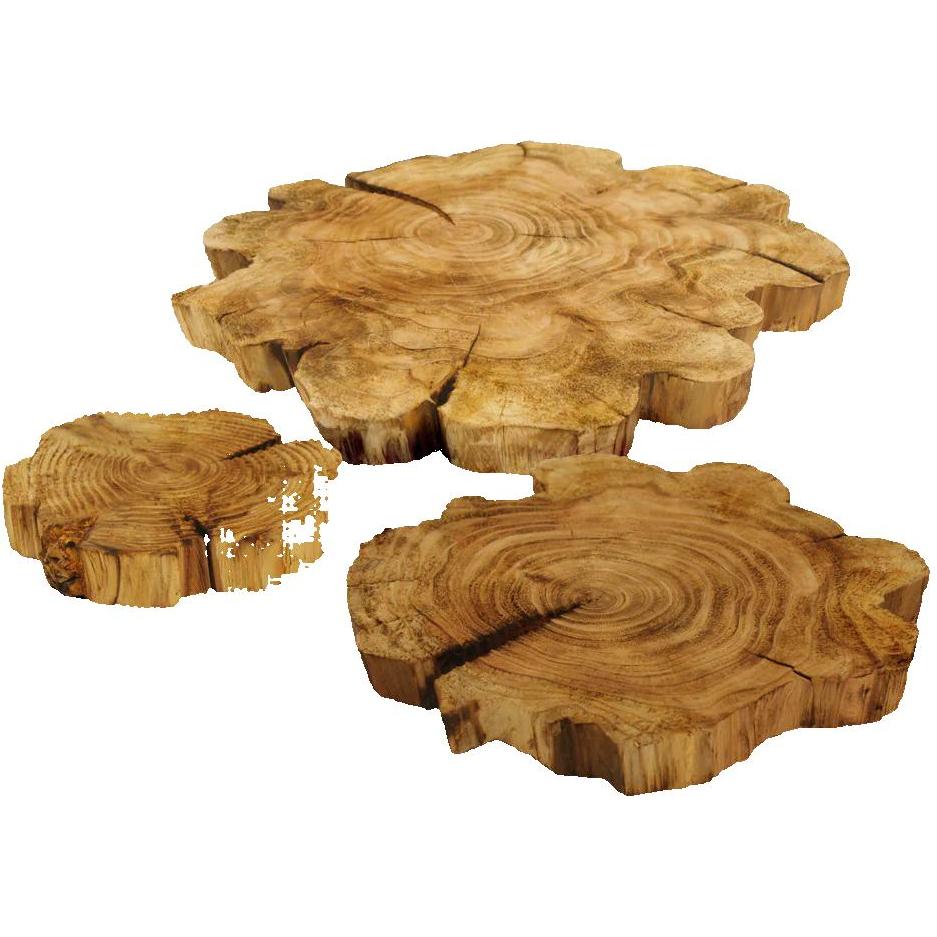} 
                    & \includegraphics[width=7cm]{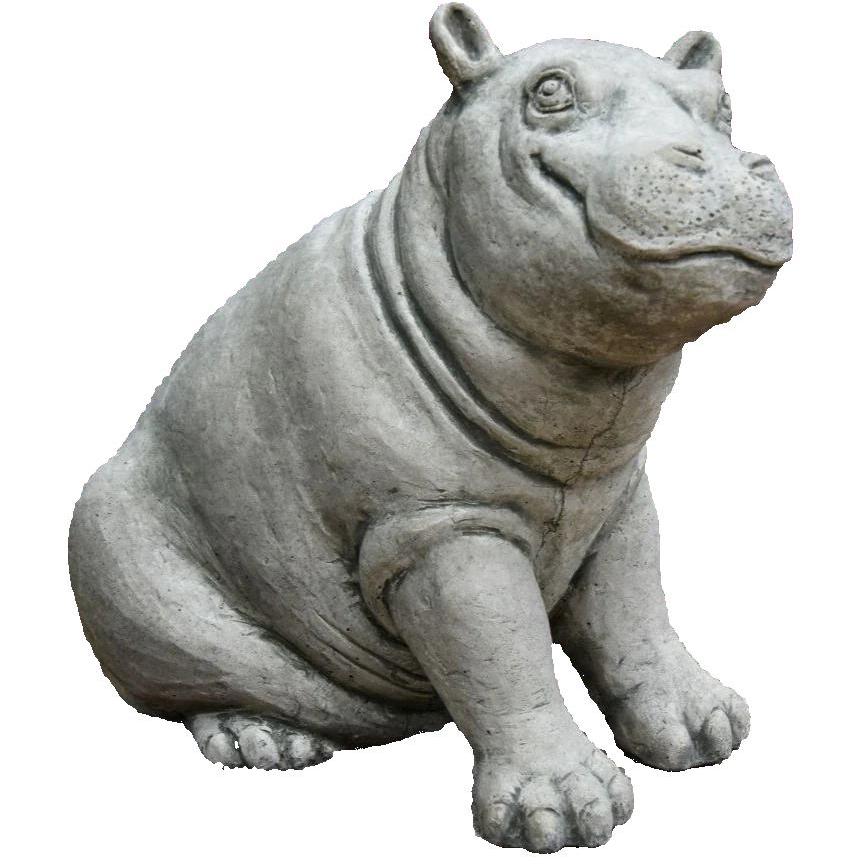} 
                    & \includegraphics[width=7cm]{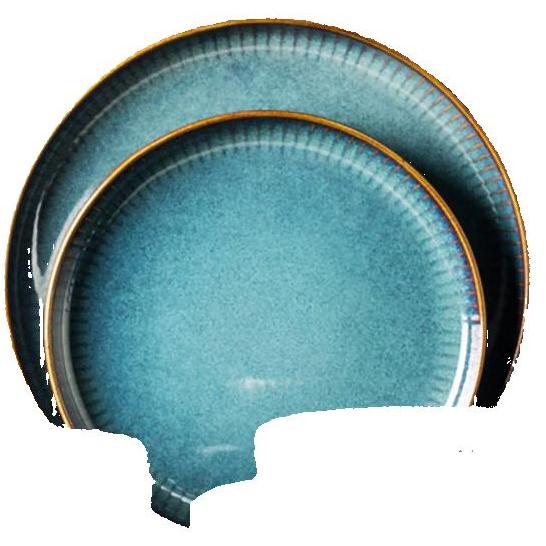} \\

                  \fontsize{26}{26}\selectfont{\emph{Plastic}} &  \fontsize{26}{26}\selectfont{\emph{Rubber}} &  \fontsize{26}{26}\selectfont{\emph{Metal}}& \fontsize{26}{26}\selectfont{\emph{Leather}} & \fontsize{26}{26}\selectfont{\emph{Fabric}} & \fontsize{26}{26}\selectfont{\emph{Wood}} & 
                \fontsize{26}{26}\selectfont{\emph{Stone}} & \fontsize{26}{26}\selectfont{\emph{Ceramic}}\\

                  \includegraphics[width=7cm]{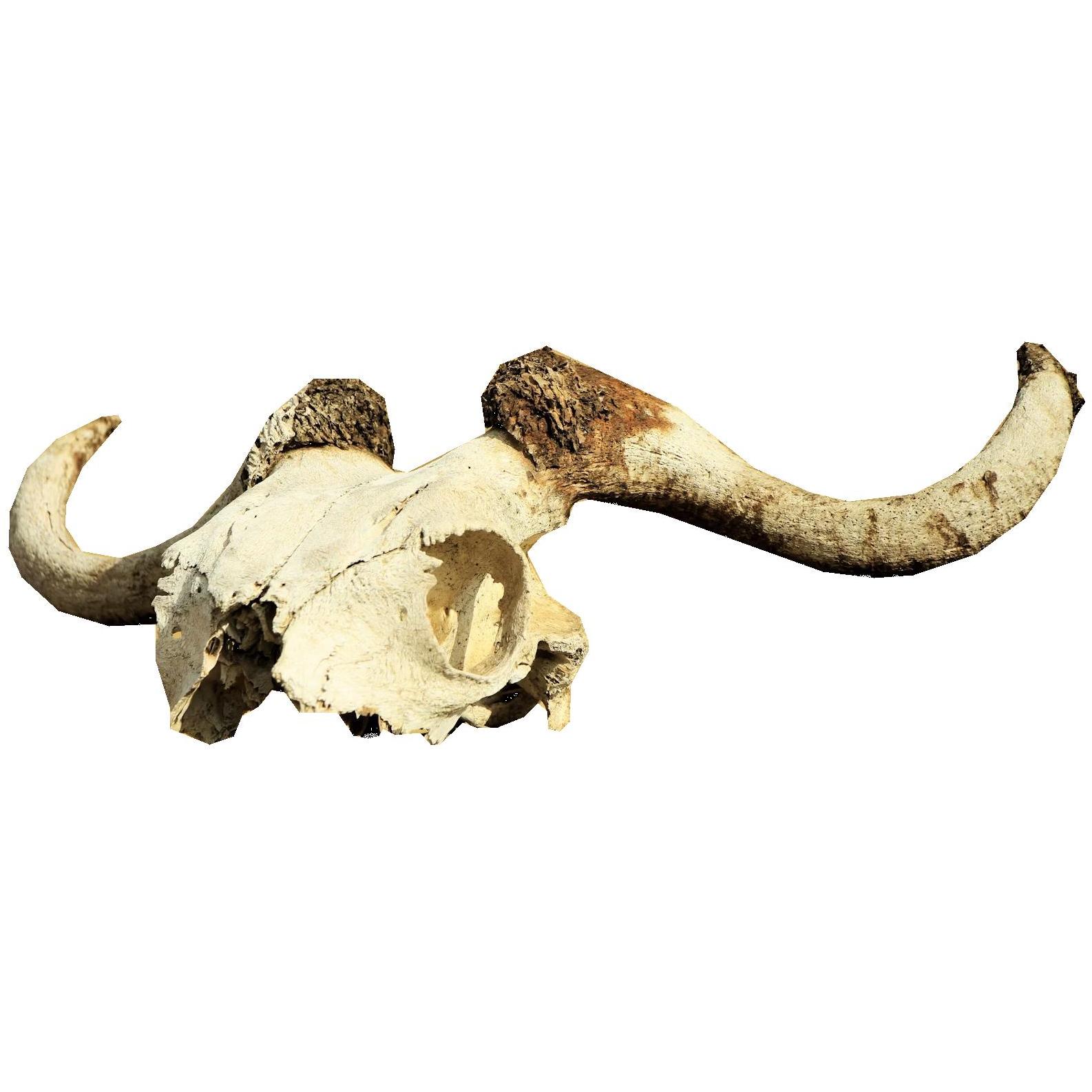} 
                 & \includegraphics[width=7cm]{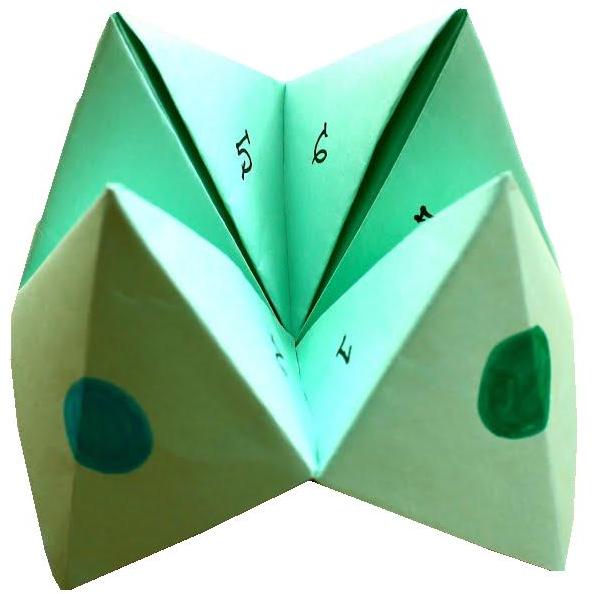} 
                    & \includegraphics[width=7cm]{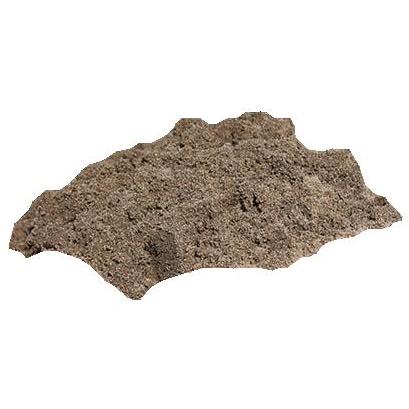} 
                    & \includegraphics[width=7cm]{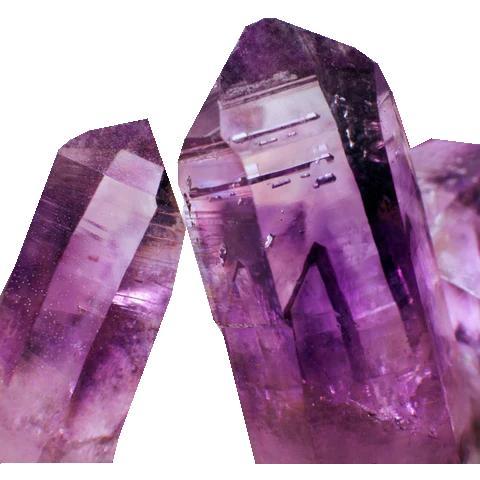} 
                    & \includegraphics[width=7cm]{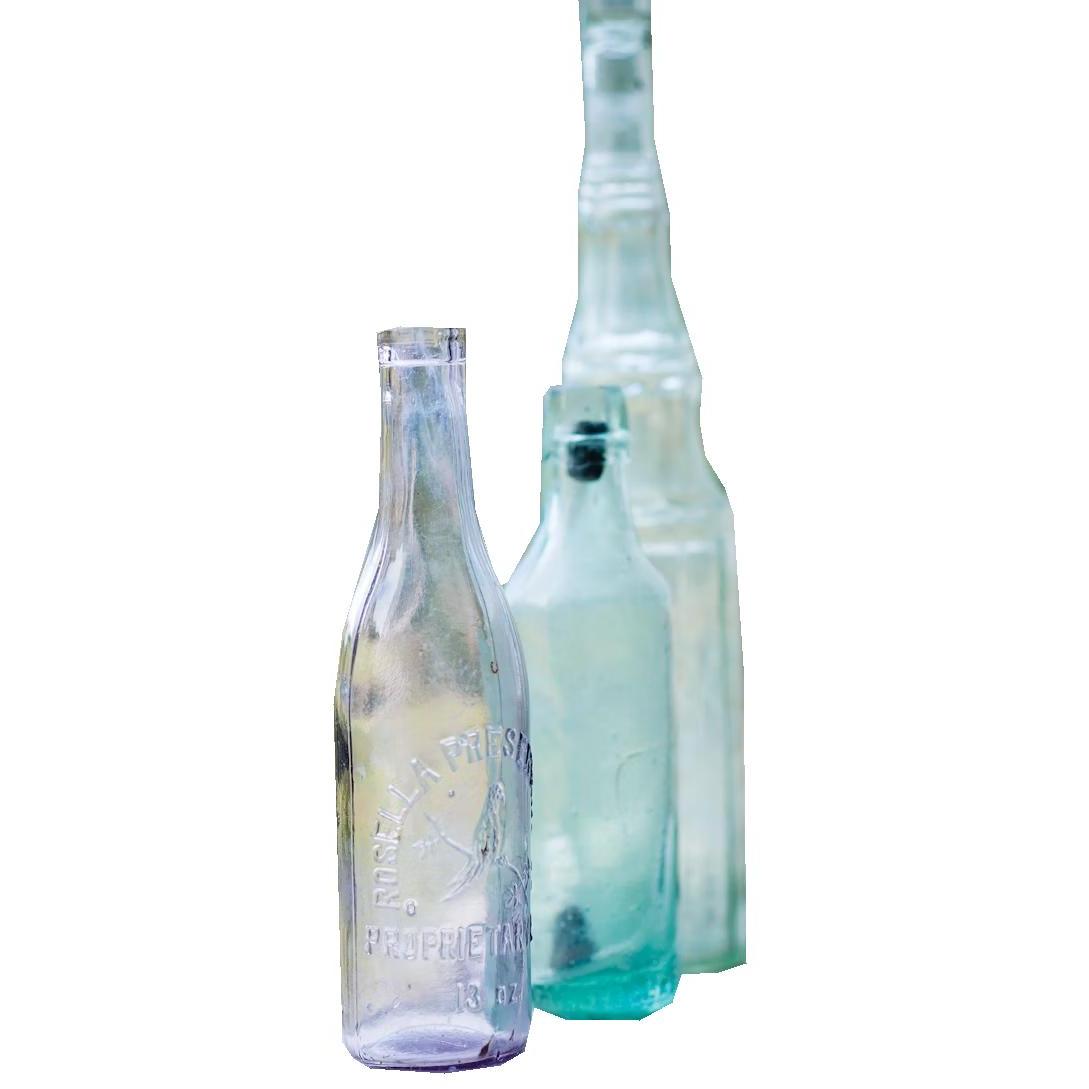} 
                    & \includegraphics[width=7cm]{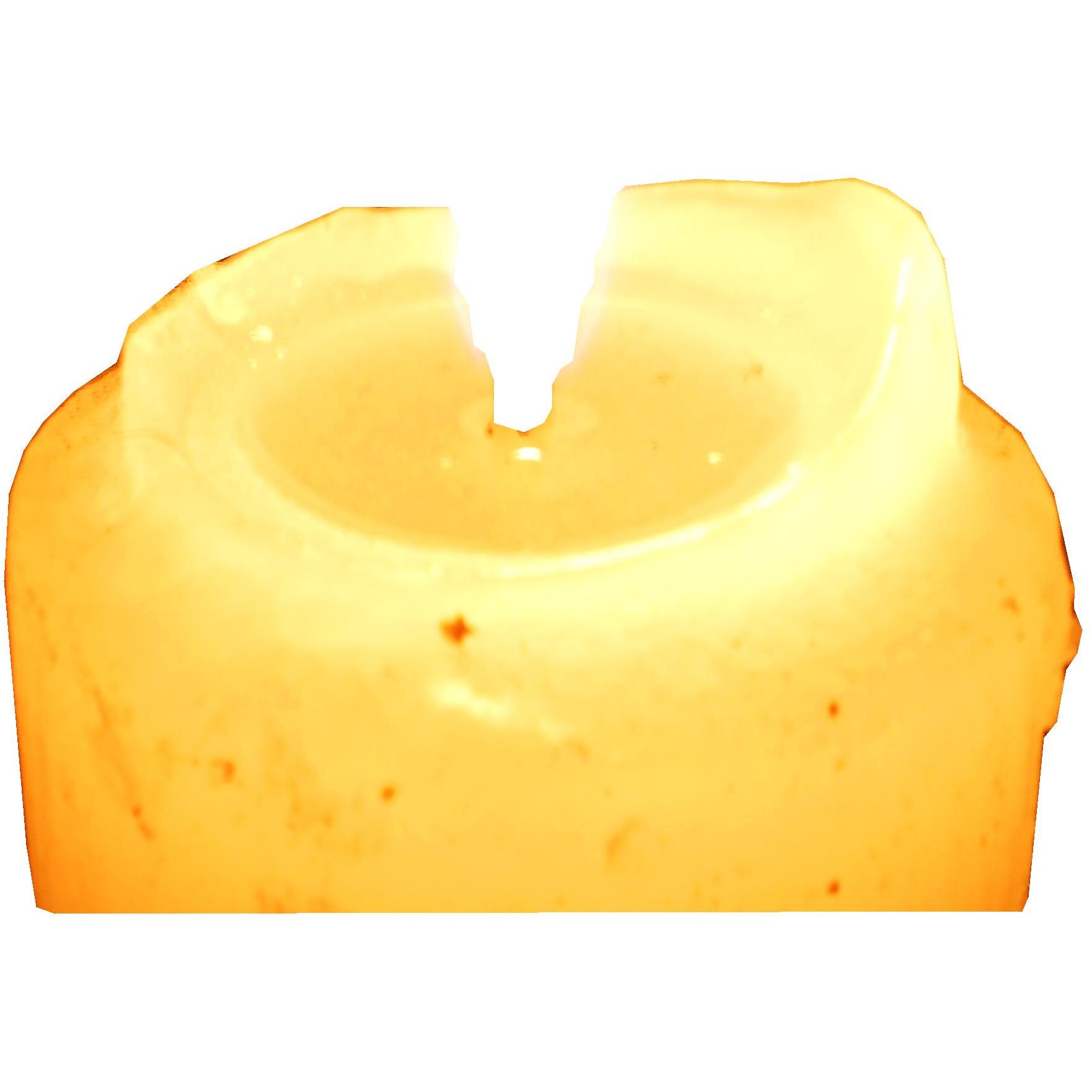} 
                    & \includegraphics[width=7cm]{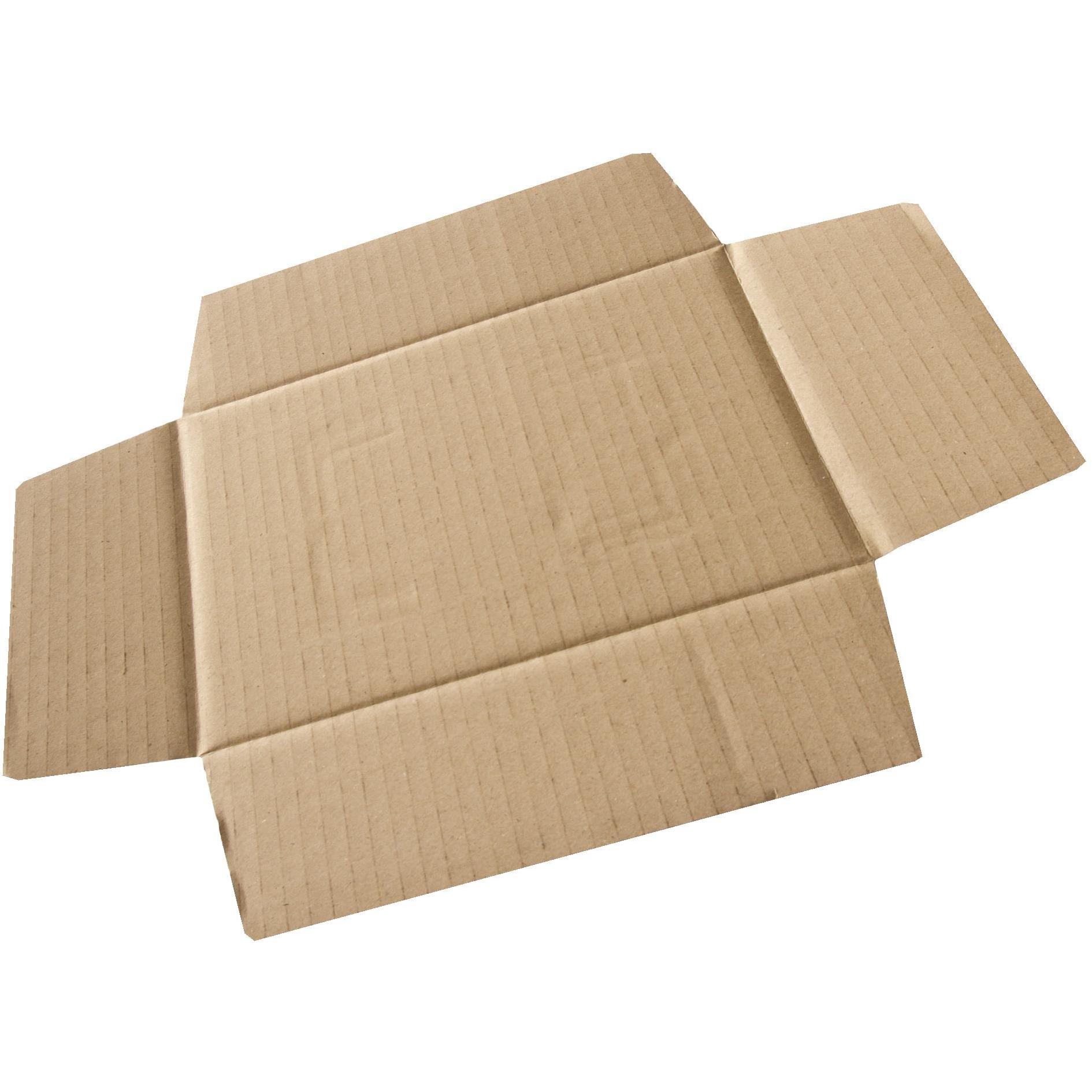} 
                    & \includegraphics[width=7cm]{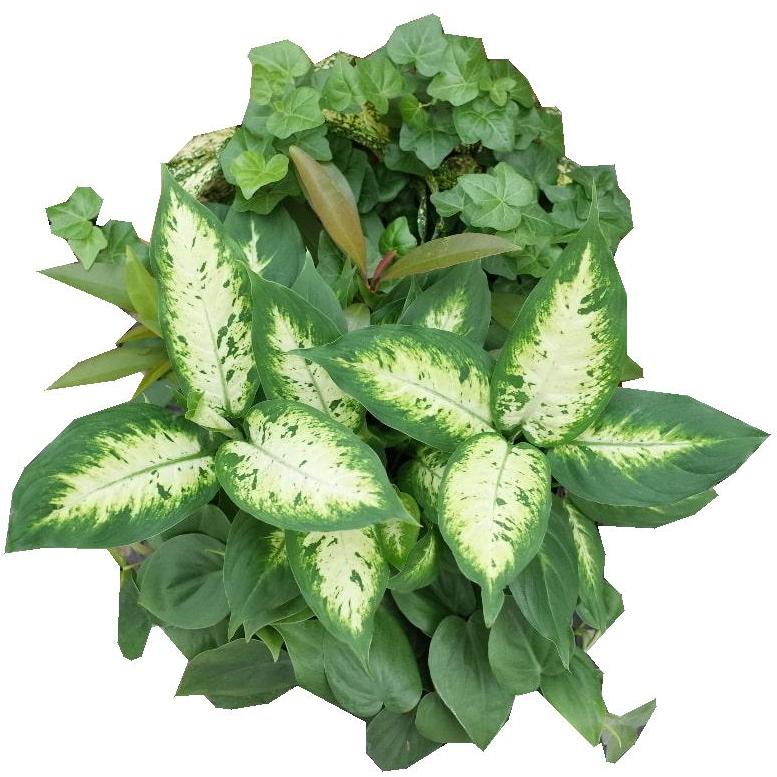} \\ 

                 \fontsize{26}{26}\selectfont{\emph{Bone}} &  \fontsize{26}{26}\selectfont{\emph{Paper}} &  \fontsize{26}{26}\selectfont{\emph{Soil}}& \fontsize{26}{26}\selectfont{\emph{Gemstone}} & \fontsize{26}{26}\selectfont{\emph{Glass}} & \fontsize{26}{26}\selectfont{\emph{Wax}} & 
                \fontsize{26}{26}\selectfont{\emph{Cardboard}} & \fontsize{26}{26}\selectfont{\emph{Foliage}}\\

                \end{tabular}

            }
               {\centering
                 \resizebox{0.625\textwidth}{!}{%
                \begin{tabular}
                {ccccc}
                \includegraphics[width=7cm]{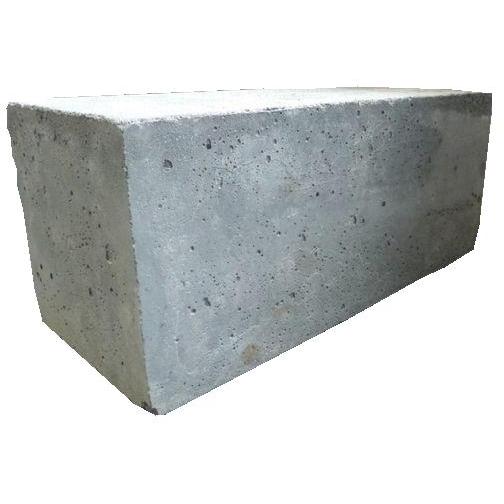} 
                    & \includegraphics[width=7cm]{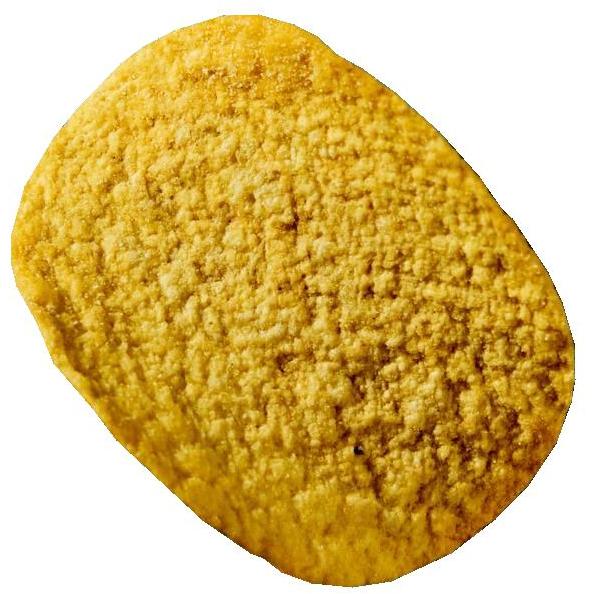} 
                    & \includegraphics[width=7cm]{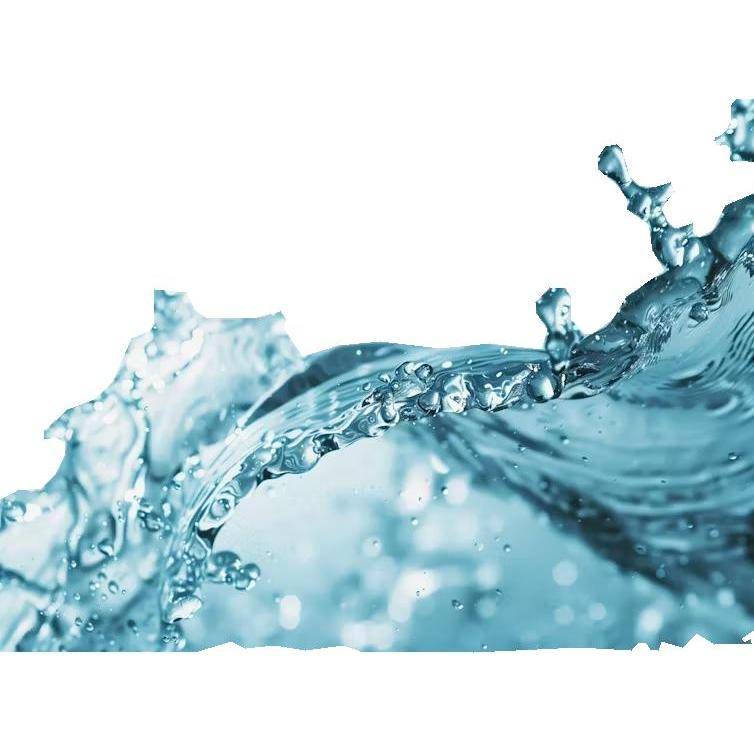} 
                    & \includegraphics[width=7cm]{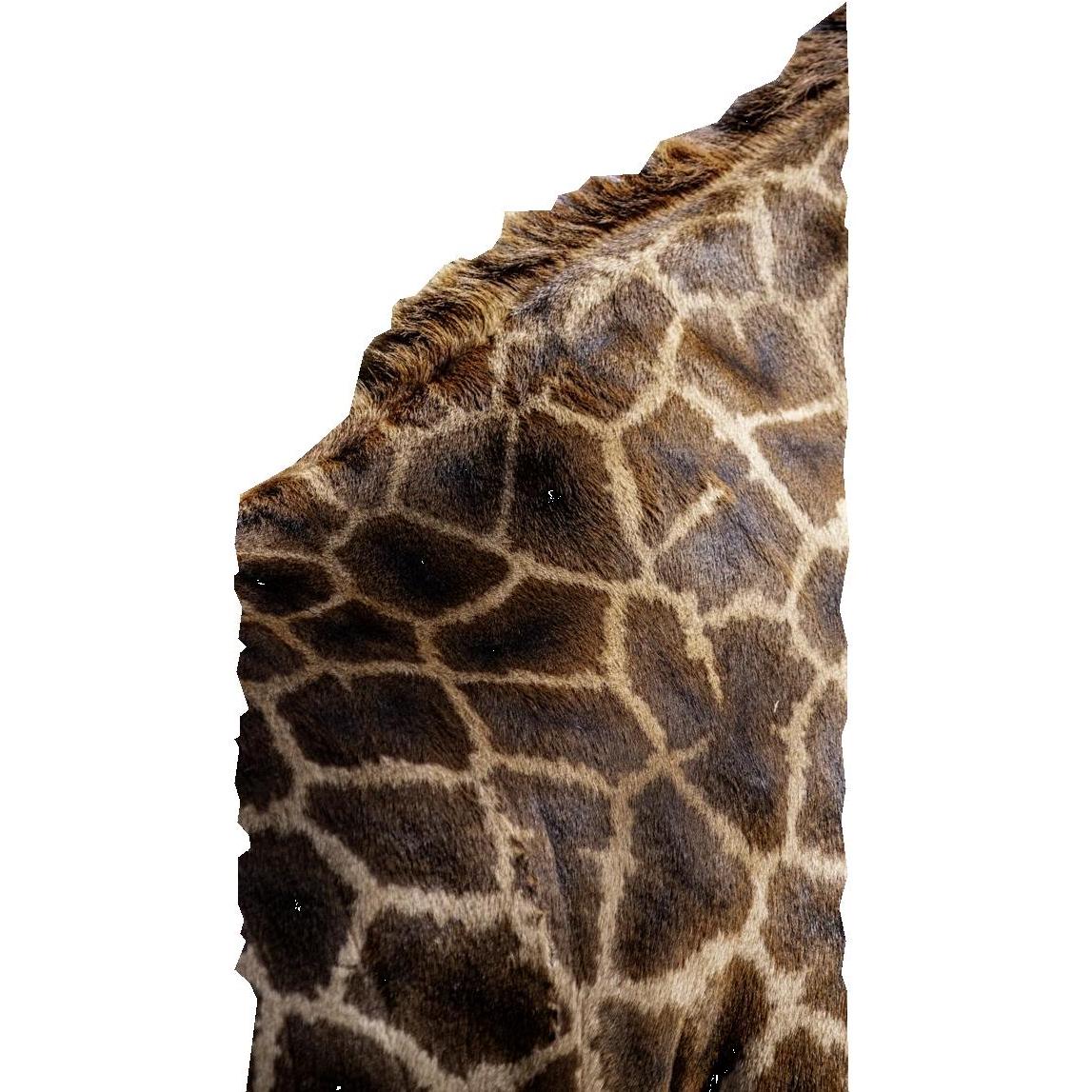} 
                    & \includegraphics[width=7cm]{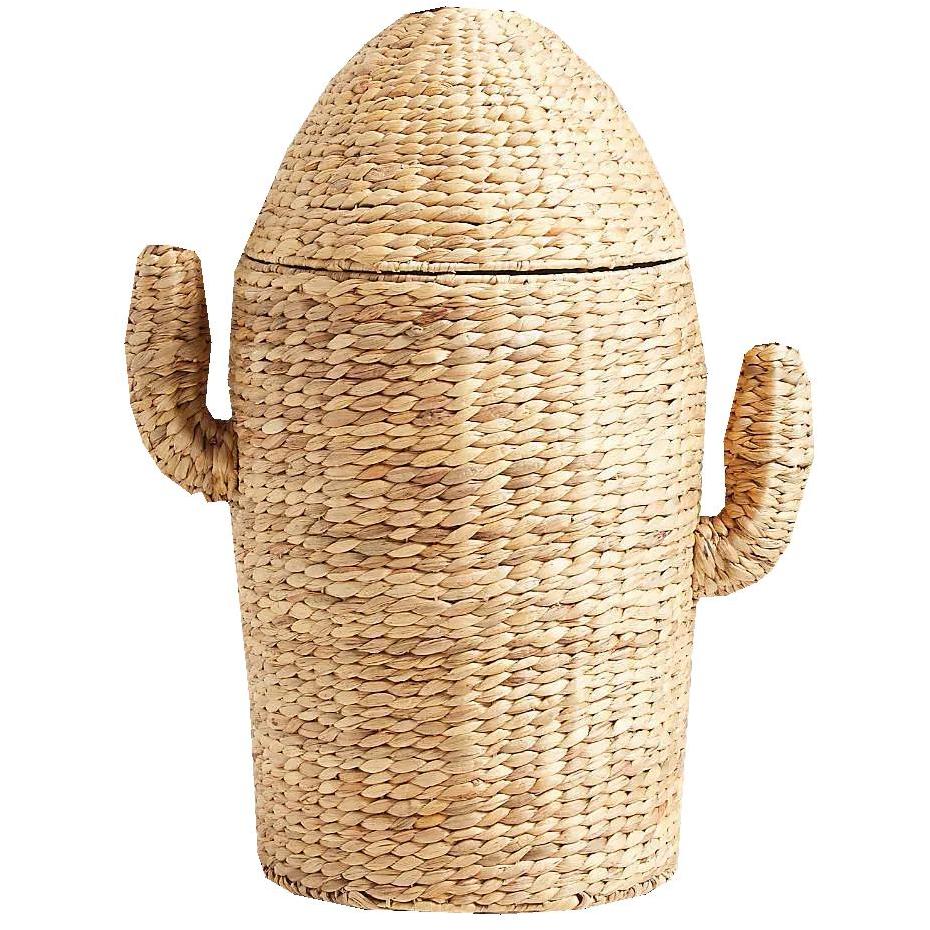} \\

                \fontsize{26}{26}\selectfont{\emph{Concrete}} &  \fontsize{26}{26}\selectfont{\emph{Sponge}} &  \fontsize{26}{26}\selectfont{\emph{Water}}& \fontsize{26}{26}\selectfont{\emph{Fur}} & \fontsize{26}{26}\selectfont{\emph{Wicker}}\\
     
                 \end{tabular}
                 }
                \par}
                \captionof{figure}{Illustration of the Google-test dataset. One representative sample is shown for each of the 21 categories.}
    \label{fig:figure8}
\vspace{-5mm}
\end{figure*}

In addition to the averaged results on the Google-test dataset reported in Table ~\ref{tab:table2} of the main text, we provide a detailed per-class breakdown here. Table ~\ref{tab:table7} presents the class-wise classification accuracy on the Google-test dataset, complementing the averaged results in the main text. This section is parallel to Section ~\ref{sec:DMS}, which reported the per-class accuracy on the DMS-test dataset. Figure ~\ref{fig:figure8} shows selected samples from the Google-test dataset, in the same manner as Figure ~\ref{fig:figure7} for DMS-test. One representative sample is shown for each of the 21 categories. Unlike DMS-test, which was constructed by filtering large connected regions from existing material datasets, Google-test was manually curated from Google Images. It also consists of 21 categories, with 30 images per category, resulting in a total of 630 test images.
\label{sec:image distribution analysis}

\begin{table}[h!]
    \caption{Comparison with state-of-the-art methods on the Google-test dataset (21 classes). All values denote classification accuracy(truncated to two decimal places), and the best result in each row is highlighted in \textbf{bold}.}
    \label{tab:table7}
    \resizebox{\columnwidth}{!}{%
    \begin{tabular}{l|c|c|c|c}
    \diagbox[width=6em]{Class}{Method}&CLIP\cite{radford2021learning} & GPT-4v \cite{achiam2023gpt}& MatSim\cite{drehwald2023one} & Ours\\
      \hline
      fabric   & 0.33         & 0.70          & 0.40             & \textbf{0.83} \\
      foliage  & \textbf{1.00}& 0.50          & 0.86             & \textbf{1.00} \\
      glass    & 0.83         & 0.76          & 0.53             & \textbf{0.90} \\
      leather  & 0.90         & \textbf{1.00} & 0.66             & \textbf{1.00} \\
      metal    & 0.70         & 0.60          & 0.36             & \textbf{1.00} \\
      paper    & 0.76         & 0.83          & 0.20             & \textbf{0.86} \\
      plastic  & 0.40         & \textbf{0.96} & 0.43             & \textbf{0.96} \\
      stone    & 0.83         & 0.76          & 0.46             & \textbf{0.96} \\
      water    & 0.86         & 0.93          & 0.90             & \textbf{1.00} \\
      wood     & 0.76         & 0.90          & 0.60             & \textbf{0.93} \\
      rubber   & 0.50         & 0.36          & 0.33             & \textbf{0.53} \\
      ceramic  & 0.56         & 0.66          & 0.33             & \textbf{0.70} \\
      sponge   & \textbf{1.00}& 0.90          & 0.80             & 0.86          \\
      bone     & \textbf{1.00}& 0.83          & 0.76             & 0.93          \\
      cardboard& 0.93         & 0.60          & 0.80             & \textbf{0.96} \\
      concrete & \textbf{0.96}& 0.90          & 0.76             & \textbf{0.96} \\
      fur      & 0.83         & 0.86          & 0.90             & \textbf{1.00} \\
      gemstone & \textbf{1.00}& 0.16          & 0.80             & \textbf{1.00} \\
      soil     & 0.93         & 0.86          & 0.66             & \textbf{1.00} \\
      wax      & \textbf{1.00}& 0.86          & 0.86             & \textbf{1.00} \\
      wicker   & \textbf{1.00}& 0.66          & 0.86             & 0.96          \\
      \hline
      {Average}& 0.81         & 0.74          & 0.63            & \textbf{0.92}  \\
    \end{tabular}
    }
     \vspace{-3mm} 
\end{table}

\section{Class image statistics} We reported detailed sample statistics for all 21 categories in our own dataset and compared them with the DMS dataset in Table ~\ref{tab:table8} . While DMS exhibits significant imbalance across classes, our generative dataset provides a more uniform distribution, enabling better supervision across rare categories.

\begin{table}[h!]
    \caption{Per-class image statistics for DMS\cite{upchurch2022dense} and our synthetic dataset. For each category, the number on the left of “\texttt{\textbar}” denotes the DMS image count, while the number on the right denotes the corresponding count in our synthetic dataset.}
    \label{tab:table8}
    \resizebox{\columnwidth}{!}{%
    \begin{tabular}{lc|lc}
     
      Class& DMS\cite{upchurch2022dense}\texttt{\textbar}Ours  & Class & DMS\cite{upchurch2022dense}\texttt{\textbar}Ours\\
       \hline
      fabric   & 31,489\texttt{\textbar} 1,345   & ceramic  & 8,314\texttt{\textbar} 740  \\
      foliage  & 11,384\texttt{\textbar} 1,000     & sponge   & 326\texttt{\textbar}  725 \\
      glass    & 28,934\texttt{\textbar}  860  &  bone     & 3,751\texttt{\textbar} 965   \\
      leather  & 7,354\texttt{\textbar}   875 & cardboard& 3,150\texttt{\textbar} 1,000  \\
      metal    & 30,504\texttt{\textbar}  2,330   & concrete & 2,853\texttt{\textbar}  955 \\
      paper    & 20,763\texttt{\textbar}  1,000   & fur      & 1,567\texttt{\textbar} 620  \\
      plastic  & 36,937\texttt{\textbar}  2,060   &  gemstone & 369 \texttt{\textbar} 615\\
      stone    & 4,206\texttt{\textbar}   730   & soil     & 1,855 \texttt{\textbar}  985  \\
      water    & 2,063\texttt{\textbar}  1020    & wax      & 1,107 \texttt{\textbar}  980 \\
      wood     & 26,274\texttt{\textbar} 1025   & wicker   & 1,895 \texttt{\textbar} 1000 \\
      rubber   & 7,811\texttt{\textbar}   645     &&        \\

      \hline
    \end{tabular}
    }
     \vspace{-3mm} 
\end{table}

\section{Samples from our generated dataset} 
\begin{figure*}[ht]
    \resizebox{\textwidth}{!}{%
 \begin{tabular}
                {cccccccc}

                   \fontsize{26}{26}\selectfont{\emph{Plastic}}
                    & \includegraphics[width=7cm]{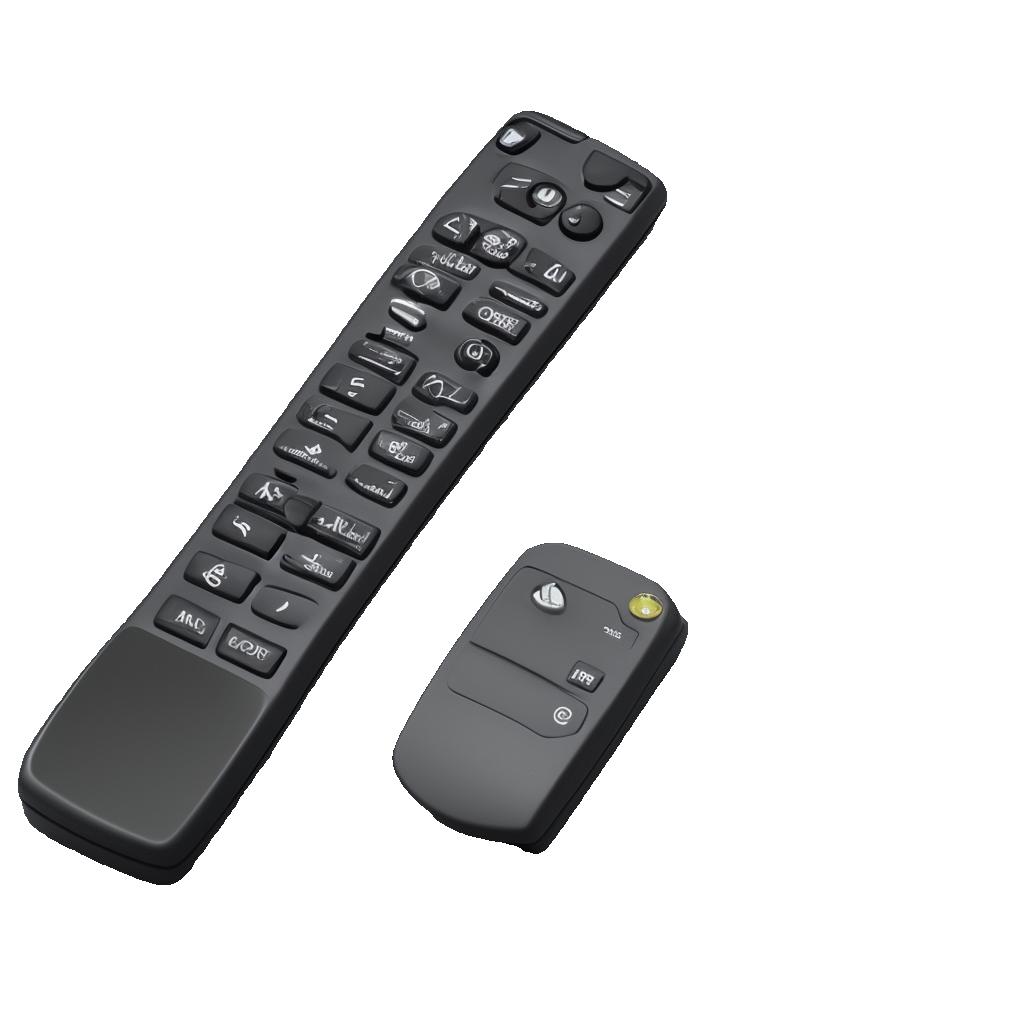} 
                    & \includegraphics[width=7cm]{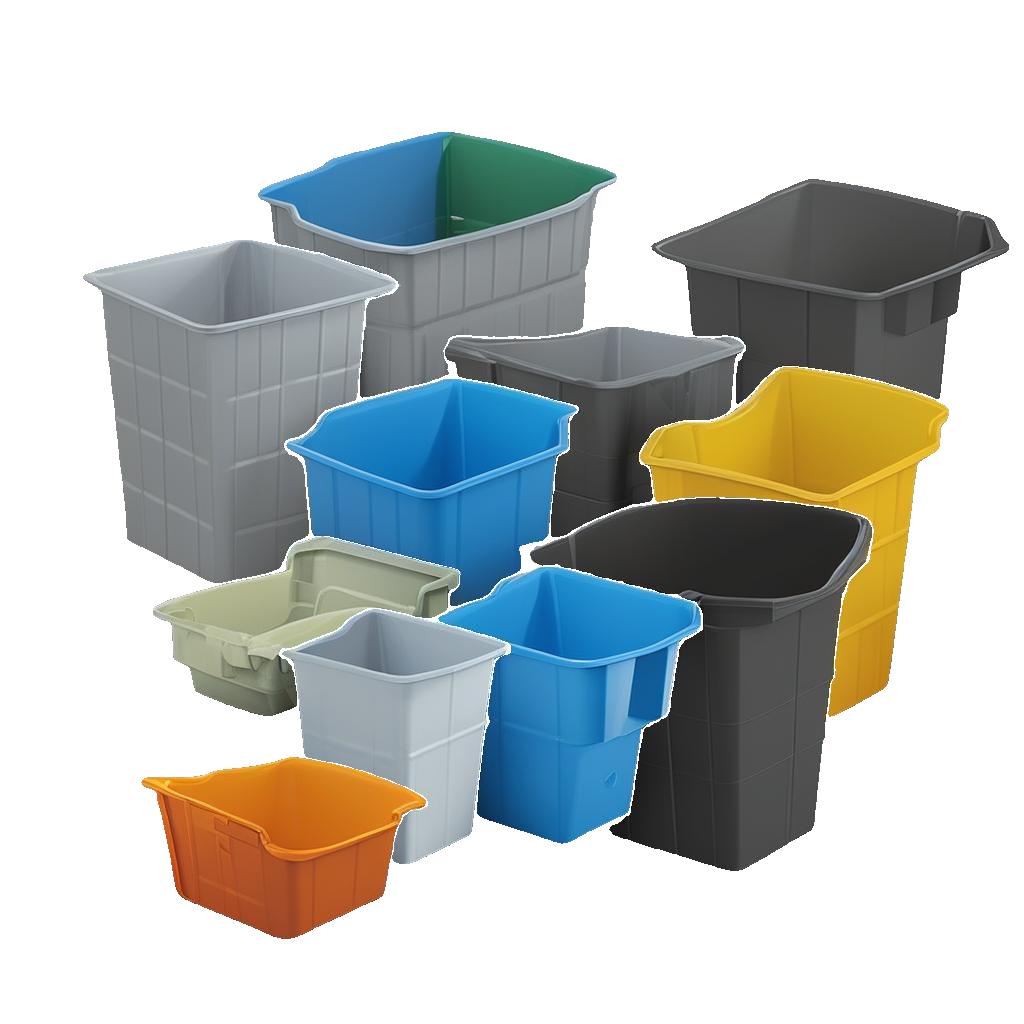} 
                    & \includegraphics[width=7cm]{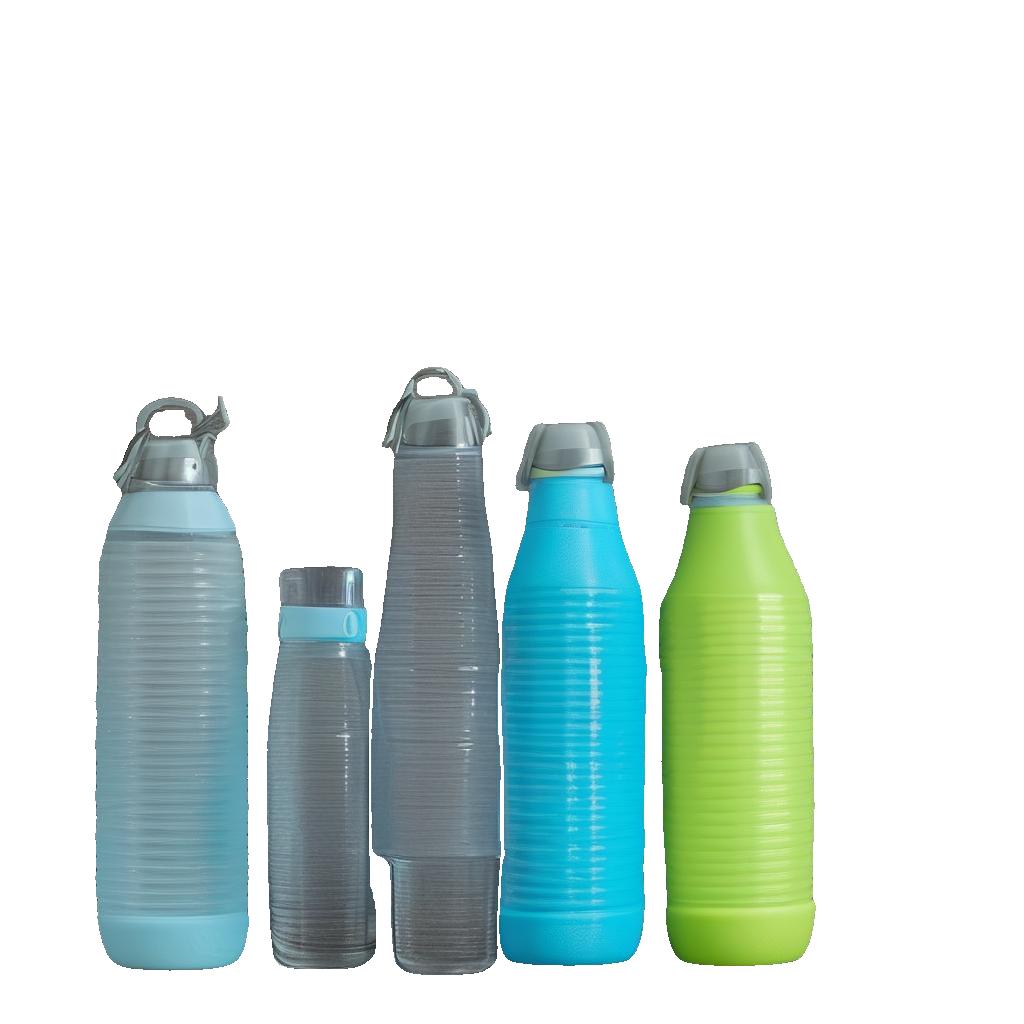} &
                    \fontsize{26}{26}\selectfont{\emph{Rubber}}
                    & \includegraphics[width=7cm]{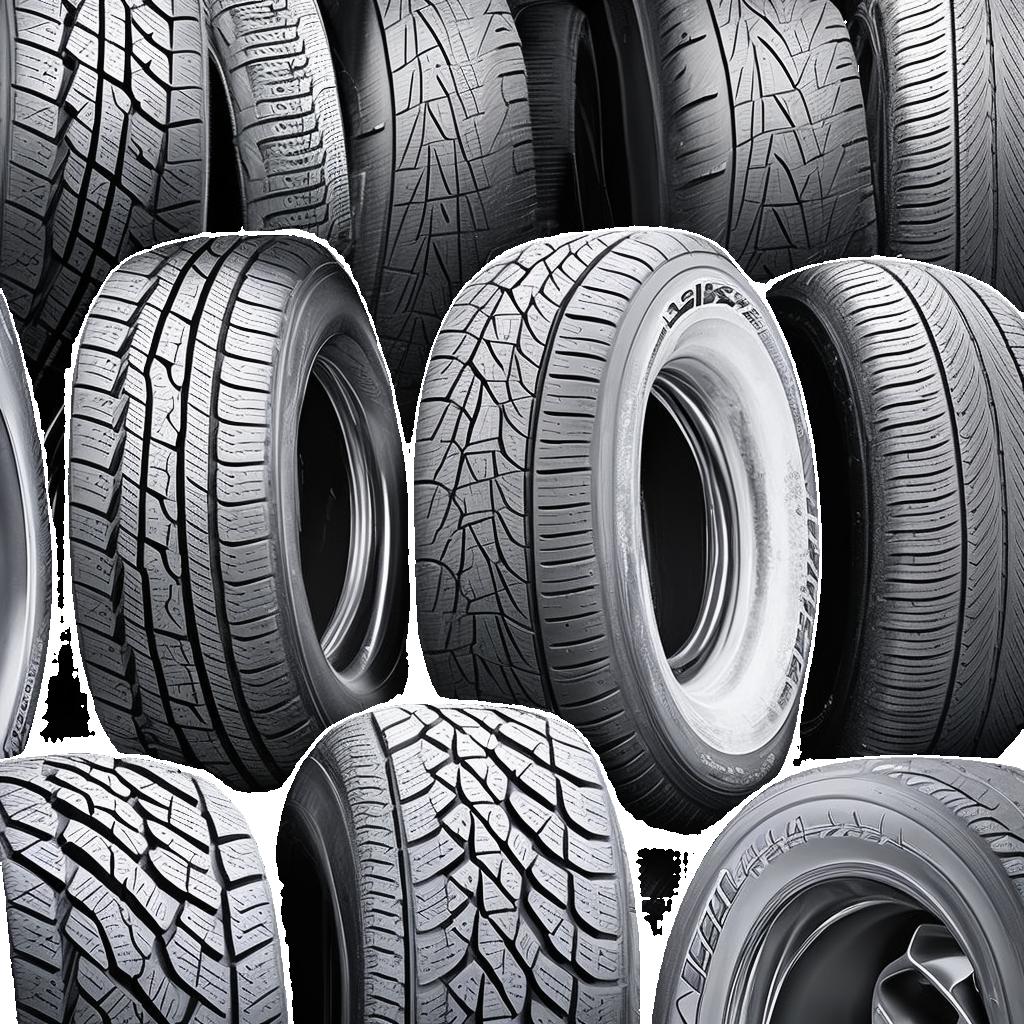} 
                    & \includegraphics[width=7cm]{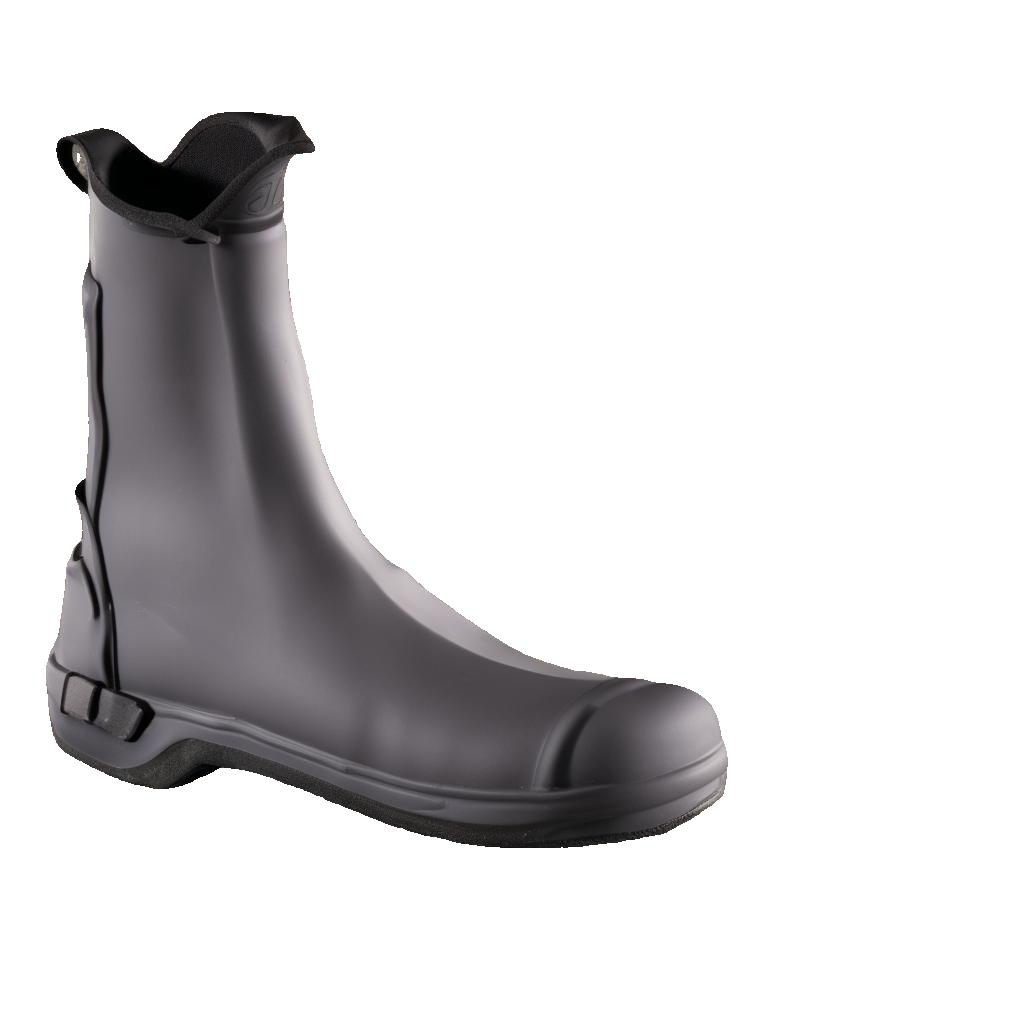} 
                    & \includegraphics[width=7cm]{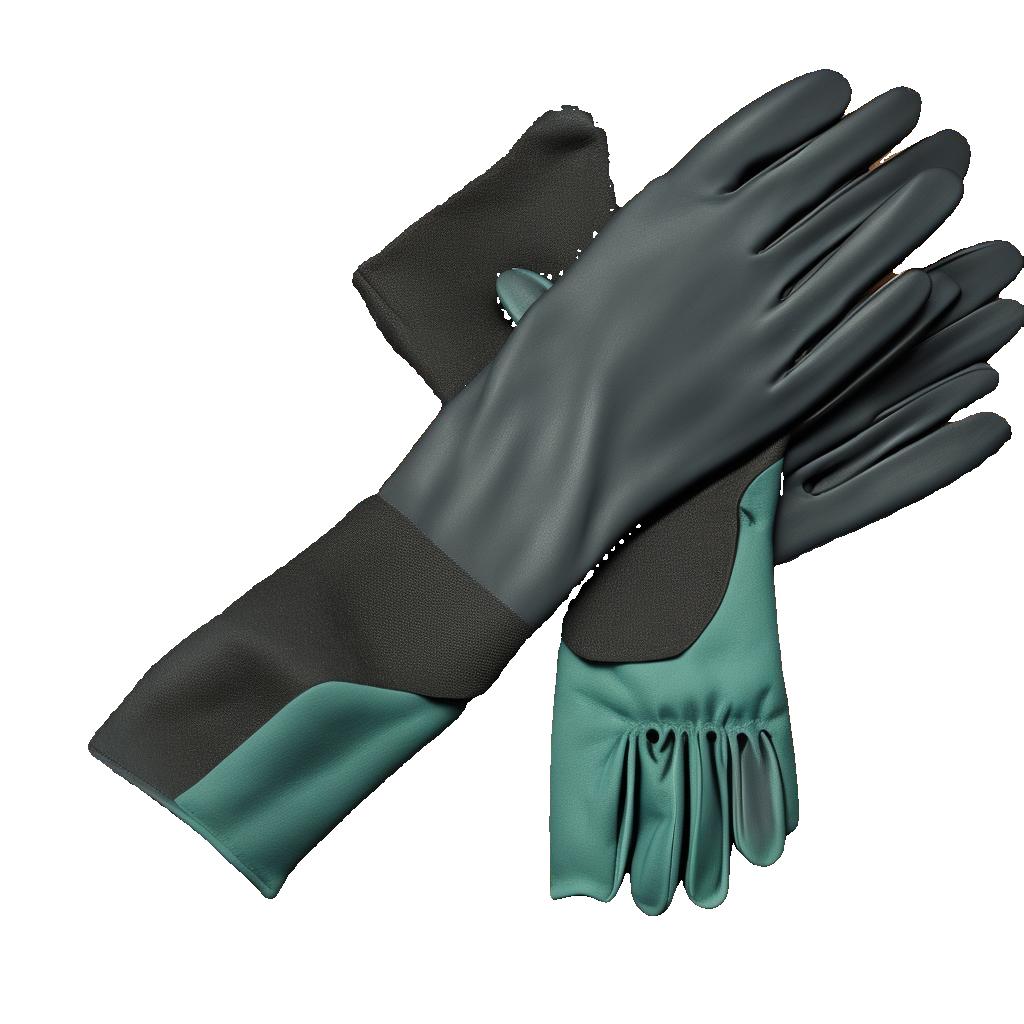} \\

                    &\fontsize{26}{26}\selectfont{\emph{remote}} &  \fontsize{26}{26}\selectfont{\emph{bin}}& \fontsize{26}{26}\selectfont{\emph{water-bottle}} &&\fontsize{26}{26}\selectfont{\emph{tire}} & 
                \fontsize{26}{26}\selectfont{\emph{boot}} & \fontsize{26}{26}\selectfont{\emph{gloves}}\\

                  \fontsize{26}{26}\selectfont{\emph{Metal}}
                    &\includegraphics[width=7cm]{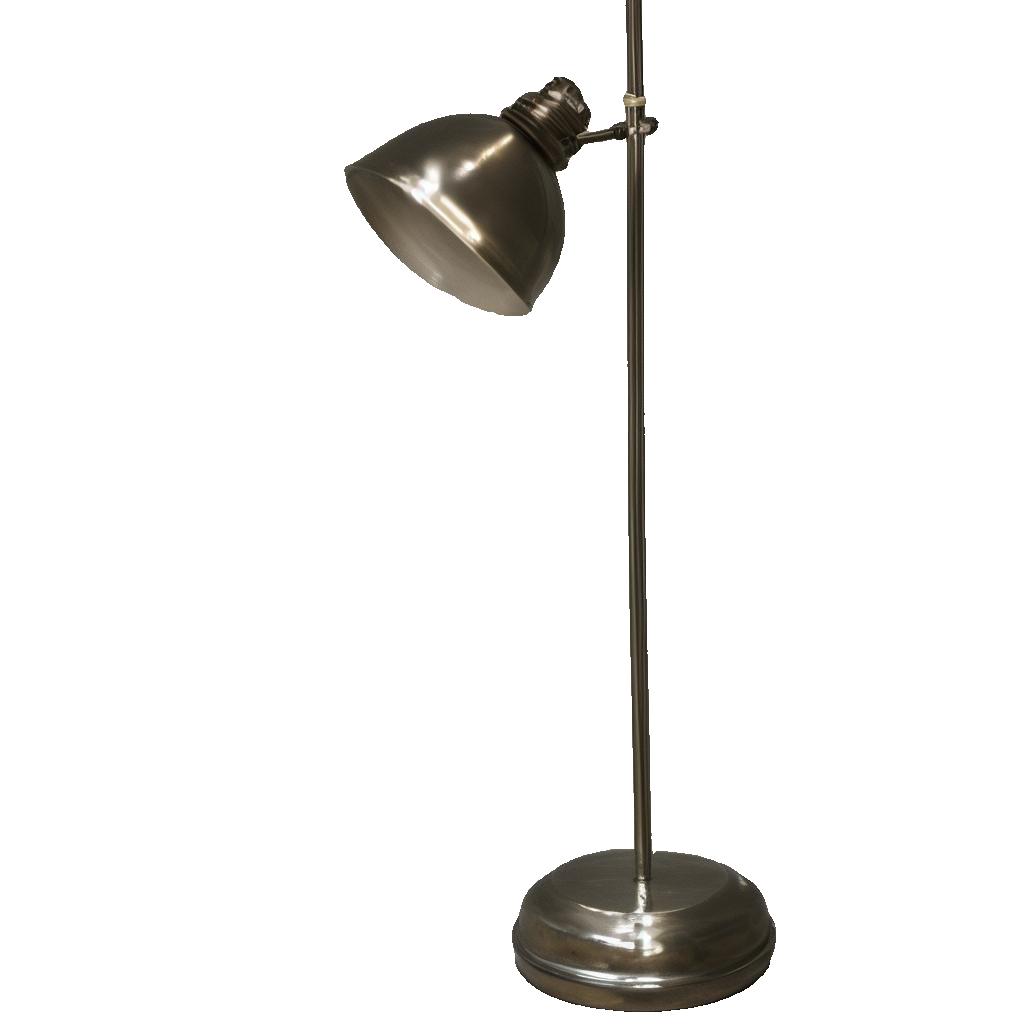} 
                 & \includegraphics[width=7cm]{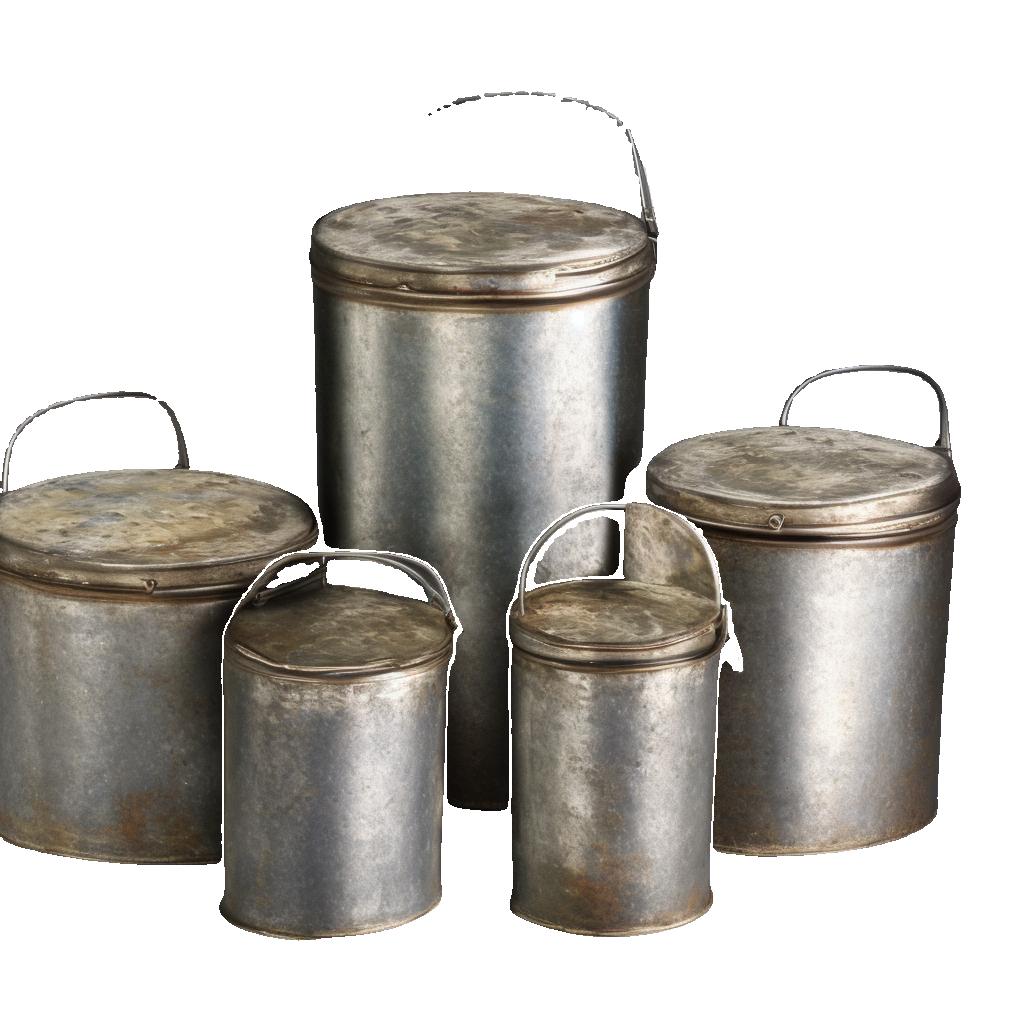} 
                    & \includegraphics[width=7cm]{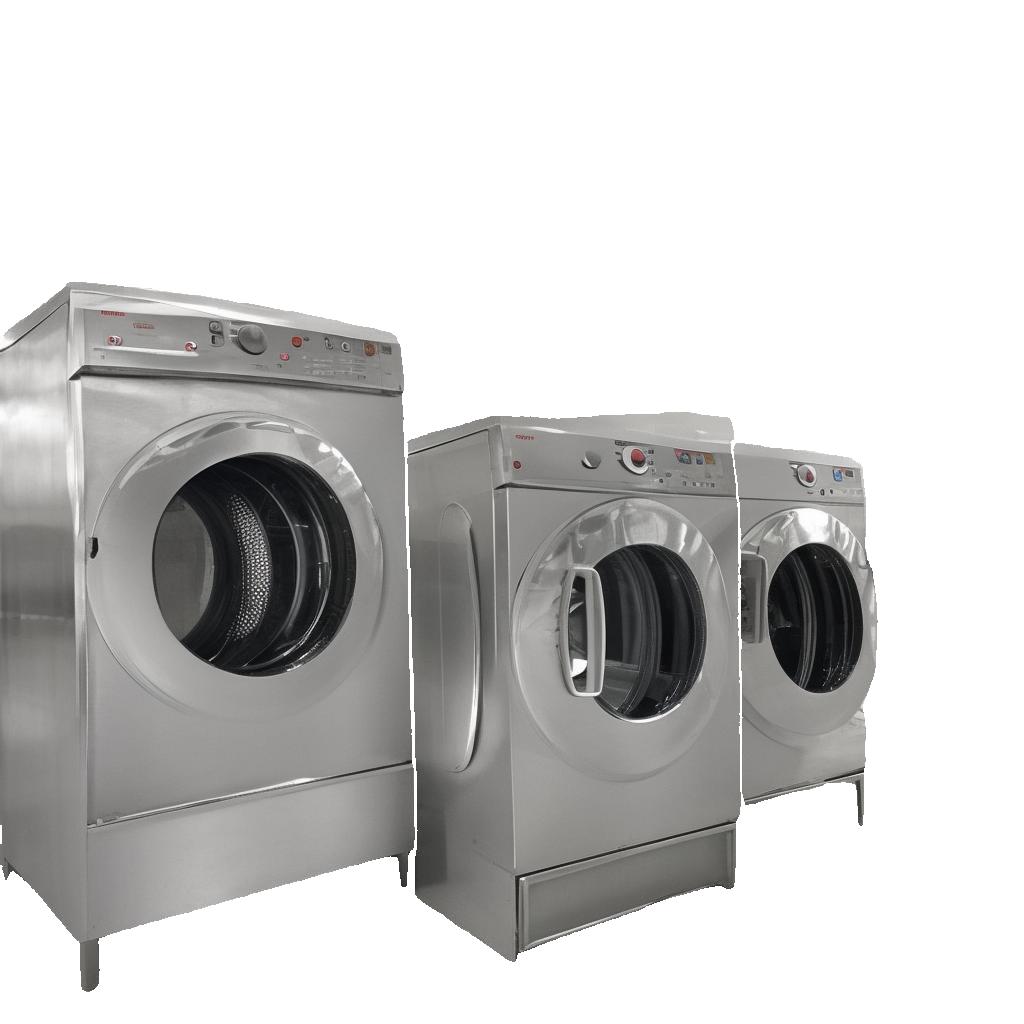} 
                    & \fontsize{26}{26}\selectfont{\emph{Leather}}
                    & \includegraphics[width=7cm]{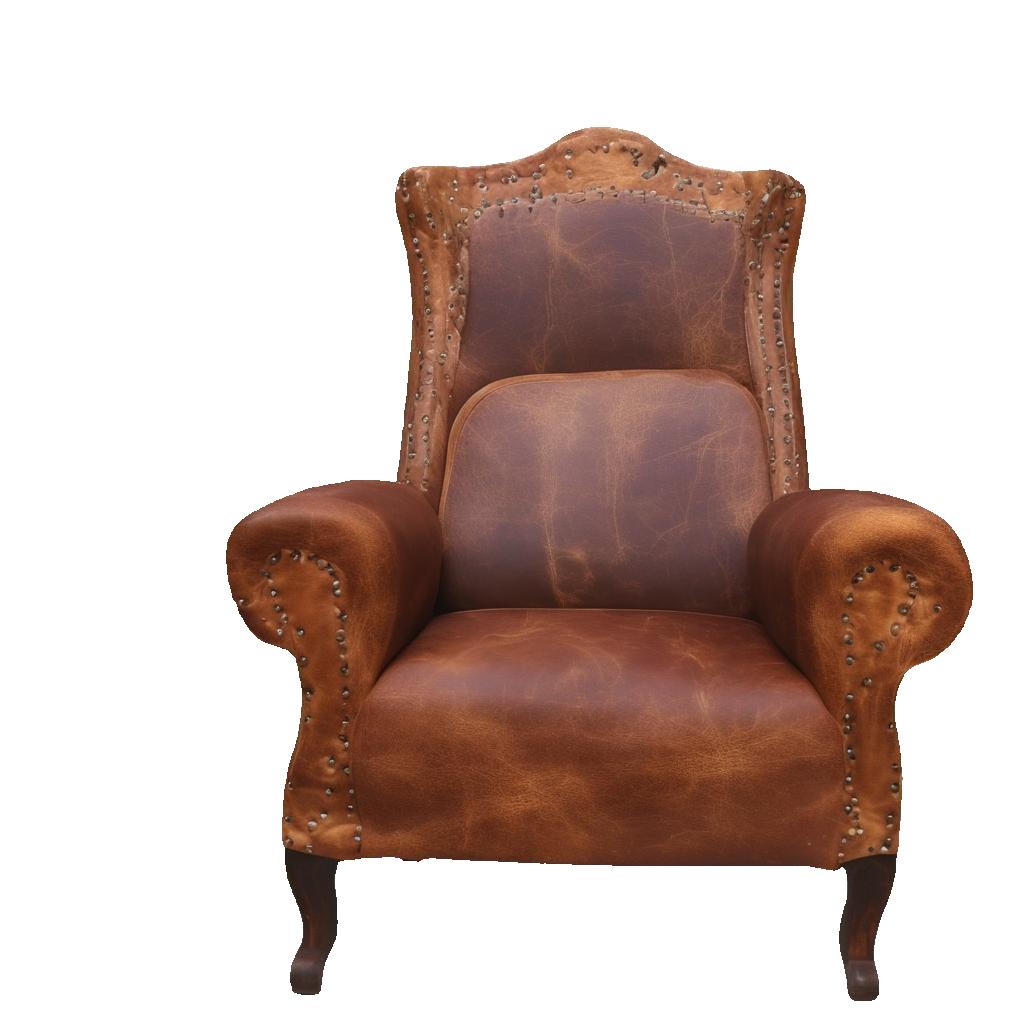} 
                    & \includegraphics[width=7cm]{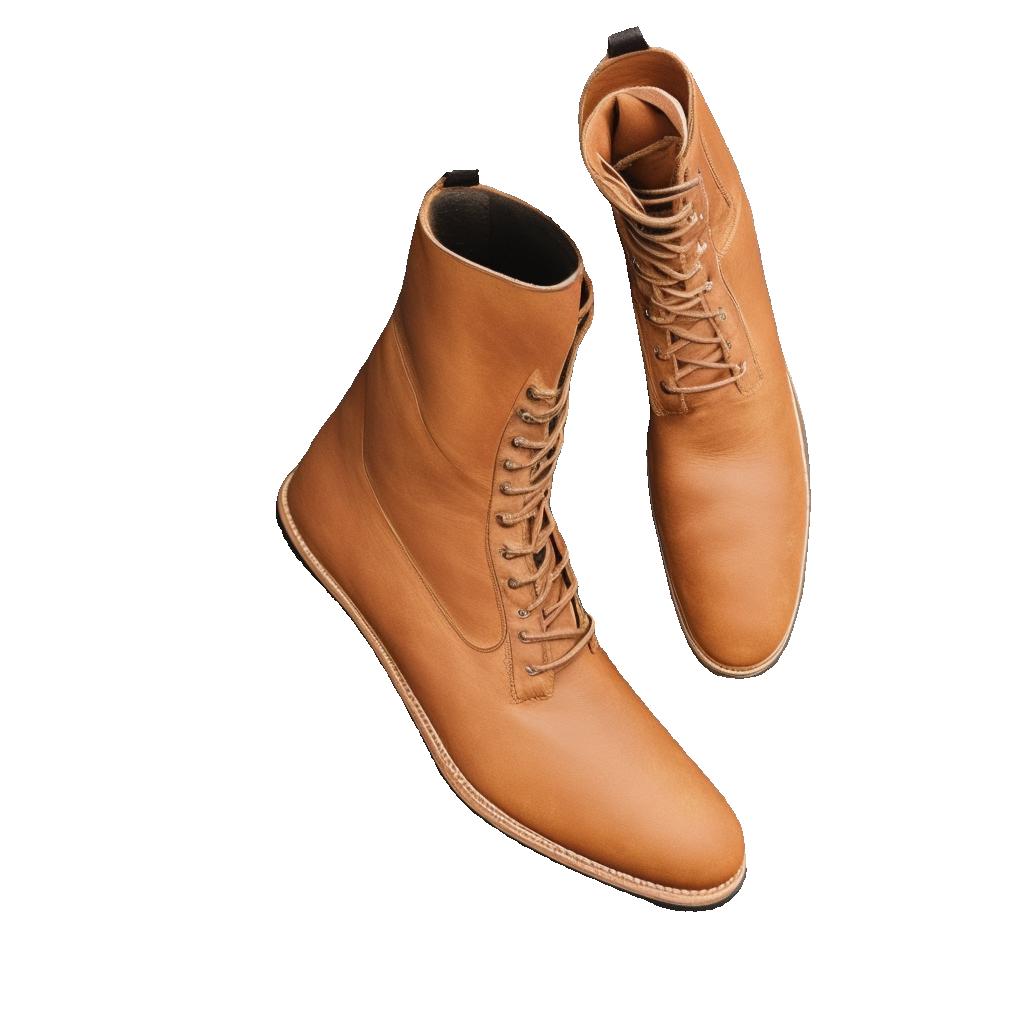} 
                    & \includegraphics[width=7cm]{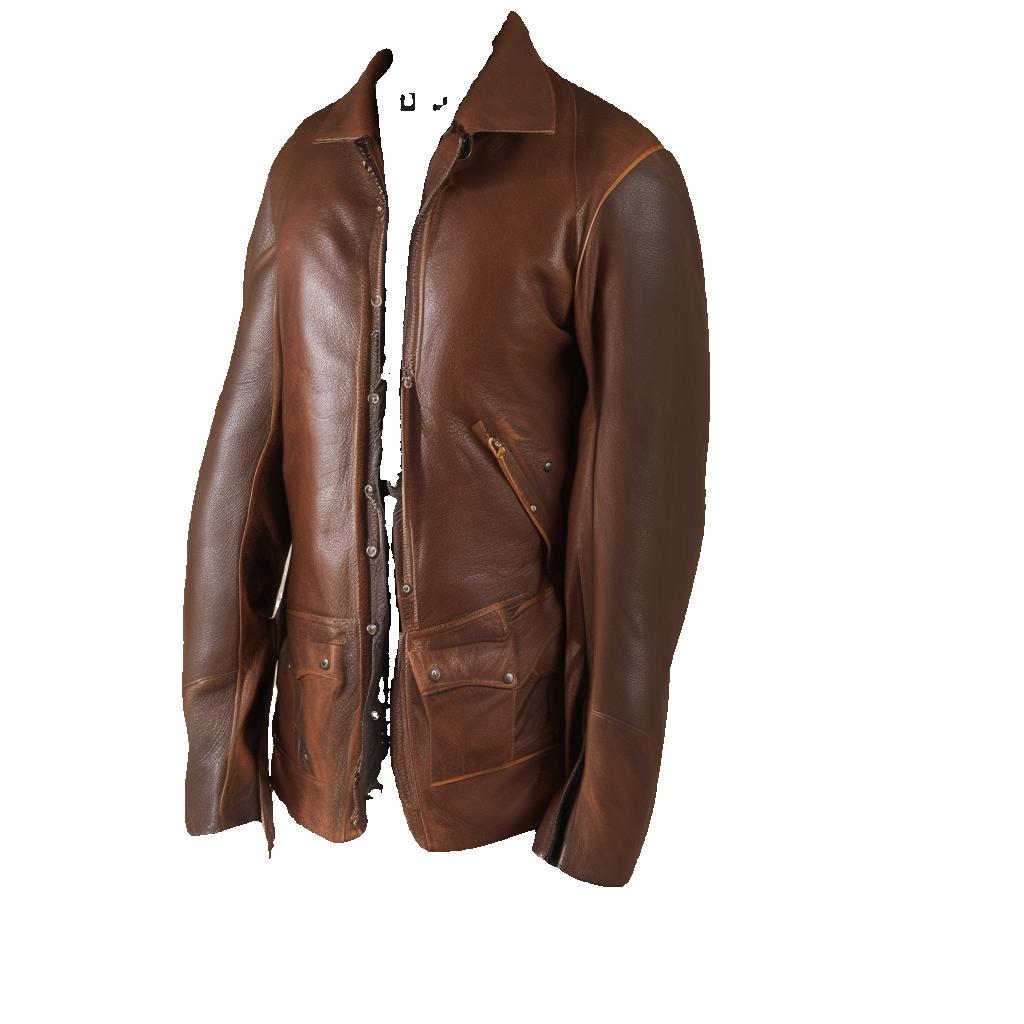} \\

                 & \fontsize{26}{26}\selectfont{\emph{lamp}} &  \fontsize{26}{26}\selectfont{\emph{canister}}& \fontsize{26}{26}\selectfont{\emph{dryer}} & & \fontsize{26}{26}\selectfont{\emph{chair}} & \fontsize{26}{26}\selectfont{\emph{boot}} & 
                \fontsize{26}{26}\selectfont{\emph{jacket}}\\

                  \fontsize{26}{26}\selectfont{\emph{Fabric}}
                    &\includegraphics[width=7cm]{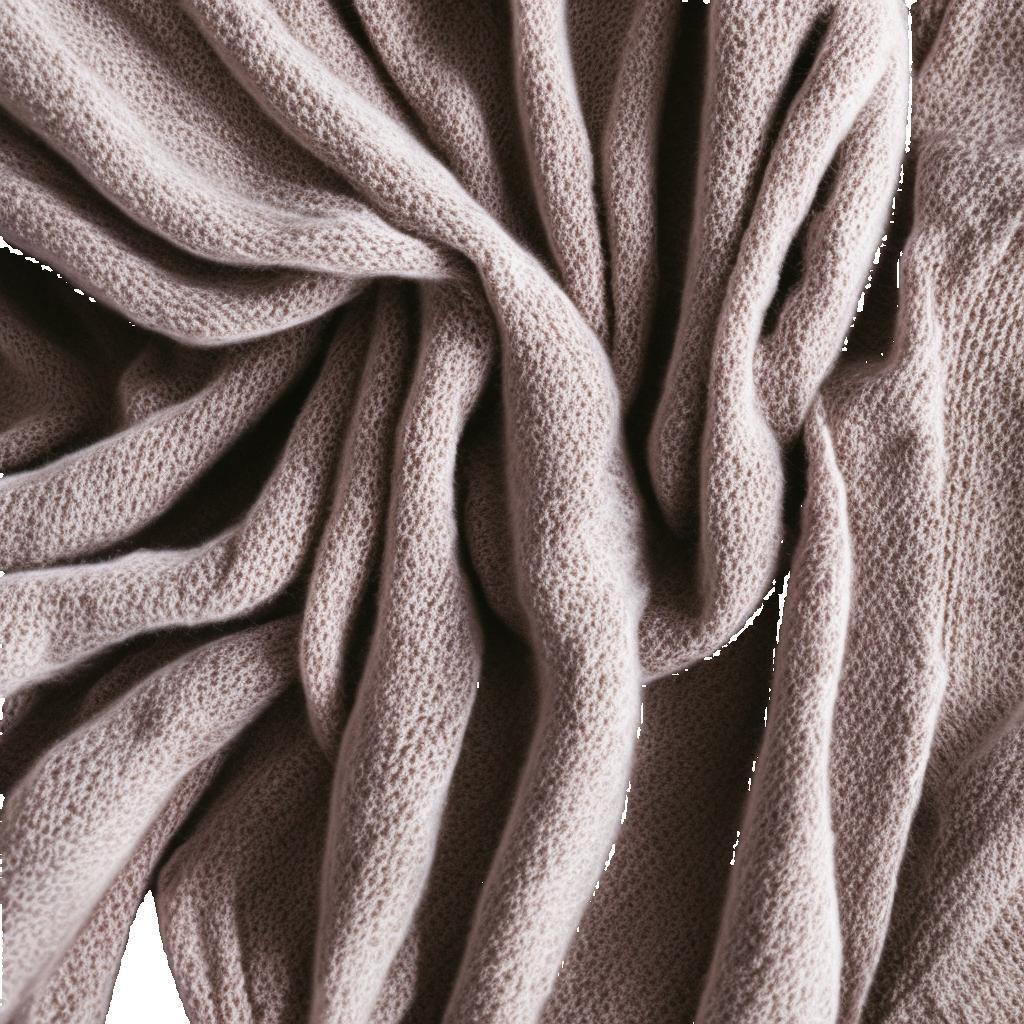} 
                 & \includegraphics[width=7cm]{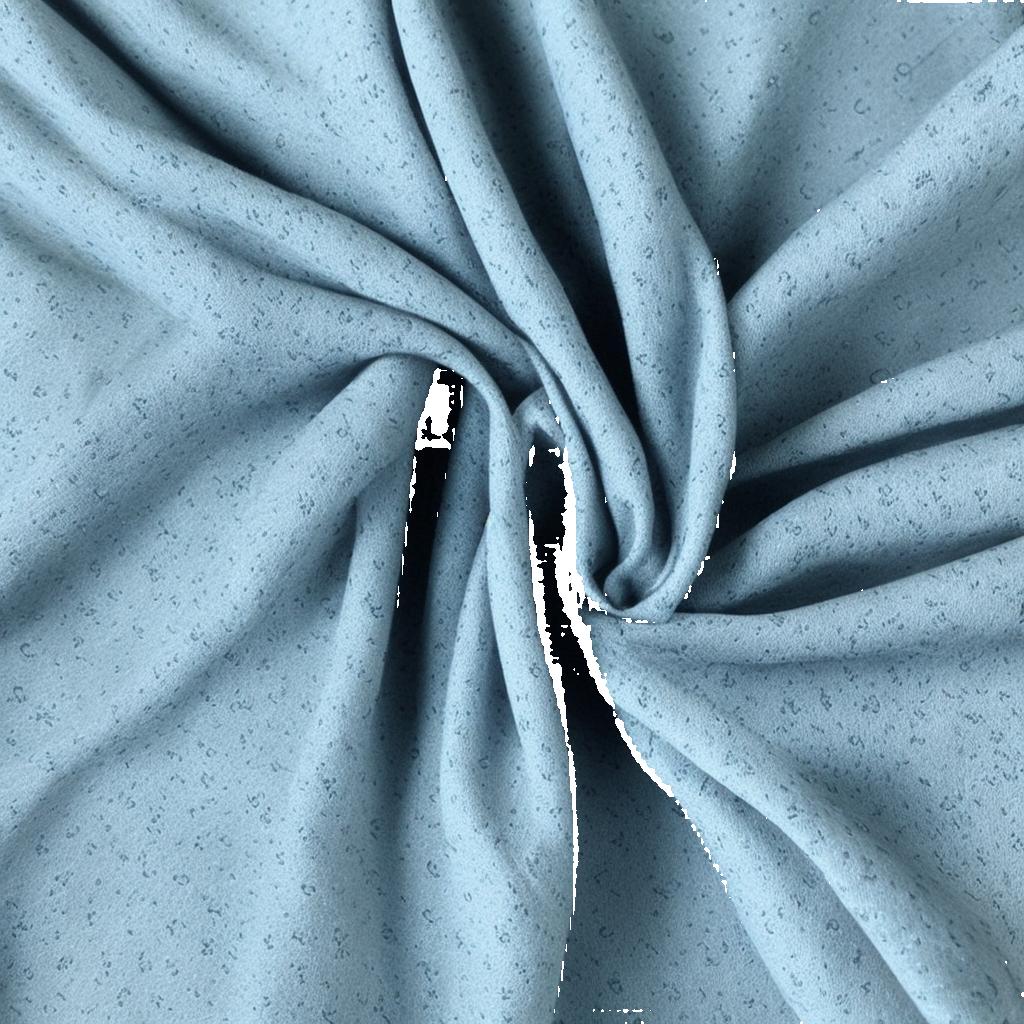} 
                    & \includegraphics[width=7cm]{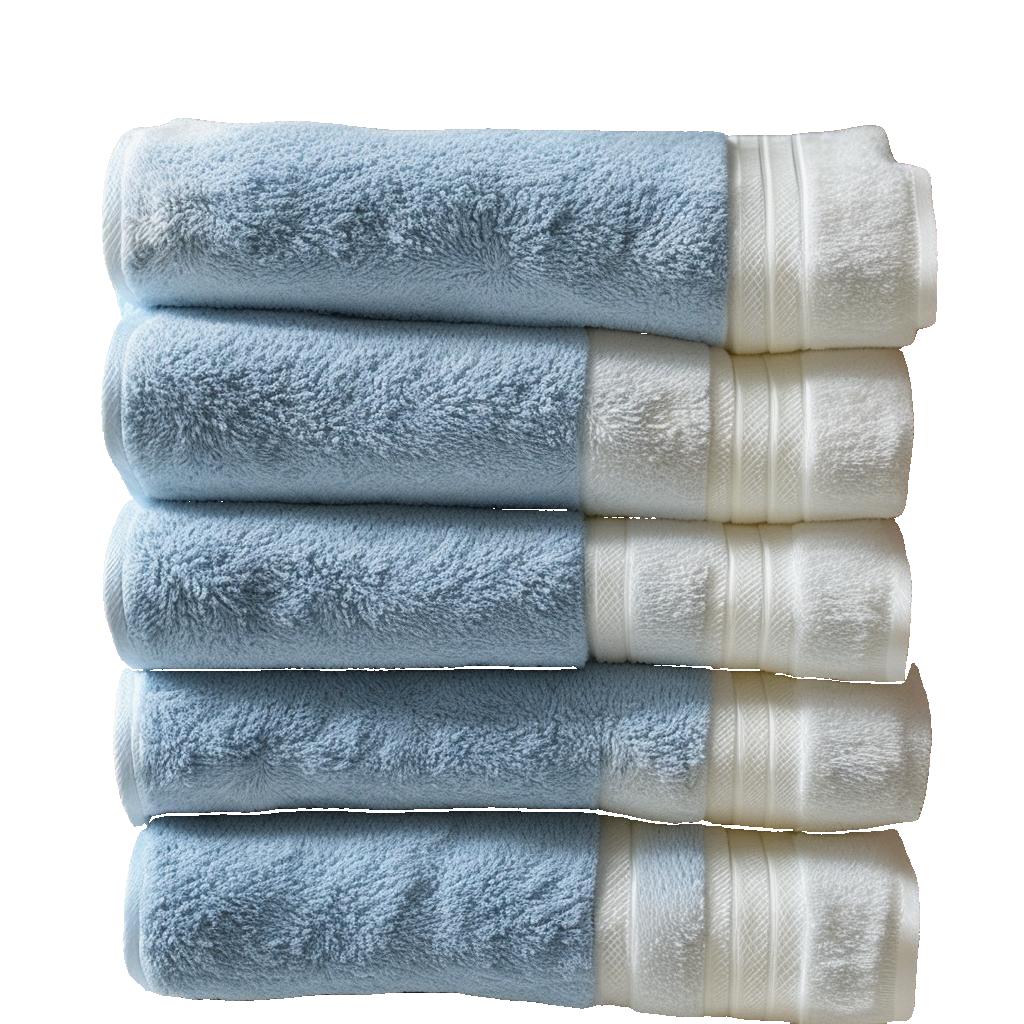} 
                    & \fontsize{26}{26}\selectfont{\emph{Wood}}
                    & \includegraphics[width=7cm]{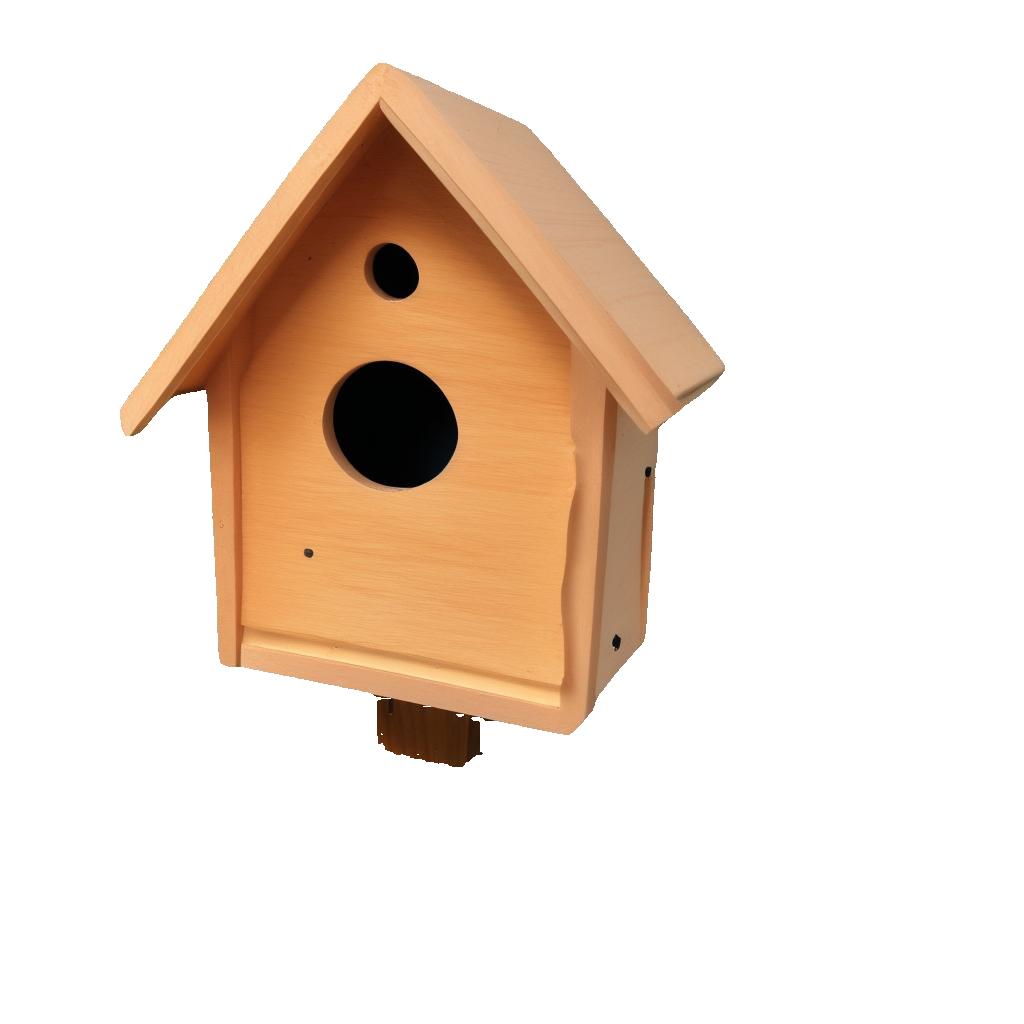} 
                    & \includegraphics[width=7cm]{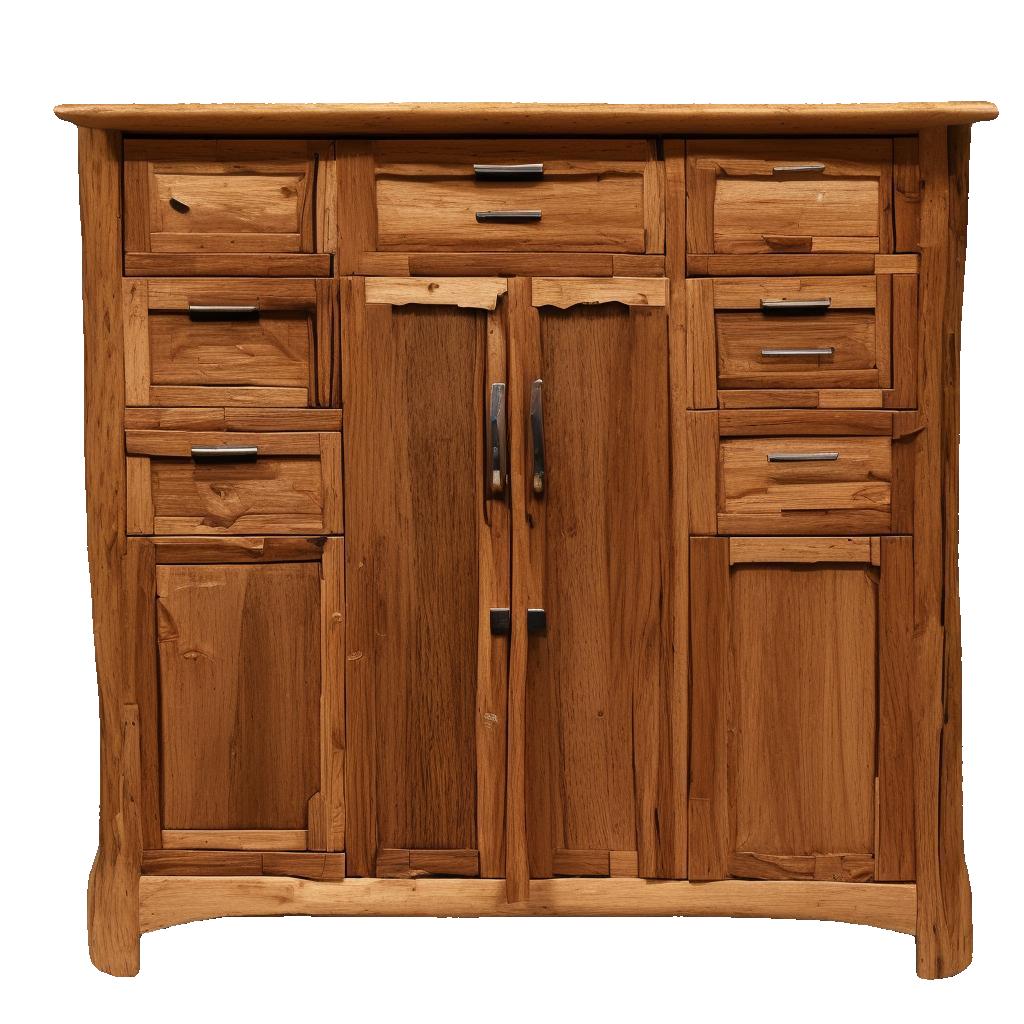} 
                    & \includegraphics[width=7cm]{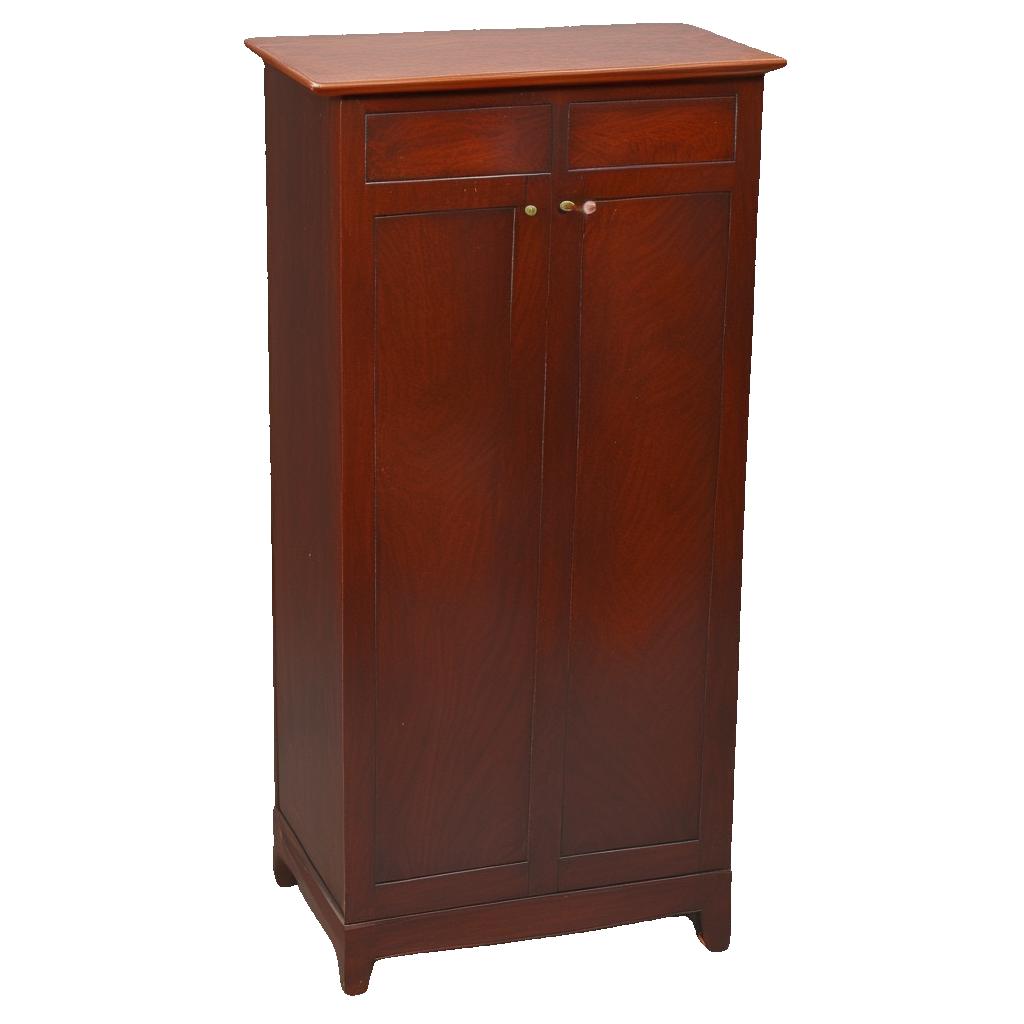} \\

                    & \fontsize{26}{26}\selectfont{\emph{blanket}} &  \fontsize{26}{26}\selectfont{\emph{underwear}}& \fontsize{26}{26}\selectfont{\emph{hand-towel}} & & \fontsize{26}{26}\selectfont{\emph{birdhouse}} & \fontsize{26}{26}\selectfont{\emph{cabinet}} & 
                \fontsize{26}{26}\selectfont{\emph{cabinet}}\\

                 \fontsize{26}{26}\selectfont{\emph{Stone}}
                    &\includegraphics[width=7cm]{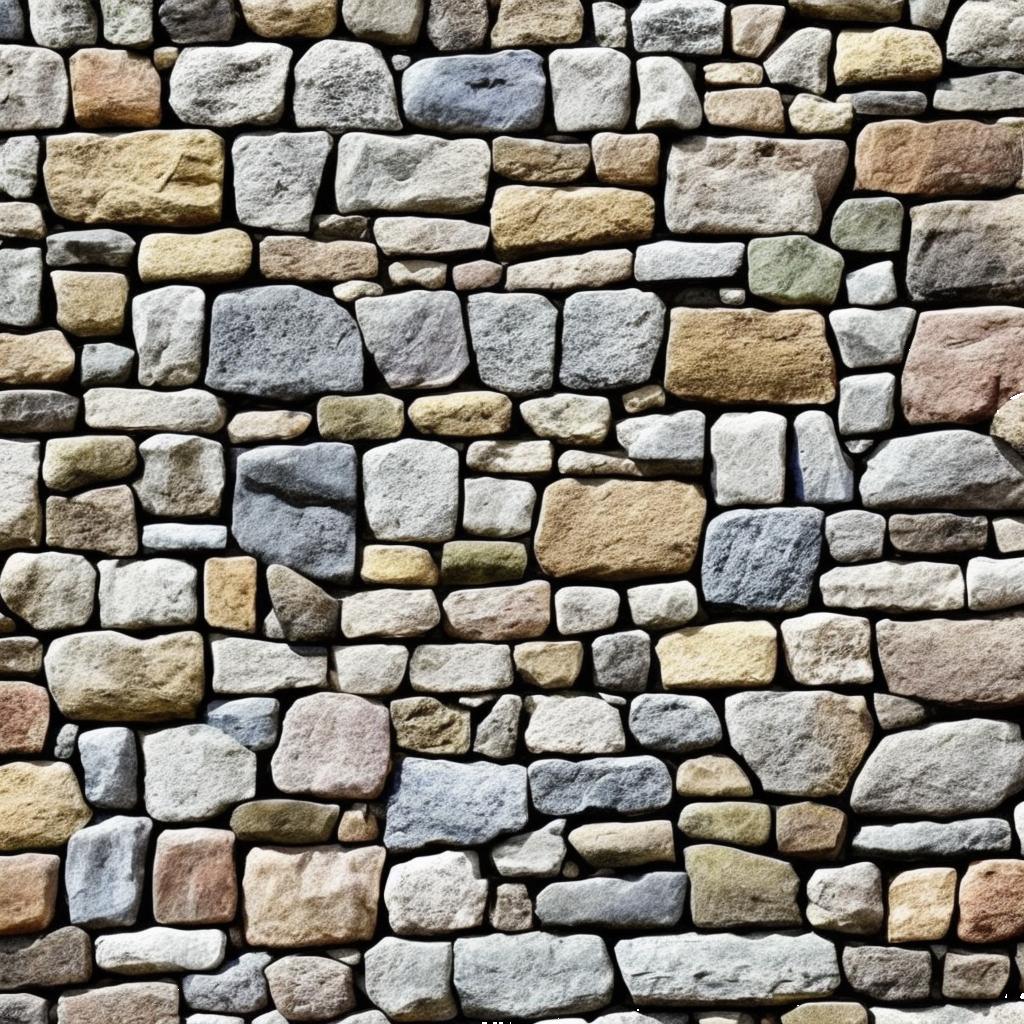} 
                 & \includegraphics[width=7cm]{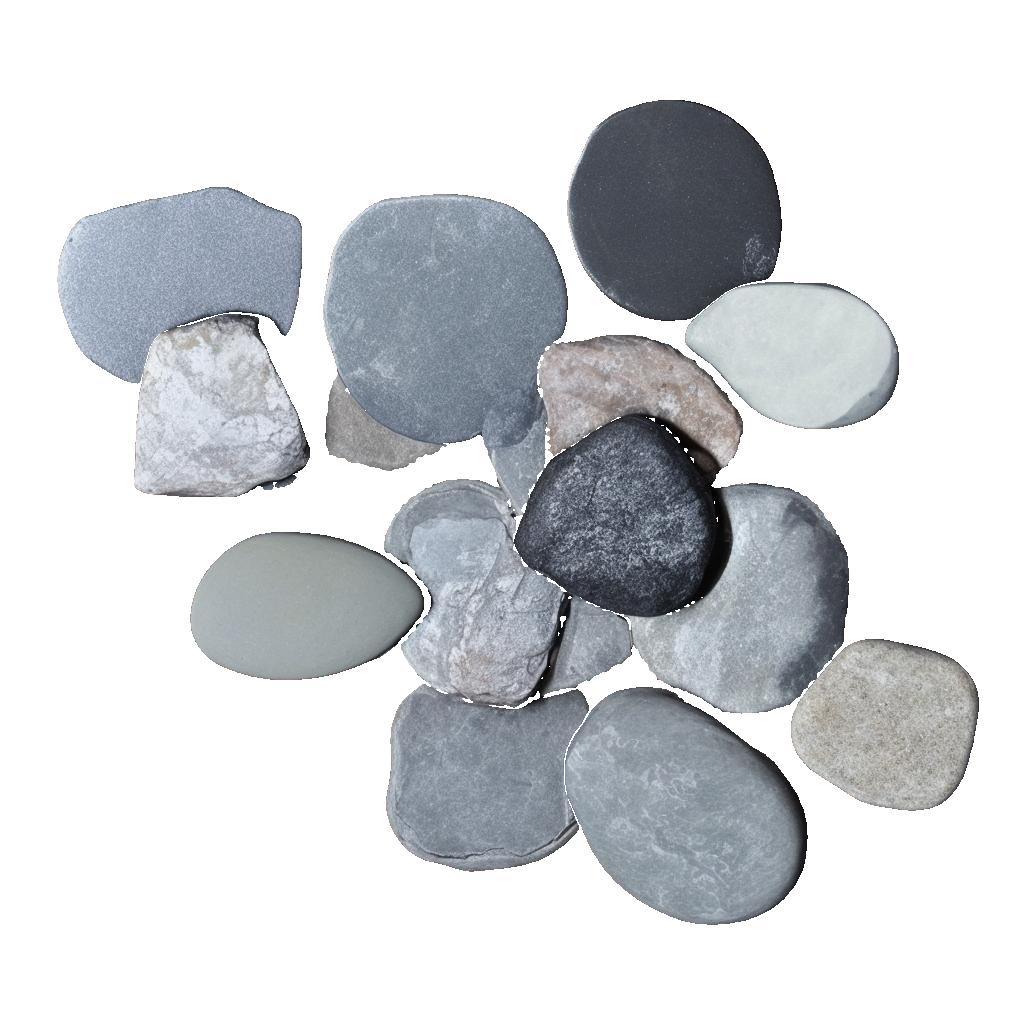} 
                    & \includegraphics[width=7cm]{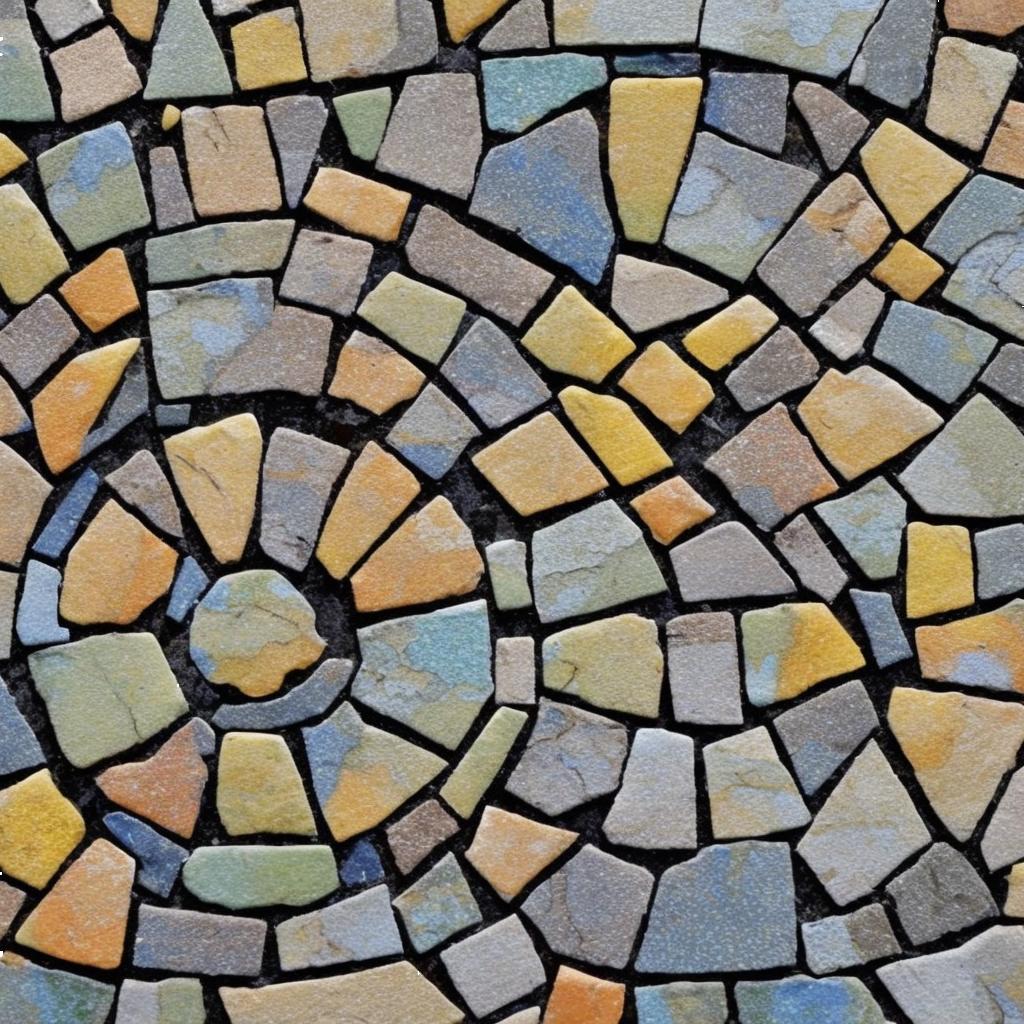} 
                    & \fontsize{26}{26}\selectfont{\emph{Ceramic}}
                    & \includegraphics[width=7cm]{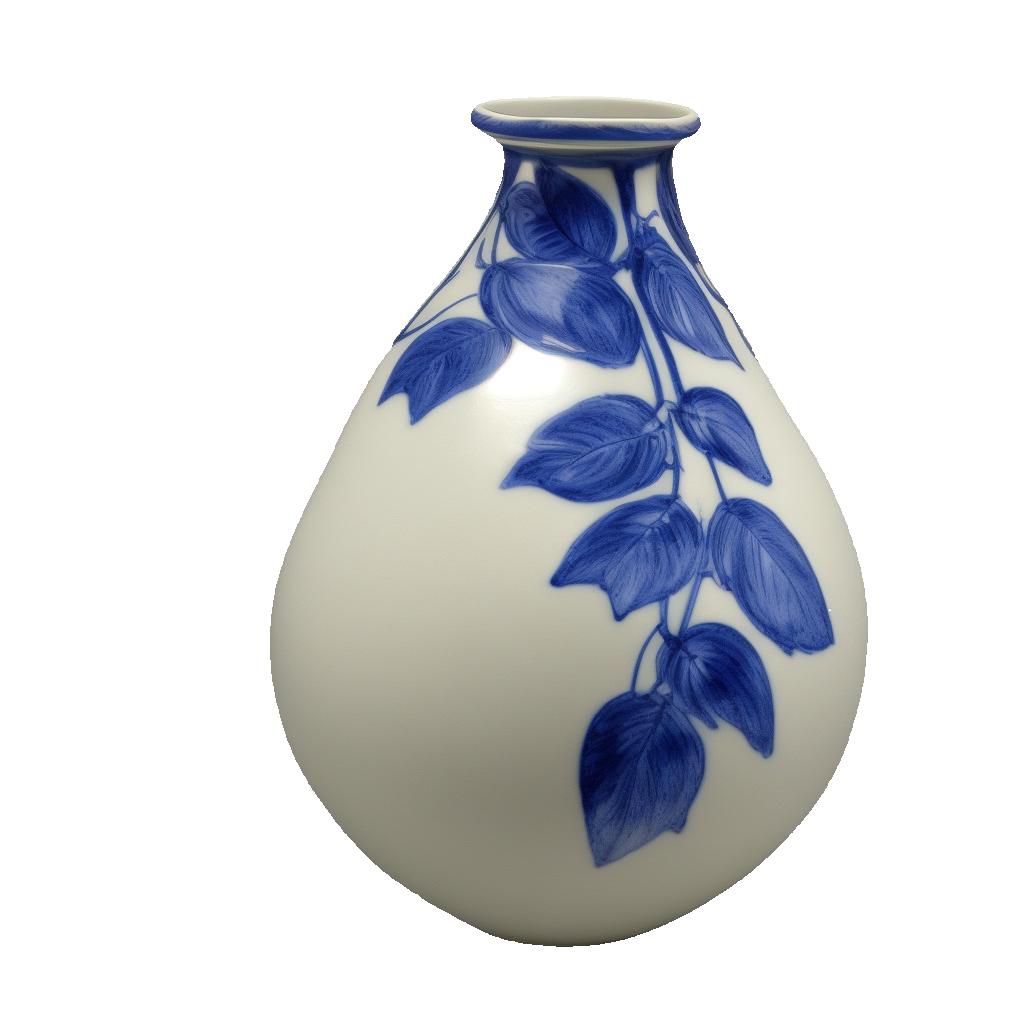} 
                    & \includegraphics[width=7cm]{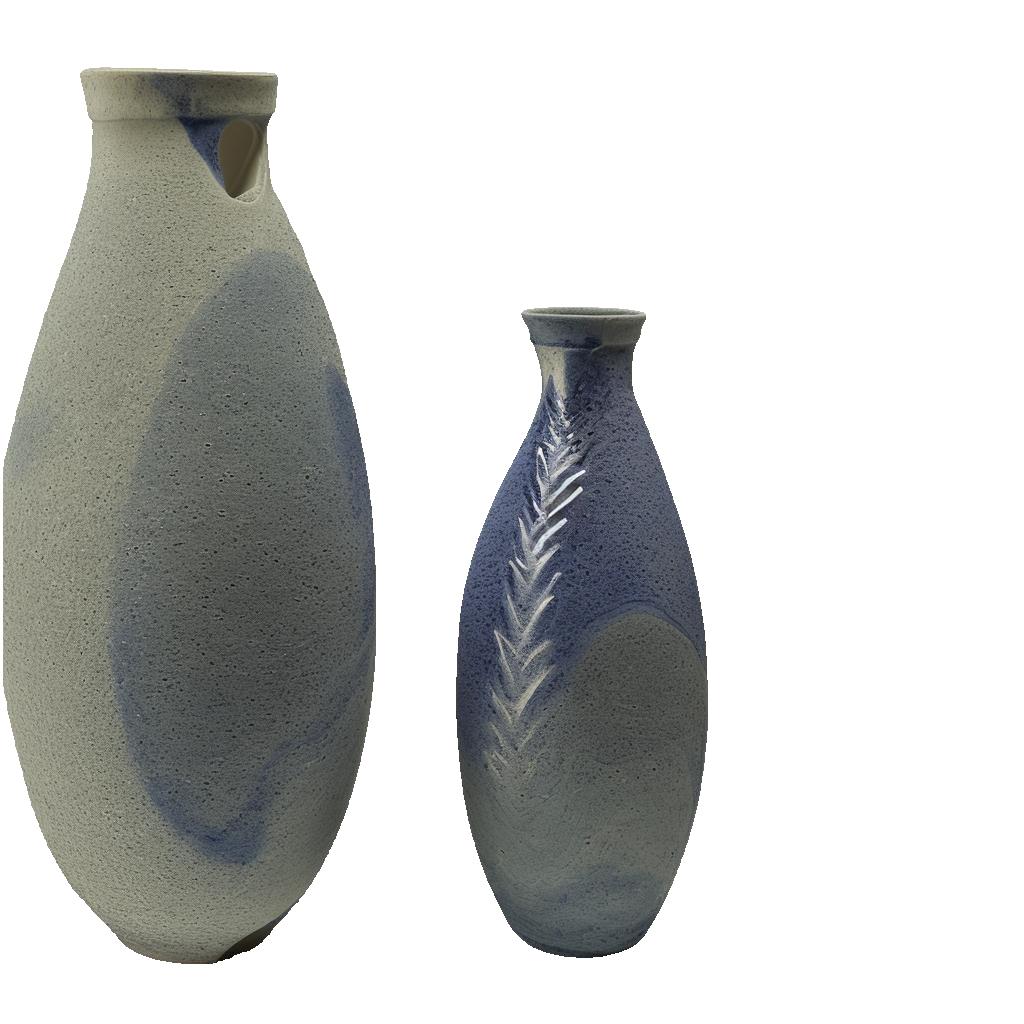} 
                    & \includegraphics[width=7cm]{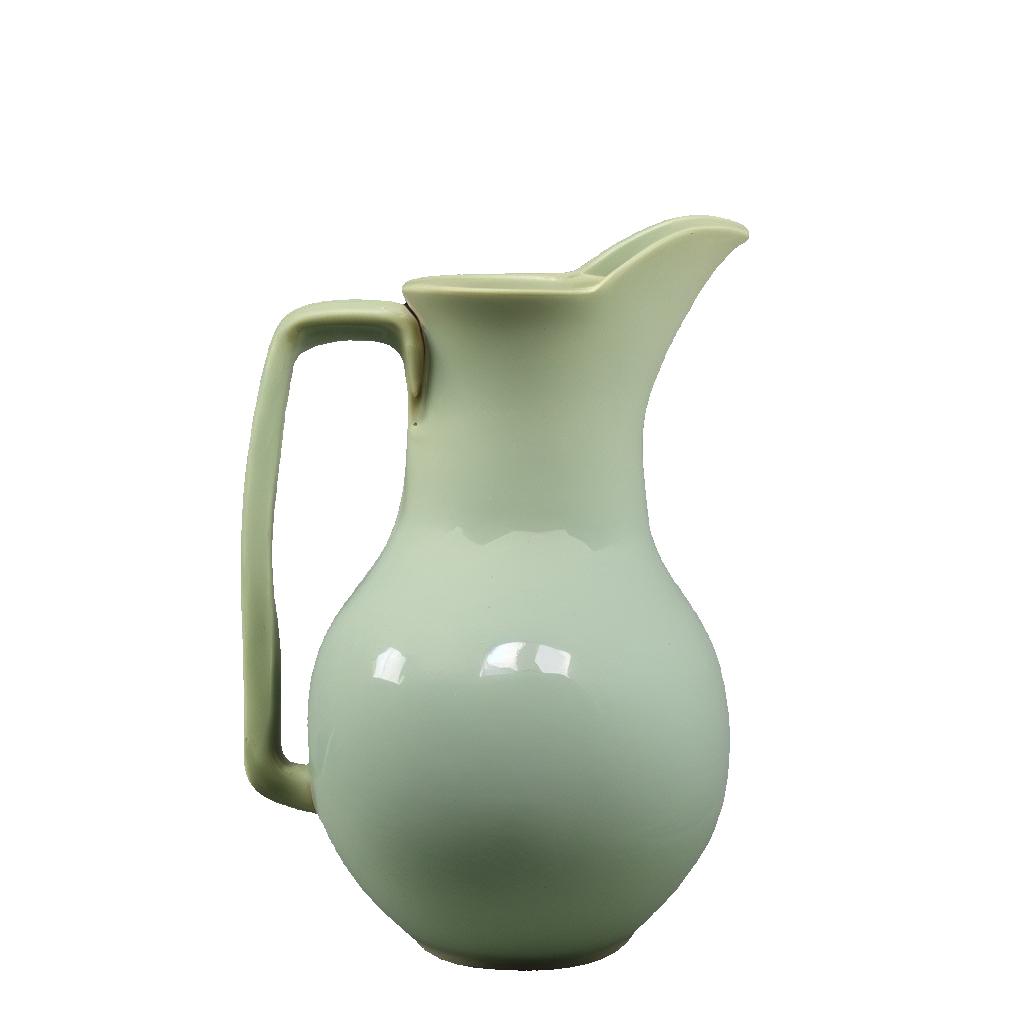} \\

                    & \fontsize{26}{26}\selectfont{\emph{wall}} &  \fontsize{26}{26}\selectfont{\emph{coaster}}& \fontsize{26}{26}\selectfont{\emph{stepping-stone}} & & \fontsize{26}{26}\selectfont{\emph{vase}} & \fontsize{26}{26}\selectfont{\emph{vase}} & 
                \fontsize{26}{26}\selectfont{\emph{pitcher}}\\

                     \fontsize{26}{26}\selectfont{\emph{Bone}}
                    &\includegraphics[width=7cm]{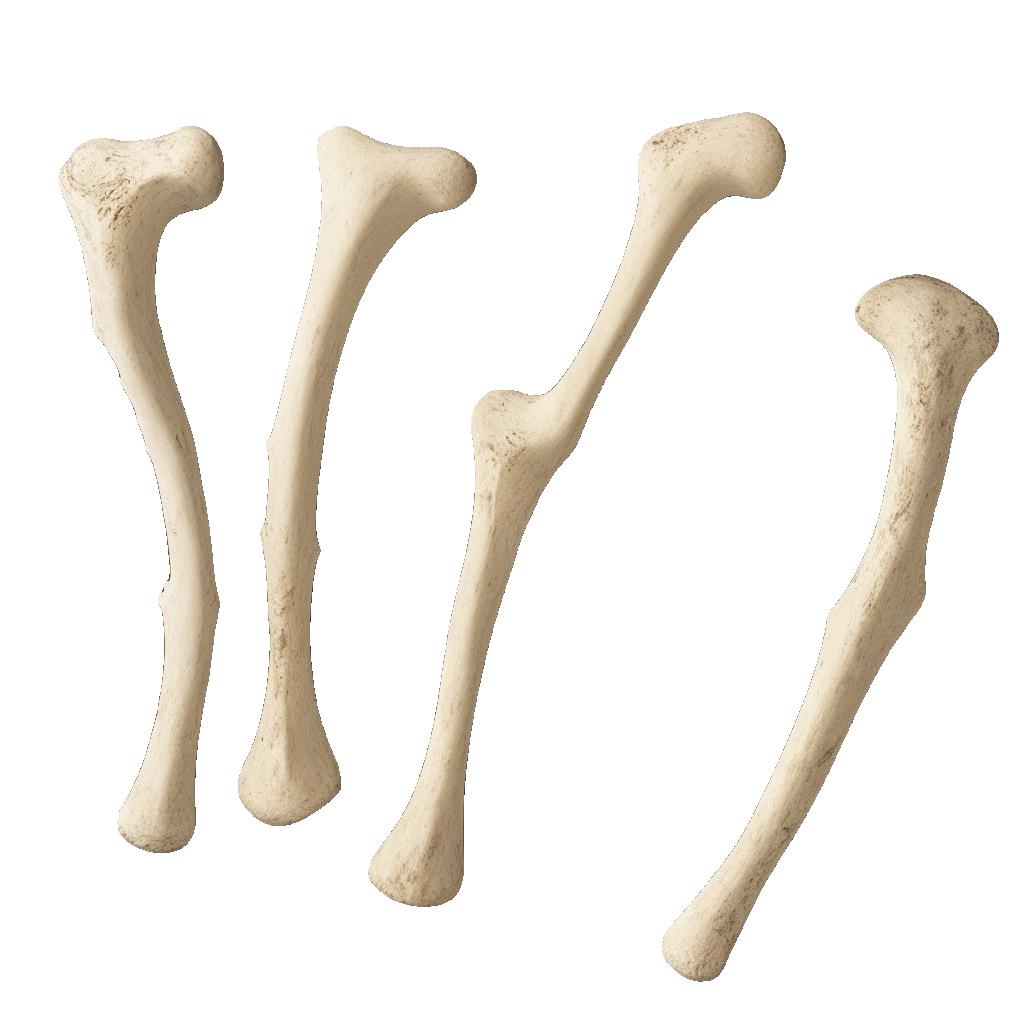} 
                 & \includegraphics[width=7cm]{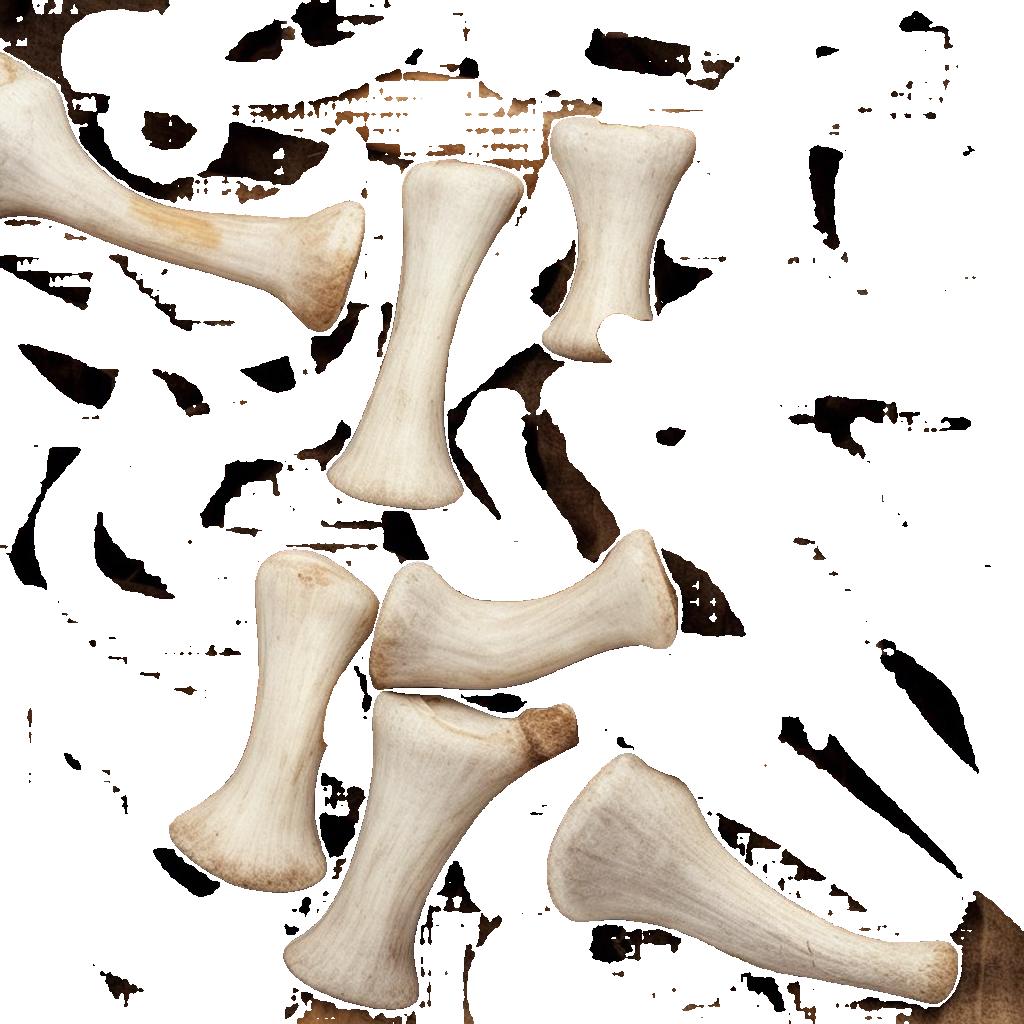} 
                    & \includegraphics[width=7cm]{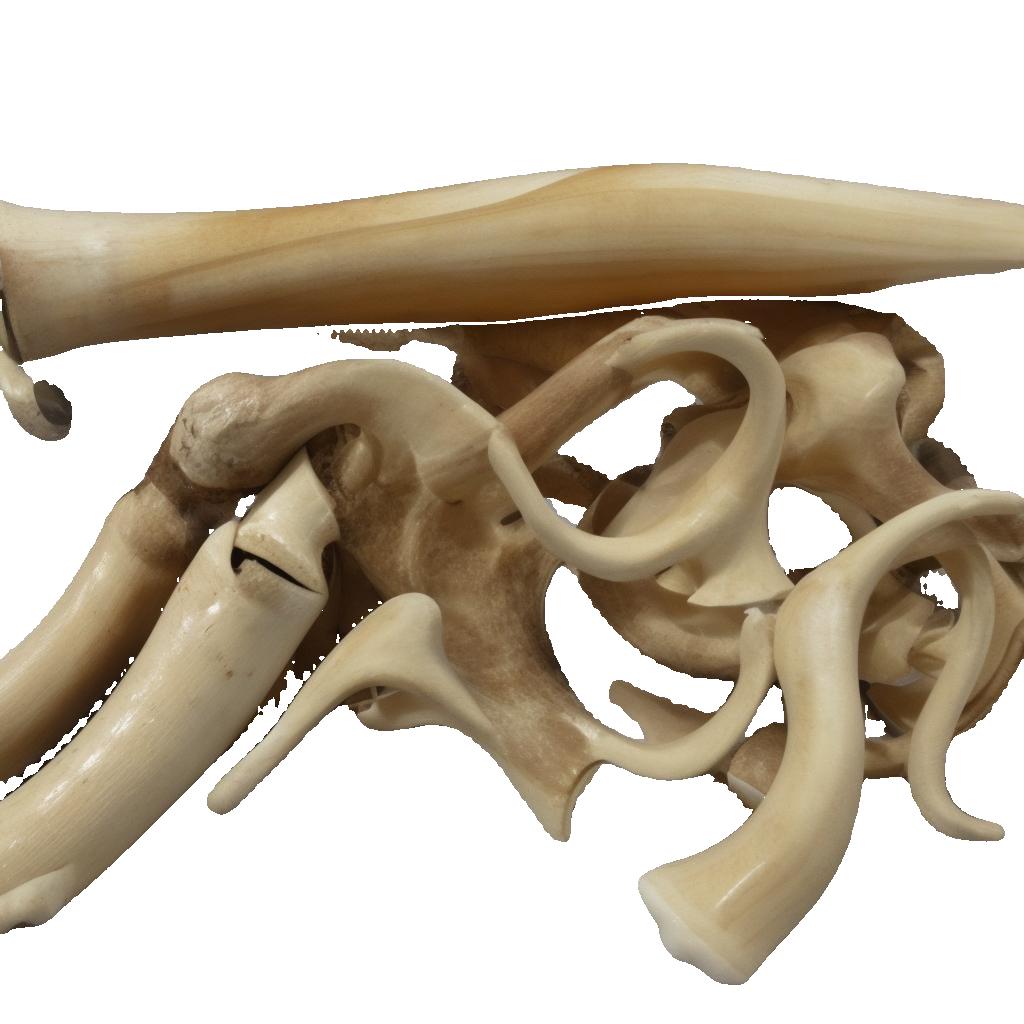} 
                    & \fontsize{26}{26}\selectfont{\emph{Paper}}
                    & \includegraphics[width=7cm]{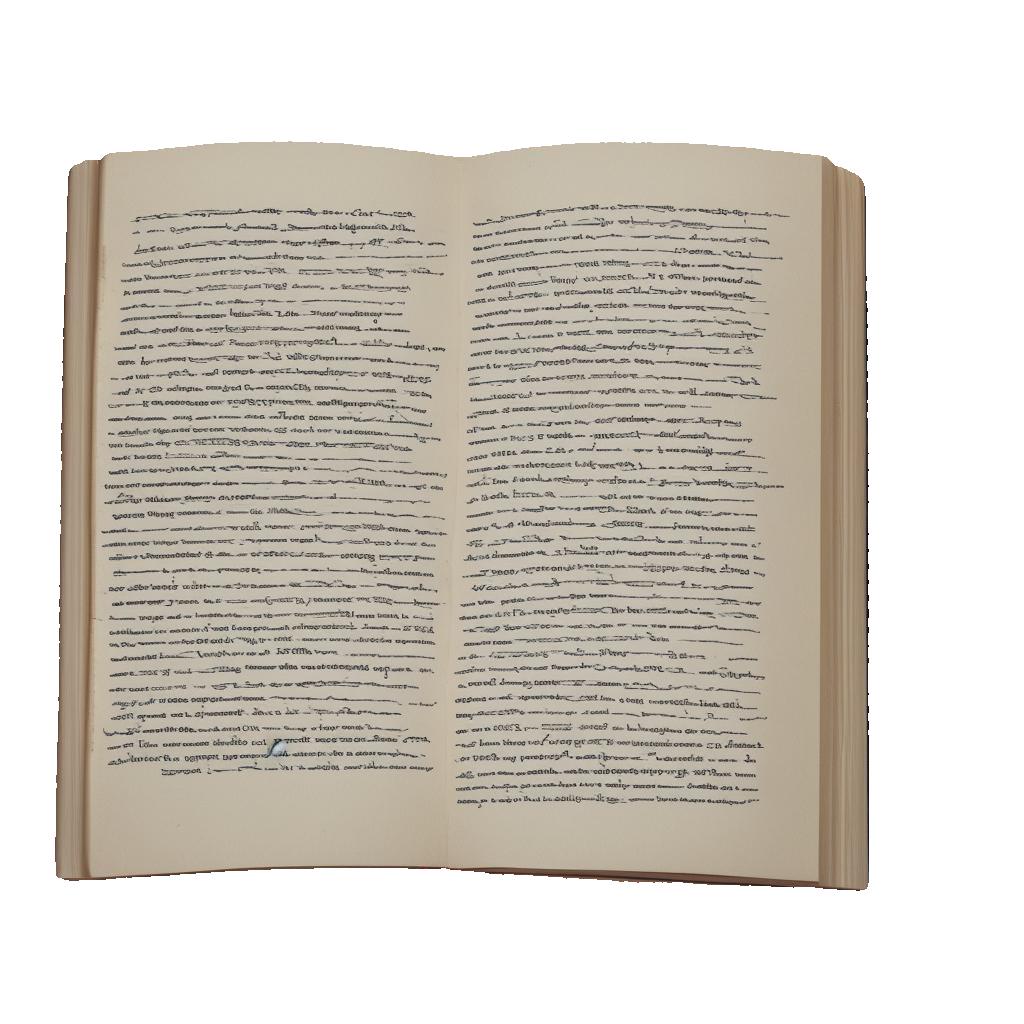} 
                    & \includegraphics[width=7cm]{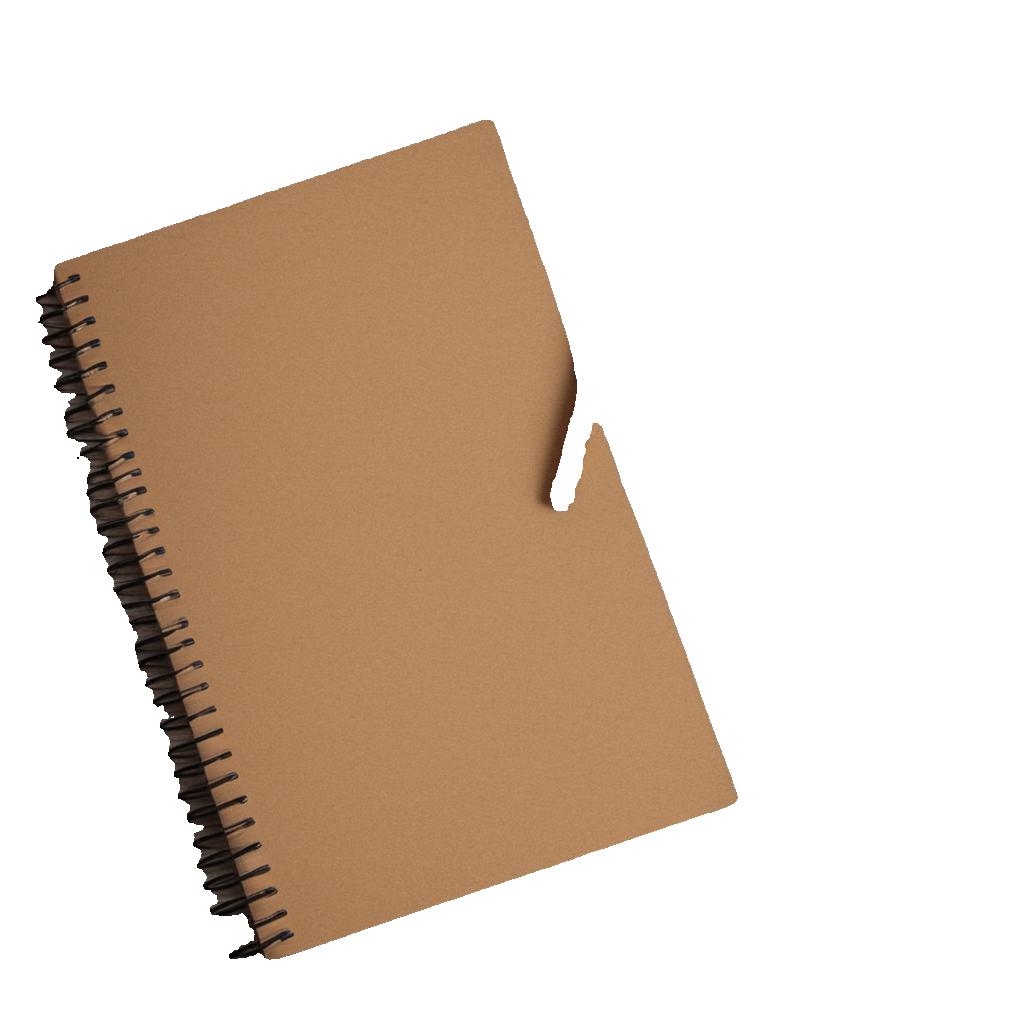} 
                    & \includegraphics[width=7cm]{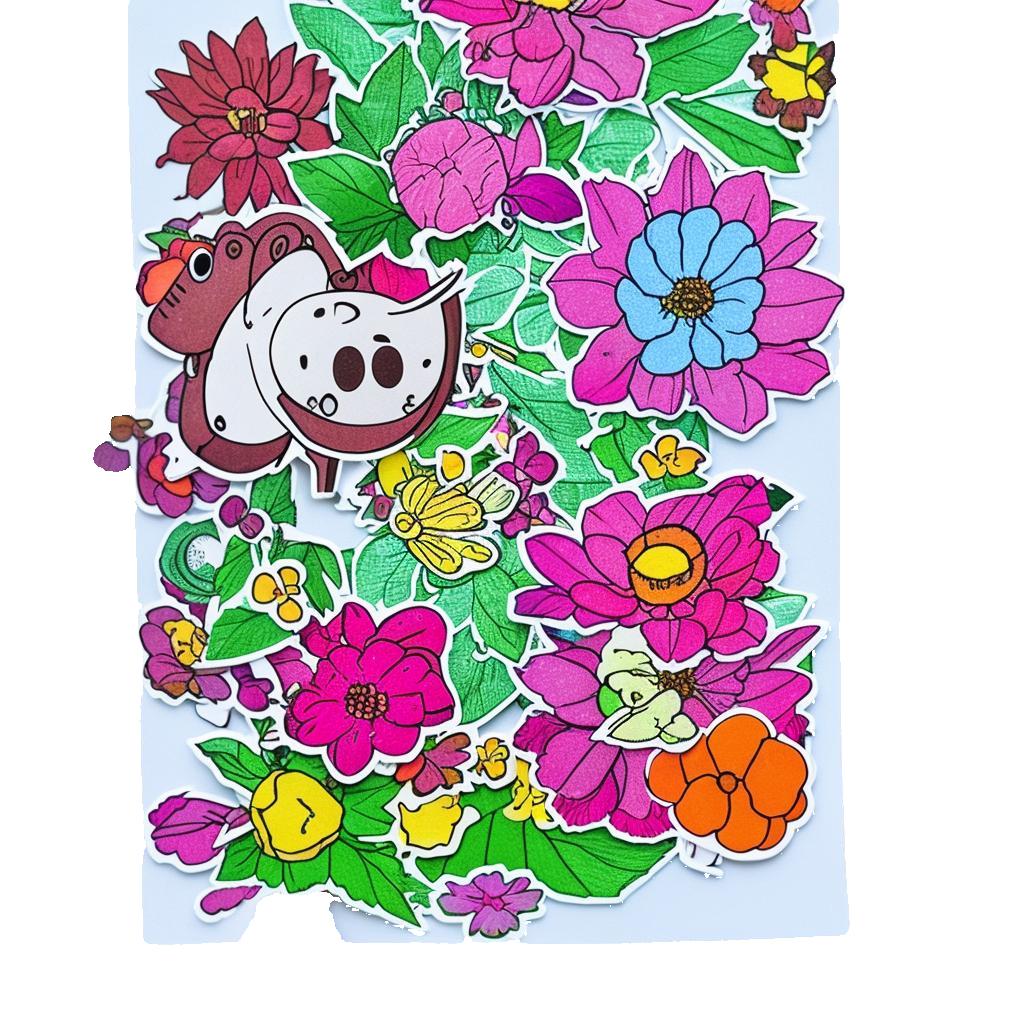} \\

                    & \fontsize{26}{26}\selectfont{\emph{grip}} &  \fontsize{26}{26}\selectfont{\emph{horn}}& \fontsize{26}{26}\selectfont{\emph{horn}} & & \fontsize{26}{26}\selectfont{\emph{book}} & \fontsize{26}{26}\selectfont{\emph{notebook}} & 
                \fontsize{26}{26}\selectfont{\emph{sticker}}\\

                  \fontsize{26}{26}\selectfont{\emph{Soil}}
                    &\includegraphics[width=7cm]{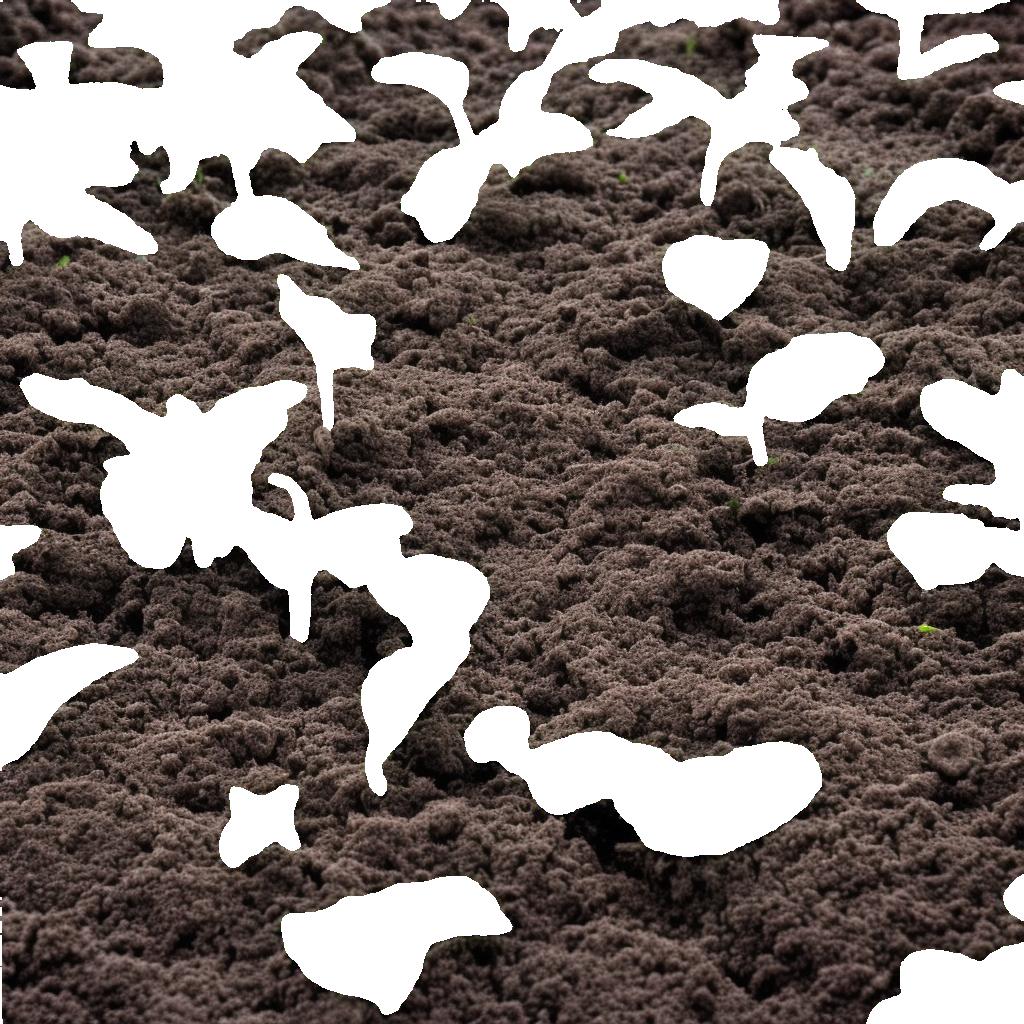} 
                 & \includegraphics[width=7cm]{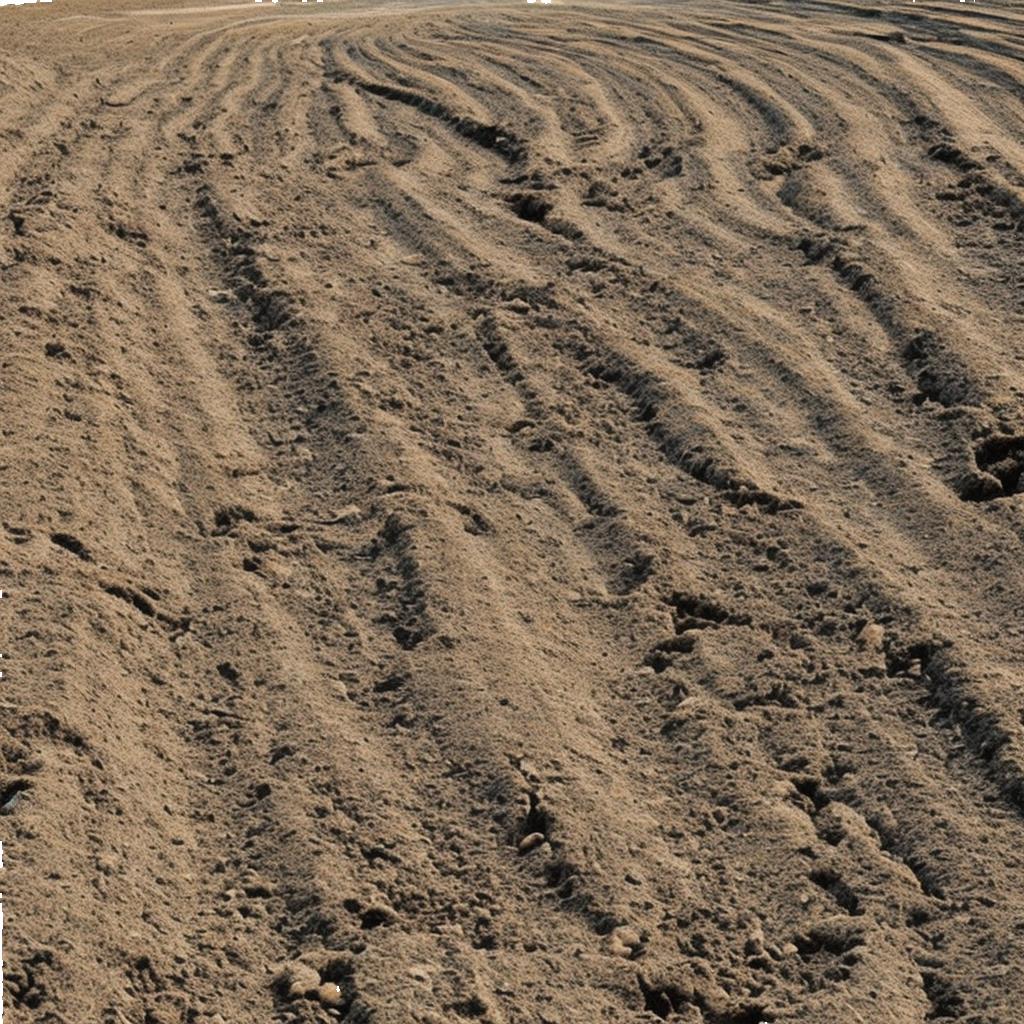} 
                    & \includegraphics[width=7cm]{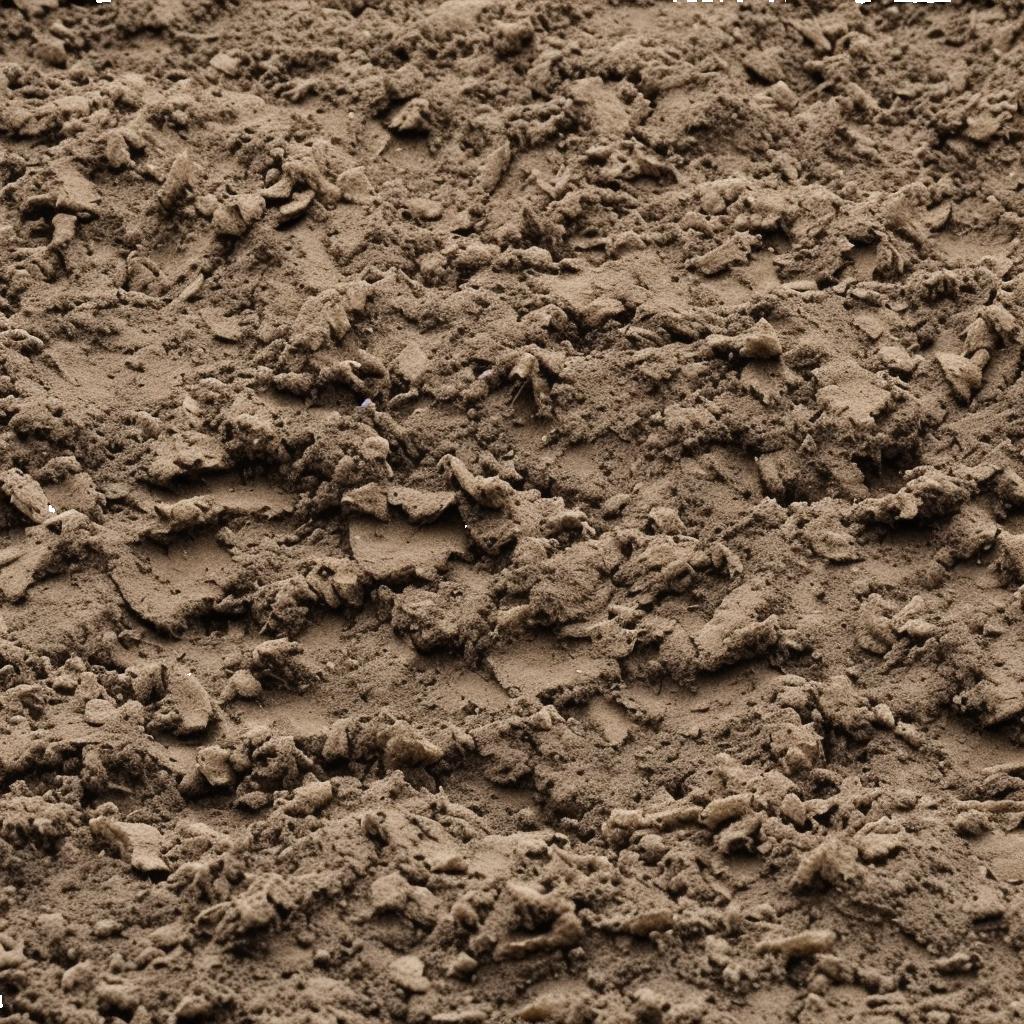} 
                    & \fontsize{26}{26}\selectfont{\emph{Gemstone}}
                    & \includegraphics[width=7cm]{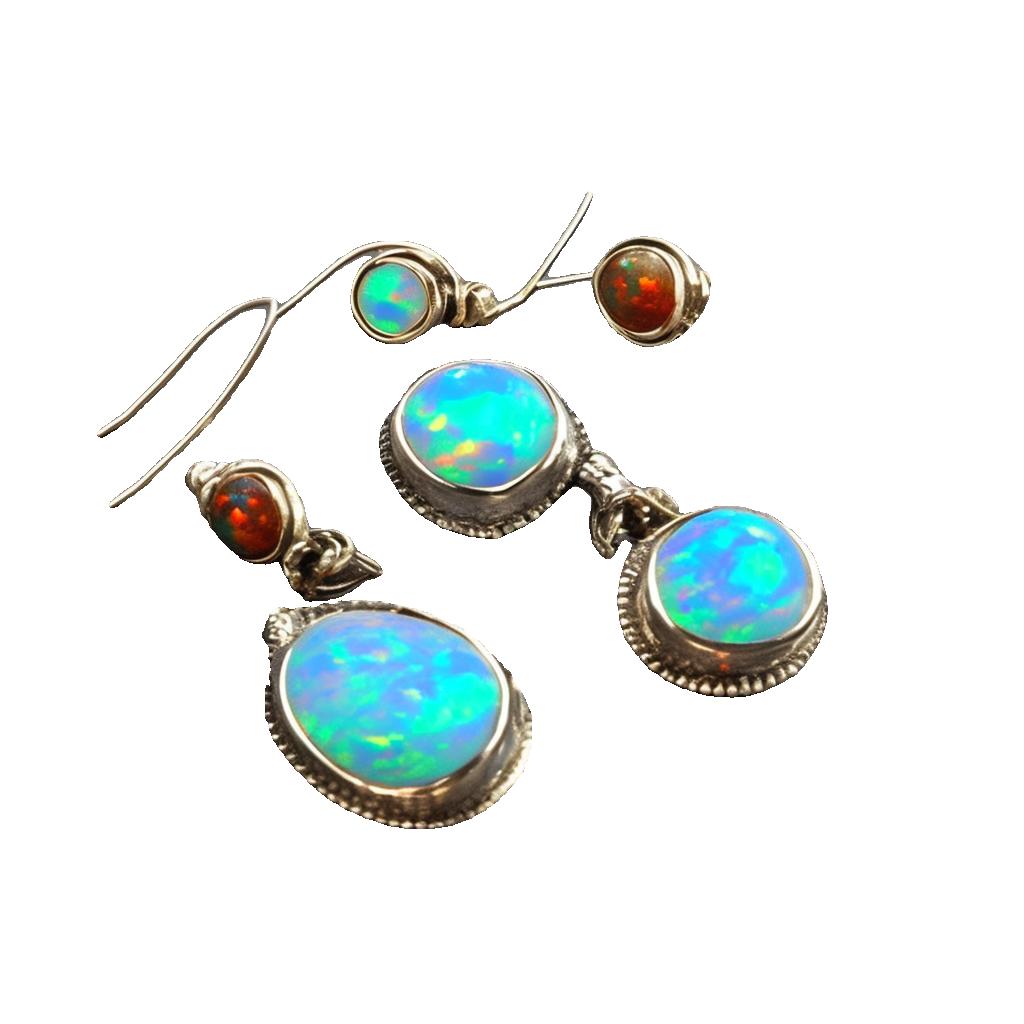} 
                    & \includegraphics[width=7cm]{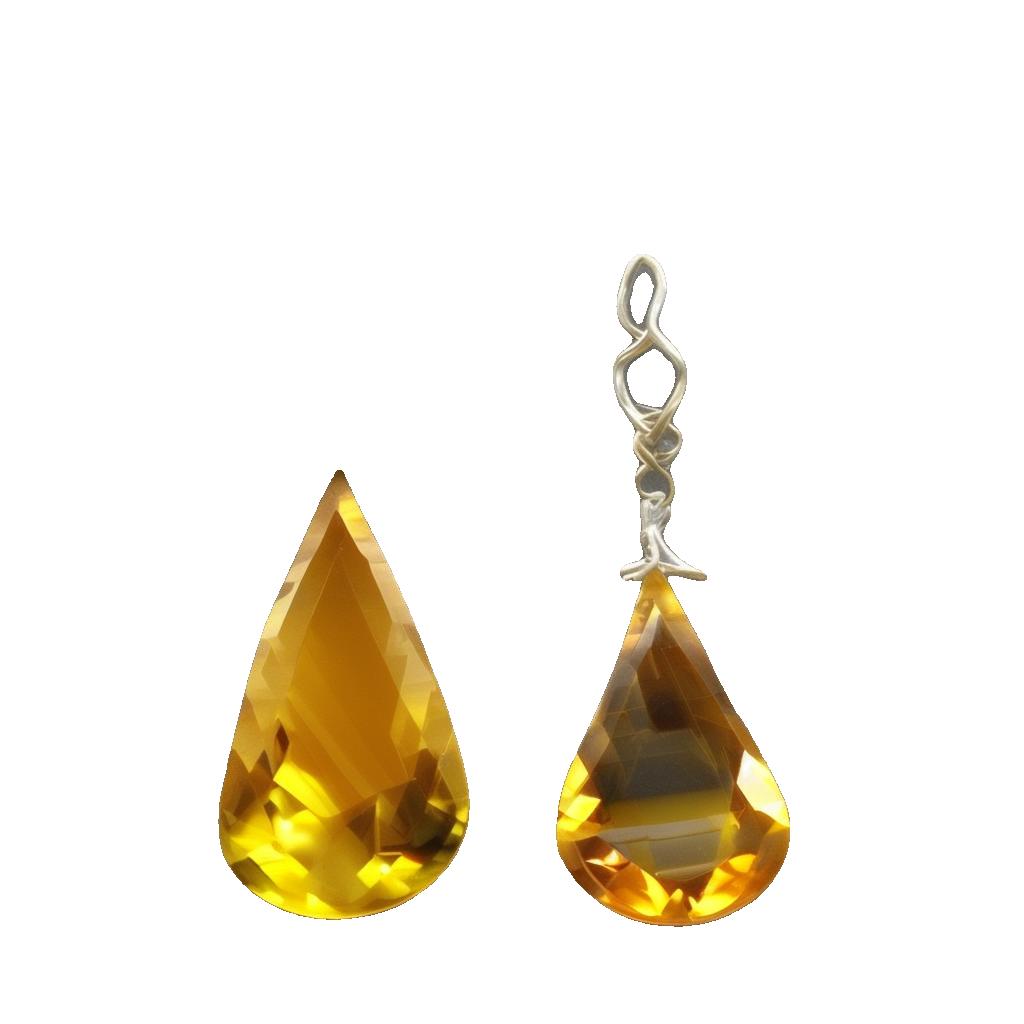} 
                    & \includegraphics[width=7cm]{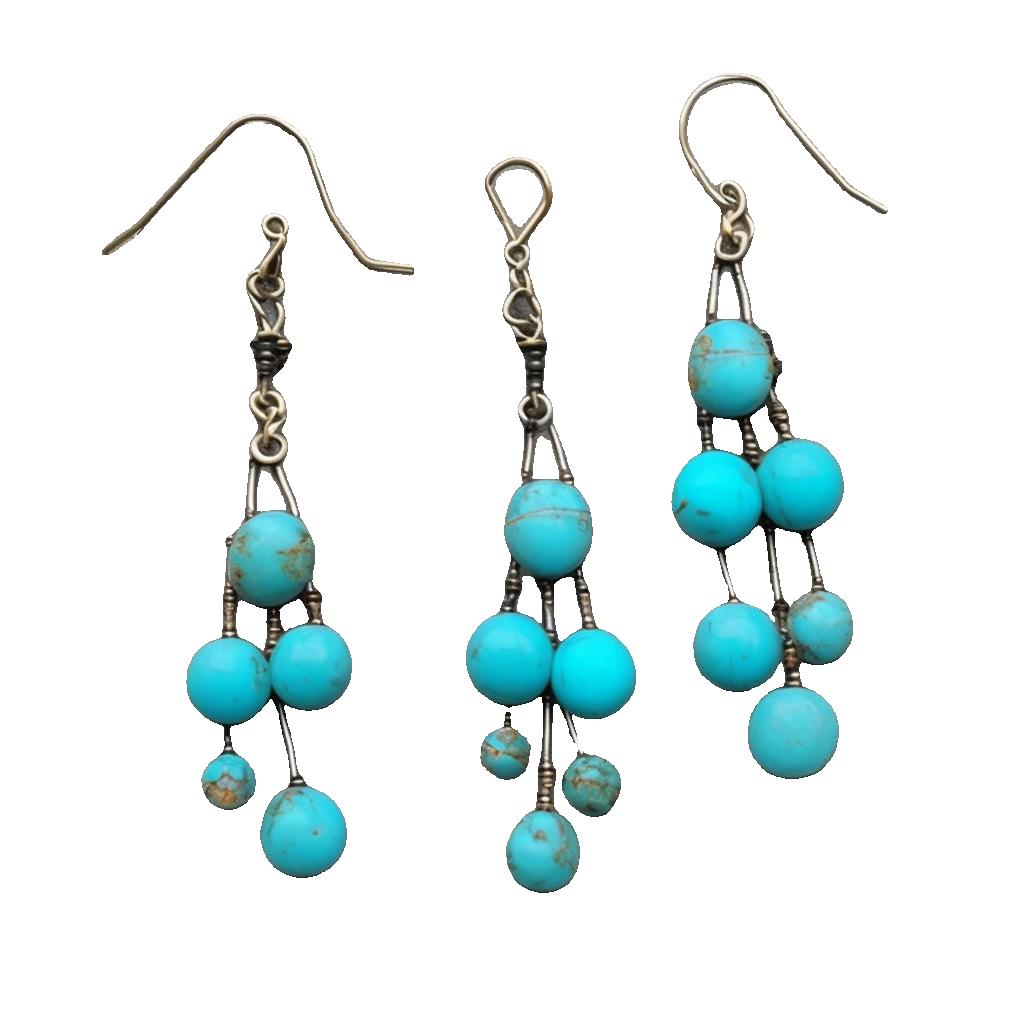} \\

                    & \fontsize{26}{26}\selectfont{\emph{yard}} &  \fontsize{26}{26}\selectfont{\emph{layer}}& \fontsize{26}{26}\selectfont{\emph{layer}} & & \fontsize{26}{26}\selectfont{\emph{earring}} & \fontsize{26}{26}\selectfont{\emph{shard}} & 
                \fontsize{26}{26}\selectfont{\emph{earring}}\\

                  \fontsize{26}{26}\selectfont{\emph{Glass}}
                    &\includegraphics[width=7cm]{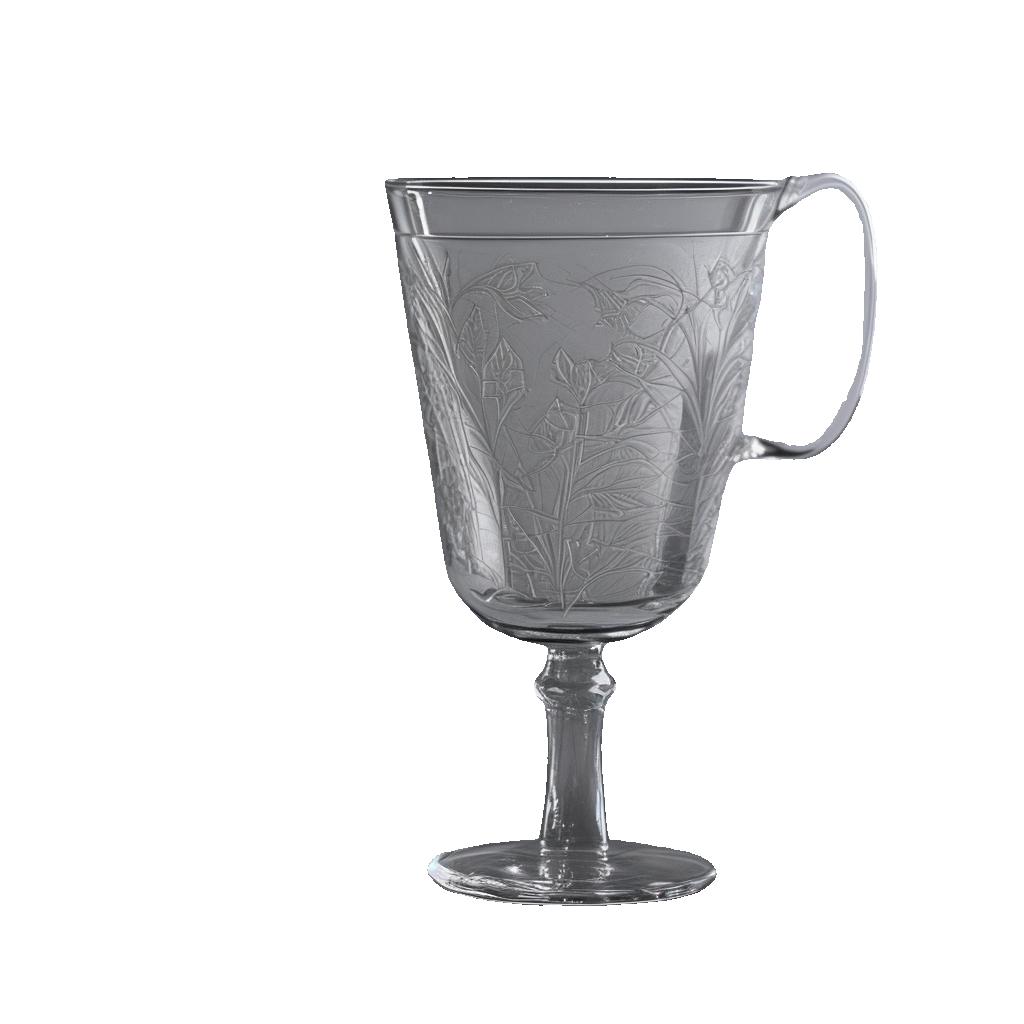} 
                 & \includegraphics[width=7cm]{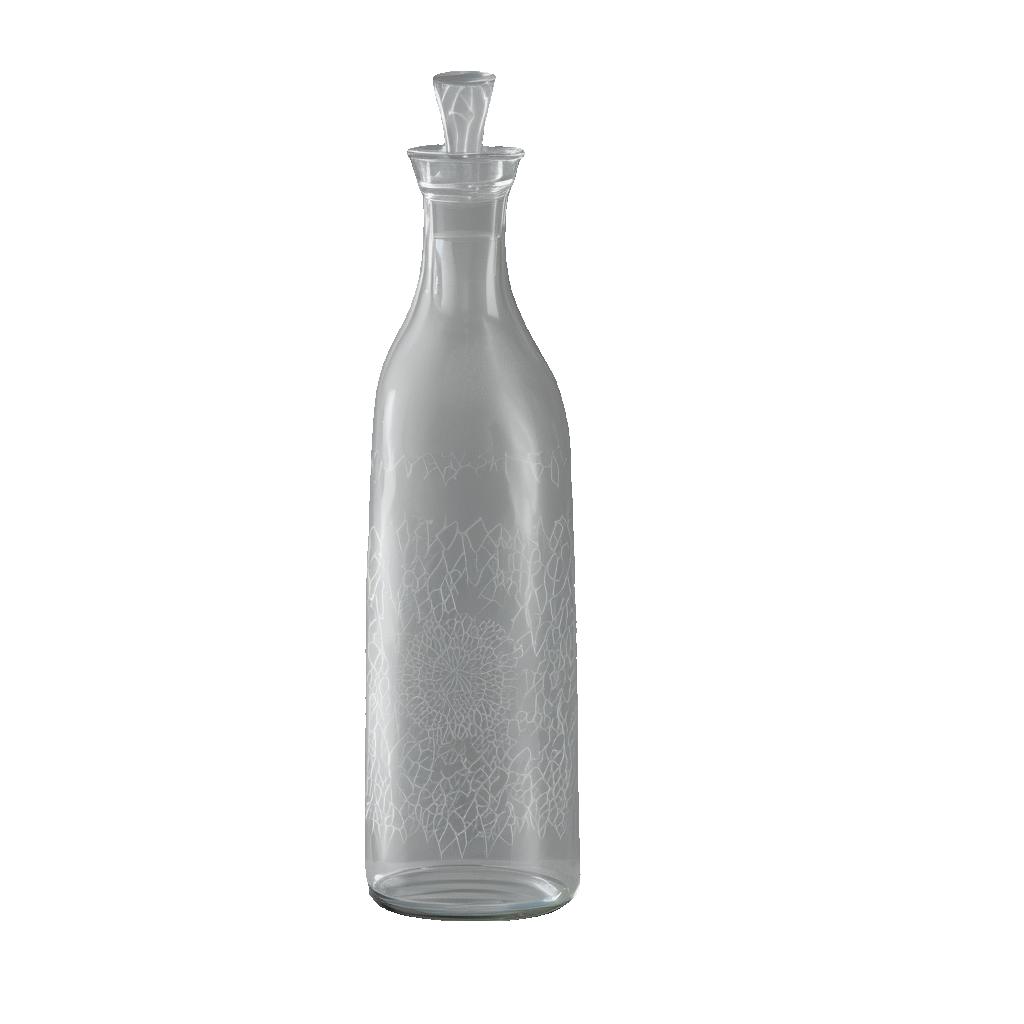} 
                    & \includegraphics[width=7cm]{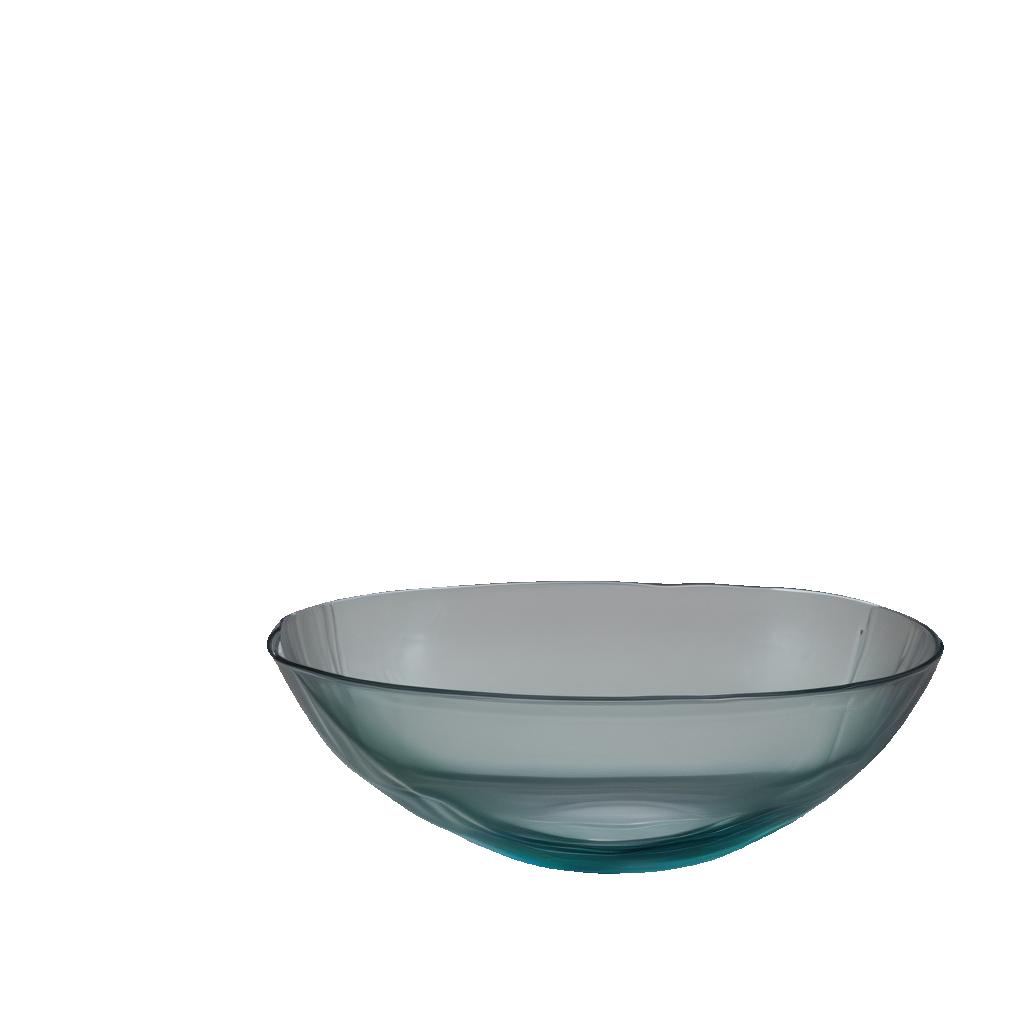} 
                    & \fontsize{26}{26}\selectfont{\emph{Wax}}
                    & \includegraphics[width=7cm]{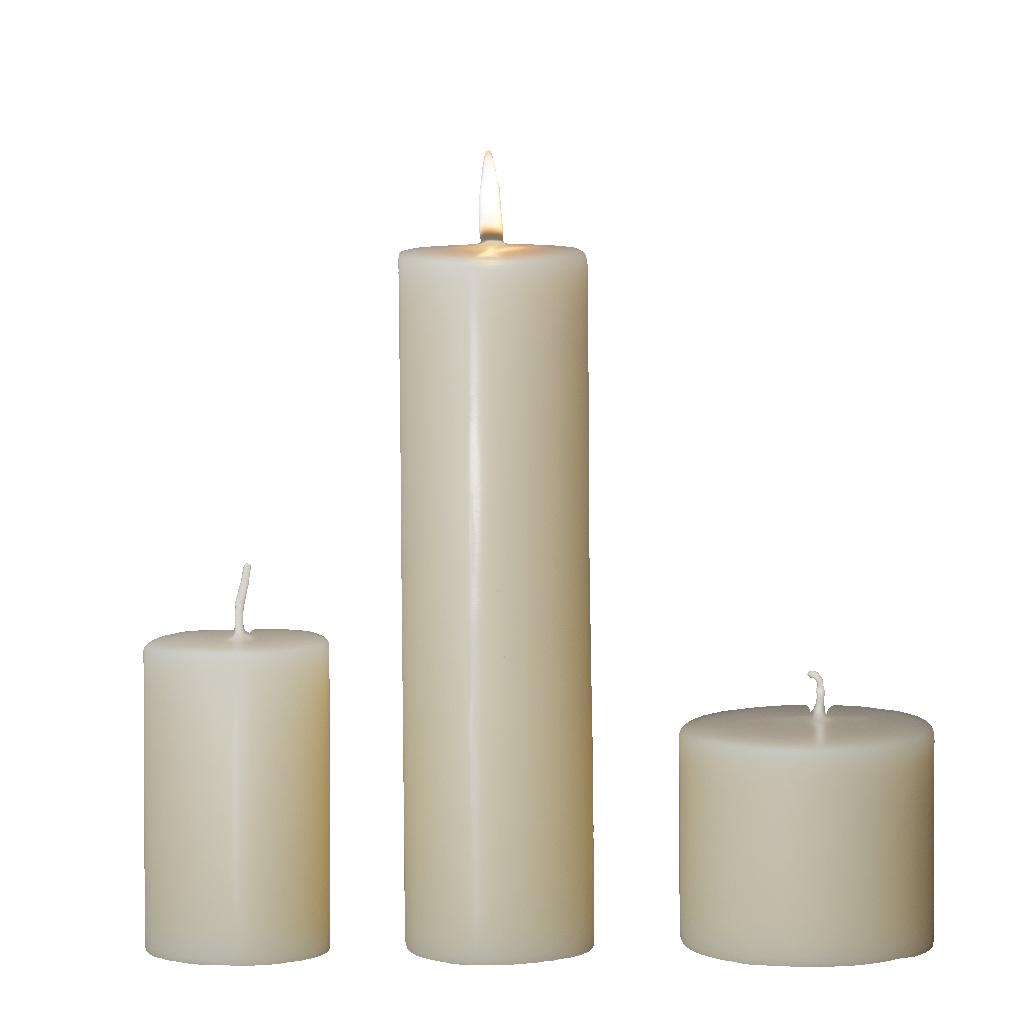} 
                    & \includegraphics[width=7cm]{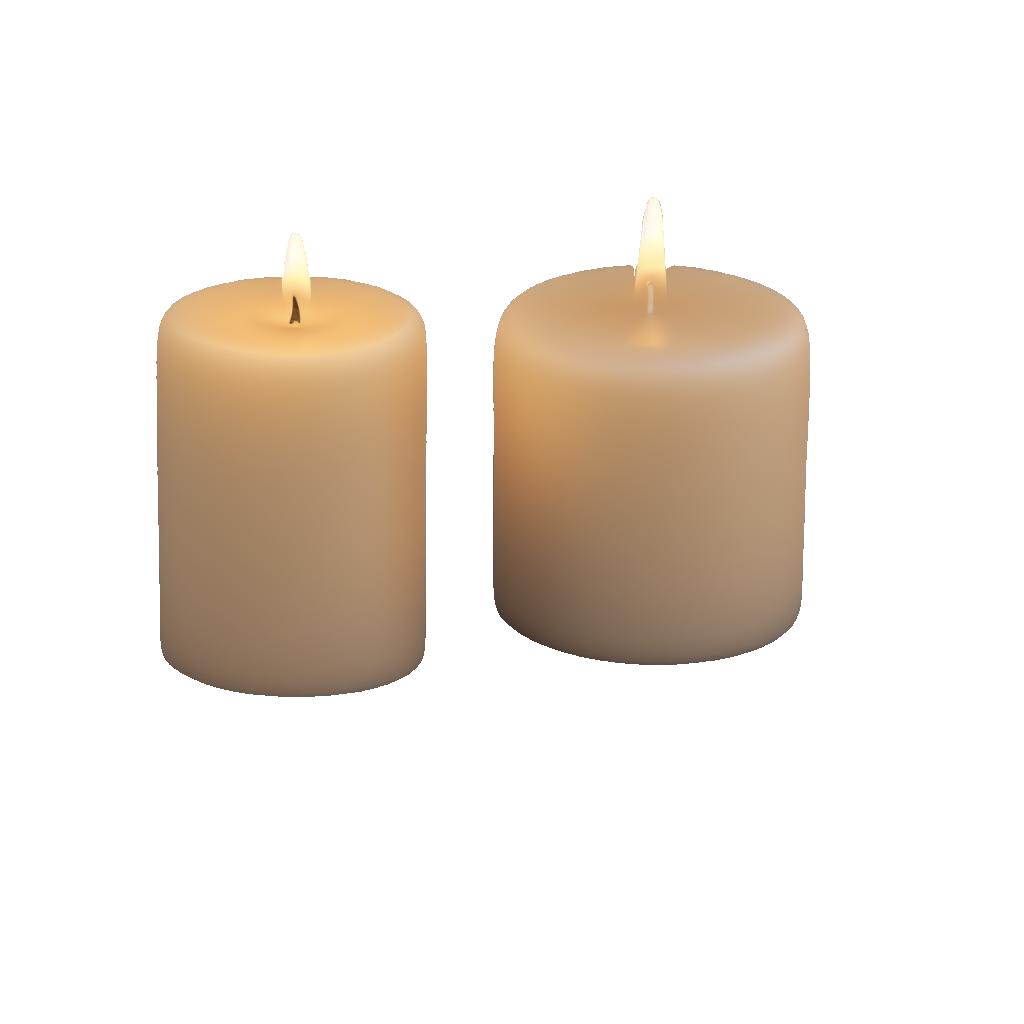} 
                    & \includegraphics[width=7cm]{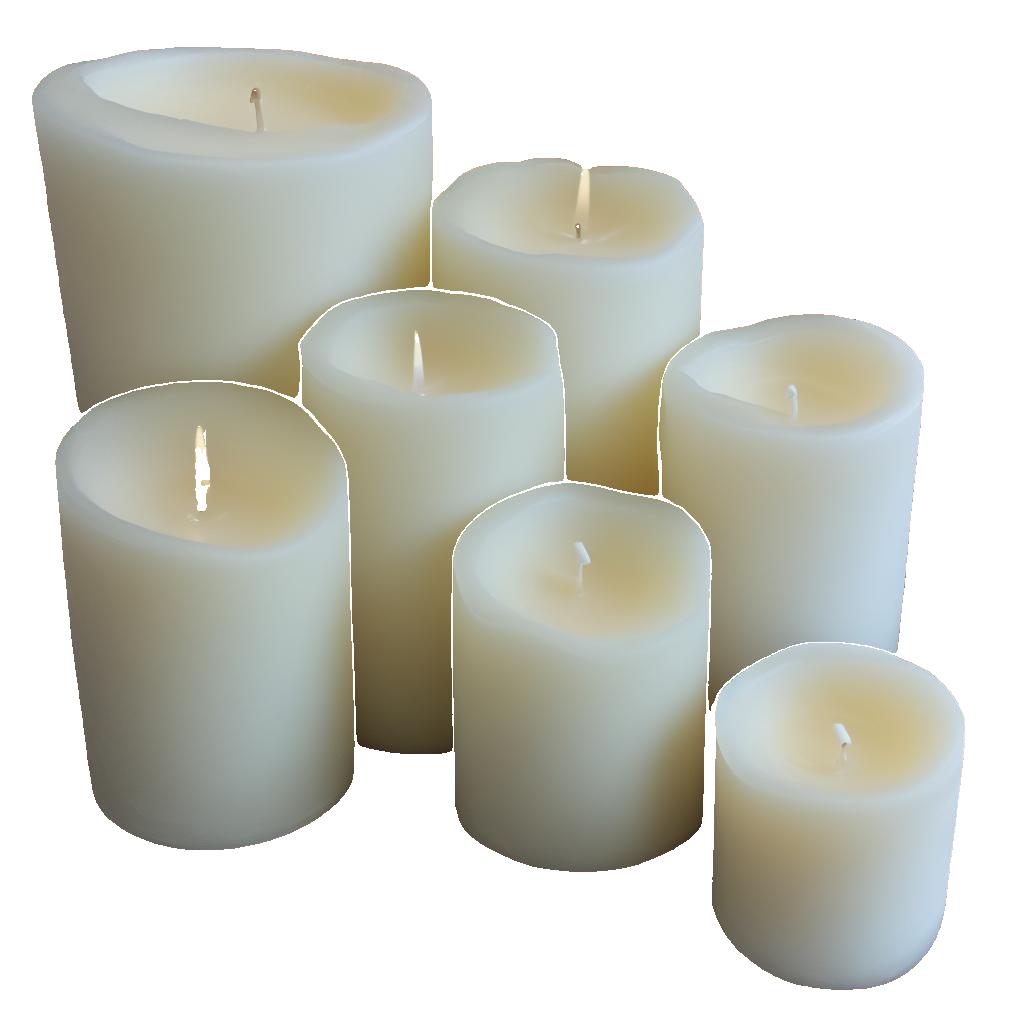} \\

                    & \fontsize{26}{26}\selectfont{\emph{cup}} &  \fontsize{26}{26}\selectfont{\emph{bottle}}& \fontsize{26}{26}\selectfont{\emph{bowl}} & & \fontsize{26}{26}\selectfont{\emph{candle}} & \fontsize{26}{26}\selectfont{\emph{candle}} & 
                \fontsize{26}{26}\selectfont{\emph{candle}}\\

                \end{tabular}

            }

                \captionof{figure}{Samples from our generated dataset and their extracted patches capture local material semantics across 21 classes.}
    \label{fig:figure9}
\vspace{-5mm}
\end{figure*}

\begin{figure*}[ht]
    \resizebox{\textwidth}{!}{%
 \begin{tabular}
                {cccccccc}
                  \fontsize{26}{26}\selectfont{\emph{Concrete}}
                    &\includegraphics[width=7cm]{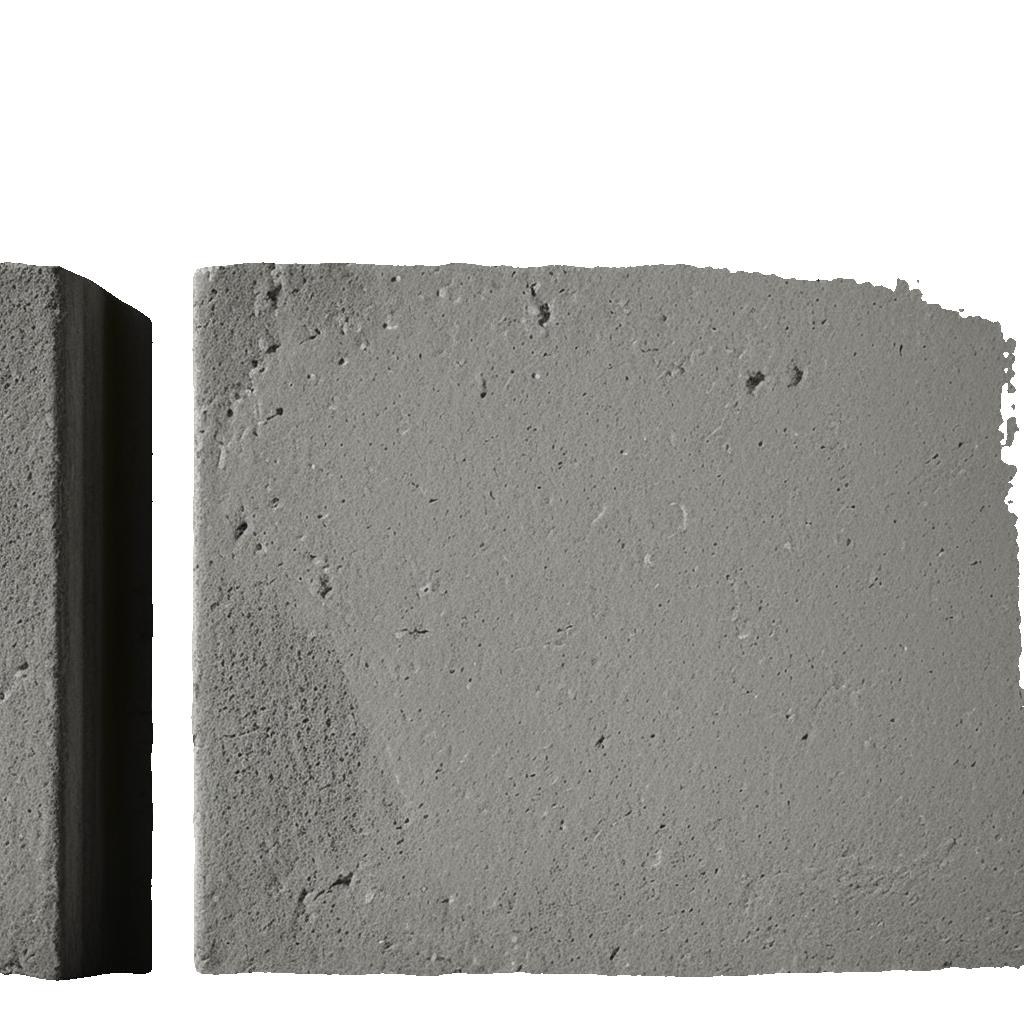} 
                 & \includegraphics[width=7cm]{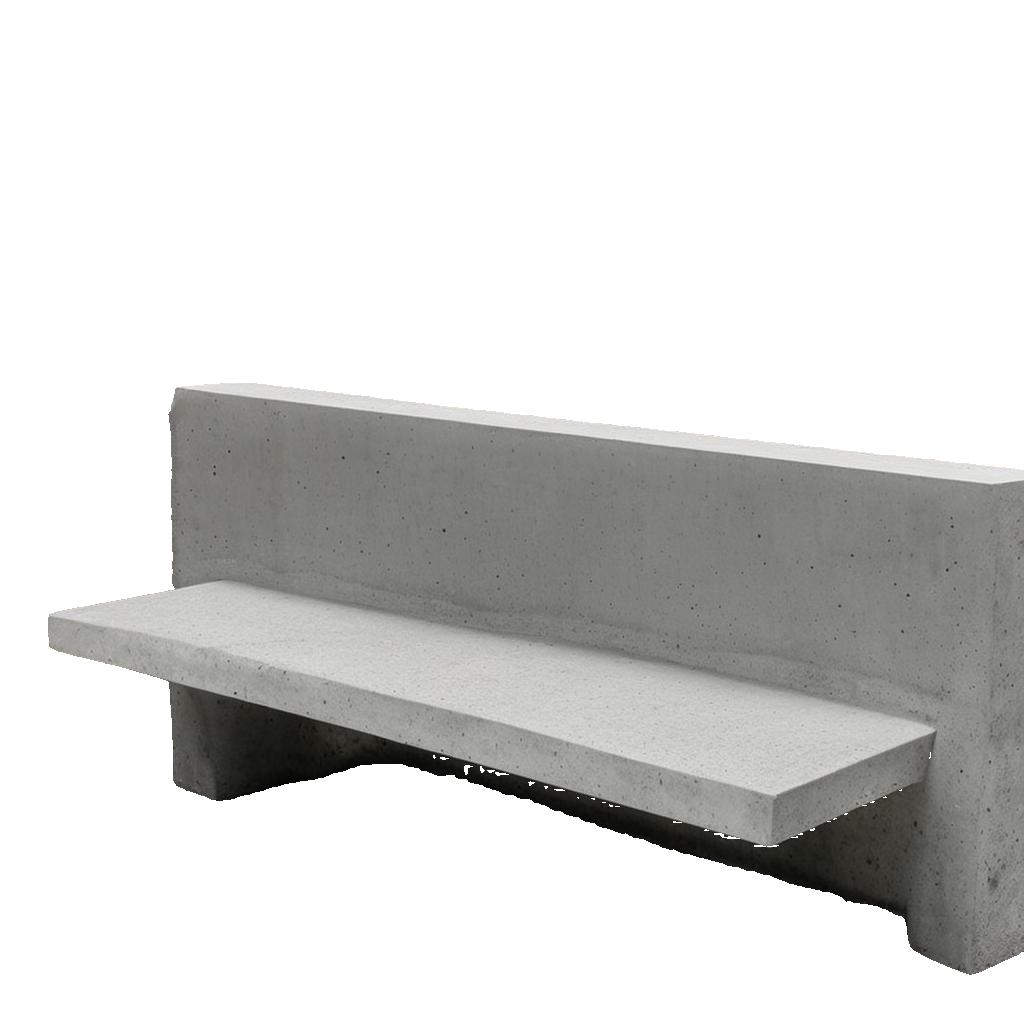} 
                    & \includegraphics[width=7cm]{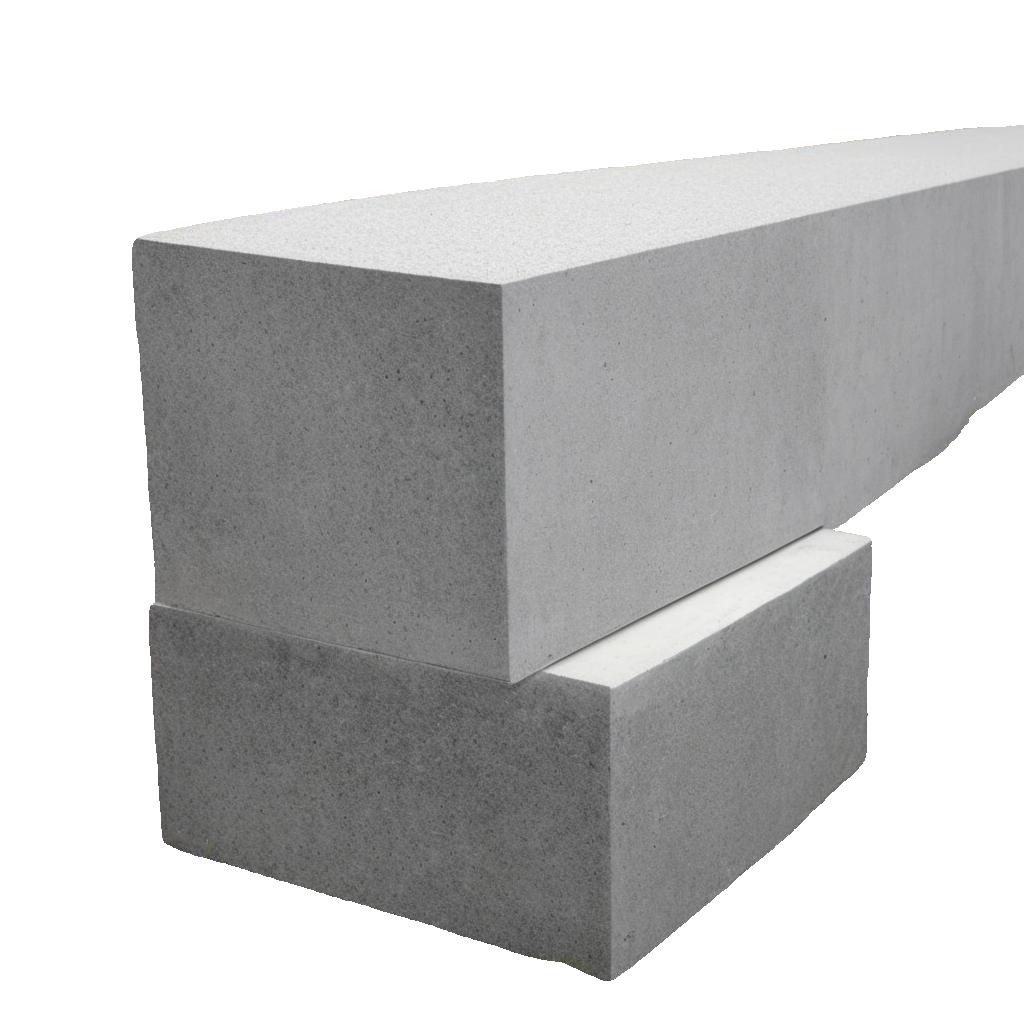} 
                    & \fontsize{26}{26}\selectfont{\emph{Sponge}}
                    & \includegraphics[width=7cm]{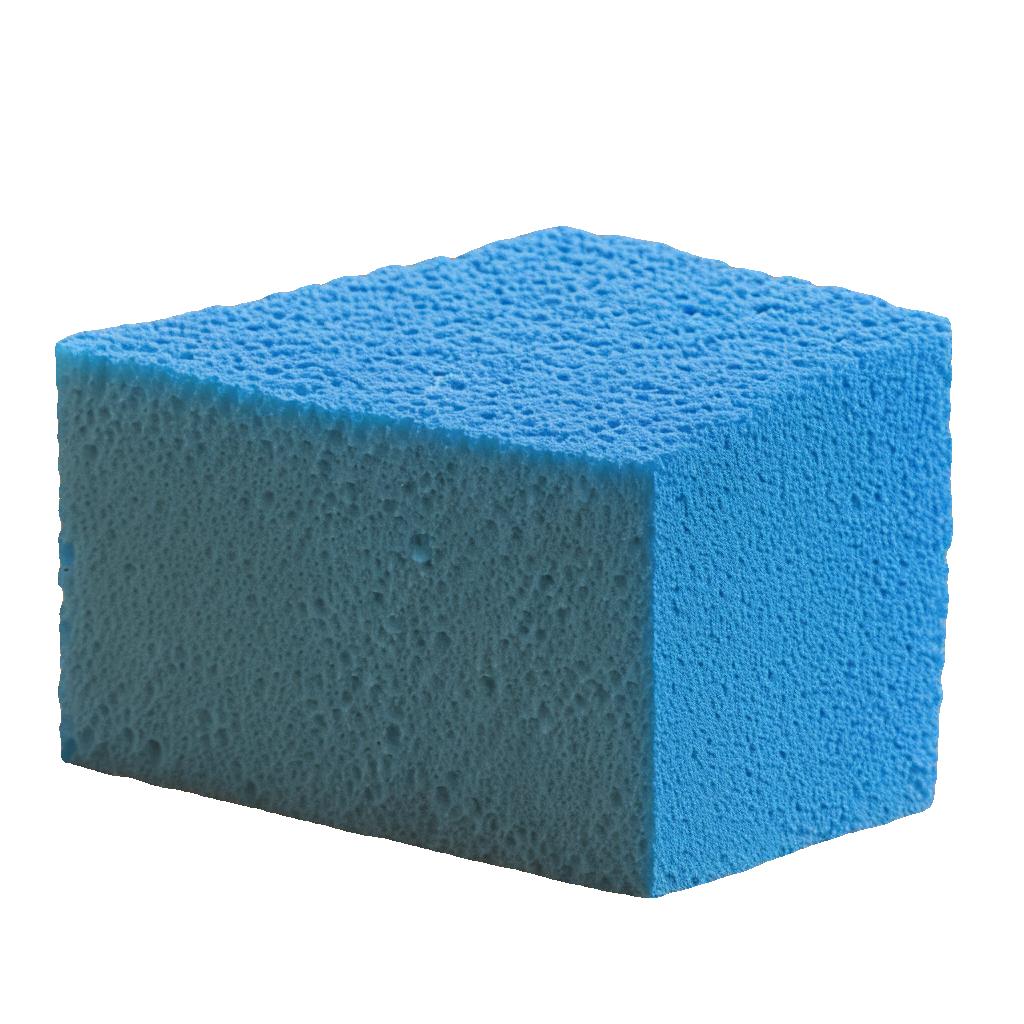} 
                    & \includegraphics[width=7cm]{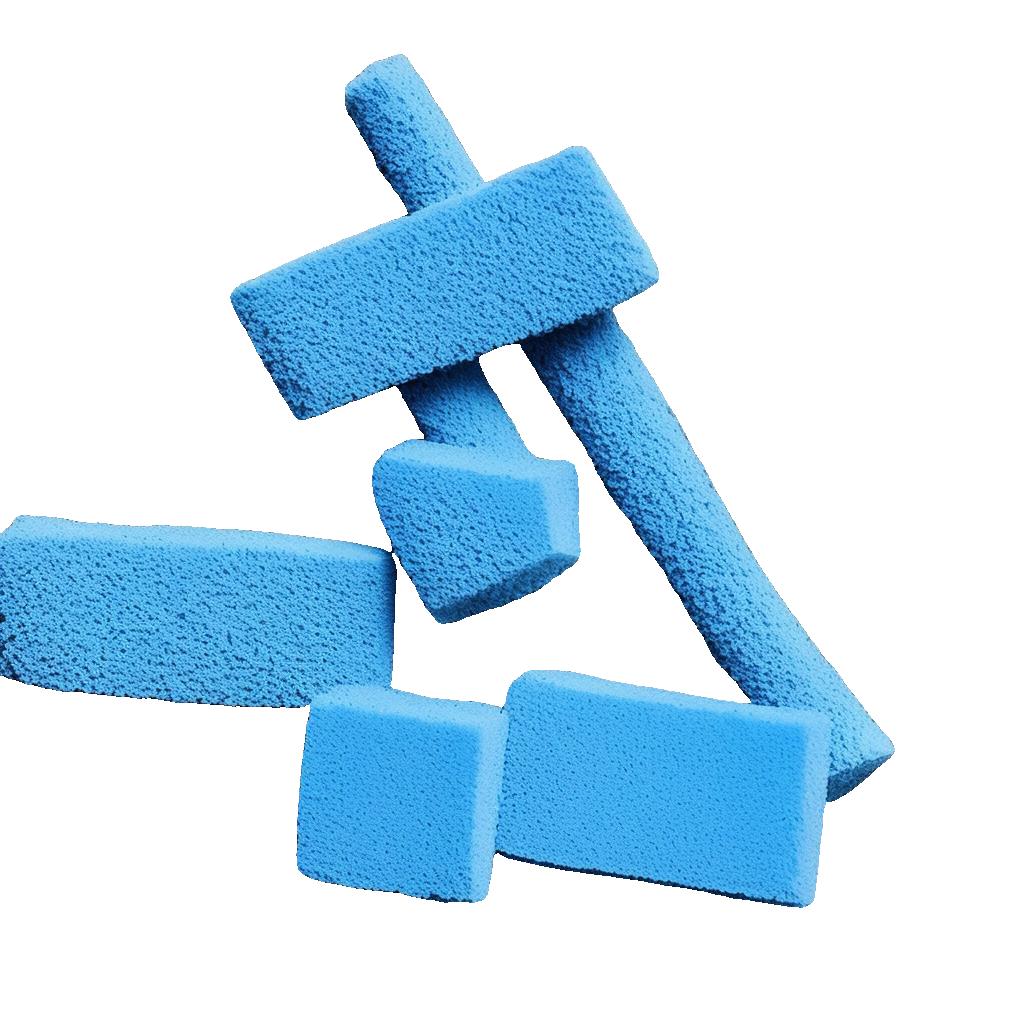} 
                    & \includegraphics[width=7cm]{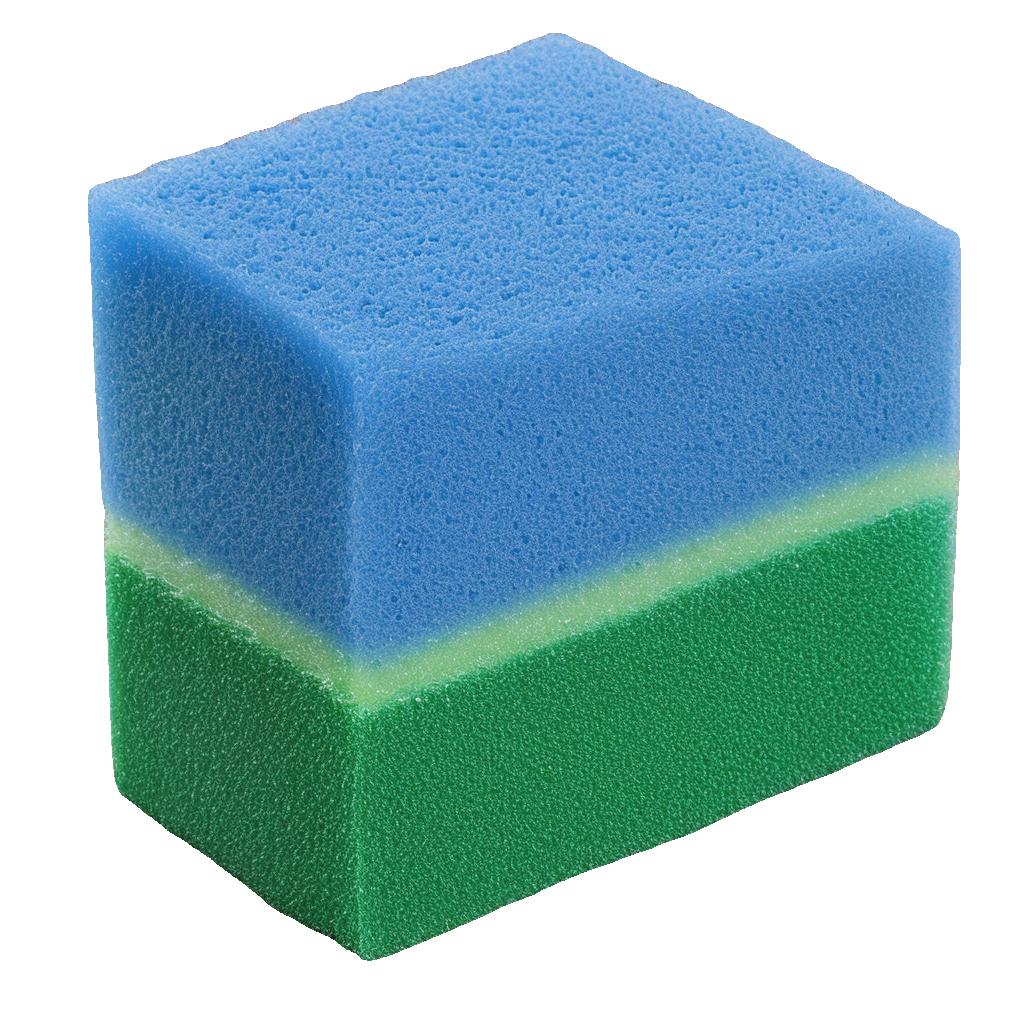} \\

                    & \fontsize{26}{26}\selectfont{\emph{post}} &  \fontsize{26}{26}\selectfont{\emph{bench}}& \fontsize{26}{26}\selectfont{\emph{plinth}} & & \fontsize{26}{26}\selectfont{\emph{block}} & \fontsize{26}{26}\selectfont{\emph{stick}} & 
                \fontsize{26}{26}\selectfont{\emph{scrubber}}\\

                  \fontsize{26}{26}\selectfont{\emph{Water}}
                    &\includegraphics[width=7cm]{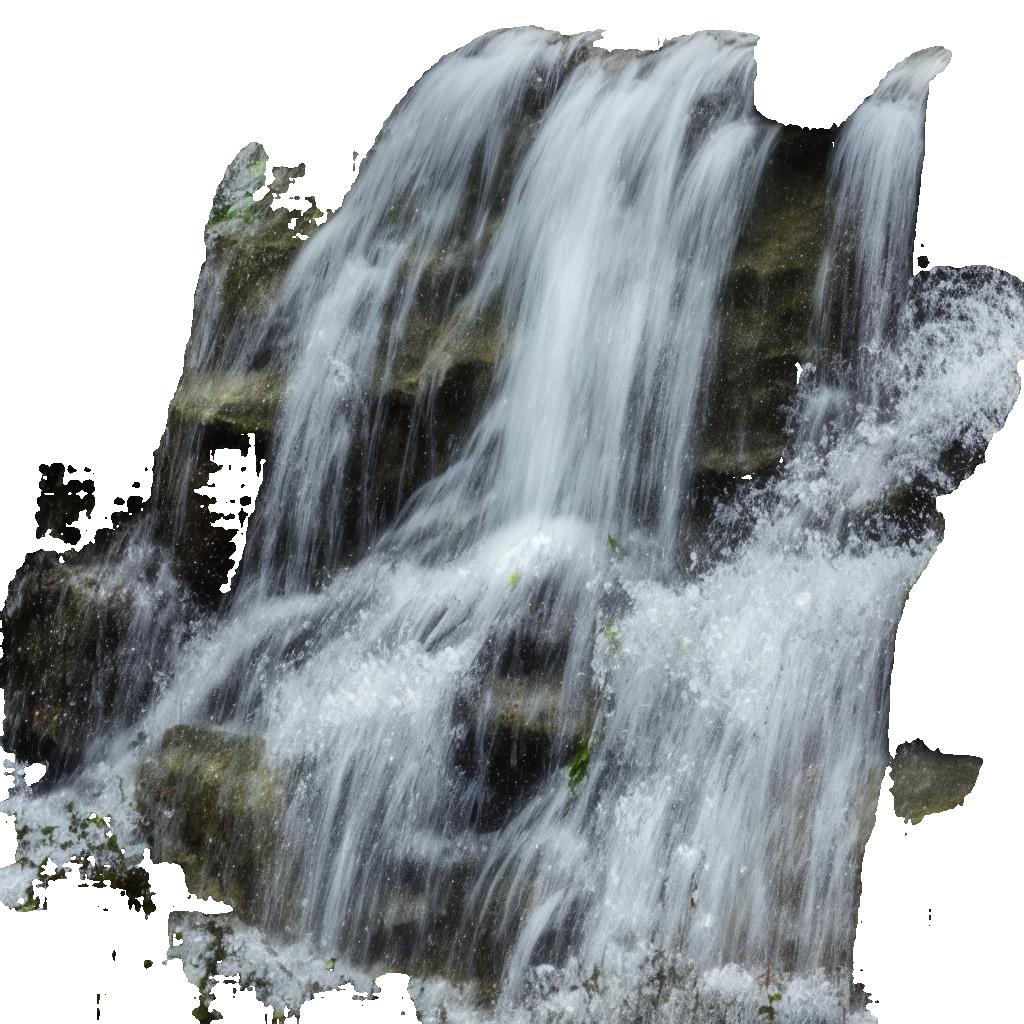} 
                 & \includegraphics[width=7cm]{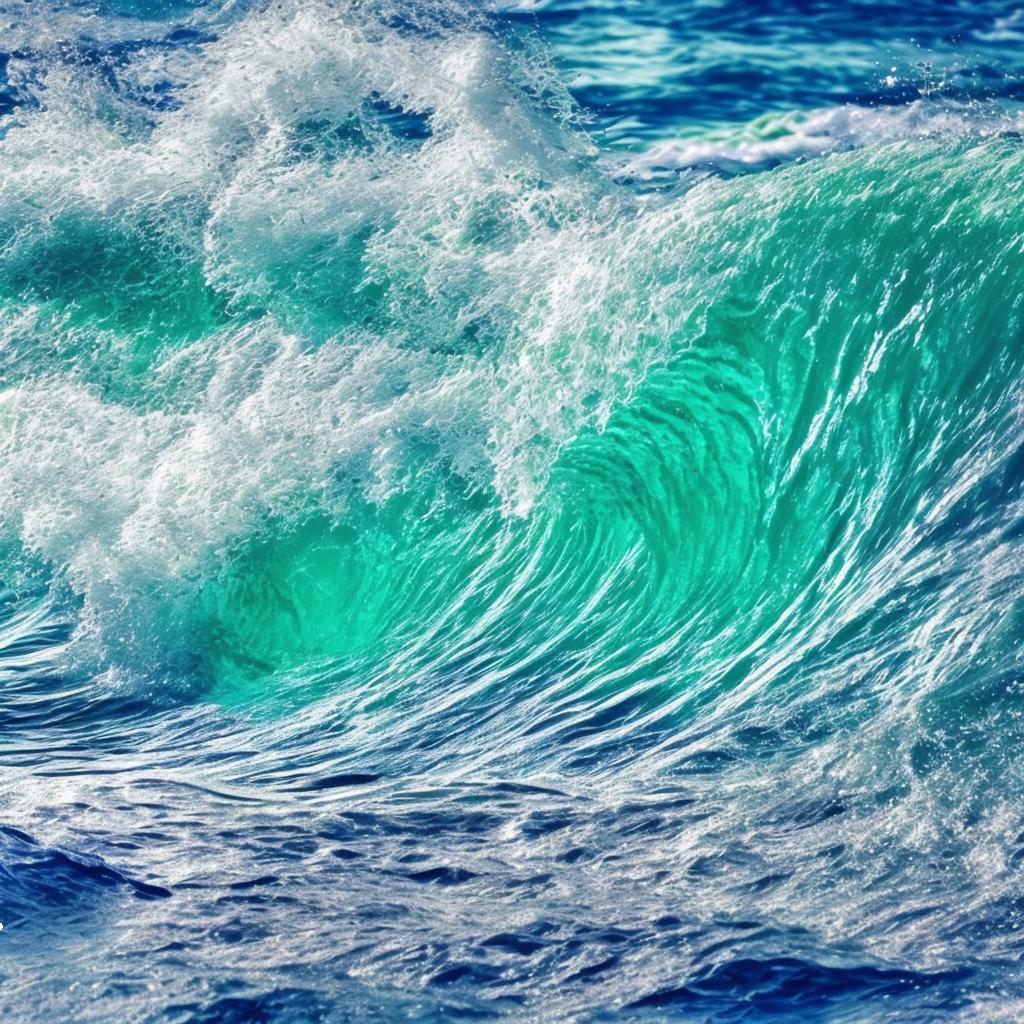} 
                    & \includegraphics[width=7cm]{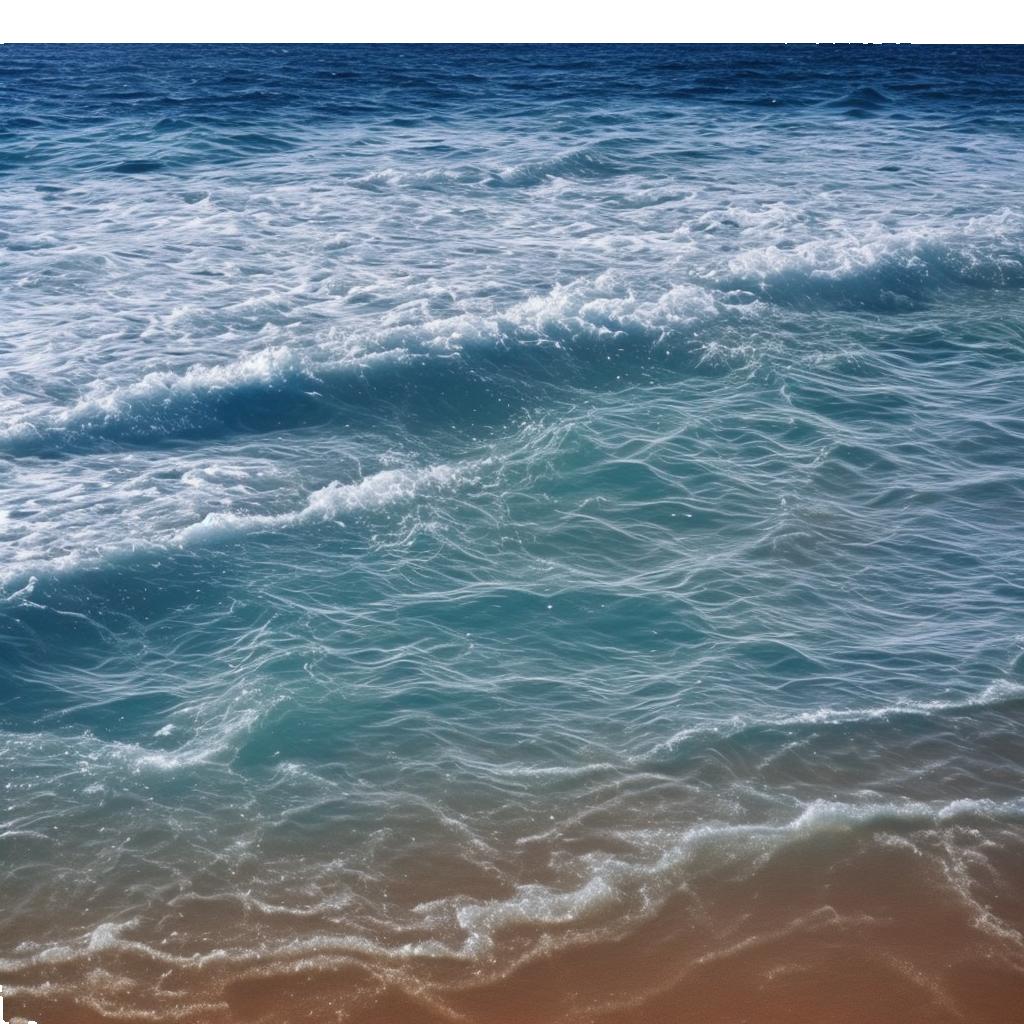} 
                    & \fontsize{26}{26}\selectfont{\emph{Fur}}
                    & \includegraphics[width=7cm]{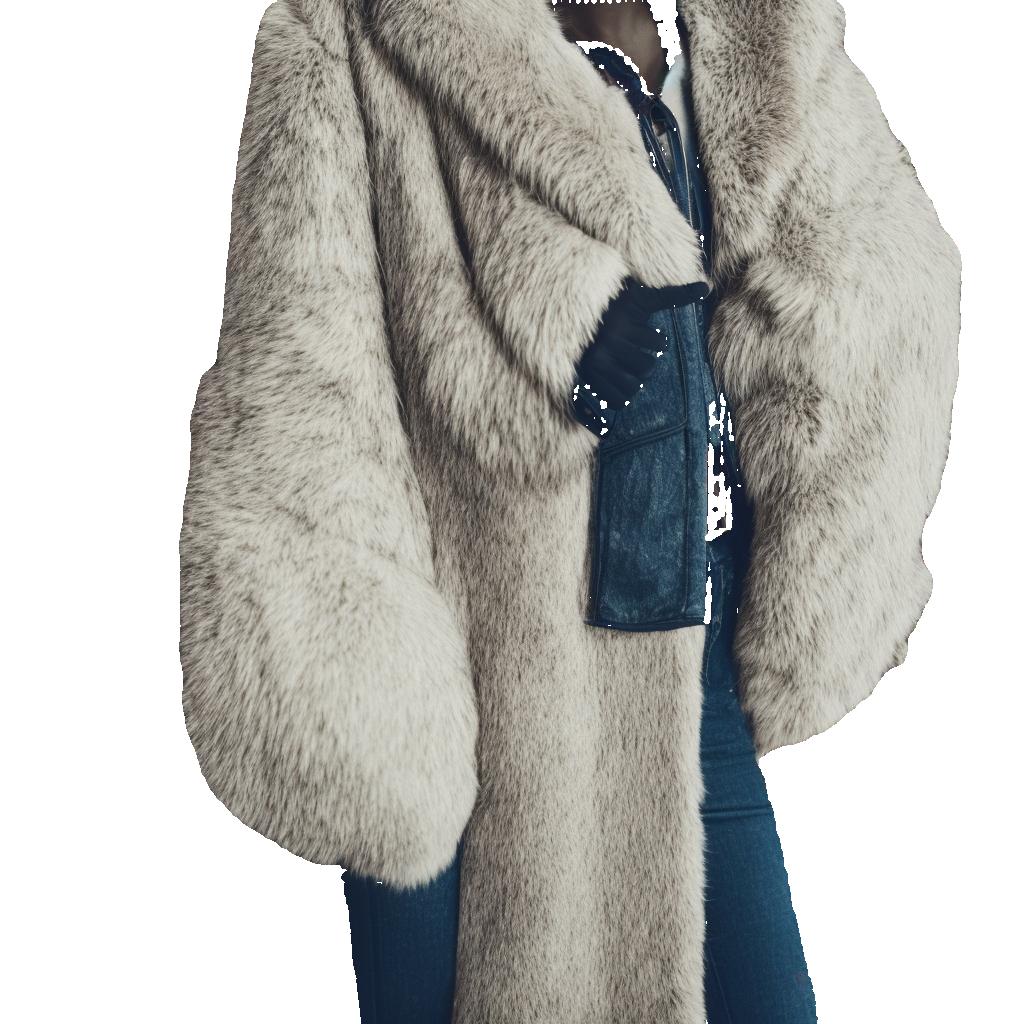} 
                    & \includegraphics[width=7cm]{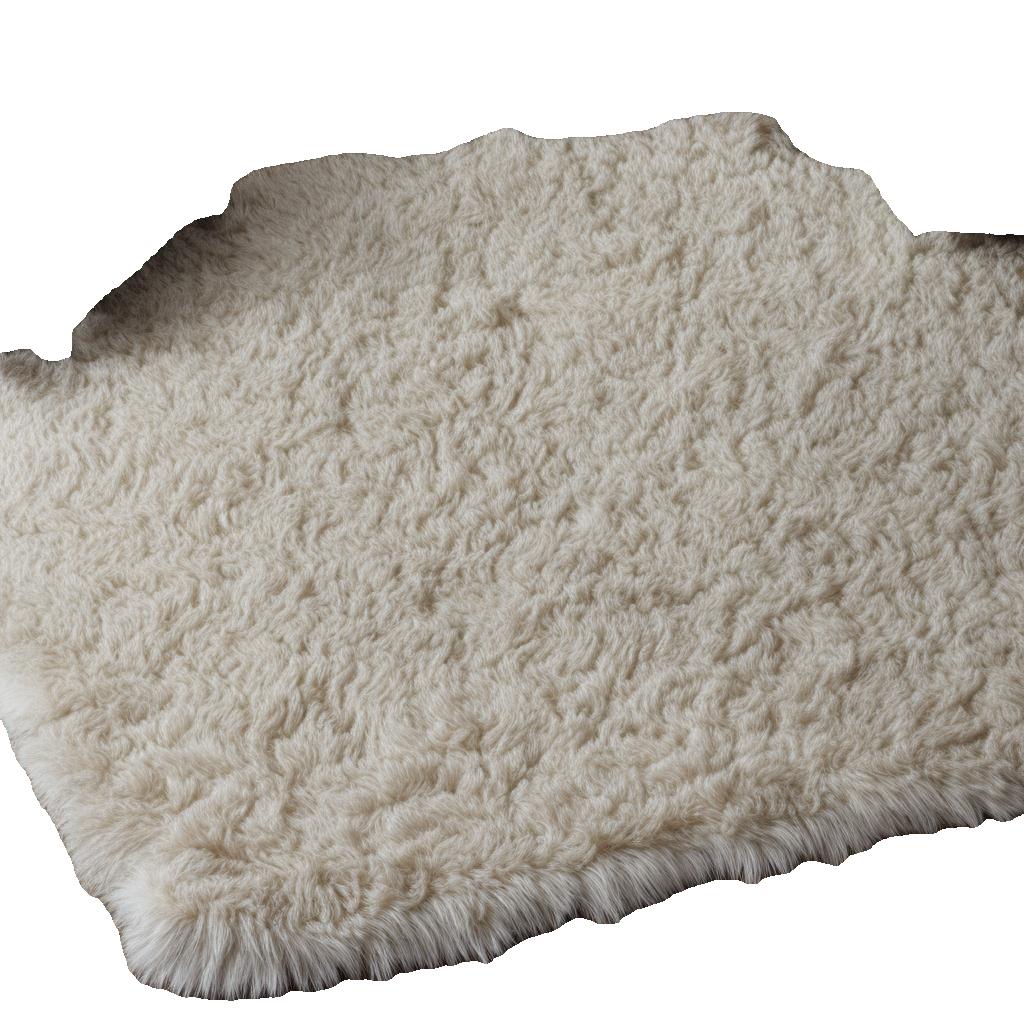} 
                    & \includegraphics[width=7cm]{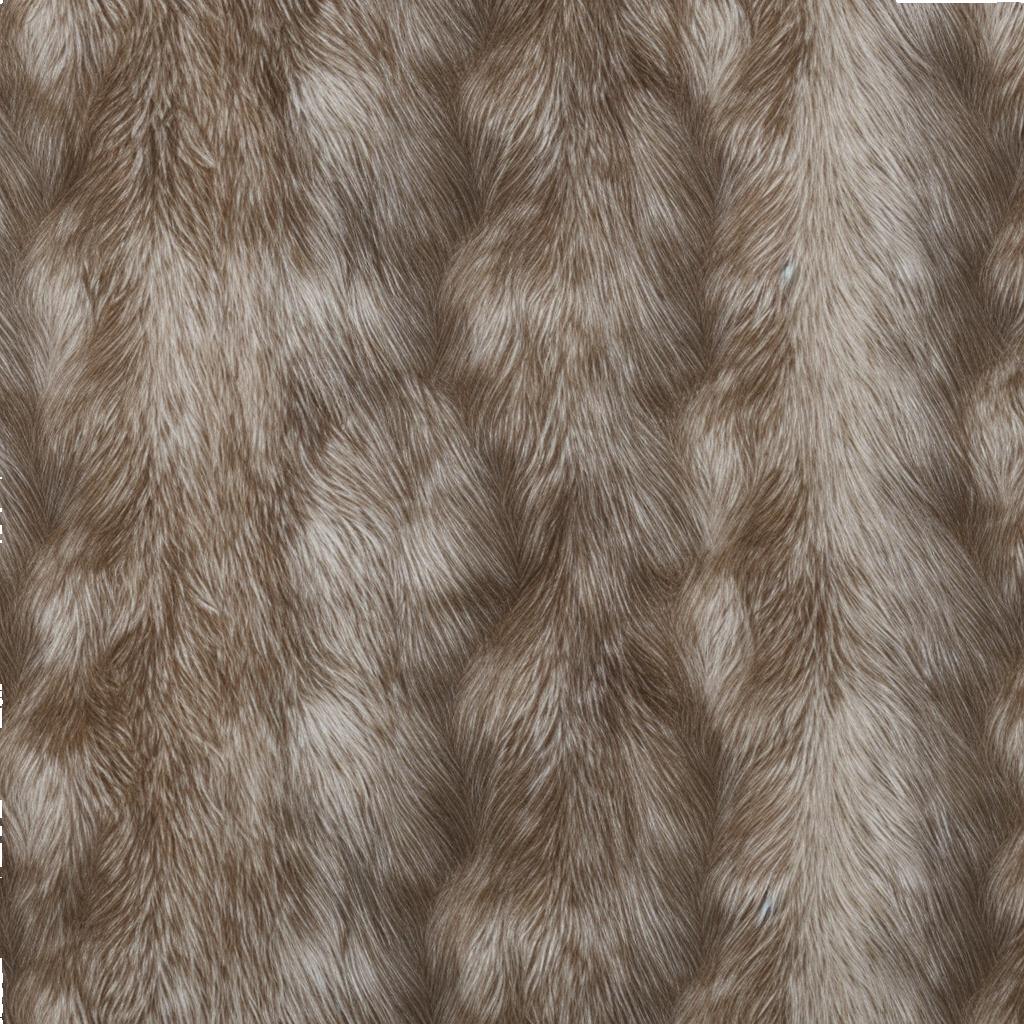} \\

                    & \fontsize{26}{26}\selectfont{\emph{waterfall}} &  \fontsize{26}{26}\selectfont{\emph{wave}}& \fontsize{26}{26}\selectfont{\emph{ocean}} & & \fontsize{26}{26}\selectfont{\emph{coat}} & \fontsize{26}{26}\selectfont{\emph{rug}} & 
                \fontsize{26}{26}\selectfont{\emph{panel}}\\

                \end{tabular}

            }
               {\centering
                 \resizebox{0.5\textwidth}{!}{%
                \begin{tabular}
                {cccc}
                  \fontsize{26}{26}\selectfont{\emph{Wicker}}
                    & \includegraphics[width=7cm]{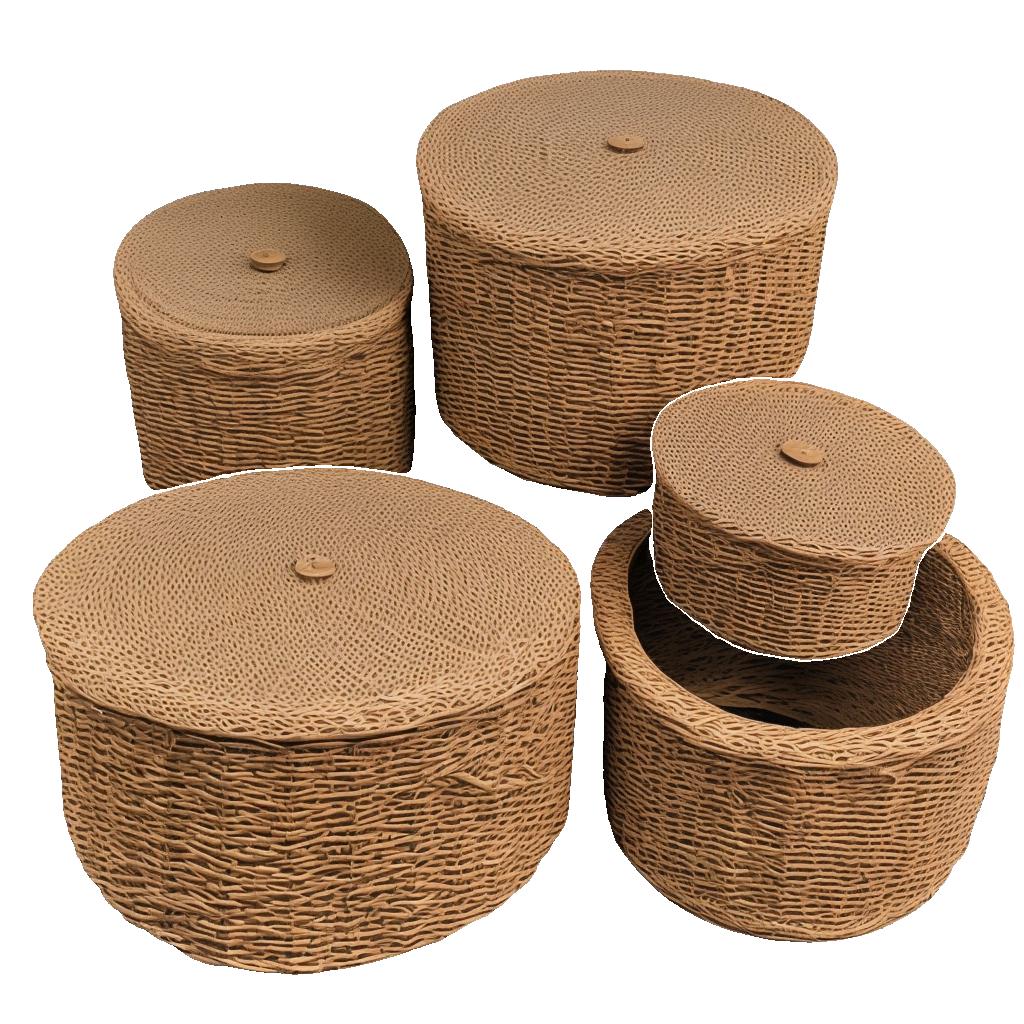} 
                    & \includegraphics[width=7cm]{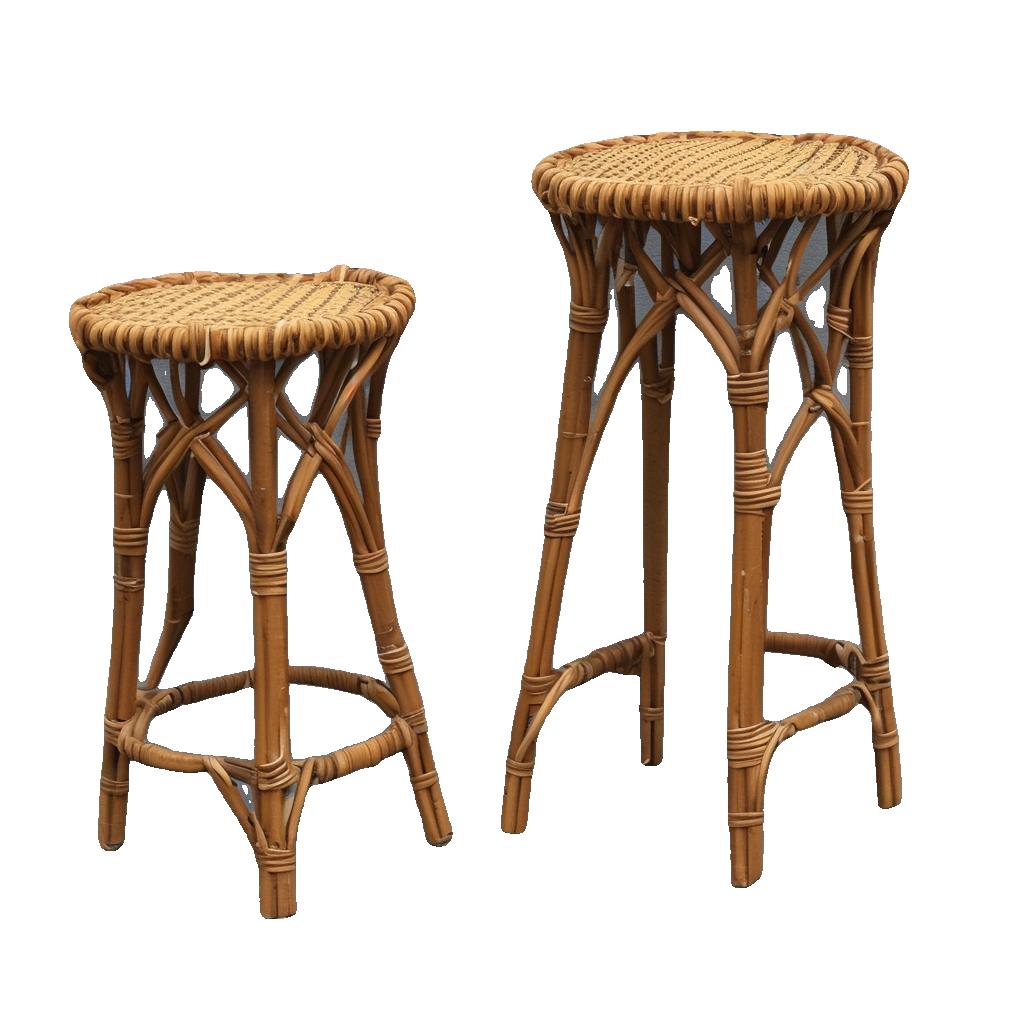} 
                    & \includegraphics[width=7cm]{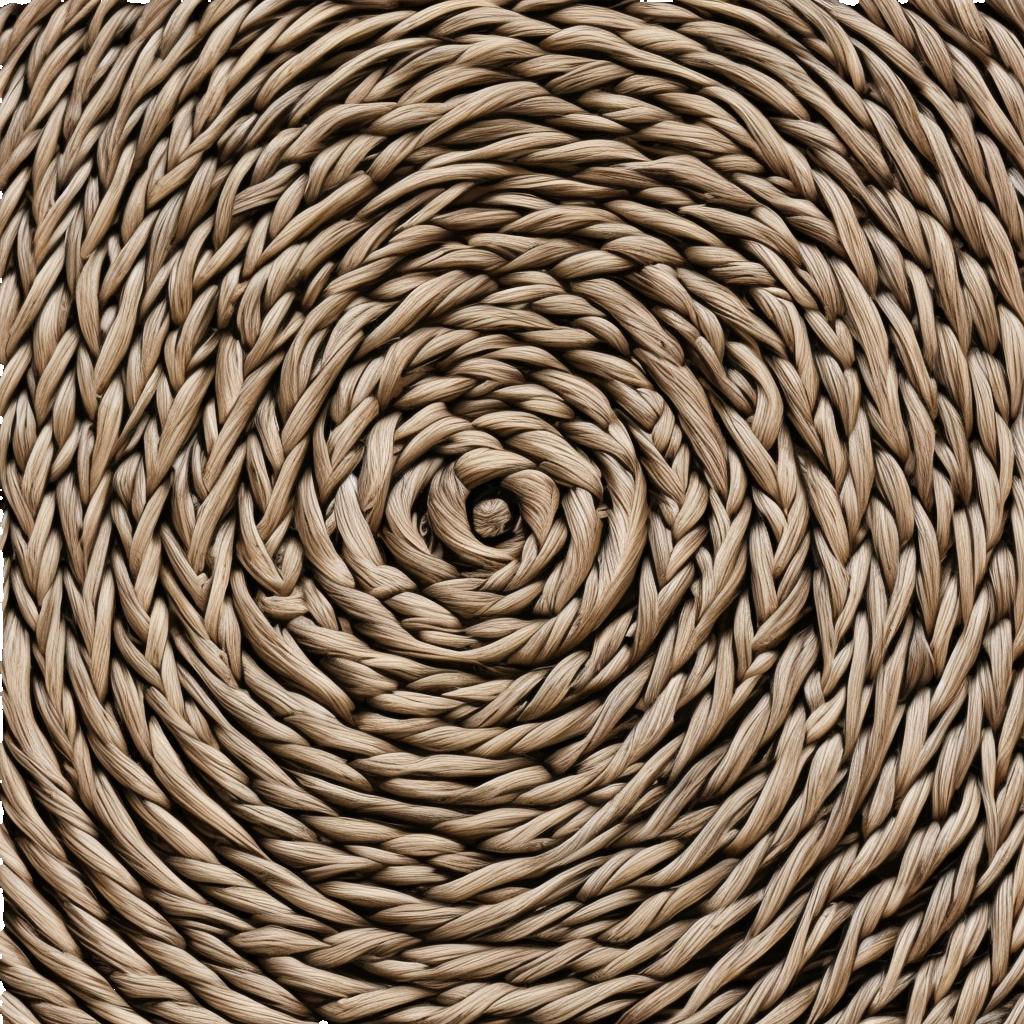} 
                   \\

              &  \fontsize{26}{26}\selectfont{\emph{box}} &  \fontsize{26}{26}\selectfont{\emph{stool}}& \fontsize{26}{26}\selectfont{\emph{tray}} \\
     
                 \end{tabular}
                 }
                \par}
                \captionof{figure}{Samples from our generated dataset and their extracted patches capture local material semantics across 21 classes.}
    \label{fig:figure10}
\vspace{-5mm}
\end{figure*}

Our generated dataset provides diverse material samples across 21 categories, each accompanied by extracted semantic patches that highlight the fine-grained local cues present in the Figure~\ref{fig:figure9} and Figure~\ref{fig:figure10}. These patches capture critical texture, reflectance, and structural details—demonstrating that the synthesized data contains rich and meaningful material semantics suitable for training and analysis.

\begin{figure*}[ht]
    \resizebox{\textwidth}{!}{%
 \begin{tabular}
                {cccccccc}

                    \includegraphics[width=7cm]{Figure_9/remote_plastic_ABS_0.jpg} 
                    & \includegraphics[width=7cm]{Figure_9/tire_rubber_butyl_3.jpg} 
                    & \includegraphics[width=7cm]{Figure_9/lamp_metal_nickel_0.jpg} 
                    & \includegraphics[width=7cm]{Figure_9/chair_leather_cowhide_2.jpg} 
                    & \includegraphics[width=7cm]{Figure_9/blanket_fabric_cashmere_3.jpg} 
                    & \includegraphics[width=7cm]{Figure_9/birdhouse_wood_cedar_2.jpg} 
                    & \includegraphics[width=7cm]{Figure_9/wall_stone_granite_4.jpg} 
                    & \includegraphics[width=7cm]{Figure_9/vase_ceramic_porcelain_2.jpg} \\

                  \fontsize{26}{26}\selectfont{\emph{Plastic}} &  \fontsize{26}{26}\selectfont{\emph{Rubber}} &  \fontsize{26}{26}\selectfont{\emph{Metal}}& \fontsize{26}{26}\selectfont{\emph{Leather}} & \fontsize{26}{26}\selectfont{\emph{Fabric}} & \fontsize{26}{26}\selectfont{\emph{Wood}} & 
                \fontsize{26}{26}\selectfont{\emph{Stone}} & \fontsize{26}{26}\selectfont{\emph{Ceramic}}\\
                
                  \includegraphics[width=7cm]{Figure_9/grip_bone_aged_1.jpg} 
                 & \includegraphics[width=7cm]{Figure_9/book_paper_lined_3.jpg} 
                    & \includegraphics[width=7cm]{Figure_9/yard_soil_wet_0.jpg} 
                    & \includegraphics[width=7cm]{Figure_9/earring_gemstone_opal_4.jpg} 
                    & \includegraphics[width=7cm]{Figure_9/cup_glass_etched_0.jpg} 
                    & \includegraphics[width=7cm]{Figure_9/candle_wax_flat_top_0.jpg} 
                    & \includegraphics[width=7cm]{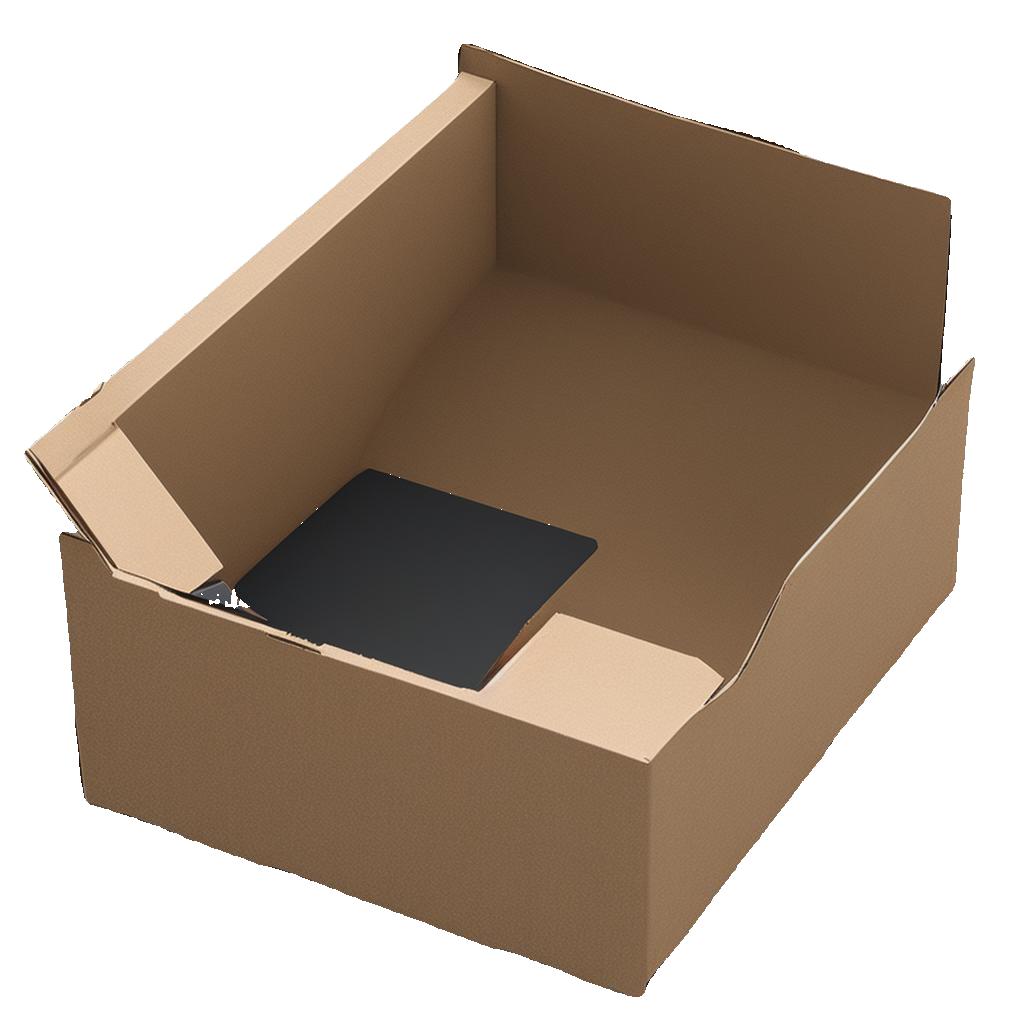} 
                    & \includegraphics[width=7cm]{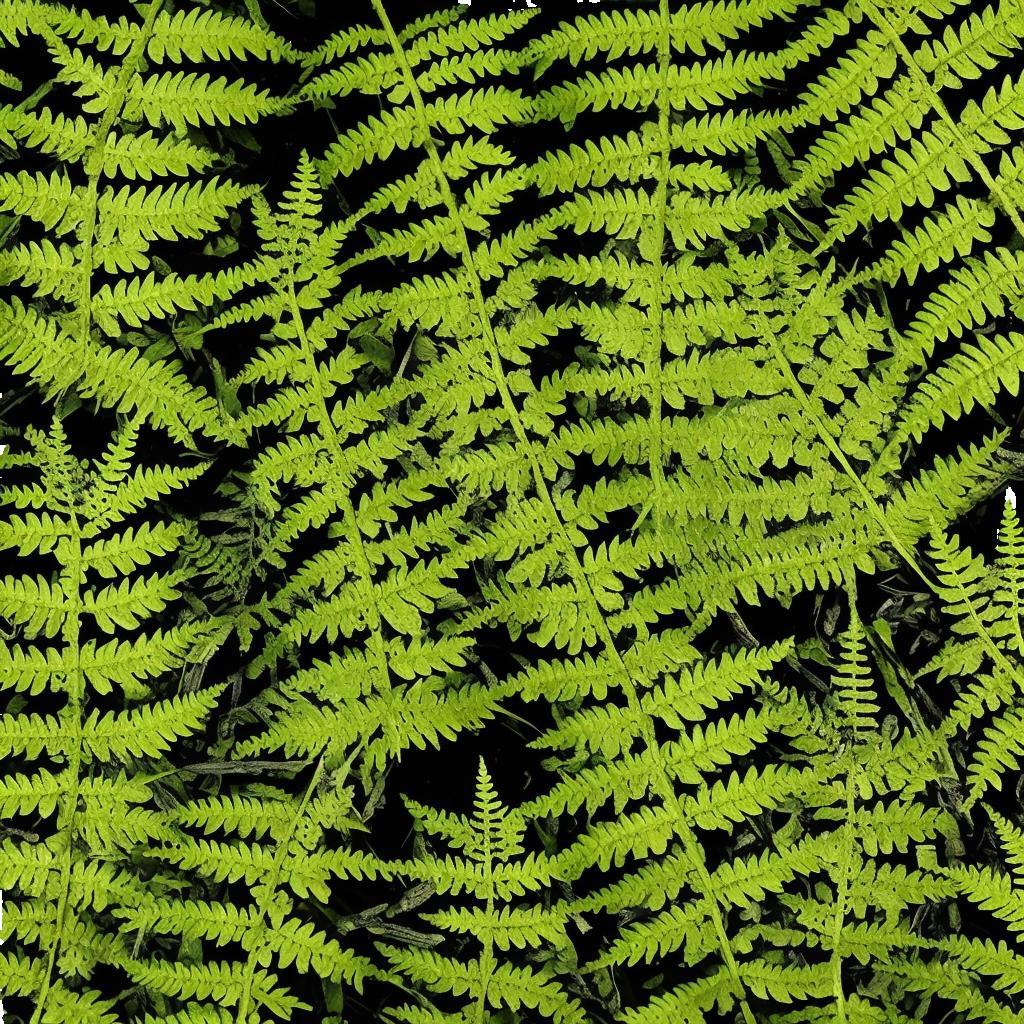} \\ 

                 \fontsize{26}{26}\selectfont{\emph{Bone}} &  \fontsize{26}{26}\selectfont{\emph{Paper}} &  \fontsize{26}{26}\selectfont{\emph{Soil}}& \fontsize{26}{26}\selectfont{\emph{Gemstone}} & \fontsize{26}{26}\selectfont{\emph{Glass}} & \fontsize{26}{26}\selectfont{\emph{Wax}} & 
                \fontsize{26}{26}\selectfont{\emph{Cardboard}} & \fontsize{26}{26}\selectfont{\emph{Foliage}}\\

                \end{tabular}

                }
               {\centering
                 \resizebox{0.625\textwidth}{!}{%
                \begin{tabular}
                {ccccc}
                \includegraphics[width=7cm]{Figure_9/post_concrete_plain_2.jpg} 
                    & \includegraphics[width=7cm]{Figure_9/block_sponge_blue_3.jpg} 
                    & \includegraphics[width=7cm]{Figure_9/waterfall_water_bubbly_1.jpg} 
                    & \includegraphics[width=7cm]{Figure_9/coat_fur_clean_3.jpg} 
                    & \includegraphics[width=7cm]{Figure_9/box_wicker_round_1.jpg} \\

                \fontsize{26}{26}\selectfont{\emph{Concrete}} &  \fontsize{26}{26}\selectfont{\emph{Sponge}} &  \fontsize{26}{26}\selectfont{\emph{Water}}& \fontsize{26}{26}\selectfont{\emph{Fur}} & \fontsize{26}{26}\selectfont{\emph{Wicker}}\\

                 \end{tabular}
                 }
                \par}
                \captionof{figure}{Generated material images using diverse prompts across 21 material categories, with one image per category.}
    \label{fig:generated_imgs}
\vspace{-5mm}
\end{figure*}

\end{document}